\RequirePackage{fix-cm}
\documentclass[
11pt, 
english, 
singlespacing, 
headsepline, 
]{MastersDoctoralThesis} 

\usepackage{multirow} 

\usepackage[utf8]{inputenc} 
\usepackage[T1]{fontenc} 
\usepackage{algorithm, algpseudocode}
\usepackage{mathpazo} 

\usepackage[backend=bibtex,style=authoryear,natbib=true]{biblatex} 

\usepackage{textpos}
\usepackage{tikz}
\usetikzlibrary{quotes,arrows.meta,positioning,decorations.pathreplacing,calc,3d,arrows}

\usepackage{tabularx}
\usepackage{mathtools}
\DeclarePairedDelimiter{\norm}{\lVert}{\rVert}

\usepackage{subfig}
\usepackage{url}
\usepackage[autostyle=true]{csquotes} 
\usepackage{xcolor}

\thesistitle{Deep Learning For Time Series Classification} 
\supervisor{
	Pr. Pierre-Alain \textsc{Muller}\\
	Pr. Lhassane \textsc{Idoumghar}\\
	Pr. Germain \textsc{Forestier}\\
	Dr. Jonathan \textsc{Weber}
} 
\advisor{
} 
\examiner{} 
\degree{Doctor of Philosophy} 
\author{Hassan \textsc{Ismail Fawaz}} 
\addresses{} 

\subject{Biological Sciences} 
\university{Université De Haute-Alsace} 
\department{{Institut de Recherche en Informatique, Mathématiques, Automatique et Signal}} 
\group{{Research Group Name}} 
\faculty{Faculté des Sciences et Techniques} 

\AtBeginDocument{
\hypersetup{pdftitle=\ttitle} 
\hypersetup{pdfauthor=\authorname} 
\hypersetup{pdfkeywords=\keywordnames} 
}

\begin{document}

\newcounter{definition}[section]
\newenvironment{definition}[1][]{\refstepcounter{definition}\par\medskip
	\noindent \textbf{Definition~\thedefinition. #1} \rmfamily}{\medskip}

\sloppy 
\frontmatter 

\pagestyle{plain} 


\begin{titlepage}
\begin{textblock}{}(0.03,0.29)
\begin{figure}
	\centering
	\includegraphics[width=3.5\linewidth]{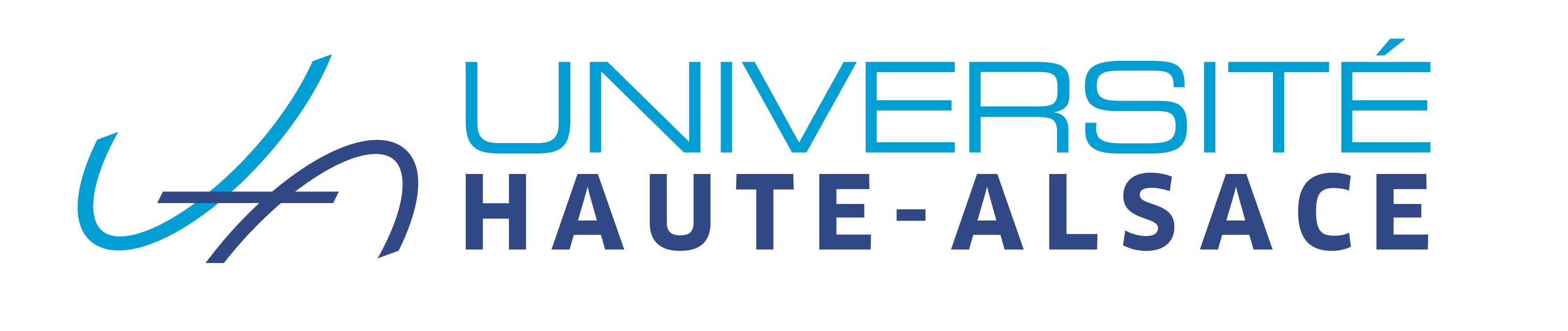}
\end{figure}
\end{textblock}
\begin{textblock}{}(8.5,0.29)
\begin{figure}
	\centering
	\includegraphics[width=1.5\linewidth]{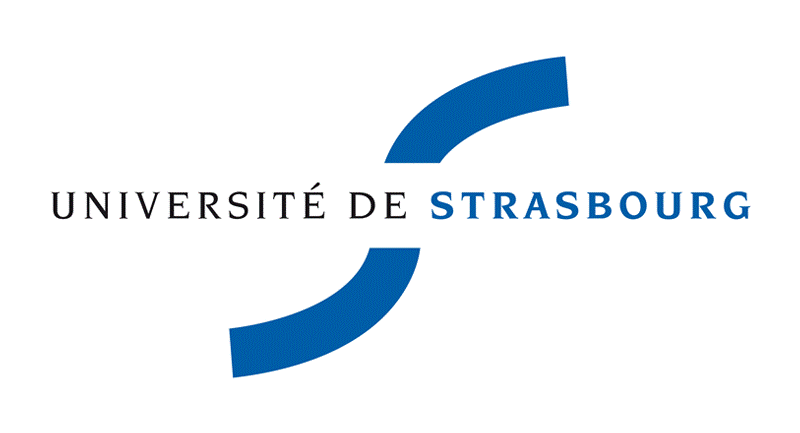}
\end{figure}
\end{textblock}
\begin{textblock}{2.0}(0.03,0.9)
Année 2020
\end{textblock}

\begin{textblock}{5.0}(5.5,0.9)
N$^\circ$  d’ordre : (attribué par le SCD)
\end{textblock}

\begin{center}

\vspace*{.09\textheight}
{\scshape\LARGE \univname}\\ 
{\scshape\small Université De Strasbourg}\vspace{0.5cm} \\
\textsc{\LARGE \textbf{Thèse}}\\[0.5cm] 
{\small Pour l'obtention du grade de \\ 
\textbf{Docteur de l'Université de Haute-Alsace \\ 
École Doctorale: Mathématiques, Sciences de l'Information et de l'Ingénieur (ED 269)}
Discipline : Informatique}\vspace{0.5cm} \\

{\small Présentée et soutenue publiquement \vspace{0.2cm} \\
par \vspace{0.2cm} \\ 
\LARGE{\authorname} \vspace{0.4cm} \\ 
\small Le 21 Septembre 2020\\ \vspace{1.5cm}
}

\HRule \\[0.4cm] 
{\huge \bfseries \ttitle\par}\vspace{0.4cm} 
\HRule \\[1.5cm] 

{\small Sous la direction du Prof. Pierre-Alain MULLER} \vspace{0.5cm} \\ 

{\normalsize Jury}

\begin{flushleft}
{\normalsize
Prof. Themis PALPANAS, Université de Paris (Rapporteur)  \vspace{0.1cm}\\
Prof. Pierre-François MARTEAU, Université Bretagne Sud (Rapporteur)  \vspace{0.1cm}\\
Prof. Anthony BAGNALL, Université d'East Anglia (Examinateur)  \vspace{0.1cm}\\
Prof. Laetitia JOURDAN, Université de Lille (Examinatrice)  \vspace{0.1cm}\\
Prof. Pierre-Alain MULLER, Université de Haute-Alsace (Directeur de thèse)  \vspace{0.1cm}\\
Prof. Lhassane IDOUMGHAR, Université de Haute-Alsace (co-Directeur de thèse)  \vspace{0.1cm}\\
Prof. Germain FORESTIER, Université de Haute-Alsace (co-Directeur de thèse)  \vspace{0.1cm}\\
Dr. Jonathan WEBER, Université de Haute-Alsace (Encadrant de thèse) \\

}
\end{flushleft}

%
%
%
%
\end{center}
\end{titlepage}

\begin{extraAbstract}
\addchaptertocentry{\resumename} 
La science des données s’intéresse aux théories et aux algorithmes permettant d’extraire des connaissances de grandes masses de données. 
L’analyse de séries temporelles est le sous-domaine de la science des données qui s’intéresse à l’analyse de données composées de suites de valeurs numériques ordonnées dans le temps. 
Les séries temporelles sont particulièrement intéressantes car elles permettent de comprendre l’évolution des états d’un processus au cours du temps. 
Leur analyse peut ainsi révéler des tendances, des relations et des similarités à travers les données. De très nombreux domaines produisent des données sous la forme de séries temporelles : données de santés (électrocardiogramme, glycémie, etc.), reconnaissance d'activités, télédétection, finance (cours de bourse), industrie (capteurs).   

Au sein de la science des données, la classification est une tâche supervisée qui consiste à apprendre un modèle à partir de données étiquetées organisées en classes afin de prédire la classe de nouvelles données. 
La classification de séries temporelles s'intéresse aux algorithmes de classification dédiés au traitement de séries temporelles. Par exemple, à l’aide d’un ensemble étiqueté d’électrocardiogrammes de patients sains ou présentant un problème cardiaque, l’objectif est d'entraîner un modèle capable de prédire si un nouvel électrocardiogramme présente ou non une pathologie. 
Les spécificités des données temporelles imposent le développement d’algorithmes dédiés au traitement de ces données, les modèles existants pour d’autres type de données (images, vidéos, etc.) n’étant pas toujours adaptés.

Dans ce contexte, l’apprentissage profond (deep learning) s’est imposé au cours des dernières années comme une des méthodes les plus performantes pour réaliser la tâche de classification, notamment dans le domaine de la vision par ordinateur. 
L’objectif principal de cette thèse a été d’étudier et de développer des modèles profonds spécifiquement construits pour la classification de séries temporelles. 
Nous avons ainsi réalisé la première étude expérimentale permettant de comparer les méthodes profondes existantes et de les positionner par rapport aux méthodes de l’état de l’art n’utilisant pas l’apprentissage profond. 
Par la suite, nous avons effectué de nombreuses contributions dans ce domaine, notamment dans le cadre de l’apprentissage par transfert, l'augmentation de données, la création d’ensembles et l'attaque adversaire. 
Enfin, nous avons également proposé une nouvelle architecture profonde, basée sur le célèbre réseau Inception (Google), qui se positionne parmis les plus performantes à ce jour.

Nos expériences menées sur des benchmarks comportant plus d’un centaine de jeux de données nous ont permis de valider les performances de nos contributions. 
Enfin, nous avons également montré la pertinence des approches profondes dans le domaine de la science des données chirurgicales (surgical data science) où nous avons proposé une approche interprétable afin d’évaluer les compétences chirurgicales à partir de données cinématiques de séries temporelles multivariées.
\end{extraAbstract}

\begin{abstract}
\addchaptertocentry{\abstractname} 

Data science is about designing algorithms and pipelines for extracting knowledge from large masses of data. 
Time series analysis is a field of data science which is interested in analyzing sequences of numerical values ordered in time. 
Time series are particularly interesting because they allow us to visualize and understand the evolution of a process over time. 
Their analysis can reveal trends, relationships and similarities across the data. 
There exists numerous fields containing data in the form of time series: health care (electrocardiogram, blood sugar, etc.), activity recognition, remote sensing, finance (stock market price), industry (sensors), etc.

In data mining, classification is a supervised task that involves learning a model from labeled data organized into classes in order to predict the correct label of a new instance.
Time series classification consists of constructing algorithms dedicated to automatically label time series data. 
For example, using a labeled set of electrocardiograms from healthy patients or patients with a heart disease, the goal is to train a model capable of predicting whether or not a new electrocardiogram contains a pathology.
The sequential aspect of time series data requires the development of algorithms that are able to harness this temporal property, thus making the existing off-the-shelf machine learning models for traditional tabular data suboptimal for solving the underlying task. 

In this context, deep learning has emerged in recent years as one of the most effective methods for tackling the supervised classification task, particularly in the field of computer vision.
The main objective of this thesis was to study and develop deep neural networks specifically constructed for the classification of time series data.
We thus carried out the first large scale experimental study allowing us to compare the existing deep methods and to position them compared other non-deep learning based state-of-the-art methods.
Subsequently, we made numerous contributions in this area, notably in the context of transfer learning, data augmentation, ensembling and adversarial attacks.
Finally, we have also proposed a novel architecture, based on the famous Inception network (Google), which ranks among the most efficient to date.

Our experiments carried out on benchmarks comprising more than a hundred data sets enabled us to validate the performance of our contributions.
Finally, we also showed the relevance of deep learning approaches in the field of surgical data science where we proposed an interpretable approach in order to assess surgical skills from kinematic multivariate time series data.

\end{abstract}


\begin{acknowledgements}
\addchaptertocentry{\acknowledgementname} 
This work would not have been done without the support of many people who I wish to thank here. 

I would like to start by thanking my supervisors: Prof. Germain Forestier, Dr. Jonathan Weber, Prof. Lhassane Idoumghar and Prof. Pierre-Alain Muller who guided me throughout this three years journey by sharing their wisdom and intellect while giving the freedom to pursue various research ideas allowing me to gain micro-managerial skills and autonomy.

I would like to express my gratitude to Prof. Themis Palpanas and Prof. Pierre-François Marteau having accepted to review this PhD. 
I would also like to thank the other members of the jury who did me the honor of judging this work: Prof. Laetitia Jourdan and Prof. Anthony Bagnall. 

I would like to thank Dr. François Petitjean and Prof. Geoff Webb, as well as all of the amazing colleagues at the Monash University in Melbourne, for hosting me as a visiting researcher, allowing me to nurture my doctorate with this exceptional international experience. 

I would like to thank every researcher who worked on providing and preparing the public dataset which have been the backbone of my PhD work: the University of California Riverside, the University of East Anglia and the John Hopkins University. 

I would like to thank the Mésocentre of Strasbourg that provided support and access to the GPU cluster, providing this huge computational resources to launch the experiments. 
Similarly, many thanks to Nvidia Corp. that granted us the GTX Quadro P6000 enabling a significant speed up in development and testing. 

Last but not least, I would like to thank all of my colleagues at the Université de Haute-Alsace for their contributions, both professionally and personally.

Finally, I would like to thank my parents, brothers and fiancée for their support during this three years journey, as well as in the previous years.

\end{acknowledgements}


\tableofcontents 

\listoffigures 

\listoftables 


\begin{abbreviations}{ll} 

\textbf{CST} & Classification de séries temporelles\\
\textbf{ECG} & Electrocardiogram\\
\textbf{TSC} & Time Series Classification\\
\textbf{NN} & Nearest Neighbor\\
\textbf{DTW} & Dynamic Time Warping\\
\textbf{ED} & Euclidean Distance\\
\textbf{BOSS} & Bag-of-SFA-Symbols\\
\textbf{DNN} & Deep Neural Networks\\
\textbf{NLP} & Natural Language Processing\\
\textbf{ResNet} & Residual Network\\
\textbf{CNN} & Convolutional Neural Networks\\
\textbf{UCR} & University of California Riverside\\
\textbf{UEA} & University of East Anglia\\
\textbf{TWE} & Time Warp Edit \\
\textbf{MSM} & Move Split Merge\\
\textbf{NN-DTW} & NN coupled with DTW\\
\textbf{SVM} & Support Vector Machine\\
\textbf{COTE} & Collective Of Transformation-based Ensembles\\
\textbf{HIVE-COTE} & Hierarchical Vote Collective of Transformation-Based Ensembles\\
\textbf{ST} & Shapelet Transform\\
\textbf{GPU} & Graphical Processing Unit\\
\textbf{MTS} & Multivariate Time Series\\
\textbf{CAM} & Class Activation Map\\
\textbf{MLP} & Multi Layer Perceptron\\
\textbf{ESN} & Echo State Network\\
\textbf{FC} & Fully Connected\\
\textbf{ReLU} & Rectified Linear Unit\\
\textbf{RNN} & Recurrent Neural Network\\
\textbf{SDAE} & Stacked Denoising Auto-Encoder\\
\textbf{DBN} & Deep Belief Network\\
\textbf{FCN} & Fully Convolutional Network\\
\textbf{GAP} & Global Average Pooling\\
\textbf{PReLU} & Parametric Rectified Linear Unit\\
\textbf{MCNN} & Multi-scale Convolutional Neural Network\\
\textbf{WS} & Window Slicing\\
\textbf{T-LeNet} & Time Le-Net\\
\textbf{WW} & Window Warping\\
\textbf{MCDCNN} & Multi Channel Deep Convolutional Neural Network\\
\textbf{MSE} & Mean Squared Error\\
\textbf{TWIESN} & Time Warping Invariant Echo State Network\\
\textbf{SGD} & Stochastic Gradient Descent\\
\textbf{EE} & Elastic Ensemble\\
\textbf{PF} & Proximity Forest\\
\textbf{MDS} & Multi-Dimensional Scaling\\
\textbf{t-SNE} & t-distributed Stochastic Neighbor Embedding\\
\textbf{TEKA} & Time Elastic Kernel Averaging\\
\textbf{NNE} & Neural Network Ensemble\\
\textbf{FGSM} & Fast Gradient Sign Method\\
\textbf{KNN} & K Nearest Neighbors\\
\textbf{AdvProp} & Adversarial Propagation\\
\textbf{BIM} & Basic Iterative Method\\
\textbf{RF} & Receptive Field\\
\textbf{SITS} & Satellite Image Time Series\\
\textbf{OSATS} & Objective Structured Assessment of Technical Skills\\
\textbf{GMF} & Global Movement Features\\
\textbf{JIGSAWS} & JHU-ISI Gesture and Skill Assessment Working Set\\
\textbf{s-HMM} & Sparse Hidden Markov Model\\
\textbf{ApEn} & Approximate Entropy\\
\textbf{NLTS} & Non-Linear Temporal Scaling

\end{abbreviations}


%
%
%


%
%
%
%




\mainmatter 

\newcommand{\ourmethod}[1]{InceptionTime}

\pagestyle{thesis} 

\addchap{Résumé des chapitres en Français}

\section*{Chapitre 1: L'état de l'art de la classification de séries temporelles}
\addcontentsline{toc}{section}{Chapitre 1: L'état de l'art de la classification des séries temporelles}

Au cours des deux dernières décennies, la Classification de Séries Temporelles (CST) a été considéré comme l'un des problèmes les plus difficiles dans la fouille de données~\citep{yang200610,esling2012time}.
Avec l'augmentation de la disponibilité des données temporelles~\citep{silva2018speeding}, des centaines d'algorithmes de CST ont été proposés depuis 2015~\citep{bagnall2017the}.
En raison de leur ordre temporel, les séries sont présentes dans presque tout problème de fouille de données~\citep{langkvist2014a}.
En fait, tout problème de classification, utilisant des données enregistrées ayant un ordre spécifique, peut être converti en un problème de CST~\citep{gamboa2017deep}.
Les séries temporelles sont présentes dans de nombreuses applications du monde réel, allant des soins de santé~\citep{gogolou2018comparing} à la reconnaissance de l'activité humaine~\citep{wang2018deepsurvey,mathis2020knowledge} jusqu'à la classification des scènes acoustiques~\citep{nwe2017convolutional} et la cybersécurité~\citep{susto2018time}.
De plus, la diversité des types d'ensembles de données dans l'archive UCR/UEA~\citep{ucrarchive, bagnall2017the} (la plus grande base de données de références de séries temporelles) montre les différentes applications de la CST.

Compte tenu de la nécessité de classer avec précision les séries temporelles, les chercheurs ont proposé des centaines de méthodes pour résoudre cette tâche~\citep{bagnall2017the}.
L'une des approches de CST les plus populaires et traditionnelles est l'utilisation d'un classifieur NN couplé à une fonction de distance~\citep{lines2015time}.
En particulier, DTW lorsqu'il est utilisé avec un classifieur NN s'est révélé être une méthode de référence très solide~\citep{bagnall2017the}.
\cite{lines2015time} ont comparé plusieurs mesures de distance - telles que TWE~\citep{marteau2009time} et MSM~\citep{stefan2013the} - montrant qu'il n'y a pas de mesure de distance unique qui surpasse DTW.
Ils ont également montré que l'ensembling des classifieurs NN individuels (avec différentes mesures de distance) surpasse toutes les composantes individuelles de l'ensemble.
Par conséquent, les contributions récentes se sont concentrées sur le développement de méthodes d'ensembling qui supplante considérablement le NN-DTW~\citep{bagnall2016time, hills2014classification, bostrom2015binary, lines2016hive, schafer2015the, kate2016using, deng2013a, baydogan2013a}.
Ces approches utilisent soit un ensemble d'arbres de décision (forêt aléatoire)~\citep{baydogan2013a, deng2013a} ou un ensemble de différents types de classifieurs (SVM, NN avec plusieurs distances) sur une ou plusieurs transformations~\citep{bagnall2016time, bostrom2015binary, schafer2015the, kate2016using}.
La plupart de ces approches obtiennent des résultats significativement meilleurs que NN-DTW~\citep{bagnall2017the} et partagent une propriété commune, qui est la phase de transformation des données où les séries temporelles sont transformées en un nouvel espace de description (par exemple en utilisant des transformations shapelets~\citep{bostrom2015binary} ou des caractéristiques DTW~\citep{kate2016using}).
Cette notion a motivé le développement d'un ensemble de 35 classifieurs nommé COTE~\citep{bagnall2016time} qui non seulement regroupe différents classifieurs sur la même transformation, mais regroupe à la place différents classifieurs utilisant plusieurs transformations latentes de séries temporelles.
\cite{lines2016hive, lines2018time} ont étendu COTE avec HIVE-COTE qui permet une amélioration significative par rapport à COTE en tirant parti d'une nouvelle structure hiérarchique avec un vote probabiliste, comprenant deux nouveaux classifieurs et deux domaines de transformation de représentation supplémentaires.
HIVE-COTE est actuellement considéré comme l'état de l'art pour la classification des séries temporelles~\citep{bagnall2017the} lorsqu'il est évalué sur les 85 jeux de données de l'archive UCR/UEA.

Pour atteindre sa haute précision, HIVE-COTE devient extrêmement intensif en calcul et peu pratique pour fonctionner sur un vrai problème d'exploration de données volumineuses~\citep{bagnall2017the}.
L'approche nécessite l'entrainement de 37 classifieurs ainsi que la validation croisée de chaque hyperparamètre de ces algorithmes, ce qui rend l'approche impossible à appliquer dans certaines situations~\citep{lucas2018proximity}.
Pour souligner cette inaptitude, notons que l'un de ces 37 classifieurs est le ST~\citep{hills2014classification} dont la complexité temporelle est $ O (n ^ 2 \cdot l ^ 4) $ avec $ n $ étant le nombre de séries temporelles dans l'ensemble de données et $ l $ étant la longueur d'une série temporelle.
À la complexité du temps d'apprentissage s'ajoute le temps de \emph{classification} élevé de l'un des 37 classifieurs: le plus proche voisin qui doit analyser l'ensemble d'apprentissage avant de prendre une décision au moment du test.
Par conséquent, étant donné que le plus proche voisin constitue une composante essentielle de HIVE-COTE, son déploiement dans un environnement en temps réel est encore limité, voire impossible.
Enfin, s'ajoutant à l'énorme temps d'exécution de HIVE-COTE, la décision prise par 37 classifieurs ne peut pas être interprétée facilement par les experts du domaine, car il est déjà difficile de comprendre les décisions prises par un unique classifieur.
Notons que récemment,~\cite{bagnall2020usage} ont proposé une nouvelle version de HIVE-COTE qui est sensiblement plus rapide, montrant l'importance de pouvoir faire évoluer les méthodes de CST.

Après avoir établi l'état de l'art actuel des classifieurs non profonds pour la CST~\citep{bagnall2017the}, nous discutons du succès de l'apprentissage profond~\citep{lecun2015deep} dans diverses tâches de classification qui ont motivé l'utilisation récente de l'apprentissage profond pour la CST~\citep{wang2017time}.
Les CNNs profonds ont révolutionné le domaine de la vision par ordinateur~\citep{krizhevsky2012imagenet}.
Par exemple, en 2015, les CNNs ont été utilisés pour atteindre les performances au niveau humain dans les tâches de reconnaissance d'image~\citep{szegedy2015going}.
Suite au succès des DNNs en vision par ordinateur, de nombreuses recherches ont proposé plusieurs architectures DNN pour résoudre de nombreuses tâches de NLP telles que la traduction automatique~\citep{sutskever2014sequence, bahdanau2015neural}, la représentation vectorielle des mots~\citep{mikolov2013distributed, mikolov2013efficient} et la classification des documents~\citep{le2014distributed, goldberg2016a}.
Les DNNs ont également eu un impact énorme sur la communauté de reconnaissance vocale~\citep{hinton2012deep,sainath2013deep}.
Il est intéressant de noter que la similitude intrinsèque entre le NLP et les tâches de reconnaissance vocale est due à l'aspect séquentiel des données qui est également l'une des principales caractéristiques des séries temporelles.

Dans ce contexte, ce premier chapitre cible les questions ouvertes suivantes: \textit{Quel est le meilleur DNN actuel pour la CST}?
\textit{Existe-t-il une approche DNN moins complexe que HIVE-COTE pouvant obtenir des résultats compétitifs avec l'état de l'art} ?
\textit{Quel type d'architectures DNN fonctionne le mieux pour la tâche de CST}?
\textit{Comment l'initialisation aléatoire affecte-t-elle les performances des classifieurs d'apprentissage profond} ?
Et enfin: \textit{L'effet boîte noire des DNNs pourrait-il être évité pour fournir une interprétabilité des résultats} ?
Étant donné que ces dernières questions n'ont pas été abordées par la communauté de CST, il est surprenant de voir que les DNNs n'ont pas été considéré comme classifieur précis potentiel des séries temporelles~\citep{lines2018time}.
En fait, une étude empirique récente~\citep{bagnall2017the} a évalué 18 algorithmes de CST sur 85 jeux de données, dont aucun n'était un modèle d'apprentissage profond.
Cela montre à quel point la communauté n'a pas une vue d'ensemble des performances actuelles des modèles d'apprentissage profond pour résoudre le problème de CST~\citep{lines2018time}.

Dans ce chapitre, nous avons effectué une étude comparative empirique des approches d'apprentissage profond les plus récentes pour la CST.
Avec la montée en puissance des GPUs, nous avons montré comment les architectures profondes peuvent être entraînées efficacement pour apprendre de bout en bout des fonctionnalités discriminantes cachées, à partir de séries temporelles brutes.
De manière similaire à~\cite{bagnall2017the}, afin d'avoir une comparaison équitable entre les approches testées, nous avons développé une plateforme en Python, Keras~\citep{chollet2015keras} et Tensorflow~\citep{tensorflow2015whitepaper} pour entrainer les modèles d'apprentissage profond sur un cluster de plus de 60 GPUs.

En plus de l'évaluation des ensembles de données univariés, nous avons testé les approches sur 12 jeux de données MTS~\citep{baydogan2015mts}.
L'évaluation multivariée montre un autre avantage des modèles d'apprentissage en profondeur, qui est la capacité de gérer la malédiction de la dimensionnalité~\citep{bellman2010dynamic, keogh2017curse} en exploitant différents degrés de fluidité dans la fonction de composition~\citep{poggio2017why} ainsi que les calculs parallèles des GPUs~\citep{lu2015a}.

Quant à la comparaison des classifieurs sur plusieurs ensembles de données, nous avons suivi les recommandations de~\cite{demsar2006statistical} et utilisé le test de Friedman~\citep{friedman1940a} pour rejeter l'hypothèse nulle.
Une fois que nous avons établi qu'il existe une différence statistique dans les performances des classifieurs, nous avons suivi l'analyse post-hoc par paires recommandée par~\cite{benavoli2016should} où la comparaison de rang moyen est remplacée par un test de rang signé Wilcoxon~\citep{wilcoxon1945individual} avec la correction alpha de Holm~\citep{holm1979a, garcia2008an}.

Dans cette étude, nous avons entrainé environ 1 milliard de paramètres dans 97 jeux de données de séries temporelles univariées et multivariées.
Malgré le fait qu'un grand nombre de paramètres risquent de sur-apprendre~\citep{zhang2017understanding} les ensembles d'entrainement relativement petit dans les archives UCR/UEA, nos expériences ont montré que non seulement les DNNs sont capables de surpasser considérablement le NN-DTW, mais sont également capables d'obtenir des résultats qui ne sont \emph{pas significativement} différents de COTE et HIVE-COTE en utilisant une architecture de réseau résiduel profond~\citep{he2016deep,wang2017time}.
Enfin, nous avons analysé comment de mauvaises initialisations aléatoires peuvent avoir un effet significatif sur les performances d'un DNN.

En conclusion, avec les problèmes de fouille de données de plus en plus fréquents, tirant parti d'architectures plus approfondies qui peuvent apprendre automatiquement de bout en bout des données annotées, l'apprentissage en profondeur est une approche très attrayante.
Dans ce chapitre, nous avons montré le potentiel des réseaux de neurones profonds pour le problème de CST, néanmoins ces modèles complexes d'apprentissage automatique peuvent encore bénéficier de nombreuses techniques de régularisation, qui est l'objectif principal du chapitre suivant.

\section*{Chapitre 2: Regularisation des réseaux de neurones profonds}
\addcontentsline{toc}{section}{Chapitre 2: Regularisation des réseaux de neurones profonds}

Les modèles d'apprentissage profond ont généralement plus de paramètres à entrainer qu'il n'y a d'instances d'entrainement.
Néanmoins, dans le chapitre précédent, nous avons montré comment ces réseaux de neurones artificiels sont capables d'atteindre de bonnes capacités de généralisation par rapport aux algorithmes de CST traditionnels.
Pourtant, la plupart de ces modèles nécessitent une sorte de régularisation afin de minimiser l'erreur de généralisation (en d'autres termes minimiser la différence entre l'erreur d'entrainement et d'erreur de test).
Dans ce chapitre, nous présentons les quatre principales techniques pour régulariser les DNNs pour la CST.

\textbf{Apprentissage par transfert}: Nous avons récemment montré que les CNNs peuvent atteindre des performances similaires à celles de l'état de l'art.
Cependant, malgré les performances élevées de ces CNNs, les modèles d'apprentissage profond sont toujours sujets au sur-apprentissage.
Un exemple où ces réseaux de neurones ne parviennent pas à généraliser est lorsque l'ensemble des données d'apprentissage est très petit.
Nous attribuons cette énorme différence de précision au phénomène de sur-apprentissage, qui reste un domaine de recherche ouvert dans la communauté ~\citep{zhang2017understanding}.
Ce problème est connu pour être atténué à l'aide de plusieurs techniques de régularisation telles que l'apprentissage par transfert~\citep{yosinski2014transferable}, où un modèle entraîné sur une première tâche est ensuite affiné sur un ensemble de données cible.
L'apprentissage par transfert est actuellement utilisé dans presque tous les modèles d'apprentissage en profondeur lorsque l'ensemble de données cible ne contient pas suffisamment de données étiquetées~\citep{yosinski2014transferable}.
Malgré son récent succès en vision par ordinateur~\citep{csurka2017domain}, l'apprentissage par transfert a rarement été appliqué aux modèles d'apprentissage profond dédiés aux séries temporelles.
L'une des raisons de cette absence est probablement le manque d'un grand ensemble de données généralistes similaire à ImageNet~\citep{russakovsky2015imagenet} ou OpenImages\citep{openimages} mais pour les séries temporelles.
De plus, ce n'est que récemment que l'apprentissage en profondeur s'est avéré efficace pour la CST~\citep{cui2016multi} et il reste encore beaucoup à explorer dans la construction de réseaux de neurones profonds pour l'analyse de séries temporelles~\citep{gamboa2017deep}.
Comme le transfert de modèles d'apprentissage en profondeur, entre les différents jeux de données des archives UCR/UEA~\citep{ucrarchive}, n'a pas été étudié de manière approfondie, nous avons décidé de nous y atteler dans le but ultime de déterminer à l'avance quels types de jeux de données pourraient bénéficier du transfert de modèles CNNs et améliorer leur précision.

\textbf{Ensembling}: Une autre façon d'améliorer les classifieurs basés sur les réseaux de neurones est de construire un ensemble de modèles d'apprentissage profond.
Cette idée semble très intéressante pour les tâches de CST car l'état de l'art évolue vers des solutions d'ensemble~\citep{lines2018time,lines2015time,bagnall2017the,baydogan2013a}.
De plus, les ensembles de réseaux de neurones profonds semblent obtenir des résultats très prometteurs dans de nombreux domaines de l'apprentissage automatique supervisé tels que la détection des lésions cutanées~\citep{goyal2018deep}, la reconnaissance d'expression faciale~\citep{wen2017ensemble} et le remplissage automatique des seaux~\citep{dadhich2018predicting}.
Par conséquent, nous proposons de regrouper les modèles actuels d'apprentissage profond pour la CST développés dans le chapitre précédent, en construisant un modèle composé de 60 réseaux de neurones profonds différents: 6 architectures différentes~\citep{wang2017time,zheng2014time,zhao2017convolutional,serra2018towards} chacun avec 10 initialisations différentes des poids du modèle.
En évaluant sur les 85 jeux de données de l'archive UCR/UEA, nous démontrons une amélioration significative par rapport aux classifieurs individuels tout en atteignant des performances très similaires à HIVE-COTE: méthode ensembliste contenant 37 classifieurs différents et représentant actuellement l'état de l'art.
Enfin, en nous inspirant de nos résultats sur l'apprentissage par transfert~\citep{IsmailFawaz2018transfer}, nous remplaçons les réseaux initialisés de manière aléatoire par un ensemble construit à partir de modèles affinés à partir des 84 autres jeux de données de l'archive, et montrons une amélioration significative pour la CST.

\textbf{Augmentation de données}: Bien que les CNNs profonds récemment proposés aient atteint des hautes performances pour la CST sur l'archive UCR/UEA~\citep{wang2017time}, ils montrent toujours de faibles capacités de généralisation sur certains petits jeux de données tels que l'ensemble de données CinCECGTorso avec 40 instances d’entraînement.
Cela est surprenant car le NN-DTW fonctionne exceptionnellement bien sur ce jeu de données, ce qui montre la relative facilité de cette tâche de classification.
Ainsi, les similitudes entre séries temporelles dans ces petits jeux de données ne peuvent pas être capturées par les CNNs en raison du manque d'instances étiquetées, ce qui pousse l'algorithme d'optimisation du réseau à être bloqué dans des minimums locaux~\citep{zhang2017understanding}.
Ce phénomène, également connu sous le nom de sur-apprentissage dans la communauté du machine learning, peut être résolu en utilisant différentes techniques telles que la régularisation ou simplement la collecte de données étiquetées supplémentaires~\citep{zhang2017understanding} (qui, dans certains domaines, sont difficiles à obtenir).
Une autre technique bien connue est l'augmentation des données, où les données synthétiques sont générées à l'aide d'une méthode spécifique.
Par exemple, les images contenant des numéros de rue sur des maisons peuvent être légèrement pivotées sans changer leur numéro réel~\citep{krizhevsky2012imagenet}.
Pour les modèles d'apprentissage en profondeur, ces méthodes sont généralement proposées pour les données d'image et se généralisent mal aux séries temporelles~\citep{um2017data}.
Cela est probablement dû au fait que pour les images, une comparaison visuelle peut confirmer si la transformation (telle que la rotation) n'a pas modifié la classe de l'image, tandis que pour les séries temporelles, on ne peut pas facilement confirmer l'effet de telles transformations ad hoc sur la nature d'une série.
Nous proposons de tirer parti d'une technique d'augmentation de données basée sur DTW spécifiquement développée pour les séries temporelles, afin d'améliorer les performances d'un ResNet profond pour la CST.
Nos expériences préliminaires révèlent que l'augmentation des données peut améliorer considérablement la précision des CNNs sur certains jeux de données tout en ayant un impact négatif faible sur d'autres jeux de données.

\textbf{Attaques adversaires}: Comme nous l'avons déjà discuté, la CST est utilisée dans diverses tâches d'exploration de données du monde réel, allant des soins de santé et de la sécurité~\citep{tan2017indexing, tobiyama2016malware} à la sécurité alimentaire et surveillance de la consommation d'énergie ~\citep{owen2012powering}.
Avec des modèles d'apprentissage en profondeur révolutionnant de nombreux domaines de l'apprentissage automatique tels que la vision par ordinateur~\citep{krizhevsky2012imagenet} et le traitement du langage naturel~\citep{yang2018investigating, wang2018hierarchical}, nous avons montré dans le chapitre précédent que ces modèles ont commencé à être adopté pour les tâches de CST~\citep{IsmailFawaz2018deep}.
Après l'avènement du deep learning, les chercheurs ont commencé à étudier la vulnérabilité des réseaux profonds aux attaques adversaires~\citep{yuan2017adversarial}.
Dans le cadre de la reconnaissance d'images, une attaque adversaire consiste à modifier une image originale afin que les changements soient quasiment indétectables par un humain~\citep{yuan2017adversarial}.
L'image modifiée est appelée une image adversaire, qui sera mal classée par le réseau de neurone, tandis que l'image d'origine est correctement classée.
L'une des attaques les plus célèbres de la vie réelle consiste à modifier une image de panneau de signalisation afin qu'elle soit mal interprétée par un véhicule autonome~\citep{eykholt2018robust}.
Une autre application est l'altération du contenu illégal pour le rendre indétectable par les algorithmes de modération automatique~\citep{yuan2017adversarial}.
Bien que ces approches aient été intensivement étudiées dans le contexte de la reconnaissance d'image, elles n'ont pas été étudiées en profondeur pour la CST.
Cela est surprenant car les modèles d'apprentissage en profondeur deviennent de plus en plus populaires pour classer les séries temporelles~\citep{IsmailFawaz2018deep}.
En outre, des attaques adverses potentielles sont présentes dans de nombreuses applications où l'utilisation de données de séries temporelles est cruciale.
Par exemple, dans ce chapitre on montre la similitude entre une série temporelle originale et perturbée de spectrographe de grains de café. 
Alors qu'un réseau de neurones profond classe correctement la série d'origine en tant que grains Robusta, l'ajout de petites perturbations le classe comme Arabica.
Par conséquent, étant donné que les grains Arabica ont plus de valeur que les grains Robusta, cette attaque pourrait être utilisée pour tromper les tests de contrôle des aliments et, éventuellement, les consommateurs.
Nous présentons, transférons et adaptons les attaques qui se sont avérées efficaces sur les images, aux données de séries temporelles.
Nous présentons également une étude expérimentale utilisant les 85 jeux de données de l'archive UCR/UEA~\citep{ucrarchive} qui révèle que les réseaux de neurones sont sensibles aux attaques adversaires.
Nous mettons en évidence des cas d'utilisation concrets spécifiques pour souligner l'importance de ces attaques dans des situations réelles, à savoir la qualité et la sécurité des aliments, les capteurs des véhicules et la consommation d'électricité.
Nos résultats montrent que les réseaux profonds pour les données de séries temporelles sont vulnérables aux attaques adverses comme leurs homologues en vision par ordinateur.
Par conséquent, ce travail met en lumière la nécessité de se protéger contre de telles attaques, en particulier lorsque l'apprentissage profond est utilisé pour les applications sensibles de CST.
Nous montrons également que les séries temporelles adverses apprises à l'aide d'une architecture de réseau peuvent être transférées à différentes architectures.
Nous discutons ensuite de certains mécanismes pour empêcher ces attaques tout en renforçant la robustesse des modèles aux attaques adversaires.
Enfin, dans un esprit de régularisation des DNNs, nous montrons comment ces séries perturbées peuvent être exploitées afin d'améliorer la capacité de généralisation d'un modèle d'apprentissage en profondeur: une technique appelée entraînement adversaire~\citep{xie2019adversarial}.

\section*{Chapitre 3: InceptionTime: Recherche d'AlexNet pour la classification de séries temporelles}
\addcontentsline{toc}{section}{Chapitre 3: InceptionTime: Recherche d'AlexNet pour la classification des séries temporelles}

Les industries allant des soins de santé~\citep{forestier2018surgical, lee2018diagnosis,IsmailFawaz2019automatic} et de la sécurité sociale~\citep{yi2018an} à la reconnaissance de l'activité humaine~\citep{yuan2018muvan} et à la télédétection~\citep{pelletier2019temporal}, produisent toutes des séries temporelles d'une échelle jamais vue auparavant --- à la fois en termes de longueur et de quantité de séries.
Cette croissance signifie également une dépendance accrue à l'égard de la classification automatique de ces données séquentielles et, idéalement, des algorithmes capables de le faire à grande échelle.

Dans les chapitres précédents, nous avons montré en quoi ces problèmes de CST diffèrent considérablement de l'apprentissage supervisé traditionnel pour les données structurées, en ce que les algorithmes doivent être capables de gérer et d'exploiter les informations temporelles présentes dans le signal.
Il est facile d'établir des parallèles entre ce scénario et des problèmes de vision par ordinateur tels que la classification d'images et la localisation d'objets, où les algorithmes efficaces apprennent des informations spatiales contenues dans une image.
En termes simples, le problème de classification des séries temporelles est essentiellement la même classe de problèmes, juste avec une dimension de moins.
Pourtant, malgré cette similitude, les algorithmes d'état de l'art actuels des deux domaines partagent peu de ressemblance~\citep{IsmailFawaz2018deep}.

Le deep learning a une longue histoire (en termes d'apprentissage automatique) en vision par ordinateur~\citep{lecun1998efficient} mais sa popularité a explosé avec AlexNet~\citep{krizhevsky2012imagenet}, après quoi il a été incontestablement la classe d'algorithmes avec le plus de réussite~\citep{lecun2015deep}.
Inversement, l'apprentissage en profondeur n'a commencé à gagner en popularité que récemment parmi les chercheurs en exploration de données temporelles~\citep{IsmailFawaz2018deep}.
Ceci est souligné par le fait que ResNet, qui est actuellement considéré comme l'architecture de réseau neurone d'état de l'art pour la CST lorsqu'elle est évaluée sur l'archive UCR/UEA~\citep{ucrarchive}, a été initialement proposé simplement comme modèle de base pour la tâche sous-jacent~\citep{wang2017time}.
Compte tenu des similitudes dans les données, il est facile de suggérer qu'il y a beaucoup d'amélioration potentielle pour l'apprentissage profond dans la CST.
Dans le chapitre précédent, nous avons montré comment il est possible d'améliorer la précision d'une architecture d'apprentissage profond donnée, en utilisant diverses techniques de régularisation telles que l'ensembling, l'apprentissage par transfert, l'augmentation des données et l'apprentissage adversaire.
Cependant, nous pensons qu'il y a encore place à l'amélioration en termes d'architecture de réseau, qui peut être considérée comme une tâche orthogonale aux différentes méthodes de régularisation des DNNs.

Dans ce chapitre, nous franchissons une étape importante vers la recherche de l'équivalent d'AlexNet pour la CST en présentant \ourmethod{} --- un nouvel ensemble d'apprentissage profond pour la CST.
\ourmethod{} atteint la précision d'état de l'art lorsqu'il est évalué sur l'archive UCR/UEA (actuellement le plus grand référentiel publiquement disponible pour la CST~\citep{ucrarchive}) et passe mieux à l'échelle dû à son coût computationnel plus faible.

\ourmethod{} est un ensemble de cinq modèles d'apprentissage profond pour la CST, chacun est créé en cascadant plusieurs modules Inception~\citep{szegedy2015going}.
Chaque classifieur individuel (modèle) aura exactement la même architecture mais avec une initialisation des poids différente et aléatoire.
L'idée centrale d'un module Inception est d'appliquer simultanément plusieurs filtres à une série temporelle en entrée.
Le module comprend des filtres de longueurs variables qui, comme nous le montrerons, permettent au réseau d'extraire automatiquement les caractéristiques pertinentes des séries temporelles longues et courtes.
En fait, \ourmethod{} suit ici l'idée d'ensemble présentée dans le chapitre précédent.

Après avoir présenté \ourmethod{} et ses résultats, nous effectuons une analyse des hyperparamètres architecturaux des réseaux de neurones profonds --- profondeur, longueur du filtre, nombre de filtres --- et les caractéristiques du module Inception --- le bottleneck et les connexions résiduelles, afin de mieux comprendre pourquoi ce modèle connaît un tel succès.
En fait, nous construisons des réseaux avec des filtres plus grands que ceux utilisés pour les tâches de vision par ordinateur, profitant directement du fait que les séries temporelles présentent une dimension de moins que les images.

En conclusion, l'apprentissage profond pour la classification des séries temporelles est toujours en retard par rapport aux réseaux de neurones pour la reconnaissance d'images en termes d'études expérimentales et de conceptions architecturales.
Dans ce chapitre, nous comblons cette lacune en introduisant \ourmethod{}, inspiré par le récent succès des réseaux basés sur Inception pour diverses tâches de vision par ordinateur.
Nous avons regroupé ces réseaux pour produire de nouveaux résultats d'état de l'art pour la CST sur les 85 jeux de données du UCR/UEA archive.
Notre approche est hautement évolutive, deux ordres de grandeur plus rapide que les modèles d'état de l'art actuels tels que HIVE-COTE.
L'ampleur de cette accélération est intéressante dans un contexte Big Data ainsi que pour des séries temporelles plus longues avec un taux d'échantillonnage élevé.
Nous étudions en outre les effets sur la précision globale de divers hyperparamètres des architectures CNN.
Pour ceux-ci, nous allons bien au-delà des pratiques standard pour les données d'images et la conception de réseaux avec de longs filtres.
Nous les examinons en utilisant un ensemble de données simulées et encadrons notre étude en termes de définition du ``receptive field'' d'un CNN pour la CST.

\section*{Chapitre 4: Analyse de séries temporelles pour la formation chirurgicale}
\addcontentsline{toc}{section}{Chapitre 4: Analyse de séries temporelles pour la formation chirurgicale}

Au cours des cent dernières années, la méthodologie d'enseignement classique de ``voir un, faire un, enseigner un'' a dominé les systèmes d'enseignement chirurgical dans le monde.
Avec l'avènement de salle d'opération 2.0, l'enregistrement des vidéos, des données cinématiques et de nombreux autres types de données au cours d'une opération chirurgicale est devenu une tâche facile, permettant ainsi aux systèmes d'intelligence artificielle d'être déployés et utilisés dans la pratique chirurgicale et médicale.
Récemment, il a été démontré que les données des capteurs de mouvement (par exemple cinématique) ainsi que les vidéos chirurgicales fournissent une structure de coaching permettant aux stagiaires débutants d'apprendre des chirurgiens expérimentés en rejouant ces vidéos et/ou trajectoires cinématiques.
Dans ce chapitre, nous abordons deux problèmes présents dans le programme actuel de formation chirurgicale.

\textbf{Évaluation des compétences chirurgicales}: L'idée principale de la méthodologie d'enseignement du Dr William Halsted est que l'étudiant pourrait devenir un chirurgien expérimenté en observant et en participant à des chirurgies encadrées~\citep{polavarapu2013100}.
Ces techniques de formation, bien que largement utilisées, sont dépourvues d'une méthode objective d'évaluation des compétences chirurgicales~\citep{kassahun2016surgical}.
L'évaluation standard des compétences chirurgicales est actuellement basée sur des listes de contrôle remplies par un expert observant la tâche chirurgicale~\citep{ahmidi2017a}.
Afin de prédire le niveau de compétence d'un stagiaire sans utiliser le jugement d'un chirurgien expert, l'OSATS a été proposé et est actuellement adopté comme pratique clinique standard~\citep{niitsu2013using}.
Hélas, ce type de notation observationnelle souffre encore de plusieurs facteurs externes et subjectifs tels que la fiabilité inter-évaluateurs, le processus de développement et le biais de la liste de contrôle et de l'évaluateur~\citep{hatala2015constructing}.
D'autres études ont démontré une corrélation entre les compétences techniques d'un chirurgien et les résultats postopératoires~\citep{bridgewater2003surgeon}.

Cette dernière approche souffre du fait que les suites d'une intervention chirurgicale dépendent des attributs physiologiques du patient~\citep{kassahun2016surgical}.
De plus, l'obtention de ce type de données est très ardue, ce qui rend ces techniques d'évaluation des compétences difficiles à mettre en œuvre pour la formation chirurgicale.
Les progrès récents en robotique chirurgicale tels que le système chirurgical \emph{da Vinci}~\citep{davinci} ont permis l'enregistrement de données vidéo et cinématiques de diverses tâches chirurgicales.
Par conséquent, un substitut des listes de contrôle et des approches basées sur les résultats, est de générer, à partir de ces cinématiques, des GMFs tels que la vitesse de la tâche chirurgicale, l'achèvement du temps, la fluidité du mouvement, la courbure et d'autres caractéristiques holistiques~\citep{zia2017automated, kassahun2016surgical}.
Bien que la plupart de ces techniques soient efficaces, il n'est pas évident de voir comment elles pourraient être utilisées pour soutenir le stagiaire avec une rétroaction détaillée et constructive, afin d'aller au-delà d'une classification naïve dans un niveau de compétence (c.-à-d. expert, intermédiaire, etc.) .
Cela est problématique car le retour d'expérience sur la pratique médicale permet aux chirurgiens d'atteindre des niveaux de compétence plus élevés tout en améliorant leurs performances~\citep{islam2016affordable}.
Dernièrement, un domaine intitulé \emph{Surgical Data Science}~\citep{maier-hein2017surgical} est apparu motivé par l'accès croissant à une énorme quantité de données complexes qui concernent le personnel, le patient et les capteurs pour capturer la procédure et les données relatives au patient telles que les variables cinématiques et les images~\citep{gao2014jhu}.
Au lieu d'extraire des GMFs, les enquêtes récentes ont tendance à décomposer manuellement les tâches chirurgicales en segments plus fins appelés ``gestes'', avant d’entraîner le modèle, et enfin à estimer les performances des stagiaires en fonction de leur évaluation au cours de ces gestes individuels~\citep{lingling2012sparse}.
Même si ces méthodes ont obtenu des résultats prometteurs et précis en termes d'évaluation des compétences chirurgicales, elles nécessitent d'étiqueter une énorme quantité de gestes avant de former l'estimateur~\citep{lingling2012sparse}.
Nous avons souligné deux limites majeures dans les techniques actuelles existantes qui estiment le niveau de compétence des chirurgiens à partir de leurs variables cinématiques correspondantes: premièrement, l'absence d'un résultat interprétable de la prédiction de compétence qui peut être utilisé par les stagiaires pour atteindre des niveaux de compétence chirurgicale plus élevés; deuxièmement, l'exigence de limites de gestes prédéfinies par les annotateurs qui est sujette à la fiabilité inter-annotateurs et qui prend du temps à préparer~\citep{vedula2016analysis}.
Dans cette première partie du chapitre, nous concevons une nouvelle architecture basée sur les FCNs, dédiée à l'évaluation des compétences chirurgicales.
En utilisant des noyaux unidimensionnels sur les séries temporelles cinématiques, nous évitons d'avoir à extraire de gestes peu fiables et sensibles.
La structure hiérarchique originale de notre modèle nous permet de capturer des informations globales spécifiques au niveau de compétence chirurgicale, ainsi que de représenter les gestes dans des caractéristiques latentes de bas niveau.
De plus, pour fournir une rétroaction interprétable, au lieu d'utiliser une couche dense comme la plupart des architectures traditionnelles d'apprentissage profond, nous plaçons une couche GAP qui nous permet de profiter de la carte d'activation de classe, proposée à l'origine par~\cite {zhou2016learning}, pour localiser quelle partie de l'exercice a eu un impact sur la décision du modèle lors de l'évaluation du niveau de compétence d'un chirurgien.
En utilisant une configuration expérimentale standard sur le plus grand ensemble de données publiques pour l'analyse des données robotiques chirurgicales: le JHU-ISI Gesture and Skill Assessment Working Set~\citep{gao2014jhu}, nous montrons la précision de notre modèle FCN.
Notre principale contribution est de démontrer que l'apprentissage en profondeur peut être mis à profit pour comprendre les structures complexes et latentes lors de la classification des compétences chirurgicales et de la prévision du score OSATS d'une chirurgie, d'autant plus qu'il y a encore beaucoup à apprendre sur ce qui constitue exactement une compétence chirurgicale~\citep{kassahun2016surgical}.

\textbf{Alignement des vidéos chirurgicales}: Les éducateurs ont toujours cherché des moyens innovants d'améliorer le taux d'apprentissage des apprenants.
Alors que les cours classiques sont encore les plus utilisés, les ressources multimédias sont de plus en plus adoptées~\citep{smith2000digital}, en particulier dans les cours en ligne ouverts et massifs~\citep{means2009evaluation}.
Dans ce contexte, les vidéos ont été considérées comme particulièrement intéressantes car elles peuvent combiner les images, textes, graphiques, audios et animations.
Le domaine médical ne fait pas exception et l'utilisation de ressources vidéo est intensivement adoptée dans le programme médical~\citep{masic2008learning}, en particulier dans le contexte de la formation chirurgicale~\citep{kneebone2002innovative}.
L'avènement de la chirurgie robotique stimule également cette tendance, car les robots chirurgicaux, comme le Da Vinci~\citep{davinci}, enregistrent généralement des flux vidéo pendant l'intervention.
Par conséquent, une grande quantité de données vidéo a été enregistrée au cours des dix dernières années~\citep{rapp2016youtube}.
Cette nouvelle source de données représente une opportunité sans précédent pour les jeunes chirurgiens d'améliorer leurs connaissances et leurs compétences~\citep{gao2014jhu}.
En outre, la vidéo peut également être un outil pour les chirurgiens seniors pendant les périodes d'enseignement pour évaluer les compétences des stagiaires.
En fait, une étude récente de~\cite{mota2018video} a montré que les résidents passent plus de temps à regarder des vidéos que des spécialistes, soulignant la nécessité pour les jeunes chirurgiens de profiter pleinement de cet outil.
Dans~\cite{herrera2016the}, les auteurs ont montré que les scores obtenus pour la tâche nœuds ainsi que leurs temps de réalisation des exercices se sont considérablement améliorés pour les sujets qui ont regardé les vidéos de leur propre performance.

Cependant, lorsque les stagiaires sont prêts à évaluer leurs progrès au cours de plusieurs essais de la même tâche chirurgicale en revoyant simultanément leurs vidéos chirurgicales enregistrées, le fait que les vidéos soient désynchronisées rend la comparaison entre les différents essais très difficile, voire impossible.
Ce problème se rencontre dans de nombreuses études de cas réels, car les experts accomplissent en moyenne les tâches chirurgicales en moins de temps que les chirurgiens débutants~\citep{mcNatt2001}.
Ainsi, lorsque les stagiaires améliorent leurs compétences, leur fournir une rétroaction qui identifie la raison de l'amélioration des compétences chirurgicales devient problématique car les vidéos enregistrées présentent une durée différente et ne sont pas parfaitement alignées.
Bien que la synchronisation des vidéos ait été le centre d'intérêt de plusieurs sites de recherche en vision par ordinateur, les contributions se concentrent généralement sur un cas particulier où plusieurs vidéos enregistrées simultanément (avec des caractéristiques différentes telles que les angles de vue et les facteurs de zoom) sont traitées~\citep{wolf2002sequence,wedge2005trajectory,padua2010linear}.
Un autre type de synchronisation de multiples vidéos utilise des fonctionnalités conçues à la main (telles que des trajectoires de points d'intérêt) à partir des vidéos~\citep{wang2014videosnapping,evangelidis2011efficient}, ce qui rend l'approche très sensible à la qualité des caractéristiques extraites.
Ce type de techniques était très efficace car les vidéos brutes étaient la seule source d'information disponible, alors que dans notre cas, l'utilisation de systèmes chirurgicaux robotisés permet de capturer un type de données supplémentaire: les variables cinématiques telles que les coordonnées cartésiennes $ x, y, z $ des effecteurs terminaux du Da Vinci~\citep{gao2014jhu}.
Dans cette deuxième partie du chapitre, nous proposons de tirer parti de l'aspect séquentiel des données cinématiques enregistrées par le système chirurgical Da Vinci, afin de synchroniser leurs images vidéo correspondantes en alignant les données de séries temporelles.
Lors de l'alignement de deux séries temporelles, l'algorithme standard est DTW~\citep{sakoe1978dynamic} que nous avons en effet utilisé pour aligner deux vidéos.
Cependant, lors de l'alignement de plusieurs séquences, cette dernière technique ne se généralise pas de manière simple et réalisable par calcul~\citep{petitjean2014dynamic}.
Par conséquent, pour la synchronisation de multiples vidéos, nous proposons d'aligner leurs séries temporelles correspondantes avec une série temporelle moyenne, calculée en utilisant l'algorithme DBA.
Ce processus est appelé NLTS et a été initialement proposé pour trouver l'alignement multiple d'un ensemble de gestes chirurgicaux discrétisés~\citep{forestier2014non}, que nous étendons dans ce travail à des données cinématiques numériques continues.

En conclusion, dans ce chapitre, nous avons abordé deux problèmes différents liés aux compétences chirurgicales.
Tout d'abord, en concevant un FCN, nous avons pu obtenir des résultats d'état de l'art pour l'évaluation des compétences chirurgicales (classification et régression).
Ainsi, nous avons pu atténuer l'effet de boîte noire des DNNs, en utilisant la technique CAM afin de mettre en évidence ce qui permet d'identifier l'expérience du chirurgien.
Le deuxième problème lié aux compétences chirurgicales était dû au fait que les vidéos de formation chirurgicale n'étaient pas synchronisées, ce qui rend difficile pour les stagiaires de comprendre et de comparer les vidéos entre différents chirurgiens avec différents niveaux de compétence.
Nous avons abordé ce dernier problème en proposant l'utilisation de NLTS afin d'aligner et de synchroniser plusieurs vidéos simultanément.
Ces deux projets étaient orthogonaux dans le sens où ils pouvaient également se compléter: la synchronisation de la vidéo et des séries temporelles pourrait être une étape de prétraitement qui améliorerait les modèles d'évaluation des compétences chirurgicales.

\addchap{Introduction}

\label{introduction}

Time series data are omnipresent in many practical data science applications ranging from health care~\citep{gao2014jhu} and stock market predictions~\citep{anghinoni2018time} to social media analysis~\citep{xu2018mnrd} and human activity recognition~\citep{xi2018deep}.  
In fact, any type of numerical acquisition of data with some notion of ordering will generate time series, making this type of data very common among data mining problems~\citep{langkvist2014a}. 
Compared to traditional tabular data, each time series can be represented (under the tabular format) as a row with each attribute corresponding to one time stamp (numerical acquisition). 
However, analyzing time series data differs significantly from its tabular counterpart, as harnessing the temporal information is usually very important for the underlying task we are trying to solve~\citep{bagnall2017the}.
A concrete example of time series data in health care would be the acquisition of ECG heart signals~\citep{rajan2018a}.
In stock market analysis, a time series element would correspond to the value of a stock at a given time stamp~\citep{anghinoni2018time}. 
In human activity recognition, the given Cartesian position of a hand in 3D space would constitute an element of a time series~\citep{ignatov2018activity}. 

Since 2006, time series analysis has been considered one of the most challenging problems in data mining~\citep{yang200610}, and in a more recent poll it has been shown that 48\% of data expert had analyzed time series data during their career, ahead of text and images~\citep{neamtu2018generalized}. 
Under the time series analysis umbrella, there exists many orthogonal tasks, that can be grouped into four main categories:

\textbf{Forecasting} consists of training a model using some historical time series data, with the goal to predict the future observations of this time series~\citep{hyndman2018forecasting}. 
Weather forecasting is one of the most common applications, where the training data consists of old observations, and the task is to predict the future weather behavior~\citep{taylor2009wind}.
In finance, predicting the value of stock using its historical values is one very common application~\citep{kim2003financial}. 
Further examples and details of time series forecasting an be found in~\cite{hyndman2018forecasting}.

\textbf{Anomaly detection} is a very special task of time series analysis. 
Unlike forecasting, the goal is not to predict the future, but rather to determine if a given time series observation is normal or not~\citep{blazquez2020review}. 
This task is also known as time series outlier detection. 
One of the most common application is called predictive maintenance, such as predicting anomalies in advance in order to prevent potential failures~\citep{rabatel2011anomaly}.
The reader is referred to this excellent review by~\cite{blazquez2020review} on time series anomaly/outlier detection. 

\textbf{Clustering} is probably one of the most widely studied problems in unsupervised learning. 
The problem can be defined as follows: given a set of data points, partition them into a set of groups which are as similar as possible~\citep{aggarwal2014data}. 
For time series data, traditional clustering approaches can provide a baseline, however many researchers found that mining the temporal information provided by the time series data can be crucial for many tasks~\citep{petitjean2012summarizing}. 
Applications range from discovering daily patterns of sales in marketing databases~\citep{sanwlani2013forecasting} to finding particular behaviors of solar magnetic wind in scientific databases~\citep{pravilovic2014wind}. 
For more details on time series clustering, we refer the interested reader to this recent survey by~\cite{aghabozorgi2015time}.

\begin{figure}
	\centering
	\includegraphics[width=0.8\linewidth]{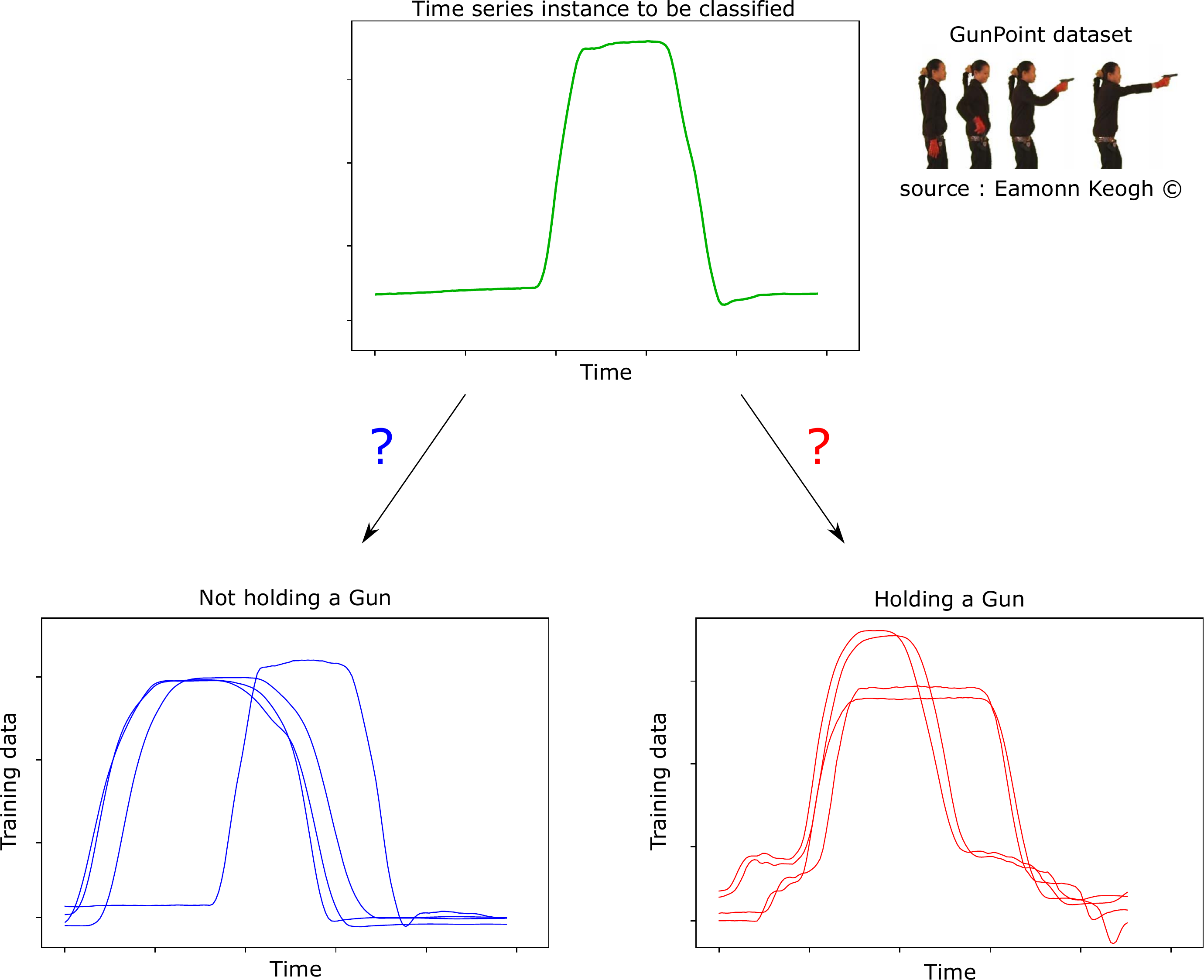}
	\caption{An example illustrating the task of classifying an input time series from the GunPointAgeSpan dataset~\citep{ucrarchive}, where the goal is to classify whether or not a person is holding a gun (a time series corresponds to the hand's x coordinate of a person holding or not a gun)}
	\label{fig-intro-explClassif}
\end{figure}

\textbf{Classification} consists of predicting the correct class of a given data point using a labeled training set. 
For TSC, this data point is by itself a whole time series, and the task consists of predicting its correct label. 
This problem is encountered in various data mining fields such as patient risk identification in health care~\citep{ma2018health}, malware detection in cyber security~\citep{tobiyama2016malware}, food safety evaluation in agriculture and livestock~\citep{nawrocka2013determination}. 
Similar to unsupervised learning, traditional classification approaches can provide a basic baseline for solving this underlying TSC task. 
However over the past two decades, research has shown that designing algorithms that can exploit the temporal information is a need in order to achieve high classification accuracy~\citep{bagnall2017the}. 
Figure~\ref{fig-intro-explClassif} illustrates an example of the TSC task. 

In this thesis, we chose to focus on the latter TSC problem. 
The reason behind choosing this field of time series analysis was motivated by our interest in a very particular problem: surgical skills evaluation from kinematic data. 
This specific task can be cast as a TSC problem: given an input time series (kinematic data registered through time) predict the correct label (skill level of the surgeon performing the surgery).
The reader can find more details on this problem in Chapter~\ref{Chapter4}. 
For solving this surgical skills evaluation problem, we started looking into the state-of-the-art TSC algorithms of that epoch. 
At the beginning of this thesis in 2017, we found that most TSC approaches were inspired by traditional machine learning algorithms such as NN classifiers coupled with a bespoke distance - such as DTW instead of ED. 
Some algorithms were inspired by traditional text mining approaches such as the Time Series Bag-Of-Features developed by~\cite{baydogan2013a}. 
Other TSC methods were inspired by Fourier analysis from signal processing such as BOSS (proposed by~\cite{schafer2015the}). 
One type of classifier focused on extracting discriminative subsequence from the time series called Shapelets, which are later used for classifying the input time series~\citep{hills2014classification}. 
Meanwhile in 2017, the computer vision community has already achieved tremendous human level performance with DNNs, followed by the much more recent success of deep learning for various NLP tasks.
However, we have noticed that the TSC community have not considered DNNs as potential classifiers of time series data, which is evident in the great TSC bake-off paper~\citep{bagnall2017the}. 
This is very surprising as given the success of deep learning with image classification problems, coupled with the intrinsic similarity between 2D patterns in images and 1D patterns in series, one should consider the potential of deep learning for TSC problems.
We therefore started investigating and benchmarking the recent work proposing the use of DNNs for classifying time series data, which is the main focus of Chapter~\ref{Chapter1}~\citep{IsmailFawaz2018deep}.

Following this thorough review of the recent advances in deep learning architectures for TSC, we started looking into the different techniques to improve the accuracy of a given neural network model. 
This type of technique is also known as regularization, which enables us to improve the generalization capabilities of a given machine learning model. 
These methods range from transfer learning~\citep{IsmailFawaz2018transfer} and ensembling~\citep{IsmailFawaz2019deep} to data augmentation~\citep{IsmailFawaz2018data} and adversarial training~\citep{IsmailFawaz2019adversarial}. 
The latter techniques were the main focus of Chapter~\ref{Chapter2} of this thesis. 

Having identified the current state-of-the-art architectures for TSC in Chapter~\ref{Chapter1}, followed by the main techniques on how to improve the generalization capability of a given deep learning model in Chapter~\ref{Chapter2}, we took a further step into designing a new type of neural network architecture for TSC based on the famous Inception module proposed by~\cite{szegedy2015going}. 
Although we achieved similar results to other non deep learning based approaches for TSC, the main focus of Chapter~\ref{Chapter3} was to motivate the use of the Inception module, with a focus on its running time, a severe bottleneck of the current state-of-the-art algorithm for TSC~\citep{bagnall2017the}. 

Finally, with the recent advances in deep learning for TSC being presented in the first three chapters, we turned our attention in Chapter~\ref{Chapter4} to our initial motivation: evaluating surgical skills from kinematic data using DNNs. 
We also focused on achieving state-of-the-art results while providing interpretability of our deep learning model, allowing us to leverage the high accuracy from DNNs while mitigating their black-box effect. 

\begin{table}
	{
		\centering
		\scriptsize
		\begin{tabularx}{\textwidth}{c|c|c|c|c|c}
			Paper & Method(s) & Data \tiny{(in paper)} & Data \tiny{(on GitHub)} & Chap. & GitHub \\ 
			\midrule
			\citep{IsmailFawaz2018deep} & 9 methods & \href{http://timeseriesclassification.com}{UCR/UEA (85)} & \href{https://www.cs.ucr.edu/~eamonn/time_series_data_2018/}{UCR (128)} & \ref{Chapter1} & \href{https://github.com/hfawaz/dl-4-tsc}{dl-4-tsc} \\ 
			\midrule 
			\citep{IsmailFawaz2018transfer} & FCN & \href{https://www.cs.ucr.edu/~eamonn/time_series_data/}{UCR (85)} & \href{https://www.cs.ucr.edu/~eamonn/time_series_data/}{UCR (85)} & \ref{Chapter2} & \href{https://github.com/hfawaz/bigdata18}{bigdata18} \\ 
			\midrule 
			\citep{IsmailFawaz2019deep} & 6 methods & \href{http://timeseriesclassification.com}{UCR/UEA (85)} & \href{http://timeseriesclassification.com}{UCR/UEA (85)} & \ref{Chapter2} & \href{https://github.com/hfawaz/ijcnn19ensemble}{ijcnn19ensemble} \\ 
			\midrule 
			\citep{IsmailFawaz2018data} & ResNet & \href{https://www.cs.ucr.edu/~eamonn/time_series_data/}{UCR (85)} & \href{https://www.cs.ucr.edu/~eamonn/time_series_data/}{UCR (85)} & \ref{Chapter2} & \href{https://github.com/hfawaz/aaltd18}{aaltd18} \\ 
			\midrule 
			\citep{IsmailFawaz2019adversarial} & ResNet & \href{https://www.cs.ucr.edu/~eamonn/time_series_data/}{UCR (85)} & \href{https://www.cs.ucr.edu/~eamonn/time_series_data/}{UCR (85)} & \ref{Chapter2} & \href{https://github.com/hfawaz/ijcnn19attacks}{ijcnn19attacks} \\ 
			\midrule 
			\citep{IsmailFawaz2020inceptionTime} & InceptionTime & \href{http://timeseriesclassification.com}{UCR/UEA (85)} & \href{https://www.cs.ucr.edu/~eamonn/time_series_data_2018/}{UCR (128)} & \ref{Chapter3} & \href{https://github.com/hfawaz/InceptionTime}{InceptionTime} \\ 
			\midrule 
			\citep{IsmailFawaz2018evaluating} & FCN & \href{https://cirl.lcsr.jhu.edu/research/hmm/datasets/jigsaws_release/}{JIGSAWS} & \href{https://cirl.lcsr.jhu.edu/research/hmm/datasets/jigsaws_release/}{JIGSAWS} & \ref{Chapter4} & \href{https://github.com/hfawaz/miccai18}{miccai18} \\ 
			\midrule 
			\citep{ismailfawaz2019accurate} & FCN & \href{https://cirl.lcsr.jhu.edu/research/hmm/datasets/jigsaws_release/}{JIGSAWS} & \href{https://cirl.lcsr.jhu.edu/research/hmm/datasets/jigsaws_release/}{JIGSAWS} & \ref{Chapter4} & \href{https://github.com/hfawaz/ijcars19}{ijcars19} \\ 
			\midrule 
			\citep{IsmailFawaz2019automatic} & NLTS & \href{https://cirl.lcsr.jhu.edu/research/hmm/datasets/jigsaws_release/}{JIGSAWS} & \href{https://cirl.lcsr.jhu.edu/research/hmm/datasets/jigsaws_release/}{JIGSAWS} & \ref{Chapter4} & \href{https://github.com/hfawaz/aime19}{aime19} \\ 
			\bottomrule
			
		\end{tabularx}
		\caption{List of papers with the corresponding public datasets used as well as the companion GitHub repository.}\label{tab-papers}
		
	}
\end{table}

In order to have a thorough and fair experimental evaluation of all approaches, we used the whole UCR/UEA archive~\citep{ucrarchive} which contained 85 univariate time series datasets at that time.
Other than the fact of being publicly available, the choice of validating on the UCR/UEA archive is motivated by having datasets from different domains which have been broken down into seven different categories (Image Outline, Sensor Readings, Motion Capture, Spectrographs, ECG, Electric Devices and Simulated Data) in~\cite{bagnall2017the}. 
In fact, the UCR/UEA archive has evolved over the years. 
In 2015 the benchmark contained 44 datasets, then in 2017 UCR and UEA collaborated and provided the community with an even larger archive containing 85 datasets, which represents the version that we have used in this work. 
In 2018,~\cite{ucrarchive} published the most recent version of the archive containing 128 datasets. 
We did our best to be consistent in the datasets used for our experiments to make our results easily comparable to other state-of-the-art methods. 
The evolution of the UCR during this thesis can bring some sort of confusion in the datasets used in our different papers. 
Therefore \tablename~\ref{tab-papers} has the objective to clarify what has been used when our papers were published and what is now available on the companion GitHub pages. 
In any case, the source code for each paper is available to the community and can be easily reused to follow further evolution of the archive or other challenges.
We are aware of the limitations of using the UCR/UEA archive as a sole reference to compare methods and we share the thoughts about this issue discussed by the authors in~\cite{ucrarchive}. 
However, we believe that having a way to objectively compare the methods between them is very important and we are thankful to all the people involved in making the UCR/UEA archive publicly available.
As for the experiments regarding the surgical evaluation skills, we have used the publicly available JIGSAWS dataset published in~\cite{gao2014jhu}.


\chapter{The state of the art for time series classification} \label{Chapter1}



\newcommand{\keyword}[1]{\textbf{#1}}
\newcommand{\tabhead}[1]{\textbf{#1}}
\newcommand{\code}[1]{\texttt{#1}}
\newcommand{\file}[1]{\texttt{\bfseries#1}}
\newcommand{\option}[1]{\texttt{\itshape#1}}



\section{Introduction}

During the last two decades, TSC has been considered as one of the most challenging problems in data mining~\citep{yang200610,esling2012time}. 
With the increase of temporal data availability~\citep{silva2018speeding}, hundreds of TSC algorithms have been proposed since 2015~\citep{bagnall2017the}.
Due to their natural temporal ordering, time series data are present in almost every task that requires some sort of human cognitive process~\citep{langkvist2014a}.  
In fact, any classification problem, using data that is registered taking into account some notion of ordering, can be  cast as a TSC problem~\citep{gamboa2017deep}. 
Time series are encountered in many real-world applications ranging from health care~\citep{gogolou2018comparing} and human activity recognition~\citep{wang2018deepsurvey,mathis2020knowledge} to acoustic scene classification~\citep{nwe2017convolutional} and cyber security~\citep{susto2018time}. 
In addition, the diversity of the datasets' types in the UCR/UEA archive~\citep{ucrarchive,bagnall2017the} (the largest repository of time series datasets) shows the different applications of the TSC problem. 

Given the need to accurately classify time series data, researchers have proposed hundreds of methods to solve this task~\citep{bagnall2017the}.
One of the most popular and traditional TSC approaches is the use of an NN classifier coupled with a distance function~\citep{lines2015time}. 
Particularly, DTW when used with an NN classifier has been shown to be a very strong baseline~\citep{bagnall2017the}. 
\cite{lines2015time} compared several distance measures - such as TWE~\citep{marteau2009time} and MSM~\citep{stefan2013the} - showing that there is no single distance measure that significantly outperforms DTW.
They also showed that ensembling the individual NN classifiers (with different distance measures) outperforms all of the ensemble's individual components.   
Hence, recent contributions have focused on developing ensembling methods that significantly outperforms the NN-DTW~\citep{bagnall2016time,hills2014classification,bostrom2015binary,lines2016hive,schafer2015the,kate2016using,deng2013a,baydogan2013a}. 
These approaches use either an ensemble of decision trees (random forest)~\citep{baydogan2013a,deng2013a} or an ensemble of different types of discriminant classifiers (SVM, NN with several distances) on one or several feature spaces~\citep{bagnall2016time,bostrom2015binary,schafer2015the,kate2016using}. 
Most of these approaches significantly outperform the NN-DTW~\citep{bagnall2017the} and share one common property, which is the data transformation phase where time series are transformed into a new feature space (for example using shapelets transform~\citep{bostrom2015binary} or DTW features~\citep{kate2016using}). 
This notion motivated the development of an ensemble of 35 classifiers named COTE~\citep{bagnall2016time} that does not only ensemble different classifiers over the same transformation, but instead ensembles different classifiers over different time series representations. 
\cite{lines2016hive,lines2018time} extended COTE with HIVE-COTE which has been shown to achieve a significant improvement over COTE by leveraging a new hierarchical structure with probabilistic voting, including two new classifiers and two additional representation transformation domains.
HIVE-COTE is currently considered the state-of-the-art algorithm for time series classification~\citep{bagnall2017the} when evaluated over the 85 datasets from the UCR/UEA archive. 

To achieve its high accuracy, HIVE-COTE becomes hugely computationally intensive and impractical to run on a real big data mining problem~\citep{bagnall2017the}.
The approach requires training 37 classifiers as well as cross-validating each hyperparameter of these algorithms, which makes the approach infeasible to train in some situations~\citep{lucas2018proximity}. 
To emphasize on this infeasibility, note that one of these 37 classifiers is the ST~\citep{hills2014classification} whose time complexity is $O(n^2\cdot l^4)$ with $n$ being the number of time series in the dataset and $l$ being the length of a time series.
Adding to the training time's complexity is the high \emph{classification} time of one of the 37 classifiers: the nearest neighbor which needs to scan the training set before taking a decision at test time.
Therefore since the nearest neighbor constitutes an essential component of HIVE-COTE, its deployment in a real-time setting is still limited if not impractical. 
Finally, adding to the huge runtime of HIVE-COTE, the decision taken by 37 classifiers cannot be interpreted easily by domain experts, since researchers already struggle with understanding the decisions taken by an individual classifier.
We should note that recently,~\cite{bagnall2020usage} proposed a new version of HIVE-COTE that is substantially faster, showing the importance of having a scalable TSC algorithm. 

After having established the current state-of-the-art of non deep classifiers for TSC~\citep{bagnall2017the}, we discuss the success of deep learning~\citep{lecun2015deep} in various classification tasks which motivated the recent utilization of deep learning models for TSC~\citep{wang2017time}. 
Deep CNNs have revolutionized the field of computer vision~\citep{krizhevsky2012imagenet}. 
For example, in 2015, CNNs were used to reach human level performance in image recognition tasks~\citep{szegedy2015going}. 
Following the success of DNNs in computer vision, a huge amount of research proposed several DNN architectures to solve many NLP tasks such as machine translation~\citep{sutskever2014sequence,bahdanau2015neural}, learning word embeddings~\citep{mikolov2013distributed,mikolov2013efficient} and document classification~\citep{le2014distributed,goldberg2016a}. 
DNNs also had a huge impact on the speech recognition community~\citep{hinton2012deep,sainath2013deep}.
Interestingly, we should note that the intrinsic similarity between the NLP and speech recognition tasks is due to the sequential aspect of the data which is also one of the main characteristics of time series data.  

In this context, this chapter targets the following open questions: \textit{What is the current state-of-the-art DNN for TSC}?
\textit{Is there a current DNN approach that reaches state-of-the-art performance for TSC and is less complex than HIVE-COTE}? 
\textit{What type of DNN architectures works best for the TSC task}? 
\textit{How does the random initialization affect the performance of deep learning classifiers}?
And finally: \textit{Could the black-box effect of DNNs be avoided to provide interpretability}? 
Given that the latter questions have not been addressed by the TSC community, it is surprising how a small amount of papers have considered DNNs to be a potential accurate classifier of time series data~\citep{lines2018time}. 
In fact, a recent empirical study~\citep{bagnall2017the} evaluated 18 TSC algorithms on 85 time series datasets, none of which was a deep learning model. 
This shows how much the community lacks of an overview of the current performance of deep learning models for solving the TSC problem~\citep{lines2018time}. 

In this chapter, we performed an empirical comparative study of the most recent deep learning approaches for TSC. 
With the rise of GPUs, we show how deep architectures can be trained efficiently to learn hidden discriminative features from raw time series in an end-to-end manner.
Similarly to~\cite{bagnall2017the}, in order to have a fair comparison between the tested approaches, we developed a common framework in Python, Keras~\citep{chollet2015keras} and Tensorflow~\citep{tensorflow2015whitepaper} to train the deep learning models on a cluster of more than 60 GPUs.

In addition to the univariate datasets' evaluation, we tested the approaches on 12 MTS datasets~\citep{baydogan2015mts}.
The multivariate evaluation shows another benefit of deep learning models, which is the ability to handle the curse of dimensionality~\citep{bellman2010dynamic,keogh2017curse} by leveraging different degrees of smoothness in compositional function~\citep{poggio2017why} as well as the parallel computations of the GPUs~\citep{lu2015a}. 

As for comparing the classifiers over multiple datasets, we followed the recommendations in~\cite{demsar2006statistical} and used the Friedman test~\citep{friedman1940a} to reject the null hypothesis. 
Once we have established that a statistical difference exists within the classifiers' performance, we followed the pairwise post-hoc analysis recommended by~\cite{benavoli2016should} where the average rank comparison is replaced by a Wilcoxon signed-rank test~\citep{wilcoxon1945individual} with Holm's alpha correction~\citep{holm1979a,garcia2008an}. 

In this study, we have trained about 1 billion parameters across 97 univariate and multivariate time series datasets. 
Despite the fact that a huge number of parameters risks overfitting~\citep{zhang2017understanding} the relatively small train set in the UCR/UEA archive, our experiments showed that not only DNNs are able to significantly outperform the NN-DTW, but are also able to achieve results that are \emph{not significantly} different than COTE and HIVE-COTE using a deep residual network architecture~\citep{he2016deep,wang2017time}.
Finally, we analyze how poor random initializations can have a significant effect on a DNN's performance.

The rest of the chapter is structured as follows. 
In Section~\ref{ch-1-sec-background}, we provide some background materials concerning the main types of architectures that have been proposed for TSC.
In Section~\ref{ch-1-sec-approaches}, the tested architectures are individually presented in details.
We describe our experimental open source framework in Section~\ref{ch-1-sec-experiment}.    
The corresponding results and the discussions are presented in Section~\ref{ch-1-sec-results}. 
In Section~\ref{ch-1-sec-visualization}, we describe in detail a couple of methods that mitigate the black-box effect of the deep learning models.  
Finally, we present a conclusion in Section~\ref{ch-1-sec-conclusion} to summarize our findings and discuss future directions. 

\smallskip

\noindent The main contributions presented in this chapter can be summarized as follows: 
\begin{itemize}
	\item We explain with practical examples, how deep learning can be adapted to one dimensional time series data. 
	\item We propose a unified taxonomy that regroups the recent applications of DNNs for TSC in various domains under two main categories: generative and discriminative models. 
	\item We detail the architecture of nine end-to-end deep learning models designed specifically for TSC. 
	\item We evaluate these models on the univariate UCR/UEA archive benchmark and 12 MTS classification datasets.
	\item We provide the community with an open source deep learning framework for TSC in which we have implemented all nine approaches.
	\item We investigate the use of CAM in order to reduce DNNs' black-box effect and explain the different decisions taken by various models.  
\end{itemize} 

\section{Background} \label{ch-1-sec-background}

In this section, we start by introducing the necessary definitions for ease of understanding. 
We then follow by an extensive theoretical background on training DNNs for the TSC task. 
Finally we present our proposed taxonomy of the different DNNs with examples of their application in various real world data mining problems. 

\subsection{Time series classification}
Before introducing the different types of neural networks architectures, we go through some formal definitions for TSC. 
\begin{definition}
	A univariate time series $X=[x_1,x_2, \dots, x_T]$ is an ordered set of real values. 
	The length of $X$ is equal to the number of real values $T$. 
\end{definition}

\begin{definition}
	An $M$-dimensional MTS, $X=[X^1,X^2, \dots, X^M]$ consists of $M$ different univariate time series with $X^i$ $\in \mathbb{R}^T$. 
\end{definition}

\begin{definition}
	A dataset $D=\{(X_1,Y_1), (X_2,Y_2), \dots, (X_N,Y_N)\}$ is a collection of pairs $(X_i,Y_i)$ where $X_i$ could either be a univariate or multivariate time series with $Y_i$ as its corresponding one-hot label vector.
	For a dataset containing $K$ classes, the one-hot label vector $Y_i$ is a vector of length $K$ where each element $j\in[1,K]$ is equal to $1$ if the class of $X_i$ is $j$ and $0$ otherwise.
\end{definition}

The task of TSC consists of training a classifier on a dataset $D$ in order to map from the space of possible inputs to a probability distribution over the class variable values (labels).

\subsection{Deep learning approaches for time series classification}

\begin{figure}
	\centering
	\includegraphics[width=0.9\linewidth]{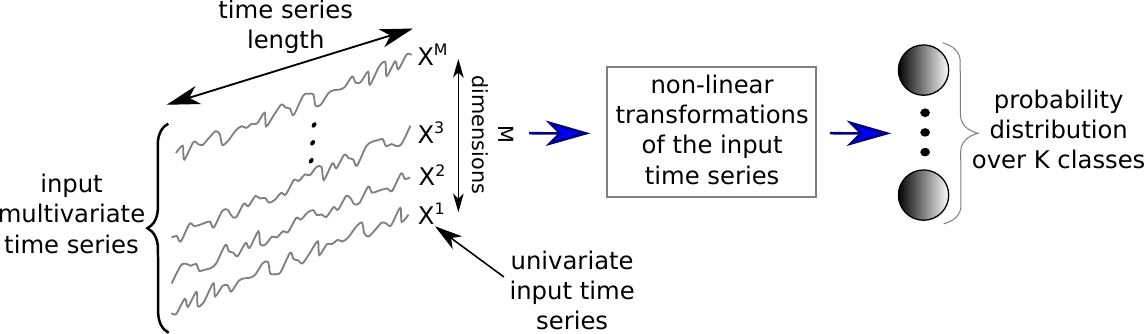}
	\caption{A unified deep learning framework for time series classification.}
	\label{fig-unified-framework}
\end{figure}

In this chapter, we focus on reviewing various approaches tackling the TSC task~\citep{bagnall2017the} using DNNs, which are considered complex machine learning models~\citep{lecun2015deep}.
A general deep learning framework for TSC is depicted in Figure~\ref{fig-unified-framework}. 
These networks are designed to learn hierarchical representations of the data. 
A deep neural network is a composition of $L$ parametric functions referred to as layers 
where each layer is considered a representation of the input domain~\citep{papernot2018deep}.
One layer $l_i$, such as $i \in 1 \dots L$, contains neurons, which are small units that compute one element of the layer's output. 
The layer $l_i$ takes as input the output of its previous layer $l_{i-1}$ and applies a non-linearity (such as the sigmoid function) to compute its own output. 
The behavior of these non-linear transformations is controlled by a set of parameters $\theta_i$ for each layer. 
In the context of DNNs, these parameters are called weights which link the input of the previous layer to the output of the current layer.
Hence, given an input $x$, a neural network performs the following computations to predict the class:
\begin{equation}\label{eq-comput-graph}
f_L(\theta_L,x) = f_{L-1}(\theta_{L-1},f_{L-2}(\theta_{L-2}, \dots ,f_1(\theta_1,x)))
\end{equation}
where $f_i$ corresponds to the non-linearity applied at layer $l_i$. 
For simplicity, we will omit the vector of parameters $\theta$ and use $f(x)$ instead of $f(\theta,x)$. 
This process is also referred to as \emph{feed-forward} propagation in the deep learning literature. 

During training, the network is presented with a certain number of known input-output (for example a dataset $D$).
First, the weights are initialized randomly~\citep{lecun1998efficient}, although a robust alternative would be to take a pre-trained model on a source dataset and fine-tune it on the target dataset~\citep{pan2010a}. 
This process is known as transfer learning which we do not study empirically, rather we discuss the transferability of each model with respect to the architecture in Section~\ref{ch-1-sec-approaches}. 
After the weight's initialization, a forward pass through the model is applied: using the function $f$ the output of an input $x$ is computed. 
The output is a vector whose components are the estimated probabilities of $x$ belonging to each class. 
The model's prediction loss is computed using a cost function, for example the negative log likelihood. 
Then, using gradient descent~\citep{lecun1998efficient}, the weights are updated in a backward pass to propagate the error. 
Thus, by iteratively taking a forward pass followed by backpropagation, the model's parameters are updated in a way that minimizes the loss on the training data. 

During testing, the probabilistic classifier (the model) is tested on unseen data which is also referred to as the inference phase: a forward pass on this unseen input followed by a class prediction.
The prediction corresponds to the class whose probability is maximum. 
To measure the performance of the model on the test data (generalization), we adopted the accuracy measure (similarly to~\cite{bagnall2017the}).
One advantage of DNNs over non-probabilistic classifiers (such as NN-DTW) is that a probabilistic decision is taken by the network~\citep{large2017the}, thus allowing to measure the confidence of a certain prediction given by an algorithm.

Although there exist many types of DNNs, in this review we focus on three main DNN architectures used for the TSC task: MLP, CNN and ESN.  
These three types of architectures were chosen since they are widely adopted for end-to-end deep learning~\citep{lecun2015deep} models for TSC. 

\subsubsection{Multi Layer Perceptrons}

An MLP constitutes the simplest and most traditional architecture for deep learning models. 
This form of architecture is also known as an FC network since the neurons in layer $l_i$ are  connected to every neuron in layer $l_{i-1}$ with $i\in [1,L]$.
These connections are modeled by the weights in a neural network.
A general form of applying a non-linearity to an input time series $X$ can be seen in the following equation: 
\begin{equation}\label{eq-activation}
A_{l_i}=f(\omega_{l_i}*X+b)
\end{equation}
with $\omega_{l_i}$ being the set of weights with length and number of dimensions identical to $X$'s, $b$ the bias term and $A_{l_i}$ the activation of the neurons in layer $l_i$. 
Note that the number of neurons in a layer is considered a hyperparameter.

\begin{figure}
	\centering
	\includegraphics[width=0.9\linewidth]{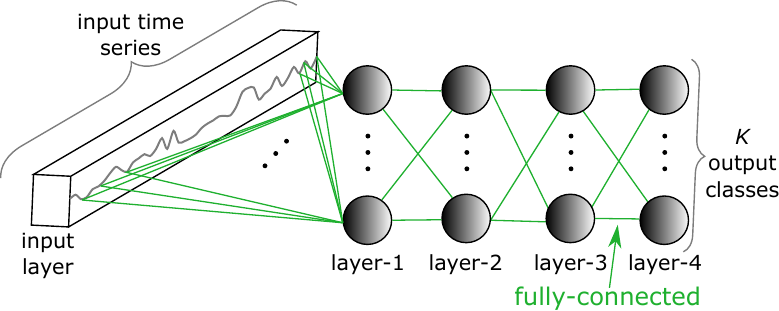}
	\caption{Multilayer perceptron for time series classification.}
	\label{fig-mlp-archi}
\end{figure}

One impediment from adopting MLPs for time series data is that they do not exhibit any spatial invariance, which can be seen in Figure~\ref{fig-mlp-archi}. 
In other words, each time stamp has its own weight and the temporal information is lost: meaning time series elements are treated independently from each other.
For example the set of weights $w_d$ of neuron $d$ contains $T\times M$ values denoting the weight of each time stamp $t$ for each dimension of the $M$-dimensional input MTS of length $T$. 
Then by cascading the layers we obtain a computation graph similar to equation~\ref{eq-comput-graph}. 

For TSC, the final layer is usually a discriminative layer that takes as input the activation of the previous layer and gives a probability distribution over the class variables in the dataset.
Most deep learning approaches for TSC employ a softmax layer which corresponds to an FC layer with softmax as activation function $f$ and a number of neurons equal to the number of classes in the dataset. 
Three main useful properties motivate the use of the softmax activation function: the sum of probabilities is guaranteed to be equal to $1$, the function is differentiable and it is an adaptation of logistic regression to the multinomial case. 
The result of a softmax function can be defined as follows: 
\begin{equation}\label{eq-softmax}
\hat{Y}_j(X) = \frac{e^{A_{L-1}*\omega_j+b_j}}{\sum_{k=1}^{K}{e^{A_{L-1}*\omega_k+b_k}}}
\end{equation}
with $\hat{Y}_j$ denoting the probability of $X$ having the class $Y$ equal to class $j$ out of $K$ classes in the dataset.
The set of weights $w_j$ (and the corresponding bias $b_j$) for each class $j$ are linked to each previous activation in layer $l_{L-1}$. 

The weights in equations~(\ref{eq-activation}) and~(\ref{eq-softmax}) should be learned automatically using an optimization algorithm that minimizes an objective cost function. 
In order to approximate the error of a certain given value of the weights, a differentiable cost (or loss) function that quantifies this error should be defined. 
The most used loss function in DNNs for the classification task is the categorical cross entropy as defined in the following equation: 
\begin{equation}
L(X)=-\sum_{j=1}^{K}{Y_j\log{\hat{Y}_j}}
\end{equation}
with $L$ denoting the loss or cost when classifying the input time series $X$. 
Similarly, the average loss when classifying the whole training set of $D$ can be defined using the following equation: 
\begin{equation}
J(\Omega)=\frac{1}{N}\sum_{n=1}^{N}{L(X_n)}
\end{equation}
with $\Omega$ denoting the set of weights to be learned by the network (in this case the weights $w$ from equations~\ref{eq-activation} and~\ref{eq-softmax}).
The loss function is minimized to learn the weights in $\Omega$ using a gradient descent method which is defined using the following equation: 
\begin{equation}
\omega = \omega - \alpha \frac{\partial J}{\partial \omega} ~ | ~ \forall ~ \omega \in \Omega
\end{equation}
with $\alpha$ denoting the learning rate of the optimization algorithm. 
By subtracting the partial derivative, the model is actually auto-tuning the parameters $\omega$ in order to reach a local minimum (or a saddle point) of $J$ in case of a non-linear classifier (which is almost always the case for a DNN). 
We should note that when the partial derivative cannot be directly computed with respect to a certain parameter $\omega$, the chain rule of derivative is employed which is in fact the main idea behind the backpropagation algorithm~\citep{lecun1998efficient}. 

\subsubsection{Convolutional Neural Networks}

Since AlexNet~\citep{krizhevsky2012imagenet} won the ImageNet competition in 2012, deep CNNs have seen a lot of successful applications in many different domains~\citep{lecun2015deep} such as reaching human level performance in image recognition problems~\citep{szegedy2015going} as well as different natural language processing tasks~\citep{sutskever2014sequence,bahdanau2015neural}. 
Motivated by the success of these CNN architectures in these various domains, researchers have started adopting them for time series analysis~\citep{gamboa2017deep}.

\begin{figure}
	\centering
	\includegraphics[width=.6\linewidth]{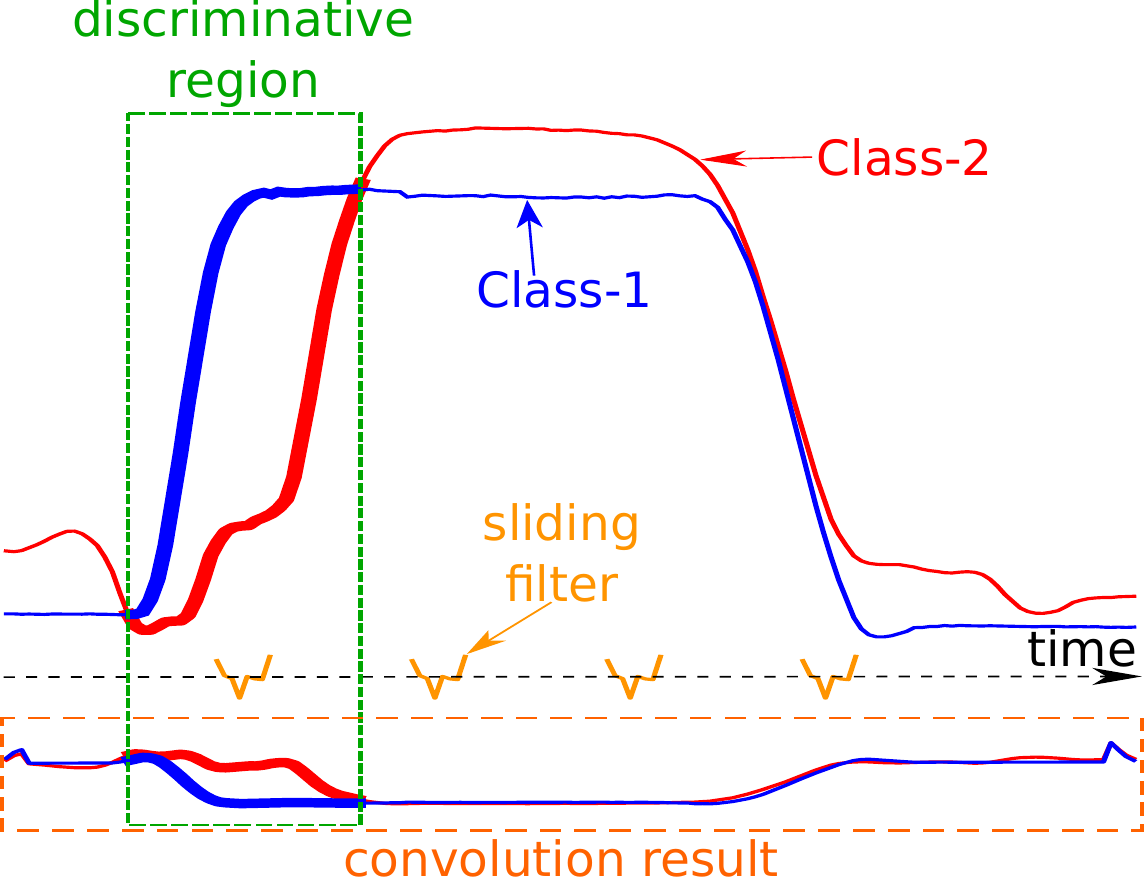}
	\caption{The result of a applying a learned discriminative convolution on the GunPoint dataset.}
	\label{fig-conv-example}
\end{figure}

A convolution can be seen as applying and sliding a filter over the time series.
Unlike images, the filters exhibit only one dimension (time) instead of two dimensions (width and height). 
The filter can also be seen as a generic non-linear transformation of a time series.
Concretely, if we are convoluting (multiplying) a filter of length 3 with a univariate time series, by setting the filter values to be equal to $[\frac{1}{3},\frac{1}{3},\frac{1}{3}]$, the convolution will result in applying a moving average with a sliding window of length $3$.
A general form of applying the convolution for a centered time stamp $t$ is given in the following equation: 
\begin{equation}
C_t = f(\omega*X_{t-l/2:t+l/2}+b)   ~  | ~ \forall ~ t \in [1,T]
\end{equation}
where $C$ denotes the result of a convolution (dot product $*$) applied on a univariate time series $X$ of length $T$ with a filter $\omega$ of length $l$, a bias parameter $b$ and a final non-linear function $f$ such as the ReLU. 
The result of a convolution (one filter) on an input time series $X$ can be considered as another univariate time series $C$ that underwent a filtering process.
Thus, applying several filters on a time series will result in a multivariate time series whose dimensions are equal to the number of filters used.  
An intuition behind applying several filters on an input time series would be to learn multiple discriminative features useful for the classification task. 

Unlike MLPs, the same convolution (the same filter values $w$ and $b$) will be used to find the result for all time stamps $t\in [1,T]$. 
This is a very powerful property (called weight sharing) of the CNNs which enables them to learn filters that are invariant across the time dimension.   

When considering an MTS as input to a convolutional layer, the filter no longer has one dimension (time) but also has dimensions that are equal to the number of dimensions of the input MTS.
Thus, the filter can be considered to be multivariate itself.

Finally, instead of setting manually the values of the filter $\omega$, these values should be learned automatically since they depend highly on the targeted dataset. 
For example, one dataset would have the optimal filter to be equal to $[1,2,2]$ whereas another dataset would have an optimal filter equal to $[2,0,-1]$. 
By \emph{optimal} we mean a filter whose application will enable the classifier to easily discriminate between the dataset classes (see Figure~\ref{fig-conv-example}).
In order to learn automatically a discriminative filter, the convolution should be followed by a discriminative classifier, which is usually preceded by a \emph{pooling} operation that can either be \emph{local} or \emph{global}. 

Local pooling such as \emph{average} or \emph{max} pooling takes an input time series and reduces its length $T$ by aggregating over a sliding window of the time series.
For example if the sliding window's length is equal to $3$ the resulting pooled time series will have a length equal to $\frac{T}{3}$ - this is only true if the stride is equal to the sliding window's length.
With a global pooling operation, the time series will be aggregated over the whole time dimension resulting in a single real value. 
In other words, this is similar to applying a local pooling with a sliding window's length equal to the length of the input time series. 
Usually a global aggregation is adopted to reduce drastically the number of parameters in a model thus decreasing the risk of overfitting while enabling the use of CAM to explain the model's decision~\citep{zhou2016learning}. 

In addition to pooling layers, some deep learning architectures include normalization layers to help the network converge quickly. 
For time series data, the batch normalization operation is performed over each channel therefore preventing the internal covariate shift across one mini-batch training of time series~\citep{ioffe2015batch}.
Another type of normalization was proposed by~\cite{ulyanov2016instance} to normalize each instance instead of a per batch basis, thus learning the mean and standard deviation of each training instance for each layer via gradient descent. 
The latter approach is called instance normalization and mimics learning the z-normalization parameters for the time series training data.

\begin{figure}[t]
	\centering
	\includegraphics[width=1.0\linewidth]{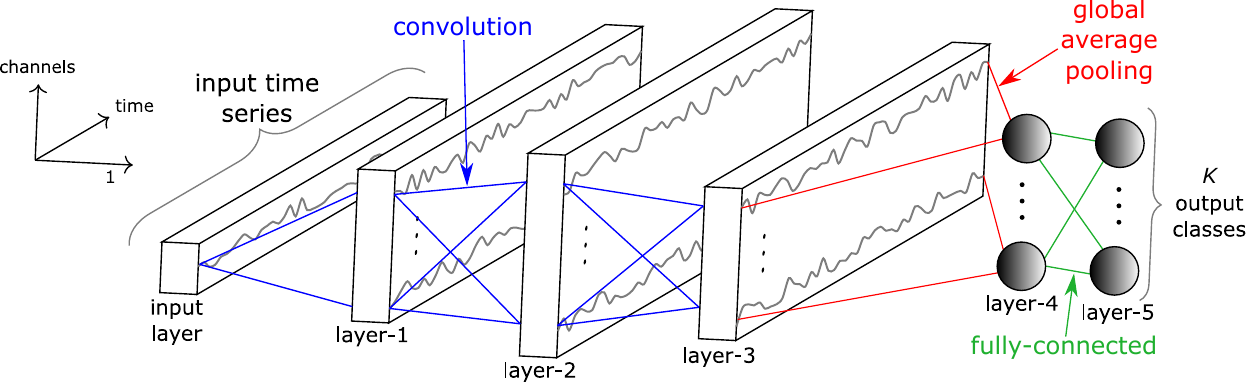}
	\caption{Fully Convolutional Neural Network architecture.}
	\label{fig-fcn-archi}
\end{figure}

The final discriminative layer takes the representation of the input time series (the result of the convolutions) and give a probability distribution over the class variables in the dataset.  
Usually, this layer is comprised of a softmax operation similarly to the MLPs. 
Note that for some approaches, we would have an additional non-linear FC layer before the final softmax layer which increases the number of parameters in a network. 
Finally in order to train and learn the parameters of a deep CNN, the process is identical to training an MLP: a feed-forward pass followed by backpropagation~\citep{lecun1998efficient}. 
An example of a CNN architecture for TSC with three convolutional layers is illustrated in Figure~\ref{fig-fcn-archi}.

\subsubsection{Echo State Networks}

Another popular type of architectures for deep learning models is the RNN. 
Apart from time series forecasting, we found that these neural networks were rarely applied for time series classification which is mainly due to three factors: (1) the type of this architecture is designed mainly to predict an output for each element (time stamp) in the time series~\citep{langkvist2014a}; (2) RNNs typically suffer from the vanishing gradient problem due to training on long time series~\citep{pascanu2012understanding}; (3) RNNs are considered hard to train and parallelize which led the researchers to avoid using them for computational reasons~\citep{pascanu2013on}.

Given the aforementioned limitations, a relatively recent type of recurrent architecture was proposed for time series: ESNs~\citep{gallicchio2017deep}. 
ESNs were first invented by~\cite{jaeger2004Harnessing} for time series prediction in wireless communication channels.
They were designed to mitigate the challenges of RNNs by eliminating the need to compute the gradient for the hidden layers which reduces the training time of these neural networks thus avoiding the vanishing gradient problem.   
These hidden layers are initialized randomly and constitutes the \emph{reservoir}: the core of an ESN which is a sparsely connected random RNN. 
Each neuron in the reservoir will create its own nonlinear activation of the incoming signal. 
The inter-connected weights inside the reservoir and the input weights are not learned via gradient descent, only the output weights are tuned using a learning algorithm such as logistic regression or Ridge classifier~\citep{hoerl1970ridge}. 

\begin{figure}
	\centering
	\includegraphics[width=0.8\linewidth]{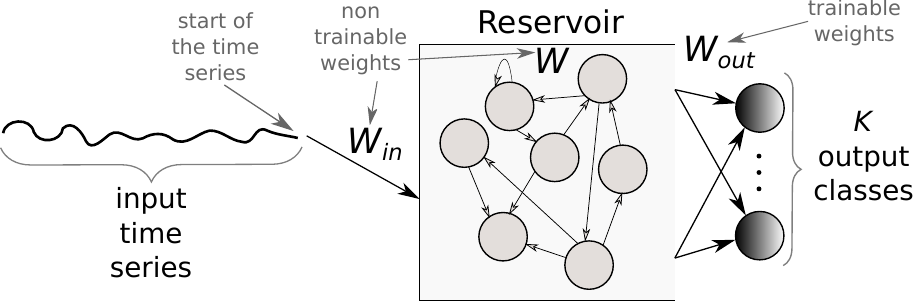}
	\caption{An Echo State Network architecture for time series classification.}
	\label{fig-esn-archi}
\end{figure}

To better understand the mechanism of these networks, consider an ESN with input dimensionality $M$, neurons in the reservoir $N_r$ and an output dimensionality $K$ equal to the number of classes in the dataset. 
Let $X(t) \in \mathbb{R}^M$, $I(t) \in \mathbb{R}^{N_r}$ and $\hat{Y}(t) \in \mathbb{R}^K$ denote the vectors of the input $M$-dimensional MTS, the internal (or hidden) state and the output unit activity for time $t$ respectively. 
Further let $W_{in} \in \mathbb{R}^{N_r\times M}$ and $W \in \mathbb{R}^{N_r\times N_r}$ and $W_{out} \in \mathbb{R}^{C\times N_r}$ denote respectively the weight matrices for the input time series, the internal connections and the output connections as seen in Figure~\ref{fig-esn-archi}.    
The internal unit activity $I(t)$ at time $t$ is updated using the internal state at time step $t-1$ and the input time series element at time $t$. 
Formally the hidden state can be computed using the following recurrence: 
\begin{equation}
I(t)=f(W_{in}X(t)+W I(t-1))   ~  | ~ \forall ~ t \in [1,T]
\end{equation}
with f denoting an activation function of the neurons, a common choice is $tanh(\cdot)$ applied element-wise~\citep{tanisaro2016time}.
The output can be computed according to the following equation: 
\begin{equation}
\hat{Y}(t)=W_{out}I(t)
\end{equation}
thus classifying each time series element $X(t)$.  
Note that ESNs depend highly on the initial values of the reservoir that should satisfy a pre-determined hyperparameter: the spectral radius. 
Figure~\ref{fig-esn-archi} shows an example of an ESN with a univariate input time series to be classified into $K$ classes.

Finally, we should note that for all types of DNNs, a set of techniques was proposed by the deep learning community to enhance neural networks' generalization capabilities. 
Regularization methods such as $l2$-norm weight decay~\citep{bishop2006pattern} or Dropout~\citep{srivastava2014a} aim at reducing overfitting by limiting the activation of the neurons. 
Another popular technique is data augmentation, which tackles the problem of overfitting a small dataset by increasing the number of training instances~\citep{baird1992document}. 
This method consists in cropping, rotating and blurring images which have been shown to improve the DNNs' performance for computer vision tasks~\citep{zhang2017understanding}. 
Although two approaches in this survey include a data augmentation technique, the study of its impact on TSC is currently limited~\citep{IsmailFawaz2018data}.
Finally we should note that Chapter~\ref{Chapter2} of this thesis is dedicated to study several regularization techniques of DNNs for the underlying TSC task.

\subsection{Generative or discriminative approaches}

\begin{figure}
	\centering
	\includegraphics[width=1.0\linewidth]{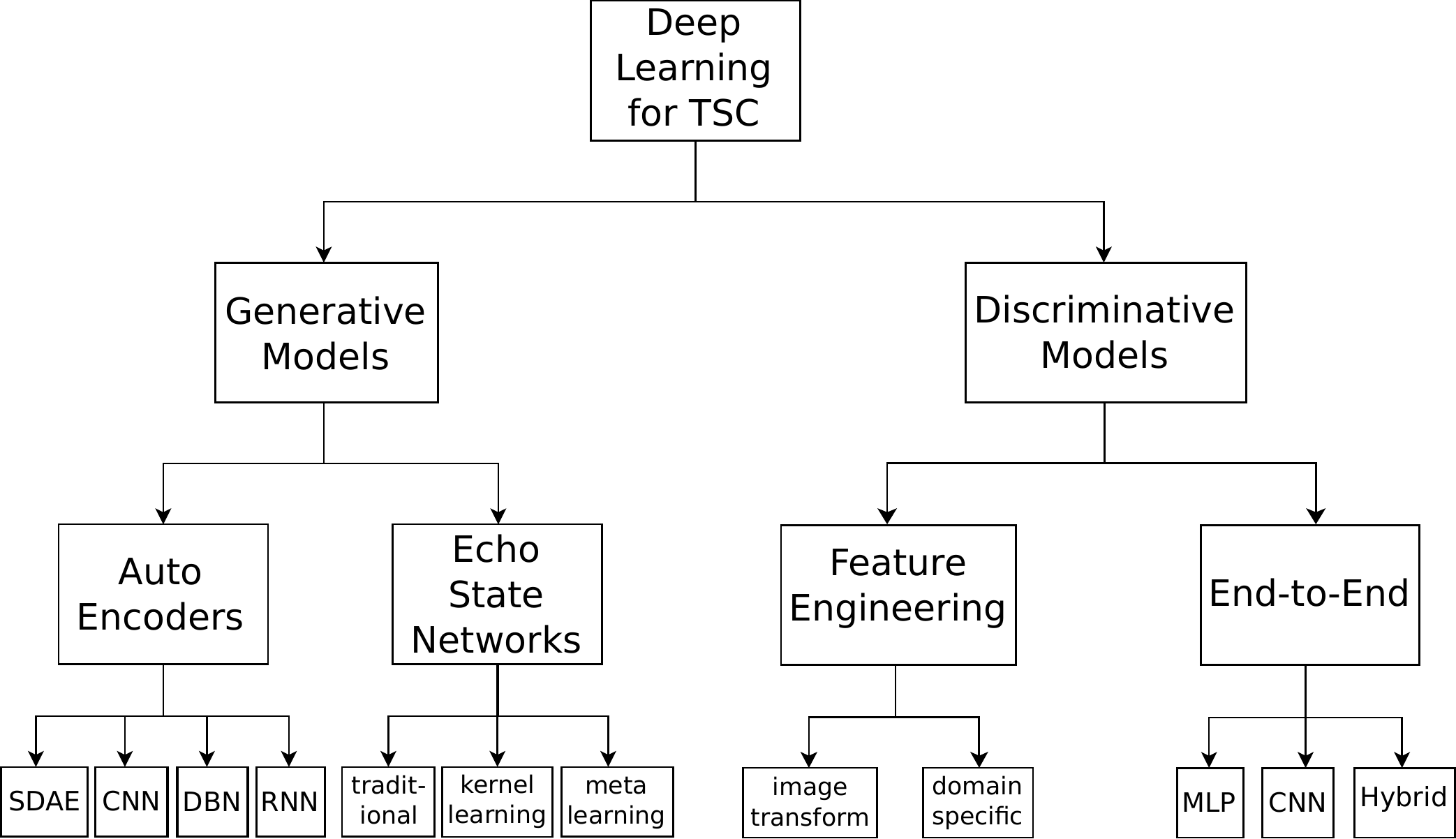}
	\caption{An overview of the different deep learning approaches for time series classification.}
	\label{fig-diagram}
\end{figure}

Deep learning approaches for TSC can be separated into two main categories: the \emph{generative} and the \emph{discriminative} models (as proposed in \cite{langkvist2014a}).
We further separate these two groups into sub-groups which are detailed in the following subsections and illustrated in Figure~\ref{fig-diagram}. 

\subsubsection{Generative models}
Generative models usually exhibit an unsupervised training step that precedes the learning phase of the classifier~\citep{langkvist2014a}.
This type of network has been referred to as \emph{Model-based} classifiers in the TSC community~\citep{bagnall2017the}. 
Some of these generative non deep learning approaches include auto-regressive models~\citep{bagnall2014a}, hidden Markov models~\citep{kotsifakos2014model} and kernel models~\citep{chen2013model}.  

For all generative approaches, the goal is to find a good representation of time series prior to training a classifier~\citep{langkvist2014a}. 
Usually, to model the time series, classifiers are preceded by an unsupervised pre-training phase such as SDAEs~\citep{bengio2013generalized,hu2016transfer}. 
A generative CNN-based model was proposed in~\cite{wang2016representation,mittelman2015time} where the authors introduced a deconvolutional operation followed by an upsampling technique that helps in reconstructing a multivariate time series. 
DBNs were also used to model the latent features in an unsupervised manner which are then leveraged to classify univariate and multivariate time series~\citep{wang2017a,banerjee2017a}.
In~\cite{mehdiyev2017time,malhotra2017timenet,rajan2018a}, an RNN auto-encoder was designed to first generate the time series then using the learned latent representation, they trained a classifier (such as SVM or Random Forest) on top of these representations to predict the class of a given input time series.

Other studies such as in~\cite{aswolinskiy2017time,bianchi2018reservoir,chouikhi2018genesis,ma2016functional} used self-predict modeling for time series classification where ESNs were first used to re-construct the time series and then the learned representation in the reservoir space was utilized for classification.  
We refer to this type of architecture by traditional ESNs in Figure~\ref{fig-diagram}. 
Other ESN-based approaches~\citep{chen2015model,chen2013model,che2017decade} define a kernel over the learned representation followed by an SVM or an MLP classifier.  
In~\cite{gong2018Multiobjective,wang2016an}, a meta-learning evolutionary-based algorithm was proposed to construct an optimal ESN architecture for univariate and multivariate time series. 
For more details concerning generative ESN models for TSC, we refer the interested reader to a recent empirical study~\citep{aswolinskiy2016time} that compared classification in reservoir and model-space for both multivariate and univariate time series.   

\subsubsection{Discriminative models}
A discriminative deep learning model is a classifier (or regressor) that directly learns the mapping between the raw input of a time series (or its hand engineered features) and outputs a probability distribution over the class variables in a dataset. 
Several discriminative deep learning architectures have been proposed to solve the TSC task, but we found that this type of model could be further sub-divided into two groups: (1) deep learning models with hand engineered features and (2) \emph{end-to-end} deep learning models. 

The most frequently encountered and computer vision inspired feature extraction method for hand engineering approaches is the transformation of time series into images using specific imaging methods such as Gramian fields~\citep{wang2015imaging,wang2015Encoding}, recurrence plots~\citep{hatami2017classification,tripathy2018use} and Markov transition fields~\citep{wang2015spatially}. 
Unlike image transformation, other feature extraction methods are not domain agnostic. 
These features are first hand-engineered using some domain knowledge, then fed to a deep learning discriminative classifier. 
For example in~\cite{uemura2018Feasibility}, several features (such as the velocity) were extracted from sensor data placed on a surgeon's hand in order to determine the skill level during surgical training. 
In fact, most of the deep learning approaches for TSC with some hand engineered features are present in human activity recognition tasks~\citep{ignatov2018activity}.
For more details on the different applications of deep learning for human motion detection using mobile and wearable sensor networks, we refer the interested reader to a recent survey by~\cite{nweke2018deep} where deep learning approaches (with or without hand engineered features) were thoroughly described specifically for the human activity recognition task.

In contrast to feature engineering, \emph{end-to-end} deep learning aims to incorporate the feature learning process while fine-tuning the discriminative classifier~\citep{nweke2018deep}. 
Since this type of deep learning approach is domain agnostic and does not include any domain specific pre-processing steps, we decided to further separate these end-to-end approaches using their neural network architectures. 

In~\cite{wang2017time,geng2018cost}, an MLP was designed to learn from scratch a discriminative time series classifier.
The problem with an MLP approach is that temporal information is lost and the features learned are no longer time-invariant.
This is where CNNs are most useful, by learning spatially invariant filters (or features) from raw input time series~\citep{wang2017time}. 
During our study, we found that CNN is the most widely applied architecture for the TSC problem, which is probably due to their robustness and the relatively small amount of training time compared to their counterpart architectures such as RNNs or MLPs. 
Several variants of CNNs have been proposed and validated on a subset of the UCR/UEA archive~\citep{ucrarchive,bagnall2017the} such as ResNets~\citep{wang2017time,geng2018cost} which add linear shortcut connections for the convolutional layers potentially enhancing the model's accuracy~\citep{he2016deep}. 
In~\cite{leguennec2016data,cui2016multi,wang2017time,zhao2017convolutional}, traditional CNNs were also validated on the UCR/UEA archive.
More recently in~\cite{wang2018multilevel}, the architectures proposed in~\cite{wang2017time} were modified to leverage a filter initialization technique based on the Daubechies 4 Wavelet values~\citep{alistair1995daubechies}. 
Outside of the UCR/UEA archive, deep learning has reached state-of-the-art performance on several datasets in different domains~\citep{langkvist2014a}. 
For spatio-temporal series forecasting problems, such as meteorology and oceanography, DNNs were proposed in~\cite{ziat2017spatio}.
~\cite{strodthoff2018detecting} proposed to detect myocardial infractions from electrocardiography data using deep CNNs.  
For human activity recognition from wearable sensors, deep learning is replacing the feature engineering approaches~\citep{nweke2018deep} where features are no longer hand-designed but rather learned by deep learning models trained through backpropagation.
One other type of time series data is present in Electronic Health Records, where a recent generative adversarial network with a CNN~\citep{che2017boosting} was trained for risk prediction based on patients historical medical records.
In~\cite{IsmailFawaz2018evaluating}, CNNs were designed to reach state-of-the-art performance for surgical skills identification. 
~\cite{liu2018time} leveraged a CNN model for multivariate and lag-feature characteristics in order to achieve state-of-the-art accuracy on the Prognostics and Health Management 2015 challenge data.
Finally, a recent review of deep learning for physiological signals classification revealed that CNNs were the most popular architecture~\citep{faust2018deep} for the considered task.  
We mention one final type of hybrid architectures that showed promising results for the TSC task on the UCR/UEA archive datasets, where mainly CNNs were combined with other types of architectures such as Gated Recurrent Units~\citep{lin2018gcrnn} and the attention mechanism~\citep{serra2018towards}.  
The reader may have noticed that CNNs appear under Auto Encoders as well as under End-to-End learning in Figure~\ref{fig-diagram}.
This can be explained by the fact that CNNs when trained as Auto Encoders have a complete different objective function than CNNs that are trained in an end-to-end fashion. 

Now that we have presented the taxonomy for grouping DNNs for TSC, we introduce in the following section the different approaches that we have included in our experimental evaluation. 
We also explain the motivations behind the selection of these algorithms.  

\section{Benchmarking deep learning for time series classification}\label{ch-1-sec-approaches}

In this section, we start by explaining the reasons behind choosing discriminative end-to-end approaches for this empirical evaluation.
We then describe in detail the nine different deep learning architectures with their corresponding advantages and drawbacks.

\subsection{Why discriminative end-to-end approaches ?}
As previously mentioned in Section~\ref{ch-1-sec-background}, the main characteristic of a generative model is fitting a time series self-predictor whose latent representation is later fed into an off-the-shelf classifier such as Random Forest or SVM. 
Although these models do sometimes capture the trend of a time series, we decided to leave these generative approaches out of our experimental evaluation for the following reasons:   
\begin{itemize}
	\item This type of method is mainly proposed for tasks other than classification or as part of a larger classification scheme~\citep{bagnall2017the};
	\item The informal consensus in the literature is that generative models are usually less accurate than direct discriminative models~\citep{bagnall2017the,nguyen2017time};
	\item The implementation of these models is usually more complicated than for discriminative models since it introduces an additional step of fitting a time series generator - this has been considered a barrier with most approaches whose code was not publicly available such as~\cite{gong2018Multiobjective,che2017decade,chouikhi2018genesis,wang2017a};
	\item The accuracy of these models depends highly on the chosen off-the-shelf classifier which is sometimes not even a neural network classifier~\citep{rajan2018a}. 
\end{itemize}

Given the aforementioned limitations for generative models, we decided to limit our experimental evaluation to discriminative deep learning models for TSC. 
In addition to restricting the study to discriminative models, we decided to only consider end-to-end approaches, thus further leaving classifiers that incorporate feature engineering out of our empirical evaluation.
We made this choice because we believe that the main goal of deep learning approaches is to remove the bias due to manually designed features~\citep{ordonez2016deep}, thus enabling the network to learn the most discriminant useful features for the classification task.
This has also been the consensus in the human activity recognition literature, where the accuracy of deep learning methods depends highly on the quality of the extracted features~\citep{nweke2018deep}.  
Finally, since our goal is to provide an empirical study of domain agnostic deep learning approaches for any TSC task, we found that it is best to compare models that do not incorporate any domain knowledge into their approach.

As for why we chose the nine approaches (described in the next section), it is first because among all the discriminative end-to-end deep learning models for TSC, we wanted to cover a wide range of architectures such as CNNs, Fully CNNs, MLPs, ResNets, ESNs, etc.
Second, since we cannot cover an empirical study of all approaches validated in all TSC domains, we decided to only include approaches that were validated on the whole (or a subset of) the univariate time series UCR/UEA archive~\citep{ucrarchive,bagnall2017the} and/or on the MTS archive~\citep{baydogan2015mts}.    
Finally, we chose to work with approaches that do not try to solve a sub task of the TSC problem such as in~\cite{geng2018cost} where CNNs were modified to classify imbalanced time series datasets. 
To justify this choice, we emphasize that imbalanced TSC problems can be solved using several techniques such as data augmentation~\citep{IsmailFawaz2018data} and modifying the class weights~\citep{geng2018cost}.
However, any deep learning algorithm can benefit from this type of modification. 
Therefore if we did include modifications for solving imbalanced TSC tasks, it would be much harder to determine if it is the choice of the deep learning classifier or the modification itself that improved the accuracy of the model.
Another sub task that has been at the center of recent studies is early time series classification~\citep{wang2016earliness} where deep CNNs were modified to include an early classification of time series. 
More recently, a deep reinforcement learning approach was also proposed for the early TSC task~\citep{martinez2018a}. 
For further details, we refer the interested reader to a recent survey on deep learning for \emph{early} time series classification~\citep{santos2017a}.  

\subsection{Compared approaches}
After having presented an overview over the recent deep learning approaches for time series classification, we present the nine architectures that we have chosen to compare in this chapter.

\subsubsection{Multi Layer Perceptron}
The MLP, which depicted in Figure~\ref{fig-mlp-archi}, is considered the most traditional form of DNNs proposed in~\cite{wang2017time} as a baseline architecture for TSC.
The network contains 4 layers in total where each one is fully connected to the output of its previous layer. 
The final layer is a softmax classifier, which is fully connected to its previous layer's output and contains a number of neurons equal to the number of classes in a dataset. 
All three hidden FC layers are composed of 500 neurons with ReLU as the activation function. 
Each layer is preceded by a dropout operation~\citep{srivastava2014a} with a rate equal to 0.1, 0.2, 0.2 and 0.3 for respectively the first, second, third and fourth layer. 
Dropout is one form of regularization that helps in preventing overfitting~\citep{srivastava2014a}. 
The dropout rate indicates the percentage of neurons that are deactivated (set to zero) in a feed forward pass during training.

MLP does not have any layer whose number of parameters is invariant across time series of different lengths (denoted by $\#invar$ in \tablename~\ref{tab-archi}) which means that the transferability of the network is not trivial: the number of parameters (weights) of the network depends directly on the length of the input time series.

\subsubsection{Fully Convolutional Neural Network}
FCNs, illustrated in Figure~\ref{fig-fcn-archi}, were first proposed in~\cite{wang2017time} for classifying univariate time series and validated on 44 datasets from the UCR/UEA archive. 
FCNs are mainly convolutional networks that do not contain any local pooling layers which means that the length of a time series is kept unchanged throughout the convolutions.
In addition, one of the main characteristics of this architecture is the replacement of the traditional final FC layer with a GAP layer which reduces drastically the number of parameters in a neural network while enabling the use of the CAM~\citep{zhou2016learning} that highlights which parts of the input time series contributed the most to a certain classification. 

The architecture proposed in~\cite{wang2017time} is first composed of three convolutional blocks where each block contains three operations: a convolution followed by a batch normalization~\citep{ioffe2015batch} whose result is fed to a ReLU activation function. 
The result of the third convolutional block is averaged over the whole time dimension which corresponds to the GAP layer. 
Finally, a traditional softmax classifier is fully connected to the GAP layer's output. 

All convolutions have a stride equal to 1 with a zero padding to preserve the exact length of the time series after the convolution. 
The first convolution contains 128 filters with a filter length equal to 8, followed by a second convolution of 256 filters with a filter length equal to 5 which in turn is fed to a third and final convolutional layer composed of 128 filters, each one with a length equal to 3. 

We can see that FCN does not hold any pooling nor a regularization operation.
In addition, one of the advantages of FCNs is the invariance (denoted by $\#invar$ in \tablename~\ref{tab-archi}) in the number of parameters for 4 layers (out of 5) across time series of different lengths. 
This invariance (due to using GAP) enables the use of a transfer learning approach where one can train a model on a certain source dataset and then fine-tune it on the target dataset~\citep{IsmailFawaz2018transfer}. 

\subsubsection{Residual Network}
The third and final proposed architecture in~\cite{wang2017time} is a relatively deep ResNet. 
For TSC, this is the deepest architecture with 11 layers of which the first 9 layers are convolutional followed by a GAP layer that averages the time series across the time dimension. 
The main characteristic of ResNet is the shortcut residual connection between consecutive convolutional layers.
In fact, the difference with the usual convolutions (such as in FCNs) is that a linear shortcut is added to link the output of a residual block to its input thus enabling the flow of the gradient directly through these connections, which makes training a DNN much easier by reducing the vanishing gradient effect~\citep{he2016deep}. 

The network is composed of three residual blocks followed by a GAP layer and a final softmax classifier whose number of neurons is equal to the number of classes in a dataset. 
Each residual block is first composed of three convolutions whose output is added to the residual block's input and then fed to the next layer. 
The number of filters for all convolutions is fixed to 64, with the ReLU activation function that is preceded by a batch normalization operation. 
In each residual block, the filter's length is set to 8, 5 and 3 respectively for the first, second and third convolution.  

\begin{figure}
	\centering
	\includegraphics[width=1.0\linewidth]{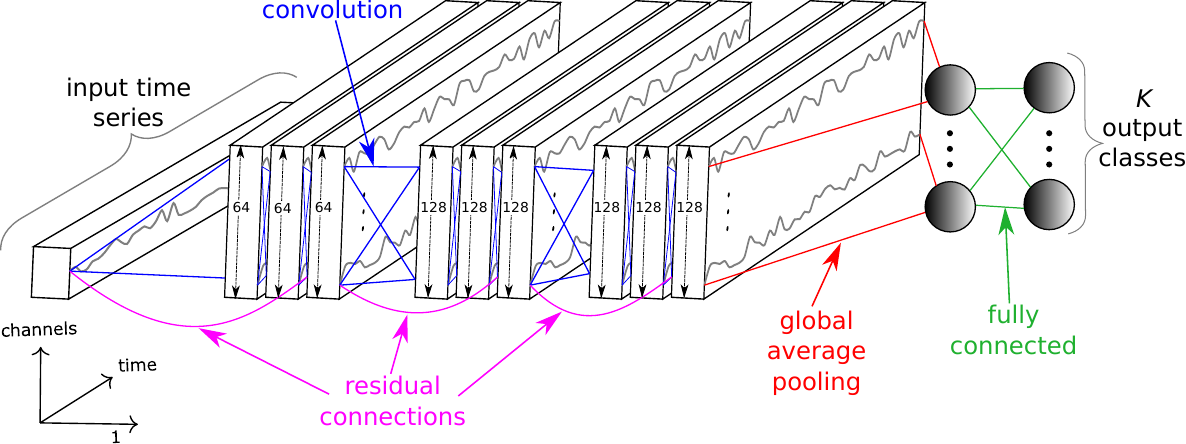}
	\caption{The Residual Network's architecture for time series classification.}
	\label{ch-1-fig-resnet-archi}
\end{figure}

Similarly to the FCN model, the layers (except the final one) in the ResNet architecture have an invariant number of parameters across different datasets. 
That being said, we can easily pre-train a model on a source dataset, then transfer and fine-tune it on a target dataset without having to modify the hidden layers of the network. 
As we have previously mentioned and since this type of transfer learning approach can give an advantage for certain types of architecture, we leave the exploration of this area of research for future work. 
The ResNet architecture proposed by~\cite{wang2017time} is depicted in Figure~\ref{ch-1-fig-resnet-archi}.

\subsubsection{Encoder}
Originally proposed by~\cite{serra2018towards}, Encoder is a hybrid deep CNN whose architecture is inspired by FCN~\citep{wang2017time} with a main difference where the GAP layer is replaced with an attention layer. 
In~\cite{serra2018towards}, two variants of Encoder were proposed: the first approach was to train the model from scratch in an end-to-end fashion on a target dataset while the second one was to pre-train this same architecture on a source dataset and then fine-tune it on a target dataset. 
The latter approach reached higher accuracy thus benefiting from the transfer learning technique. 
On the other hand, since almost all approaches can benefit to certain degree from a transfer learning method, we decided to implement only the end-to-end approach (training from scratch) which already showed high performance in the author's original paper. 

Similarly to FCN, the first three layers are convolutional with some relatively small modifications. 
The first convolution is composed of 128 filters of length 5; the second convolution is composed of 256 filters of length 11; the third convolution is composed of 512 filters of length 21. 
Each convolution is followed by an instance normalization operation~\citep{ulyanov2016instance} whose output is fed to the PReLU~\citep{he2015delving} activation function. 
The output of PReLU is followed by a dropout operation (with a rate equal to 0.2) and a final max pooling of length 2. 
The third convolutional layer is fed to an attention mechanism~\citep{bahdanau2015neural} that enables the network to learn which parts of the time series (in the time domain) are important for a certain classification. 
More precisely, to implement this technique, the input MTS is multiplied with a second MTS of the same length and number of channels, except that the latter has gone through a softmax function.
Each element in the second MTS will act as a weight for the first MTS, thus enabling the network to learn the importance of each element (time stamp).  
Finally, a traditional softmax classifier is fully connected to the latter layer with a number of neurons equal to the number of classes in the dataset. 

\begin{figure}
	\centering
	\includegraphics[width=1.0\linewidth]{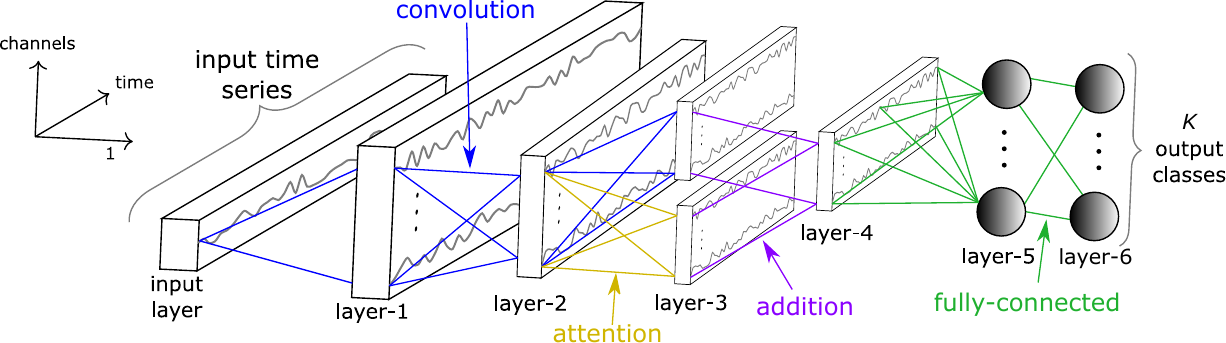}
	\caption{Encoder's architecture for time series classification.}
	\label{ch-1-fig-encoder-archi}
\end{figure}

In addition to replacing the GAP layer with the attention layer, Encoder differs from FCN in three main core changes: (1) the PReLU activation function where an additional parameter is added for each filter to enable learning the slope of the function, (2) the dropout regularization technique and (3) the max pooling operation. 
One final note is that the careful design of Encoder's attention mechanism enabled the invariance across all layers which encouraged the authors to implement a transfer learning approach.
Figure~\ref{ch-1-fig-encoder-archi} illustrates Encoder's architecture for TSC.

\subsubsection{Multi-scale Convolutional Neural Network}
Originally proposed by~\cite{cui2016multi}, MCNN is the earliest approach to validate an end-to-end deep learning architecture on the UCR Archive. 
MCNN's architecture is very similar to a traditional CNN model: with two convolutions (and max pooling) followed by an FC layer and a final softmax layer. 
On the other hand, this approach is very complex with its heavy data pre-processing step. 
\cite{cui2016multi} were the first to introduce the WS method as a data augmentation technique.
WS slides a window over the input time series and extract subsequences, thus training the network on the extracted subsequences instead of the raw input time series.
Following the extraction of a subsequence from an input time series using the WS method, a transformation stage is used. 
More precisely, prior to any training, the subsequence will undergo three transformations: (1) identity mapping; (2)  down-sampling and (3) smoothing; thus, transforming a univariate input time series into a multivariate input time series. 
This heavy pre-processing would question the end-to-end label of this approach, but since their method is generic enough we incorporated it into our developed framework. 

For the first transformation, the input subsequence is left unchanged and the raw subsequence will be used as an input for an independent first convolution. 
The down-sampling technique (second transformation) will result in shorter subsequences with different lengths which will then undergo another independent convolutions in parallel to the first convolution. 
As for the smoothing technique (third transformation), the result is a smoothed subsequence whose length is equal to the input raw subsequence which will also be fed to an independent convolution in parallel to the first and the second convolutions. 

The output of each convolution in the first convolutional stage is concatenated to form the input of the subsequent convolutional layer. 
Following this second layer, an FC layer is deployed with 256 neurons using the sigmoid activation function. 
Finally, the usual softmax classifier is used with a number of neurons equal to the number of classes in the dataset.  

\begin{figure}
	\centering
	\includegraphics[width=1.0\linewidth]{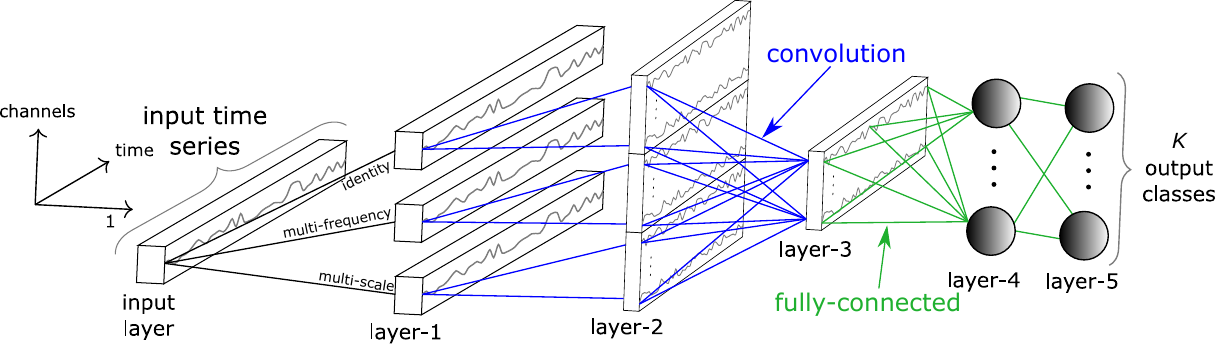}
	\caption{MCNN's architecture for time series classification.}
	\label{ch-1-fig-mcnn-archi}
\end{figure}

Note that each convolution in this network uses 256 filters with the sigmoid as an activation function, followed by a max pooling operation. 
Two architecture hyperparameters are cross-validated, using a grid search on an unseen split from the training set: the filter length and the pooling factor which determines the pooling size for the max pooling operation.   
The total number of layers in this network is 4, out of which only the first two convolutional layers are invariant (transferable).  
Finally, since the WS method is also used at test time, the class of an input time series is determined by a majority vote over the extracted subsequences' predicted labels. 
Figure~\ref{ch-1-fig-mcnn-archi} shows the MCNN's architecture for TSC. 

\subsubsection{Time Le-Net}
T-LeNet was originally proposed by~\cite{leguennec2016data} and inspired by the great performance of LeNet's architecture for the document recognition task~\citep{lecun1998gradient}. 
This model can be considered as a traditional CNN with two convolutions followed by an FC layer and a final softmax classifier. 
There are two main differences with the FCNs: (1) an FC layer and (2) local max-pooling operations.   
Unlike GAP, local pooling introduces invariance to small perturbations in the activation map (the result of a convolution) by taking the maximum value in a local pooling window.
Therefore for a pool size equal to 2, the pooling operation will halve the length of a time series by taking the maximum value between each two time steps. 

For both convolutions, the ReLU activation function is used with a filter length equal to 5. 
For the first convolution, 5 filters are used and followed by a max pooling of length equal to 2. 
The second convolution uses 20 filters followed by a max pooling of length equal to 4. 
Thus, for an input time series of length $l$, the resulting output of these two convolutions will divide the length of the time series by $8=4\times2$. 
The convolutional blocks are followed by a non-linear fully connected layer which is composed of 500 neurons, each one using the ReLU activation function.
Finally, similarly to all previous architectures, the number of neurons in the final softmax classifier is equal to the number of classes in a dataset. 

Unlike ResNet and FCN, this approach does not have much invariant layers (2 out of 4) due to the use of an FC layer instead of a GAP layer, thus increasing drastically the number of parameters needed to be trained which also depends on the length of the input time series. 
Thus, the transferability of this network is limited to the first two convolutions whose number of parameters depends solely on the number and length of the chosen filters. 

\begin{figure}
	\centering
	\includegraphics[width=\linewidth]{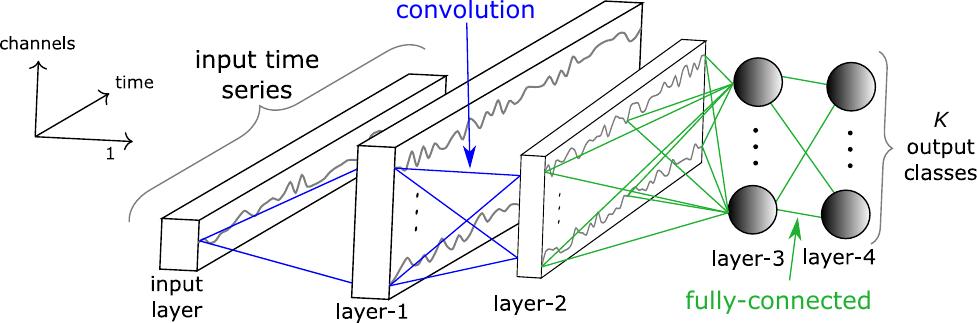}
	\caption{T-LeNet's architecture for time series classification.}
	\label{ch-1-fig-t-lenet-archi}
\end{figure}

We should note that t-LeNet is one of the approaches adopting a data augmentation technique to prevent overfitting especially for the relatively small time series datasets in the UCR/UEA archive. 
Their approach uses two data augmentation techniques: WS and WW. 
The former method is identical to MCNN's data augmentation technique originally proposed in~\cite{cui2016multi}.  
As for the second data augmentation technique, WW employs a warping technique that squeezes or dilates the time series. 
In order to deal with multi-length time series the WS method is adopted to ensure that subsequences of the same length are extracted for training the network. 
Therefore, a given input time series of length $l$ is first dilated ($\times2$) then squeezed ($\times\frac{1}{2}$) resulting in three time series of length $l$, $2l$ and $\frac{1}{2}l$ that are fed to WS to extract equal length subsequences for training.
Note that in their original paper~\citep{leguennec2016data}, WS' length is set to $0.9l$. 
Finally similarly to MCNN, since the WS method is also used at test time, a majority vote over the extracted subsequences' predicted labels is applied. 
Figure~\ref{ch-1-fig-t-lenet-archi} shows the details of T-LeNet's architecture.

\subsubsection{Multi Channel Deep Convolutional Neural Network}
MCDCNN was originally proposed and validated on two multivariate time series datasets~\citep{zheng2014time,zheng2016exploiting}.
The proposed architecture is mainly a traditional deep CNN with one modification for MTS data: the convolutions are applied independently (in parallel) on each dimension (or channel) of the input MTS.  

\begin{figure}
	\centering
	\includegraphics[width=\linewidth]{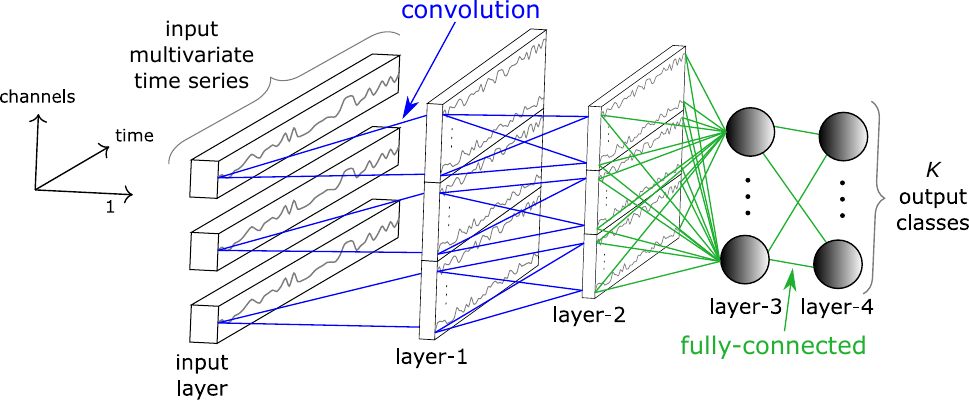}
	\caption{MCDCNN's architecture for time series classification.}
	\label{ch-1-fig-mcdcnn-archi}
\end{figure}

Each dimension for an input MTS will go through two convolutional stages with 8 filters of length 5 with ReLU as the activation function.
Each convolution is followed by a max pooling operation of length 2. 
The output of the second convolutional stage for all dimensions is concatenated over the channels axis and then fed to an FC layer with 732 neurons with ReLU as the activation function. 
Finally, the softmax classifier is used with a number of neurons equal to the number of classes in the dataset. 
By using an FC layer before the softmax classifier, the transferability of this network is limited to the first and second convolutional layers. 
MCDCNN's architecture is depicted in Figure~\ref{ch-1-fig-mcdcnn-archi}. 

\subsubsection{Time Convolutional Neural Network}
Time-CNN approach was originally proposed by~\cite{zhao2017convolutional} for both univariate and multivariate TSC. 
There are three main differences compared to the previously described networks.
The first characteristic of Time-CNN is the use of the MSE instead of the traditional categorical cross-entropy loss function, which has been used by all the deep learning approaches we have mentioned so far.
Hence, instead of a softmax classifier, the final layer is a traditional FC layer with sigmoid as the activation function, which does not guarantee a sum of probabilities equal to 1.  
Another difference to traditional CNNs is the use of a local \emph{average} pooling operation instead of local \emph{max} pooling. 
In addition, unlike MCDCNN, for MTS data they apply one convolution for all the dimensions of a multivariate classification task. 
Another unique characteristic of this architecture is that the final classifier is fully connected directly to the output of the second convolution, which removes completely the GAP layer without replacing it with an FC non-linear layer. 

\begin{figure}
	\centering
	\includegraphics[width=\linewidth]{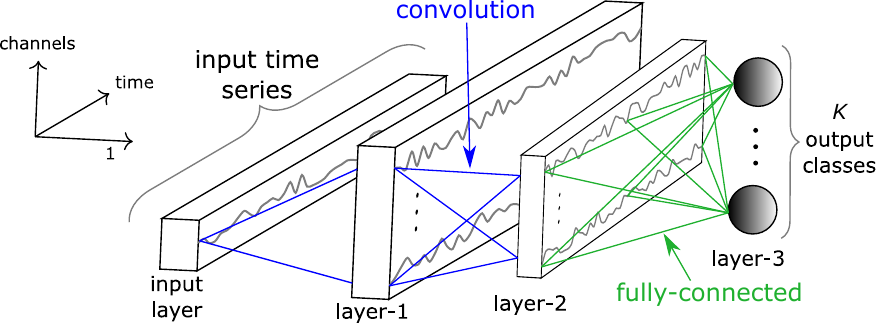}
	\caption{Time-CNN's architecture for time series classification.}
	\label{ch-1-fig-time-cnn-archi}
\end{figure}

The network is composed of two consecutive convolutional layers with respectively 6 and 12 filters followed by a local average pooling operation of length 3.
The convolutions adopt the sigmoid as the activation function.   
The network's output consists of an FC layer with a number of neurons equal to the number of classes in the dataset. 
Figure~\ref{ch-1-fig-time-cnn-archi} shows the relatively shallow architecture of Time-CNN. 

\subsubsection{Time Warping Invariant Echo State Network}
TWIESN~\citep{tanisaro2016time} is the only non-convolutional \emph{recurrent} architecture tested and re-implemented in our study.
Although ESNs were originally proposed for time series forecasting,~\cite{tanisaro2016time} proposed a variant of ESNs that uses directly the raw input time series and predicts a probability distribution over the class variables.

In fact, for each element (time stamp) in an input time series, the reservoir space is used to project this element into a higher dimensional space. 
Thus, for a univariate time series, the element is projected into a space whose dimensions are inferred from the size of the reservoir. 
Then for each element, a Ridge classifier~\citep{hoerl1970ridge} is trained to predict the class of each time series element. 
During test time, for each element of an input test time series, the already trained Ridge classifier will output a probability distribution over the classes in a dataset. 
Then the a posteriori probability for each class is averaged over all time series elements, thus assigning for each input test time series the label for which the averaged probability is maximum.
Following the original paper of~\cite{tanisaro2016time}, using a grid-search on an unseen split ($20\%$) from the training set, we optimized TWIESN's three hyperparameters: the reservoir's size, sparsity and spectral radius.
An example of TWIESN's architecture is depicted in Figure~\ref{fig-esn-archi}. 

\subsection{Hyperparameters}

Tables~\ref{tab-archi} and~\ref{tab-optim} show respectively the architecture and the optimization hyperparameters for all the described approaches except for TWIESN, since its hyperparameters are not compatible with the eight other algorithms' hyperparameters. 
We should add that for all the other deep learning classifiers (with TWIESN omitted), a model checkpoint procedure was performed either on the training set or a validation set (split from the training set).
Which means that if the model is trained for 1000 epochs, the best one on the validation set (or the train set) loss will be chosen for evaluation. 
This characteristic is included in \tablename~\ref{tab-optim} under the ``valid'' column. 
In addition to the model checkpoint procedure, we should note that all deep learning models in \tablename~\ref{tab-archi} were initialized randomly using Glorot's uniform initialization method~\citep{glorot2010understanding}. 
All models were optimized using a variant of SGD such as Adam~\citep{kingma2015adam} and AdaDelta~\citep{zeiler2012adadelta}.
We should add that for FCN, ResNet and MLP proposed in~\cite{wang2017time}, the learning rate was reduced by a factor of $0.5$ each time the model's training loss has not improved for $50$ consecutive epochs (with a minimum value equal to $0.0001$).  
One final note is that we have no way of controlling the fact that those described architectures might have been overfitted for the UCR/UEA archive and designed empirically to achieve a high performance, which is always a risk when comparing classifiers on a benchmark~\citep{bagnall2017the}. 
We therefore think that challenges where only the training data is publicly available and the testing data are held by the challenge organizer for evaluation might help in mitigating this problem. 

\begin{table}
	\centering
	\setlength\tabcolsep{4.5pt}
	{ \scriptsize
		\begin{tabularx}{0.85\textwidth}{Xcccccccc}
			\midrule
			\multirow{3}{*}{Methods} &
			\multicolumn{8}{c}{Architecture}\\ \cline{2-9}\\
			
			& \#layers & \#conv & \#invar & normalize & pooling & feature & activate & regularize \\ 
			\midrule
			
			MLP
			& 4 & 0 & 0 & none & none & FC & ReLU & dropout \\
			
			FCN
			& 5 & 3 & 4 & batch & none & GAP & ReLU & none\\
			
			ResNet
			& 11 & 9 & 10 & batch & none & GAP & ReLU & none\\
			
			Encoder
			& 5 & 3 & 4 & instance & max & Att & PReLU & dropout\\
			
			MCNN
			& 4 & 2 & 2 & none & max & FC & sigmoid & none \\
			
			t-LeNet
			& 4 & 2 & 2 & none & max & FC & ReLU & none\\
			
			MCDCNN
			& 4 & 2 & 2 & none & max & FC & ReLU & none\\
			
			Time-CNN
			& 3 & 2 & 2 & none & avg & Conv & sigmoid & none\\
			\hline
		\end{tabularx}
	}
	\caption{Architecture's hyperparameters for the deep learning approaches.}\label{tab-archi}
\end{table}

\begin{table}
	\centering
	{\scriptsize
		\begin{tabularx}{0.85\textwidth}{Xccccccc}
			\midrule
			\multirow{3}{*}{Methods} &
			
			\multicolumn{7}{c}{Optimization} \\ \cline{2-8} \\
			
			& algorithm & valid & loss & epochs & batch & learning rate & decay \\ 
			\midrule
			
			MLP
			& AdaDelta & train & entropy & 5000 & 16 & 1.0 & 0.0  \\
			
			FCN
			& Adam & train & entropy & 2000 & 16 & 0.001 & 0.0 \\
			
			ResNet
			& Adam & train & entropy & 1500 & 16 & 0.001 & 0.0\\
			
			Encoder
			& Adam & train & entropy & 100 & 12 & 0.00001 & 0.0 \\
			
			MCNN
			& Adam & split$_{20\%}$ & entropy & 200 & 256 & 0.1 & 0.0 \\
			
			t-LeNet
			& Adam & train & entropy & 1000 & 256 & 0.01 & 0.005  \\
			
			MCDCNN
			& SGD & split$_{33\%}$ & entropy & 120 & 16 & 0.01 & 0.0005 \\
			
			Time-CNN
			& Adam & train & mse & 2000 & 16 & 0.001 & 0.0 \\
			\hline
		\end{tabularx}
	}
	\caption{Optimization's hyperparameters for the deep learning approaches.}\label{tab-optim}
\end{table}

\section{Experimental setup} \label{ch-1-sec-experiment}
We first start by presenting the datasets' properties we have adopted in this empirical study. 
We then describe in details our developed open-source framework of deep learning for time series classification. 

\subsection{Datasets}

\subsubsection{Univariate archive}
In order to have a thorough and fair experimental evaluation of all approaches, we tested each algorithm on the whole UCR/UEA archive~\citep{ucrarchive} which contains 85 univariate time series datasets.
The datasets possess different varying characteristics such as the length of the series which has a minimum value of 24 for the ItalyPowerDemand dataset and a maximum equal to 2,709 for the HandOutLines dataset. 
One important characteristic that could impact the DNNs' accuracy is the size of the training set which varies between 16 and 8926 for respectively DiatomSizeReduction and ElectricDevices datasets.   
We should note that twenty datasets contains a relatively small training set (50 or fewer instances) which surprisingly was not an impediment for obtaining high accuracy when applying a very deep architecture such as ResNet. 
Furthermore, the number of classes varies between 2 (for 31 datasets) and 60 (for the ShapesAll dataset). 
Note that the time series in this archive are already z-normalized~\citep{bagnall2017the}.

Other than the fact of being publicly available, the choice of validating on the UCR/UEA archive is motivated by having datasets from different domains which have been broken down into seven different categories (Image Outline, Sensor Readings, Motion Capture, Spectrographs, ECG, Electric Devices and Simulated Data) in~\cite{bagnall2017the}. 
Further statistics, which we do not repeat for brevity, were conducted on the UCR/UEA archive in~\cite{bagnall2017the}. 

\subsubsection{Multivariate archive}
We also evaluated all deep learning models on Baydogan's archive~\citep{baydogan2015mts} that contains 13 MTS classification datasets. 
For memory usage limitations over a single GPU, we left the MTS dataset Performance Measurement System out of our experimentations.
This archive also exhibits datasets with different characteristics such as the length of the time series which, unlike the UCR/UEA archive, varies among the same dataset.
This is due to the fact that the datasets in the UCR/UEA archive are already re-scaled to have an equal length among one dataset~\citep{bagnall2017the}.

In order to solve the problem of unequal length time series in the MTS archive we decided to linearly interpolate the time series of each dimension for every given MTS, thus each time series will have a length equal to the longest time series' length.
This form of pre-processing has also been used by~\cite{ratanamahatana2005three} to show that the length of a time series is not an issue for TSC problems.
This step is very important for deep learning models whose architecture depends on the length of the input time series (such as an MLP) and for parallel computation over the GPUs.
We did not z-normalize any time series, but we emphasize that this traditional pre-processing step~\citep{bagnall2017the} should be further studied for univariate as well as multivariate data, especially since normalization is known to have a huge effect on DNNs' learning capabilities~\citep{zhang2017understanding}. 
Note that this process is only true for the MTS datasets whereas for the univariate benchmark, the time series are already z-normalized.
Since the data is pre-processed using the same technique for all nine classifiers, we can safely say, to some extent, that the accuracy improvement of certain models can be solely attributed to the model itself. 
\tablename~\ref{tab-mts} shows the different characteristics of each MTS dataset used in our experiments. 

\begin{table}
	{
		\centering
		\scriptsize
		\begin{tabularx}{0.8\textwidth}{Xcccccc}
			\hline\\
			\multirow{1}{*}{Dataset} 
			& old length & new length & classes & dimensions & train & test \\ 
			\midrule
			ArabicDigits & 4-93 & 93 & 10 & 13 & 6600 & 2200 \\
			AUSLAN & 45-136 & 136 & 95 & 22 & 1140 & 1425 \\
			CharacterTrajectories & 109-205 & 205 & 20 & 3 & 300 & 2558 \\
			CMUsubject16 & 127-580 & 580 & 2 & 62 & 29 & 29 \\
			ECG & 39-152 & 152 & 2 & 2 & 100 & 100 \\ 
			JapaneseVowels & 7-29 & 29 & 9 & 12 & 270 & 370 \\
			KickVsPunch & 274-841 & 841 & 2 & 62 & 16 & 10 \\
			Libras & 45-45 & 45 & 15 & 2 & 180 & 180 \\
			Outflow & 50-997 & 997 & 2 & 4 & 803 & 534 \\
			UWave & 315-315 & 315 & 8 & 3 & 200 & 4278 \\ 
			Wafer & 104-198 & 198 & 2 & 6 & 298 & 896 \\
			WalkVsRun & 128-1918 & 1919 & 2 & 62 & 28 & 16 \\    
			\hline
		\end{tabularx}
		\caption{The multivariate time series classification archive.}\label{tab-mts}
		
	}
\end{table}

\subsection{Experiments}
For each dataset in both archives (97 datasets in total), we have trained the nine deep learning models (presented in the previous section) with 10 different runs each.
Each run uses the same original train/test split in the archive but with a different random weight initialization, which enables us to take the mean accuracy over the 10 runs in order to reduce the bias due to the weights' initial values.
In total, we have performed 8730 experiments for the 85 univariate and 12 multivariate TSC datasets.
Thus, given the huge number of models that needed to be trained, we ran our experiments on a cluster of 60 GPUs. 
These GPUs were a mix of four types of Nvidia graphic cards: GTX 1080 Ti, Tesla K20, K40 and K80. 
The total sequential running time was approximately 100 days, that is if the computation has been done on a single GPU. 
However, by leveraging the cluster of 60 GPUs, we managed to obtain the results in less than one month.
We implemented our framework using the open source deep learning library Keras~\citep{chollet2015keras} with the Tensorflow~\citep{tensorflow2015whitepaper} back-end\footnote{The implementations are available on \url{https://github.com/hfawaz/dl-4-tsc}}.

Following~\cite{lucas2018proximity,forestier2017generating,petitjean2016faster,grabocka2014learning} we used the mean accuracy measure averaged over the 10 runs on the test set. 
When comparing with the state-of-the-art results published in~\cite{bagnall2017the} we averaged the accuracy using the median test error. 
Following the recommendation in~\cite{demsar2006statistical} we used the Friedman test~\citep{friedman1940a} to reject the null hypothesis. 
Then we performed the pairwise post-hoc analysis recommended by~\cite{benavoli2016should} where the average rank comparison is replaced by a Wilcoxon signed-rank test~\citep{wilcoxon1945individual} with Holm's alpha ($5\%$) correction~\citep{holm1979a,garcia2008an}. 
To visualize this type of comparison we used a critical difference diagram proposed by~\cite{demsar2006statistical}, where a thick horizontal line shows a group of classifiers (a clique) that are not-significantly different in terms of accuracy. 

\section{Results}\label{ch-1-sec-results}
In this section, we present the accuracies for each one of the nine approaches. 
All accuracies are absolute and not relative to each other that is if we claim algorithm A is 5\% better than algorithm B, this means that the average accuracy is 0.05 higher for algorithm A than B.

\begin{figure}
	\centering
	\includegraphics[width=\linewidth]{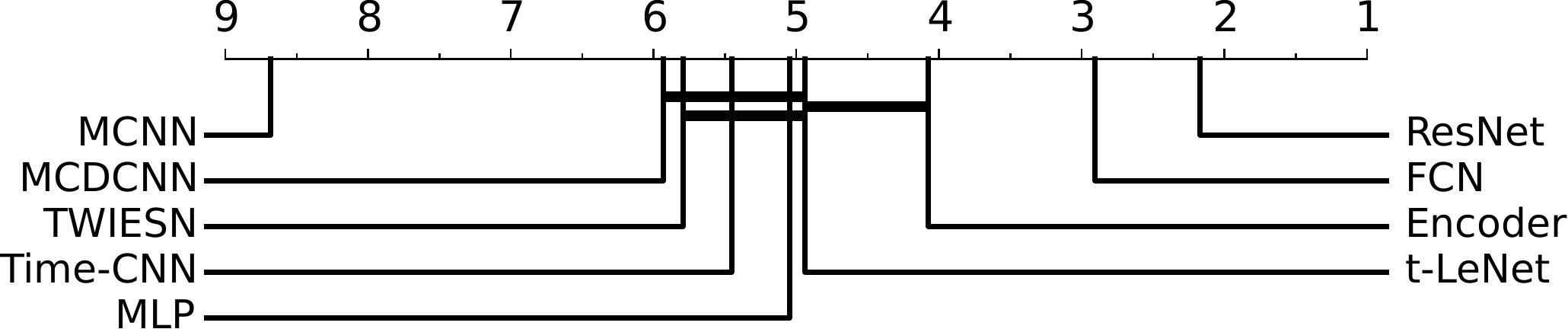}
	\caption{Critical difference diagram showing pairwise statistical difference comparison of nine deep learning classifiers on the univariate UCR/UEA time series classification archive.}
	\label{fig-cd-diagram-ucr}
\end{figure}

\subsection{Results for univariate time series}

We provide on the companion GitHub repository the raw accuracies over the 10 runs for the nine deep learning models we have tested on the 85 univariate time series datasets: the UCR/UEA archive~\citep{ucrarchive}.
The corresponding critical difference diagram is shown in Figure~\ref{fig-cd-diagram-ucr}.
The ResNet significantly outperforms the other approaches with an average rank of almost 2. 
ResNet wins on 34 problems out of 85 and significantly outperforms the FCN architecture. 
This is in contrast to the original paper's results where FCN was found to outperform ResNet on 18 out of 44 datasets, which shows the importance of validating on a larger archive in order to have a robust statistical significance. 

We believe that the success of ResNet is highly due to its deep flexible architecture.
First of all, our findings are in agreement with the deep learning for computer vision literature where deeper neural networks are much more successful than shallower architectures~\citep{he2016deep}.
In fact, in a space of 4 years, neural networks went from 7 layers in AlexNet 2012~\citep{krizhevsky2012imagenet} to \emph{1000} layers for ResNet 2016~\citep{he2016deep}.  
These types of deep architectures generally need a huge amount of data in order to generalize well on unseen examples~\citep{he2016deep}.
Although the datasets used in our experiments are relatively small compared to the billions of labeled images (such as {ImageNet}~\citep{russakovsky2015imagenet} and {OpenImages}~\citep{openimages} challenges), the deepest networks did reach competitive accuracies on the UCR/UEA archive benchmark. 

We give two potential reasons for this high generalization capabilities of deep CNNs on the TSC tasks. 
First, having seen the success of convolutions in classification tasks that require learning features that are spatially invariant in a \emph{two} dimensional space (such as width and height in images), it is only natural to think that discovering patterns in a \emph{one} dimensional space (time) should be an easier task for CNNs thus requiring less data to learn from.   
The other more direct reason behind the high accuracies of deep CNNs on time series data is its success in other sequential data such as speech recognition~\citep{hinton2012deep} and sentence classification~\citep{kim2014convolutional} where text and audio, similarly to time series data, exhibit a natural temporal ordering.

MCNN yielded very low accuracy compared to t-LeNet.
The main common idea between both of these approaches is extracting subsequences to augment the training data. 
Therefore the model learns to classify a time series from a shorter subsequence instead of the whole one, then with a majority voting scheme the time series at test time are assigned a class label. 
The poor performance of MCNN compared to t-LeNet suggest that the architecture itself is not very well optimized for the underlying task. 
In addition to MCNN's WS technique, t-LeNet uses the WW data augmentation method. 
This suggest that t-LeNet benefited even more from the warping time series data augmentation that shows significant improvement compared to MCNN's simple window slicing technique.
Nevertheless, more architecture independent experiments are needed to further verify these findings regarding data augmentation~\citep{IsmailFawaz2018data}.

Although MCDCNN and Time-CNN were originally proposed to classify MTS datasets, we have evaluated them on the univariate UCR/UEA archive.
The MCDCNN did not manage to beat any of the cl	assifiers except for the ECG5000 dataset which is already a dataset where almost all approaches reached the highest accuracy.
This low performance is probably due to the non-linear FC layer that replaces the GAP pooling of the best performing algorithms (FCN and ResNet). 
This FC layer reduces the effect of learning time invariant features which explains why MLP, Time-CNN and MCDCNN exhibit very similar performance.  

One approach that shows relatively high accuracy is Encoder~\citep{serra2018towards}. 
The statistical test indicates a significant difference between Encoder, FCN and ResNet. 
FCN wins on 17 datasets whereases Encoder wins only on 9 which suggests the superiority of the GAP layer compared to Encoder's attention mechanism. 

\subsection{Comparing with state-of-the-art approaches}

In this subsection, we compared ResNet (the most accurate DNN of our study) with the current state-of-the-art classifiers evaluated on the UCR/UEA archive in the great time series classification bake off~\citep{bagnall2017the}. 
Note that our empirical study strongly suggests to use ResNet instead of any other deep learning algorithm - it is the most accurate one with similar runtime to FCN (the second most accurate DNN). 
Finally, since ResNet's results were averaged over ten different random initializations, we chose to take one iteration of ResNet (the median) and compare it to other state-of-the-art algorithms that were executed once over the original train/test split provided by the UCR/UEA archive. 

\begin{figure}
	\centering
	\includegraphics[width=\linewidth]{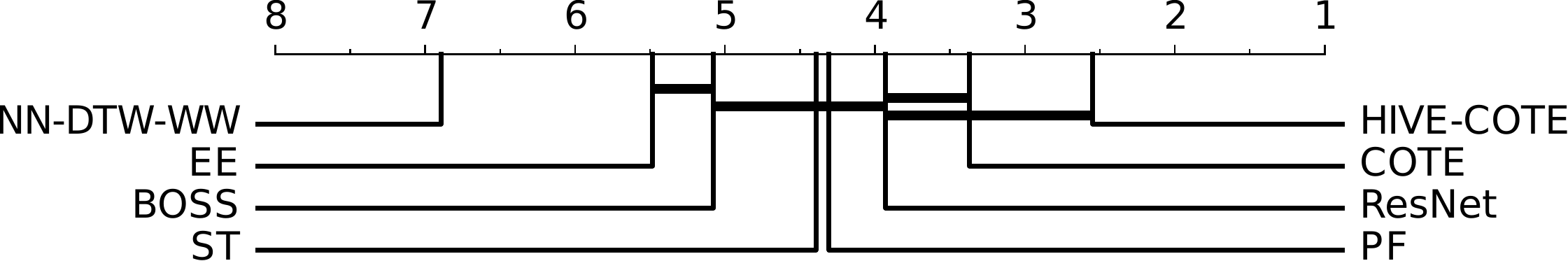}
	\caption{Critical difference diagram showing pairwise statistical difference comparison of state-of-the-art classifiers on the univariate UCR/UEA time series classification archive.}
	\label{fig-cd-diagram-uea}
\end{figure}

Out of the 18 classifiers evaluated by~\cite{bagnall2017the}, we have chosen the four best performing algorithms: (1)  EE proposed by~\cite{lines2015time} is an ensemble of nearest neighbor classifiers with 11 different time series similarity measures; (2) BOSS published in~\cite{schafer2015the} forms a discriminative bag of words by discretizing the time series using a Discrete Fourier Transform and then building a nearest neighbor classifier with a bespoke distance measure; (3) ST developed by~\cite{hills2014classification} extracts discriminative subsequences (shapelets) and builds a new representation of the time series that is fed to an ensemble of 8 classifiers; (4) COTE proposed by~\cite{bagnall2017the} is basically a weighted ensemble of 35 TSC algorithms including EE and ST.
We also include HIVE-COTE (proposed by~\cite{lines2018time}) which improves significantly COTE's performance by leveraging a hierarchical voting system as well as adding two new classifiers and two additional transformation domains. 
In addition to these five state-of-the-art classifiers, we have included the classic nearest neighbor coupled with DTW and a warping window set through cross-validation on the training set (denoted by NN-DTW-WW), since it is still one of the most popular methods for classifying time series data~\citep{bagnall2017the}.
Finally, we  added a recent approach named PF which is similar to Random Forest but replaces the attribute based splitting criteria by a random similarity measure chosen out of EE's elastic distances~\citep{lucas2018proximity}. 
Note that we did not implement any of the non-deep TSC algorithms. 
We used the results provided by~\cite{bagnall2017the} and the other corresponding papers to construct the critical difference diagram in Figure~\ref{fig-cd-diagram-uea}.

Figure~\ref{fig-cd-diagram-uea} shows the critical difference diagram over the UEA benchmark with ResNet added to the pool of six classifiers.
As we have previously mentioned, the state-of-the-art classifiers are compared to ResNet's median accuracy over the test set.
Nevertheless, we generated the ten different average ranks for each iteration of ResNet and observed that the ranking of the compared classifiers is stable for the ten different random initializations of ResNet. 
The statistical test failed to find any significant difference between COTE/HIVE-COTE and ResNet which is the only TSC algorithm that was able to reach similar performance to COTE. 
Note that for the ten different random initializations of ResNet, the pairwise statistical test always failed to find any significance between ResNet and COTE/HIVE-COTE.
PF, ST, BOSS and ResNet showed similar performances according to the Wilcoxon signed-rank test, but the fact that ResNet is not significantly different than COTE suggests that more datasets would give a better insight into these performances~\citep{demsar2006statistical}.
NN-DTW-WW and EE showed the lowest average rank suggesting that these methods are no longer competitive with current state-of-the-art algorithms for TSC. 
It is worthwhile noting that cliques formed by the Wilcoxon Signed Rank Test with Holm's alpha correction do not necessary reflect the rank order~\citep{lines2018time}. 
For example, if we have three classifiers ($C_1,C_2,C_3$) with average ranks ($C_1>C_2>C_3$), one can still encounter a case where $C_1$ is not significantly worse than $C_2$ and $C_3$ with $C_2$ and $C_3$ being significantly different. 
In our experiments, when comparing to state-of-the-art algorithms, we have encountered this problem with (ResNet$>$COTE$>$HIVE-COTE). 
Therefore we should emphasize that HIVE-COTE and COTE are \emph{significantly} different when performing the pairwise statistical test.

Although HIVE-COTE is still the most accurate classifier (when evaluated on the UCR/UEA archive) its use in a real data mining application is limited due to its huge training time complexity which is $O(N^2\cdot T^4)$ corresponding to the training time of one of its individual classifiers ST.
However, we should note that the recent work of~\cite{bostrom2015binary} showed that it is possible to use a random sampling approach to decrease significantly the running time of ST (HIVE-COTE's choke-point) without any loss of accuracy.
On the other hand, DNNs offer this type of scalability evidenced by its revolution in the field of computer vision when applied to images, which are thousand times larger than time series data~\citep{russakovsky2015imagenet}. 
In addition to the huge training time, HIVE-COTE's \emph{classification} time is bounded by a linear scan of the training set due to employing a nearest neighbor classifier, whereas the trivial GPU parallelization of DNNs provides instant classification.
Finally we should note that unlike HIVE-COTE, ResNet's hyperparameters were not tuned for each dataset but rather the same architecture was used for the whole benchmark suggesting further investigation of these hyperparameters should improve DNNs' accuracy for TSC.
These results should give an insight of deep learning for TSC therefore encouraging researchers to consider DNNs as robust real time classifiers for time series data.

\subsubsection{The need of a fair comparison}

In this paragraph, we highlight the fairness of the comparison to other machine learning TSC algorithms. 
Since we did not train nor test any of the state-of-the-art non deep learning algorithms, it is possible that we allowed much more training time for the described DNNs. 
For example, for a lazy machine learning algorithm such as NN-DTW, training time is zero when allowing maximum warping whereas it has been shown that judicially setting the warping window~\cite{dau2017judicious} can lead to a significant increase in accuracy. 
Therefore, we believe that allowing a much more thorough search of DTW's warping window would lead to a fairer comparison between deep learning approaches and other state-of-the-art TSC algorithms.  
In addition to cross-validating NN-DTW's hyper-parameters, we can imagine spending more time on data pre-processing and cleansing (e.g. smoothing the input time series) in order to improve the accuracy of NN-DTW~\citep{hoppner2016improving,ucrarchive}. 
Ultimately, in order to obtain a fair comparison between deep learning and current state-of-the-art algorithms for TSC, we think that the time spent on optimizing a network's weights should be also spent on optimizing non deep learning based classifiers especially lazy learning algorithms such as the $K$ nearest neighbor coupled with any similarity measure. 

\subsection{Results for multivariate time series}

\begin{figure}
	\centering
	\includegraphics[width=\linewidth]{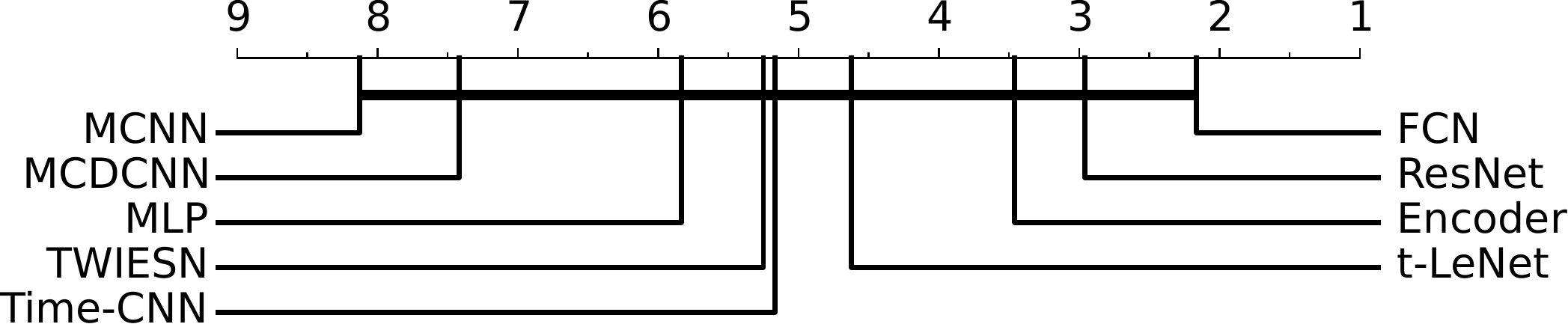}
	\caption{Critical difference diagram showing pairwise statistical difference comparison of nine deep learning classifiers on the multivariate time series classification archive.}
	\label{fig-cd-diagram-mts}
\end{figure}

\begin{figure}
	\centering
	\includegraphics[width=\linewidth]{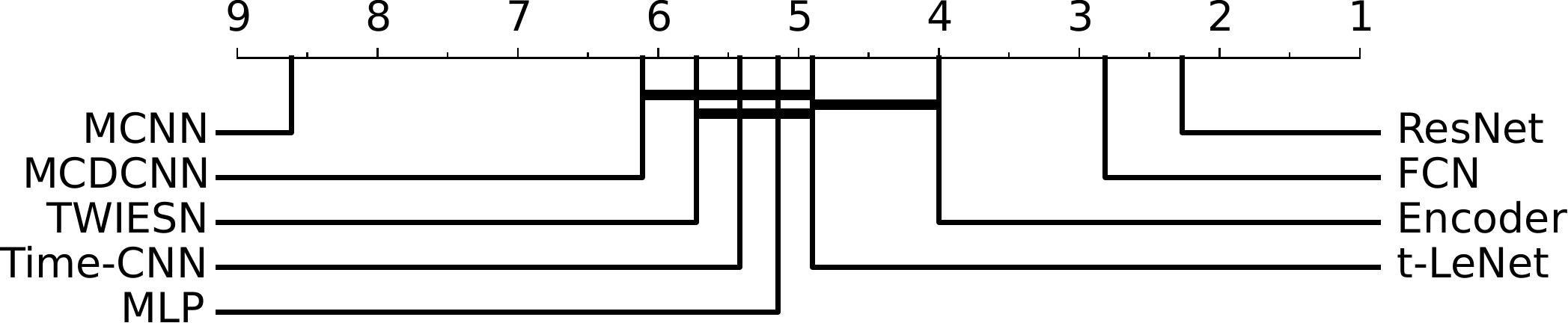}
	\caption{Critical difference diagram showing pairwise statistical difference comparison of nine deep learning classifiers on both univariate and multivariate time series classification archives.}
	\label{fig-cd-diagram-mts-ucr}
\end{figure}

We provide on our companion repository\footnote{\url{www.github.com/hfawaz/dl-4-tsc}} the detailed performance of the nine deep learning classifiers for 10 different random initializations over the 12 MTS classification datasets~\citep{baydogan2015mts}. 
Although Time-CNN and MCDCNN are the only architectures originally proposed for MTS data, they were outperformed by the three deep CNNs (ResNet, FCN and Encoder), which shows the superiority of these approaches on the MTS classification task. 
The corresponding critical difference diagram is depicted in Figure~\ref{fig-cd-diagram-mts}, where the statistical test failed to find any significant difference between the nine classifiers which is mainly due to the small number of datasets compared to their univariate counterpart. 
Therefore, we illustrated in Figure~\ref{fig-cd-diagram-mts-ucr} the critical difference diagram when both archives are combined (evaluation on 97 datasets in total).
At first glance, we can notice that when adding the MTS datasets to the evaluation, the critical difference diagram in Figure~\ref{fig-cd-diagram-mts-ucr} is not significantly different than the one in Figure~\ref{fig-cd-diagram-ucr} (where only the univariate UCR/UEA archive was taken into consideration). 
This is probably due to the fact that the algorithms' performance over the 12 MTS datasets is negligible to a certain degree when compared to the performance over the 85 univariate datasets. 
These observations reinforces the need to have an equally large MTS classification archive in order to evaluate hybrid univariate/multivariate time series classifiers.
The rest of the analysis is dedicated to studying the effect of the datasets' characteristics on the algorithms' performance.

\subsection{What can the dataset's characteristics tell us about the best architecture?}
The first dataset characteristic we have investigated is the problem's domain. 
\tablename~\ref{tab-perf-theme} shows the algorithms' performance with respect to the dataset's theme.
These themes were first defined in~\cite{bagnall2017the}. 
Again, we can clearly see the dominance of ResNet as the best performing approach across different domains.
One exception is the ECG datasets (7 in total) where ResNet was drastically beaten by the FCN model in 71.4\% of ECG datasets.
However, given the small sample size (only 7 datasets), we cannot conclude that FCN will almost always outperform the ResNet model for ECG datasets~\citep{bagnall2017the}. \\

\begin{table}
	\centering
	\setlength\tabcolsep{4.0pt}
	{ \scriptsize
		\begin{tabularx}{0.8\textwidth}{lccccccccc}
			\hline \\
			\tiny Themes (\#)     & \tiny MLP  & \tiny FCN  & \tiny ResNet & \tiny Encoder & \tiny MCNN & \tiny t-LeNet & \tiny MCDCNN & \tiny Time-CNN & \tiny TWIESN  \\
			\midrule \\
			\tiny DEVICE (6)    & 0.0  & 50.0 & \textbf{83.3}   & 0.0     & 0.0  & 16.7    & 0.0    & 0.0  & 0.0 \\
			\tiny ECG (7)       & 14.3 & \textbf{71.4} & 28.6   & 42.9    & 0.0  & 0.0    & 14.3   & 0.0  & 0.0     \\
			\tiny IMAGE (29)    & 6.9  & 34.5 & \textbf{48.3}   & 10.3    & 0.0  & 10.3    & 3.4    & 10.3 & 0.0     \\
			\tiny MOTION (14)   & 0.0 & 21.4 & \textbf{57.1}   & 0.0    & 0.0  & 42.9    & 0.0    & 0.0  & 0.0     \\
			\tiny SENSOR (16)   & 6.2  & 31.2 & \textbf{68.8}   & 31.2    & 6.2  & 37.5    & 6.2    & 0.0  & 12.5    \\
			\tiny SIMULATED (6) & 0.0  & 33.3 & \textbf{83.3}  & 33.3    & 0.0  & 33.3    & 0.0    & 0.0  & 0.0    \\
			\tiny SPECTRO (7) & 14.3 & 14.3 & \textbf{71.4}   & 0.0     & 0.0  & 0.0    & 0.0    & 28.6 & 28.6    \\
			\hline
		\end{tabularx}
	}
	\caption{Deep learning algorithms' performance grouped by themes. 
		Each entry is the percentage of dataset themes an algorithm is most accurate for. 
		Bold indicates the best model.}\label{tab-perf-theme}
\end{table}

The second characteristic which we have studied is the time series length.
Similar to the findings for non deep learning models in~\cite{bagnall2017the}, the time series length does not give information on deep learning approaches' performance.
\tablename~\ref{tab-perf-lengths} shows the average rank of each DNN over the univariate datasets grouped by the datasets' lengths. 
One might expect that the relatively short filters (3) might affect the performance of ResNet and FCN since longer patterns cannot be captured by short filters.
However, since increasing the number of convolutional layers will increase the path length viewed by the CNN model~\citep{vaswani2017attention}, ResNet and FCN managed to outperform other approaches whose filter length is longer (21) such as Encoder.
For the recurrent TWIESN algorithm, we were expecting a poor accuracy for very long time series since a recurrent model may ``forget'' a useful information present in the early elements of a long time series.
However, TWIESN did reach competitive accuracies on several long time series datasets such as reaching a 96.8\% accuracy on Meat whose time series length is equal to 448. 
This would suggest that ESNs can solve the vanishing gradient problem especially when learning from long time series.  

\begin{table}
	\centering
	\setlength\tabcolsep{4pt}
	{ \scriptsize
		\begin{tabularx}{0.8\textwidth}{Xccccccccc}
			\hline \\
			\tiny Length     & \tiny MLP  & \tiny FCN  & \tiny ResNet & \tiny Encoder & \tiny MCNN & \tiny t-LeNet & \tiny MCDCNN & \tiny Time-CNN & \tiny TWIESN \\
			\midrule \\ 
			\tiny $<$81   & 5.57 & 3.5  & \textbf{2.64}                                              & 2.93    & 8.93 & 6.14   & 3.29   & 3.86 & 5.86   \\
			\tiny 81-250  & 4.53 & \textbf{1.74} & 1.84                                              & 3.74    & 8.74 & 5.05   & 5.74   & 4.89 & 5.95   \\
			\tiny 251-450  & 4.32 & 3.05 & \textbf{1.86}                                              & 3.68    & 8.82 & 4.73   & 6.55   & 5.14 & 5.41   \\
			\tiny 451-700  & 5.23 & 2.92 & \textbf{2.0}                                               & 4.31    & 7.77 & 4.62   & 6.31   & 5.38 & 4.77   \\
			\tiny 701-1000 & 5.4  & 2.1  & \textbf{1.7}                                               & 4.6     & 8.4  & 2.9    & 6.1    & 6.8  & 5.5    \\
			\tiny $>$1000 & 3.71 & 3.0  & \textbf{1.57}                                              & 4.0     & 8.29 & 3.29   & 5.71   & 6.43 & 7.0   \\
			\hline
		\end{tabularx}
	}
	\caption{Deep learning algorithms' average ranks grouped by the datasets' length.
		Bold indicates the best model.}\label{tab-perf-lengths}
\end{table}

\begin{figure}
	\centering
	\includegraphics[width=0.7\linewidth]{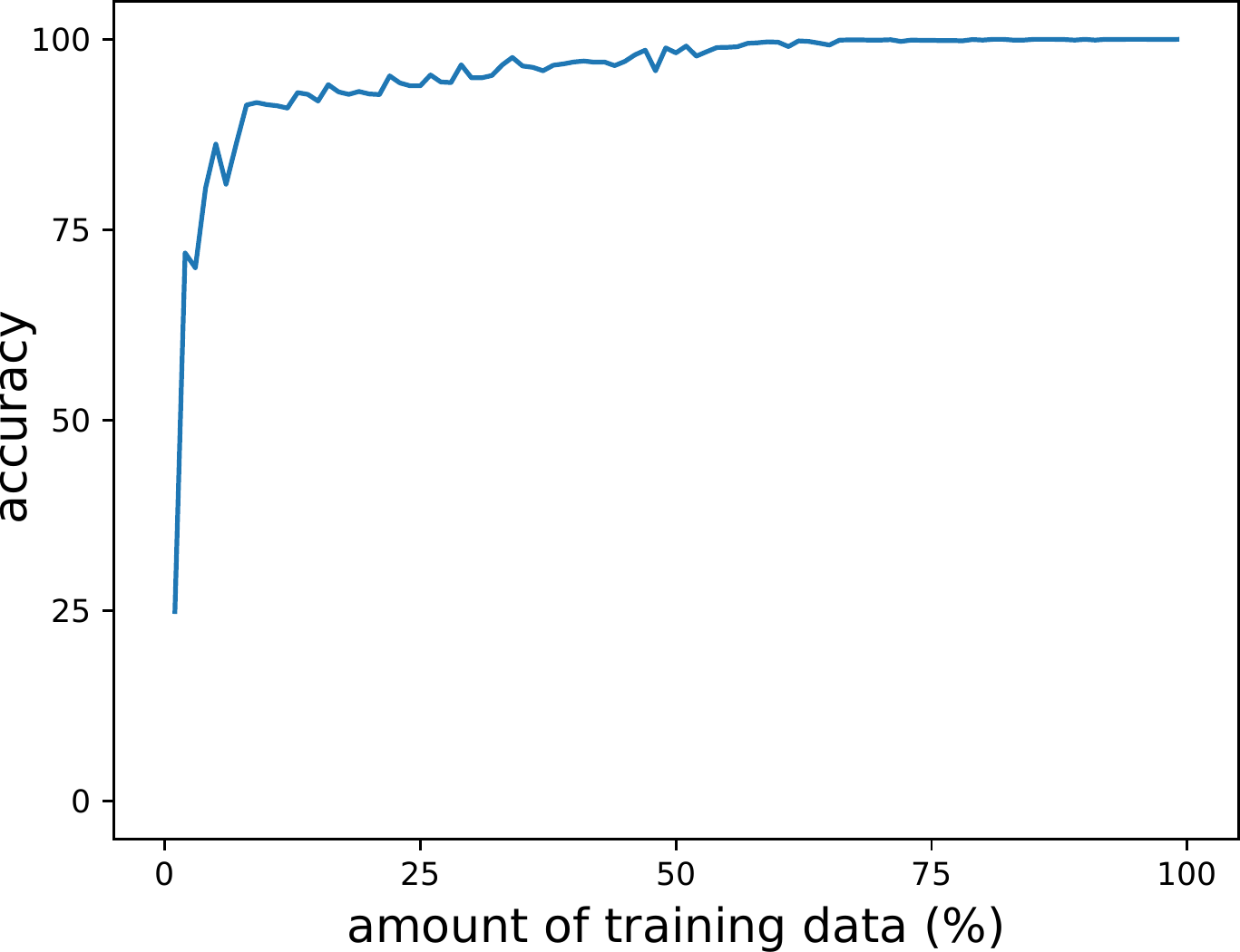}
	\caption{ResNet's accuracy variation with respect to the amount of training instances in the TwoPatterns dataset.}
	\label{fig-train-size-two-patterns}
\end{figure}

A third important characteristic is the training size of datasets and how it affects a DNN's performance. 
\tablename~\ref{tab-perf-train-size} shows the average rank for each classifier grouped by the train set's size. 
Again, ResNet and FCN still dominate with not much of a difference. 
However we found one very interesting dataset: DiatomSizeReduction. 
ResNet and FCN achieved the worst accuracy (30\%) on this dataset while Time-CNN reached the best accuracy (95\%).
Interestingly, DiatomSizeReduction is the smallest datasets in the UCR/UEA archive (with 16 training instances), which suggests that ResNet and FCN are easily overfitting this dataset. 
This suggestion is also supported by the fact that Time-CNN is the smallest model: it contains a very small number of parameters by design with only 18 filters compared to the 512 filters of FCN.
This simple architecture of Time-CNN renders overfitting the dataset much harder. 
Therefore, we conclude that the small number of filters in Time-CNN is the main reason behind its success on small datasets, however this shallow architecture is unable to capture the variability in larger time series datasets which is modeled efficiently by the FCN and ResNet architectures.
One final observation that is in agreement with the deep learning literature is that in order to achieve high accuracies while training a DNN, a large training set is needed. 
Figure~\ref{fig-train-size-two-patterns} shows the effect of the training size on ResNet's accuracy for the TwoPatterns dataset: the accuracy increases significantly when adding more training instances until it reaches 100\% for 75\% of the training data.    

\begin{table}
	\centering
	\setlength\tabcolsep{5pt}
	{\scriptsize
		\begin{tabularx}{0.8\textwidth}{Xccccccccc}
			\hline\\
			\tiny Train size   & \tiny MLP  & \tiny FCN  & \tiny ResNet & \tiny Encoder & \tiny MCNN & \tiny t-LeNet & \tiny MCDCNN & \tiny Time-CNN & \tiny TWIESN \\
			\midrule \\
			\tiny $<$100   & 4.64 & 2.21 & \textbf{1.93}                                              & 4.57    & 8.5  & 5.0    & 6.71   & 4.39 & 5.32   \\
			\tiny 100-399   & 5.21 & 2.5  & \textbf{1.86}                                              & 3.64    & 8.29 & 4.71   & 5.93   & 6.25 & 4.71   \\
			\tiny 400-799   & 4.73 & 3.07 & \textbf{2.33}                                              & 3.67    & 8.93 & 6.0    & 3.67   & 4.33 & 6.0    \\
			\tiny $>$799 & 4.29 & 3.64 & \textbf{1.86}                                              & 2.71    & 8.86 & 2.57   & 5.21   & 5.71 & 7.79   \\
			\hline 
		\end{tabularx}
	}
	\caption{Deep learning algorithms' average ranks grouped by the training sizes.
		Bold indicates the best model.}\label{tab-perf-train-size}
\end{table}

Finally, we should note that the number of classes in a dataset - although it yielded some variability in the results for the recent TSC experimental study conducted by~\cite{bagnall2017the} - did not show any significance when comparing the classifiers based on this characteristic. 
In fact, most DNNs architectures, with the categorical cross-entropy as their cost function, employ mainly the same classifier: \emph{softmax} which is basically designed for multi-class classification. 

Overall, our results show that, on average, ResNet is the best architecture with FCN and Encoder following as second and third respectively.
ResNet performed very well in general except for the ECG datasets where it was outperformed by FCN.
MCNN was significantly worse than all the other approaches, with t-LeNet showing a superiority with the additional WW technique. 
We found small variance between the approaches that replace the GAP layer with an FC dense layer (MCDCNN, CNN) which also showed similar performance to TWIESN, MLP and t-LeNet. 

\subsection{Effect of random initializations}
The initialization of deep neural networks has received a significant amount of interest from many researchers in the field~\citep{lecun2015deep}. 
These advancement have contributed to a better understanding and initialization of deep learning models in order to maximize the quality of non-optimal solutions found by the gradient descent algorithm~\citep{glorot2010understanding}. 
Nevertheless, we observed in our experiments, that DNNs for TSC suffer from a significant decrease (increase) in accuracy when initialized with bad (good) random weights.  
Therefore, we study in this section, how random initializations can affect the performance of ResNet and FCN on the whole benchmark in a best and worst case scenario.  

\begin{figure}
	\centering
	\includegraphics[width=0.7\linewidth]{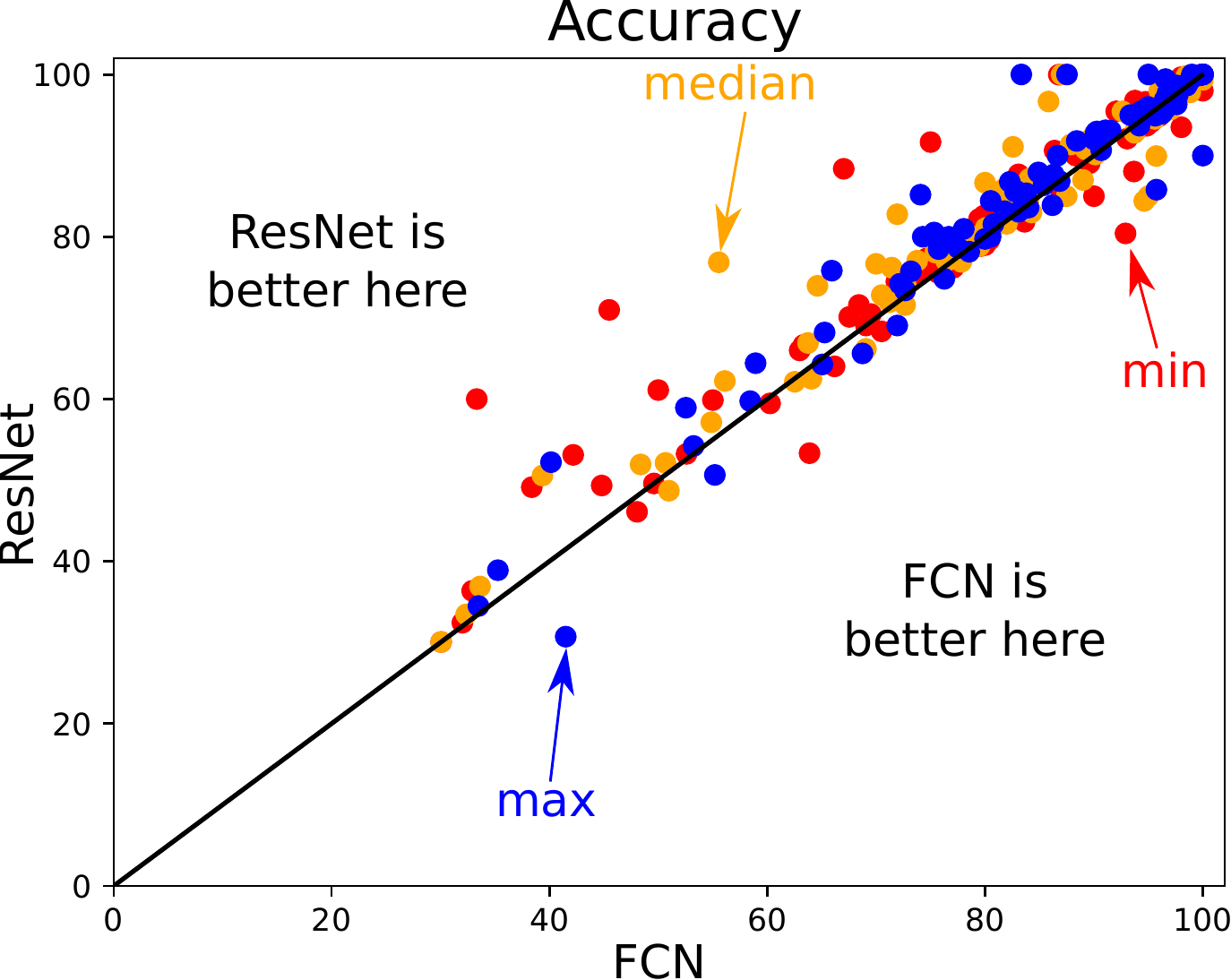}
	\caption{Accuracy of ResNet versus FCN over the UCR/UEA archive when three different aggregations are taken: the minimum, median and maximum.}
	\label{fig-resnet-vs-fcn}
\end{figure}

Figure~\ref{fig-resnet-vs-fcn} shows the accuracy plot of ResNet versus FCN on the 85 univariate time series datasets when aggregated over the 10 random initializations using three different functions: the minimum, median and maximum. 
When first observing Figure~\ref{fig-resnet-vs-fcn} one can easily conclude that ResNet has a better performance than FCN across most of the datasets regardless of the aggregation method.
This is in agreement with the critical difference diagram as well as the analysis conducted in the previous subsections, where ResNet was shown to achieve higher performance on most datasets with different characteristics.   
A deeper look into the \emph{minimum} aggregation (red points in Figure~\ref{fig-resnet-vs-fcn}) shows that FCN's performance is less stable compared to ResNet's. 
In other words, the weight's initial value can easily decrease the accuracy of FCN whereas ResNet maintained a relatively high accuracy when taking the worst initial weight values.
This is also in agreement with the average standard deviation of ResNet (1.48) which is less than FCN's (1.70).  
These observations would encourage a practitioner to avoid using a complex deep learning model since its accuracy may be unstable. 
Nevertheless we think that investigating different weight initialization techniques such as leveraging the weights of a pre-trained neural network would yield better and much more stable results~\citep{IsmailFawaz2018transfer} or even simply leveraging this high variance by ensembling DNNs~\citep{IsmailFawaz2019deep}. 

\section{Visualization}\label{ch-1-sec-visualization}

In this section, we start by investigating the use of Class Activation Map to provide an interpretable feedback that highlights the reason for a certain decision taken by the classifier.   
We then propose another visualization technique which is based on Multi-Dimensional Scaling~\citep{kruskal1978multidimensional} to understand the latent representation that is learned by the DNNs.    

\subsection{Class Activation Map}

We investigate the use of CAM which was first introduced by~\cite{zhou2016learning} to highlight the parts of an image that contributed the most for a given class identification.
\cite{wang2017time} later introduced a one-dimensional CAM with an application to TSC. 
This method explains the classification of a certain deep learning model by highlighting the subsequences that contributed the most to a certain classification. 
Figure~\ref{fig-cam-gunpoint} and~\ref{fig-cam-meat} show the results of applying CAM respectively on GunPoint and Meat datasets.  
Note that employing the CAM is only possible for the approaches with a GAP layer preceding the softmax classifier~\citep{zhou2016learning}.
Therefore, we only considered in this section the ResNet and FCN models, who also achieved the best accuracies overall. 
Note that~\cite{wang2017time} was the only paper to propose an interpretable analysis of TSC with a DNN.
We should emphasize that this is a very important research area which is usually neglected for the sake of improving accuracy: only 2 out of the 9 approaches provided a method that explains the decision taken by a deep learning model.
In this section, we start by presenting the CAM method from a mathematical point of view and follow it with two interesting case studies on Meat and GunPoint datasets. 

By employing a Global Average Pooling layer, ResNet and FCN benefit from the CAM method~\citep{zhou2016learning}, which makes it possible to identify which regions of an input time series constitute the reason for a certain classification.
Formally, let $A(t)$ be the result of the last convolutional layer which is an MTS with $M$ variables. 
$A_m(t)$ is the univariate time series for the variable $m\in [1,M]$, which is in fact the result of applying the $m^{th}$ filter. 
Now let $w_m^{c}$ be the weight between the $m^{th}$ filter and the output neuron of class $c$. 
Since a GAP layer is used then the input to the neuron of class $c$ ($z_c$) can be computed by the following equation: 
\begin{equation}
z_c=\sum_{m}{w_m^{c}}\sum_{t}{A_m(t)}
\end{equation}
The second sum constitutes the averaged time series over the whole time dimension but with the denominator omitted for simplicity.
The input $z_c$ can be also written by the following equation: 
\begin{equation}
z_c=\sum_{t}\sum_{m}{w_m^{c}A_m(t)}
\end{equation}
Finally the Class Activation Map ($CAM_c$) that explains the classification as label $c$ is given in the following equation: 
\begin{equation}
CAM_c(t)=\sum_{m}w_m^{c}A_m(t)
\end{equation}
CAM is actually a univariate time series where each element (at time stamp $t\in [1,T]$) is equal to the weighted sum of the $M$ data points at $t$, with the weights being learned by the neural network. 

\subsubsection{GunPoint dataset}
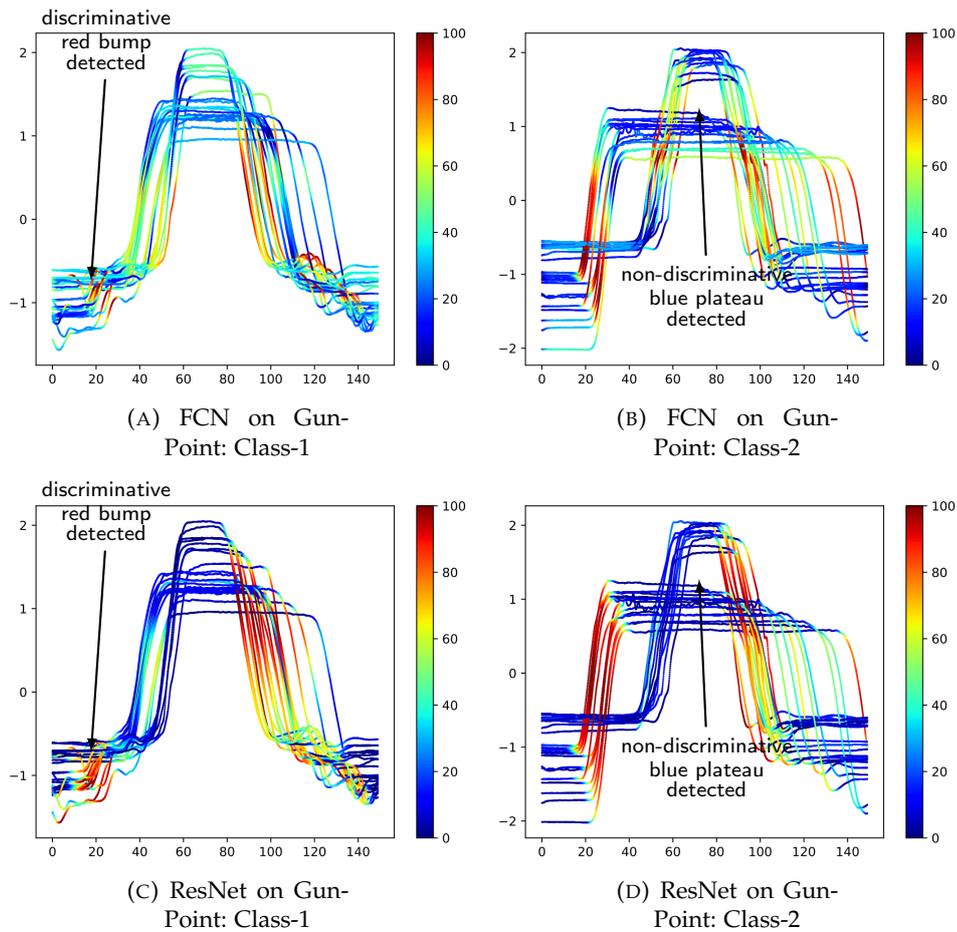
\begin{figure}
	\centering
	\subfloat[FCN on GunPoint: Class-1]{
		\includegraphics[width=0.44\linewidth]{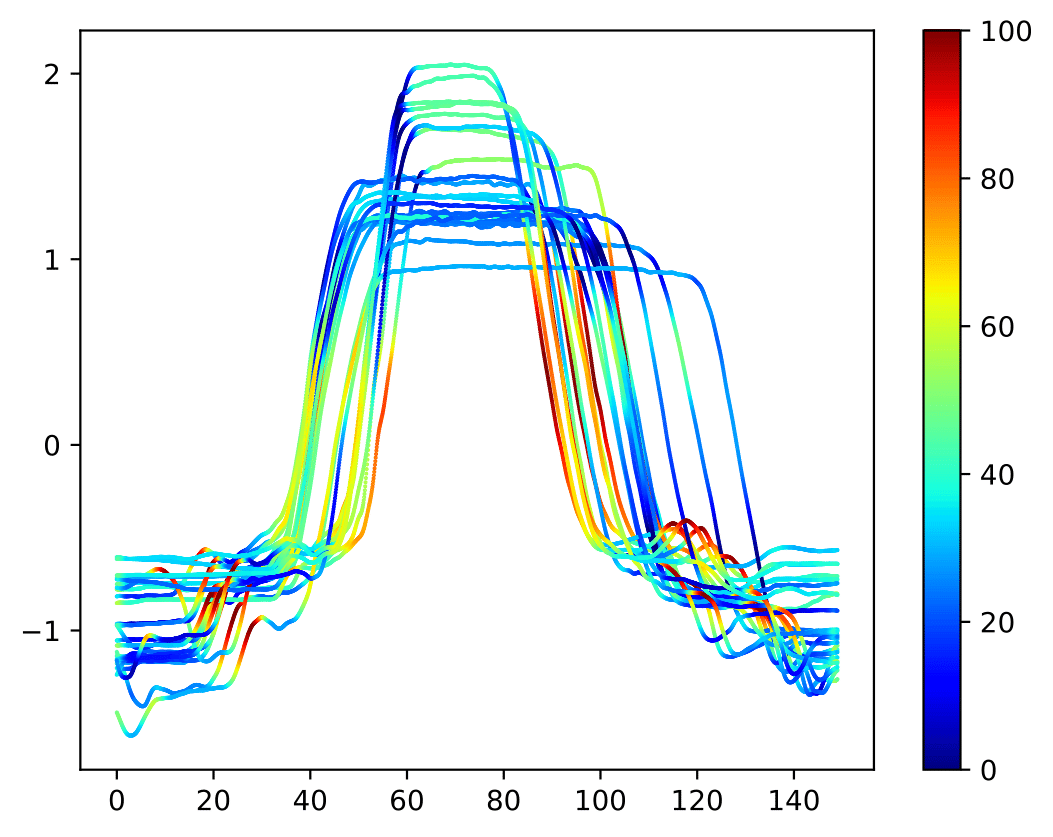}
		\label{sub-fcn-cam-GunPoint-class-1}}
	\subfloat[FCN on GunPoint: Class-2]{
		\includegraphics[width=0.44\linewidth]{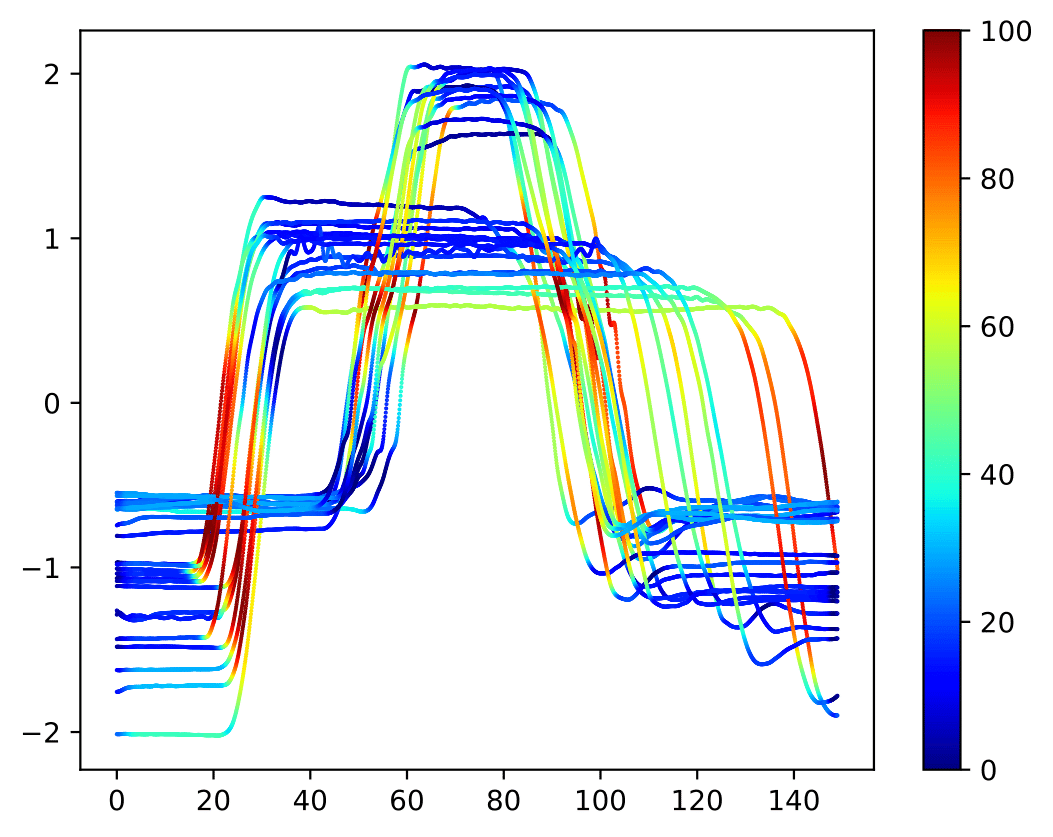}
		\label{sub-fcn-cam-GunPoint-class-2}}\\
	\subfloat[ResNet on GunPoint: Class-1]{
		\includegraphics[width=0.44\linewidth]{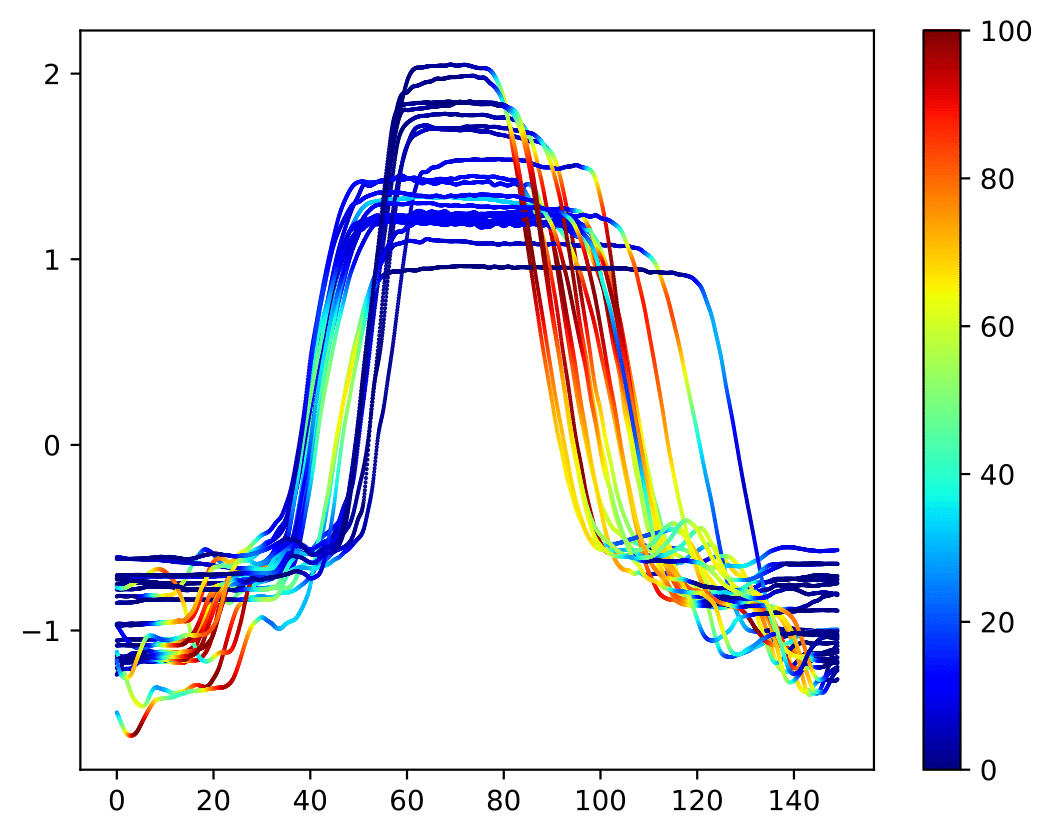}
		\label{sub-resnet-cam-GunPoint-class-1}}
	\subfloat[ResNet on GunPoint: Class-2]{
		\includegraphics[width=0.44\linewidth]{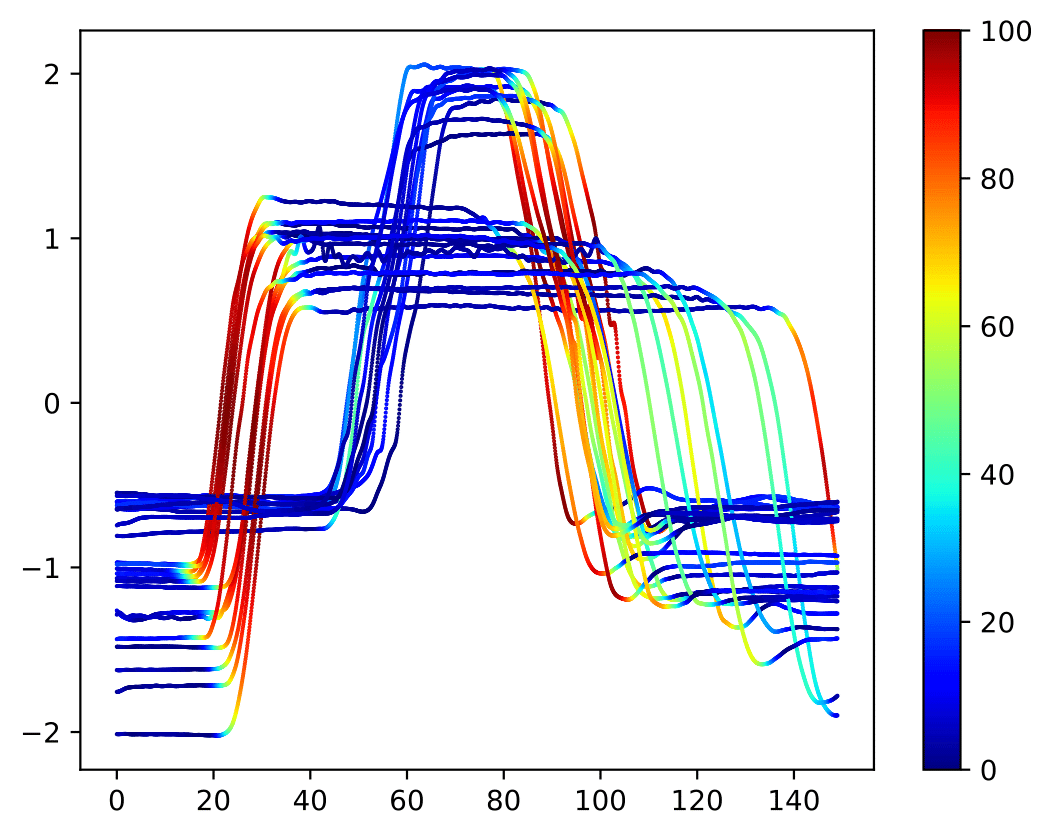}
		\label{sub-resnet-cam-GunPoint-class-2}}
	\caption{Highlighting with the Class Activation Map the contribution of each time series region for both classes in GunPoint when using the FCN and ResNet classifiers.
		Red corresponds to high contribution and blue to almost no contribution to the correct class identification (smoothed for visual clarity and best viewed in color).}
	\label{fig-cam-gunpoint}
	\usetikzlibrary{quotes,arrows.meta,positioning,decorations.pathreplacing,calc,3d,arrows}
	\begin{tikzpicture}[overlay]
	\node at (-4.9,15) (s1) {\scriptsize \textsf{discriminative}};
	\node [below=-0.15 of s1] (s2) {\scriptsize \textsf{red bump}};
	\node [below=0.12 of s1] (s3) {\scriptsize \textsf{detected}};
	\draw [-latex,thick,below left=0 of s3]edge[color=black]+(-5.1,11.5)coordinate(x1);
	\node at (3.0,11.6) (s1) {\scriptsize \textsf{non-discriminative}};
	\node [below=-0.15 of s1] (s2) {\scriptsize \textsf{blue plateau}};
	\node [below=0.12 of s1] (s3) {\scriptsize \textsf{detected}};
	\draw [-latex,thick,above=0 of s1]edge[color=black]+(2.9,13.8)coordinate(x1);
	\node at (-4.9,15-6.25) (s1) {\scriptsize \textsf{discriminative}};
	\node [below=-0.15 of s1] (s2) {\scriptsize \textsf{red bump}};
	\node [below=0.12 of s1] (s3) {\scriptsize \textsf{detected}};
	\draw [-latex,thick,below left=0 of s3]edge[color=black]+(-5.1,11.5-6.25)coordinate(x1);
	\node at (3.0,11.6-6.25) (s1) {\scriptsize \textsf{non-discriminative}};
	\node [below=-0.15 of s1] (s2) {\scriptsize \textsf{blue plateau}};
	\node [below=0.12 of s1] (s3) {\scriptsize \textsf{detected}};
	\draw [-latex,thick,above=0 of s1]edge[color=black]+(2.9,13.8-6.25)coordinate(x1);
	\end{tikzpicture}
\end{figure} 

The GunPoint dataset was first introduced by~\cite{ratanamahatana2005three} as a TSC problem. 
This dataset involves one male and one female actor performing two actions (Gun-Draw and Point) which makes it a binary classification problem.
For Gun-Draw (Class-1 in Figure~\ref{fig-cam-gunpoint}), the actors have first their hands by their sides, then draw a replicate gun from  hip-mounted holster, point it towards the target for one second, then finally place the gun in the holster and their hands to their initial position. 
Similarly to Gun-Draw, for Point (Class-2 in Figure~\ref{fig-cam-gunpoint}) the actors follow the same steps but instead of pointing a gun they point their index finger. 
For each task, the centroid of the actor's right hands on both $X$ and $Y$ axes were tracked and seemed to be very correlated, therefore the dataset contains only one univariate time series: the $X$-axis. 

We chose to start by visualizing the CAM for GunPoint for three main reasons. 
First, it is easy to visualize unlike other noisy datasets.
Second, both FCN and ResNet models achieved almost $100\%$ accuracy on this dataset which will help us to verify if both models are reaching the same decision for the same reasons.
Finally, it contains only two classes which allow us to analyze the data much more easily.  

Figure~\ref{fig-cam-gunpoint} shows the CAM's result when applied on each time series from both classes in the training set while classifying using the FCN model (Figure~\ref{sub-fcn-cam-GunPoint-class-1} and~\ref{sub-fcn-cam-GunPoint-class-2}) and the ResNet model (Figure~\ref{sub-resnet-cam-GunPoint-class-1} and~\ref{sub-resnet-cam-GunPoint-class-2}). 
At first glance, we can clearly see how both DNNs are neglecting the plateau non-discriminative regions of the time series when taking the classification decision. 
It is depicted by the blue flat parts of the time series which indicates no contribution to the classifier's decision. 
As for the highly discriminative regions (the red and yellow regions) both models were able to select the same parts of the time series which correspond to the points with high derivatives. 
Actually, the first most distinctive part of class-1 discovered by both classifiers is almost the same: the little red bump in the bottom left of Figure~\ref{sub-fcn-cam-GunPoint-class-1} and~\ref{sub-resnet-cam-GunPoint-class-1}.
Finally, another interesting observation is the ability of CNNs to localize a given discriminative shape regardless where it appears in the time series, which is evidence for CNNs' capability of learning time-invariant warped features. 

An interesting observation would be to compare the discriminative regions identified by a deep learning model with the most discriminative shapelets extracted by other shapelet-based approaches.
This observation would also be backed up by the mathematical proof provided by~\cite{cui2016multi}, that showed how the learned filters in a CNN can be considered a generic form of shapelets extracted by the learning shapelets algorithm~\citep{grabocka2014learning}.  
\cite{ye2011time} identified that the most important shapelet for the Gun/NoGun classification occurs when the actor's arm is lowered (about 120 on the horizontal axis in Figure~\ref{fig-cam-gunpoint}). 
\cite{hills2014classification} introduced a shapelet transformation based approach that discovered shapelets that are similar to the ones identified by~\cite{ye2011time}. 
For ResNet and FCN, the part where the actor lowers his/her arm (bottom right of Figure~\ref{fig-cam-gunpoint}) seems to be also identified as potential discriminative regions for some time series. 
On the other hand, the part where the actor raises his/her arm seems to be also a discriminative part of the data which suggests that the deep learning algorithms are identifying more ``shapelets''. 
We should note that this observation cannot confirm which classifier extracted the most discriminative subsequences especially because all algorithms achieved similar accuracy on GunPoint dataset. 
Perhaps a bigger dataset might provide a deeper insight into the interpretability of these machine learning models. 
Finally, we stress that the shapelet transformation classifier~\citep{hills2014classification} is an ensemble approach, which makes unclear how the shapelets affect the decision taken by the individual classifiers whereas for an end-to-end deep learning model we can directly explain the classification by using the Class Activation Map. 

\subsubsection{Meat dataset}
\begin{figure}
	\centering
	\subfloat[FCN On Meat: Class-1]{
		\includegraphics[width=0.44\linewidth]{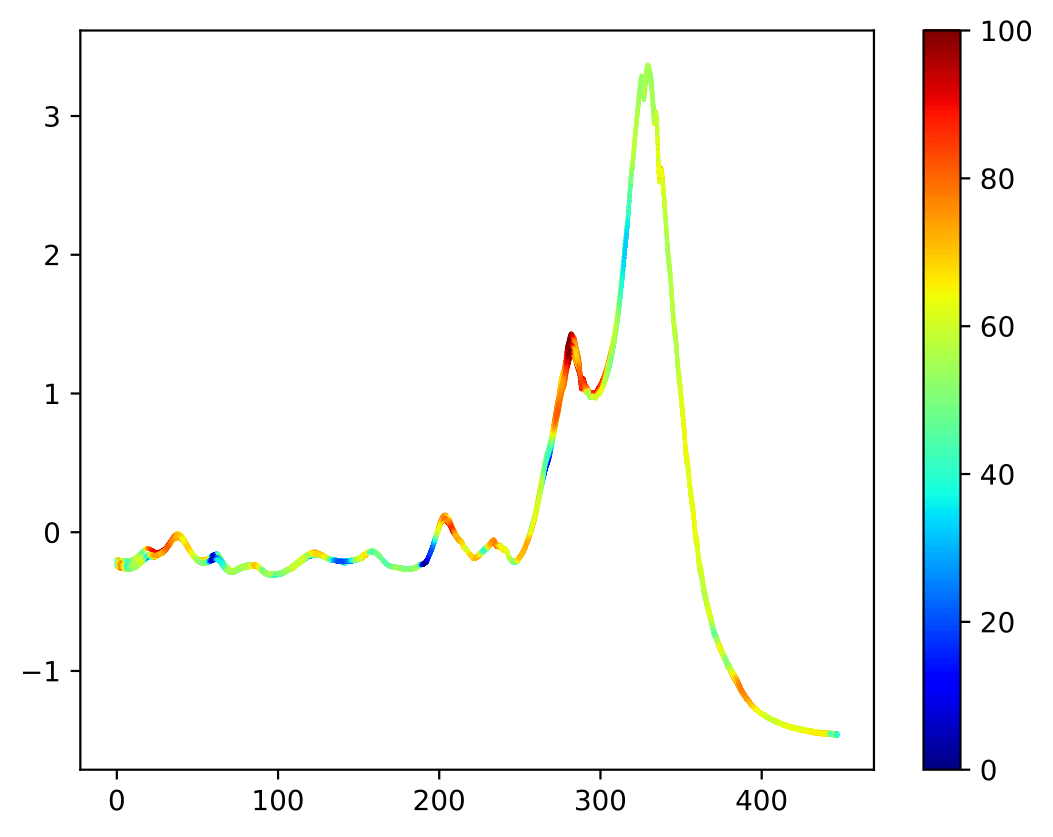}
		\label{sub-fcn-cam-Meat-class-1}}
	\subfloat[ResNet On Meat: Class-1]{
		\includegraphics[width=0.44\linewidth]{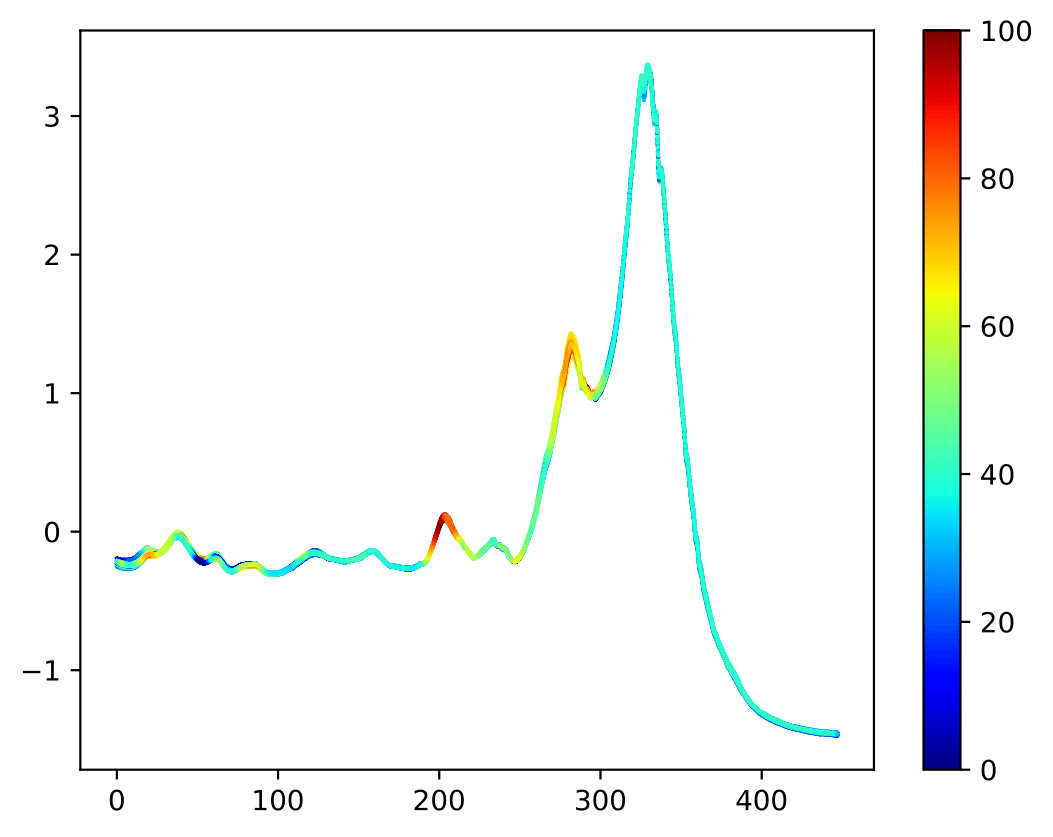}
		\label{sub-resnet-cam-Meat-class-1}}
	\\
	\subfloat[FCN On Meat: Class-2]{
		\includegraphics[width=0.44\linewidth]{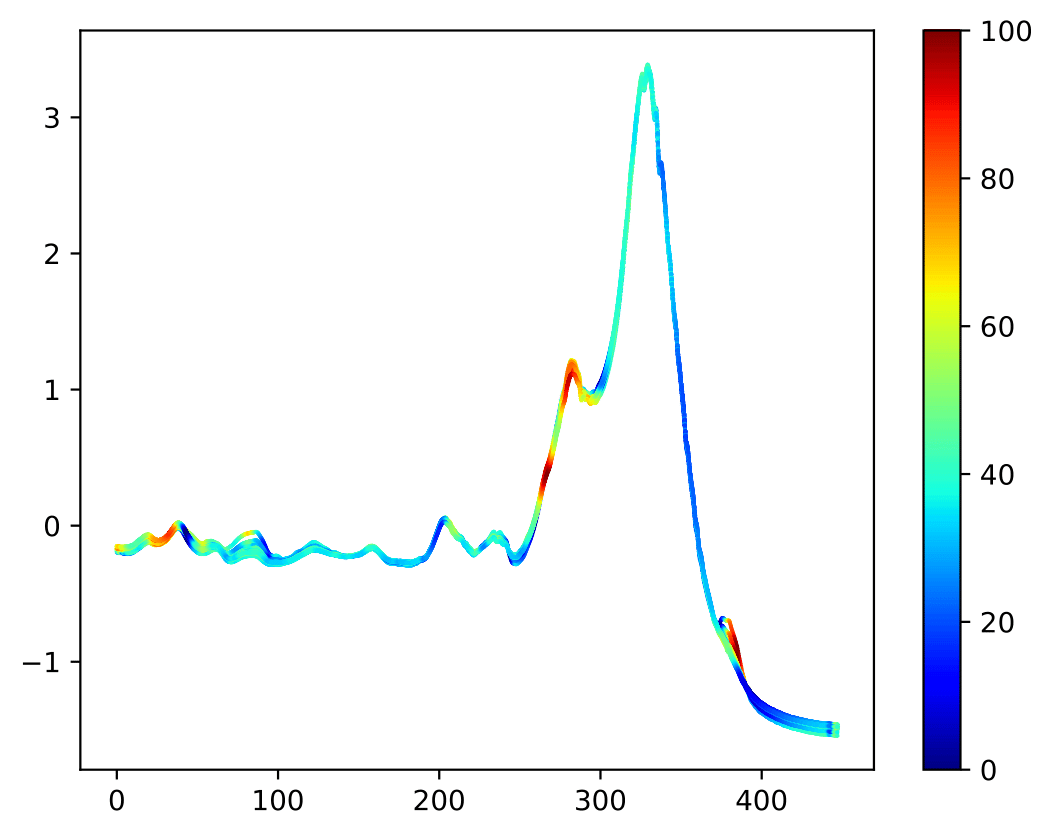}
		\label{sub-fcn-cam-Meat-class-2}}
	\subfloat[ResNet On Meat: Class-2]{
		\includegraphics[width=0.44\linewidth]{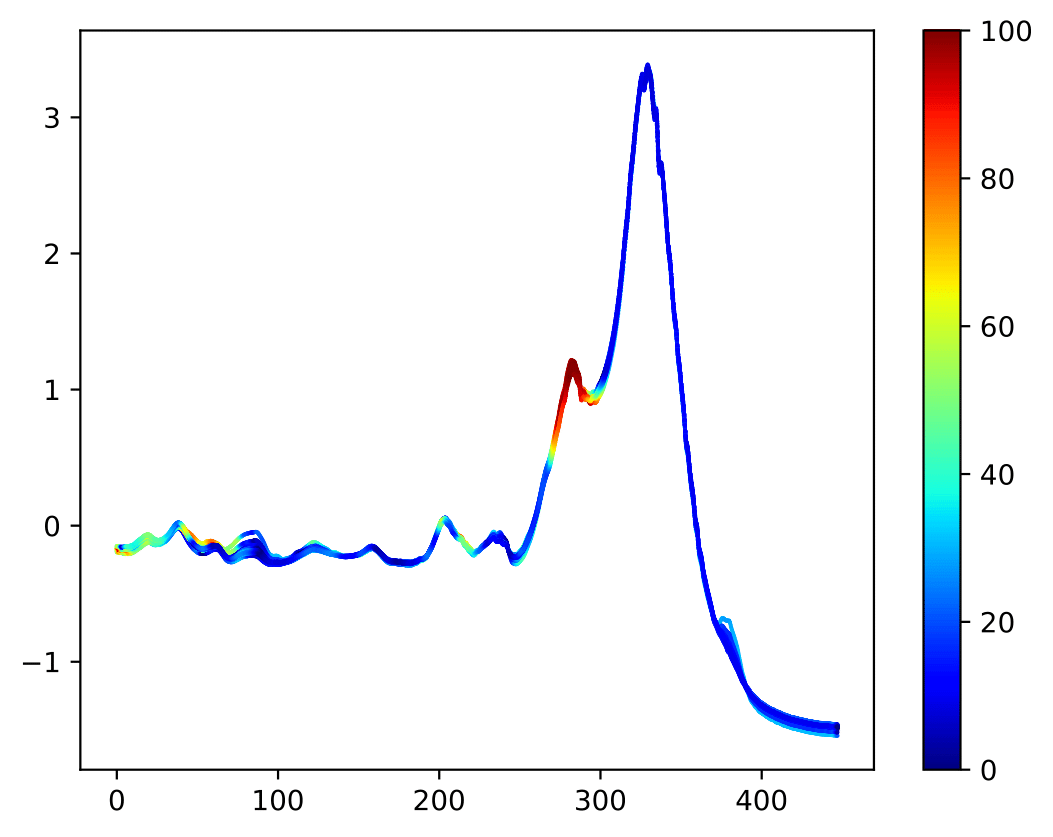}
		\label{sub-resnet-cam-Meat-class-2}}
	\\
	\subfloat[FCN On Meat: Class-3]{
		\includegraphics[width=0.44\linewidth]{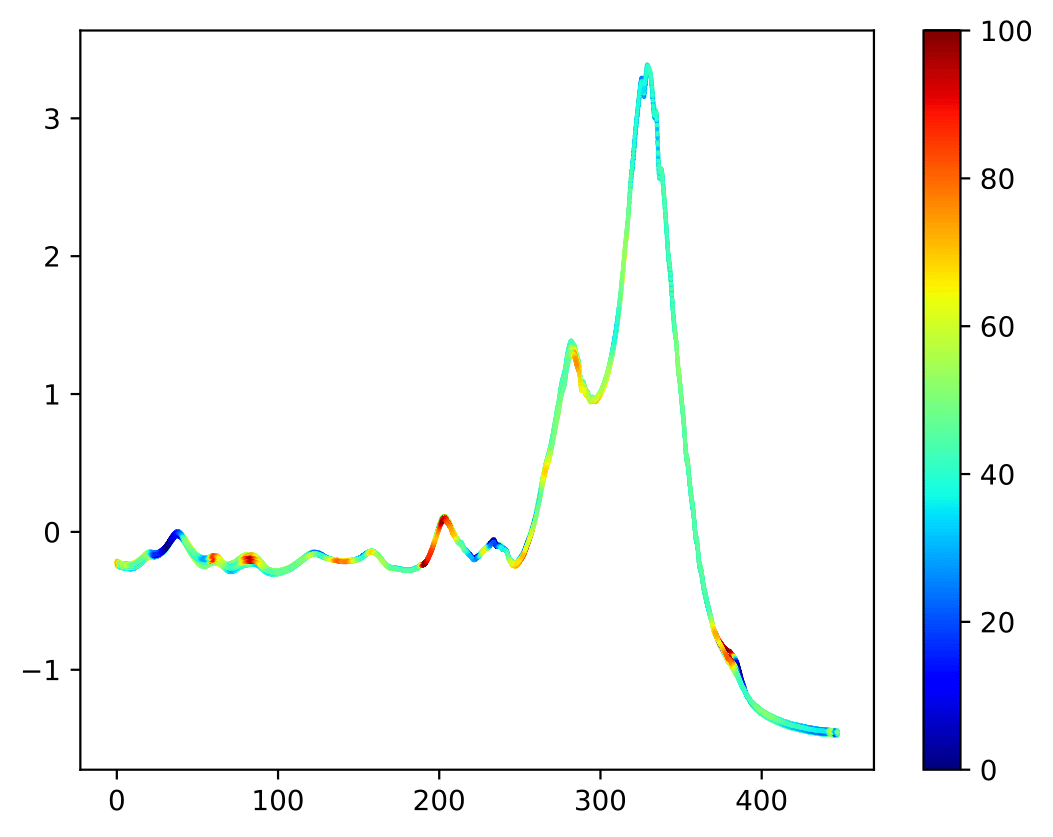}
		\label{sub-fcn-cam-Meat-class-3}}
	\subfloat[ResNet On Meat: Class-3]{
		\includegraphics[width=0.44\linewidth]{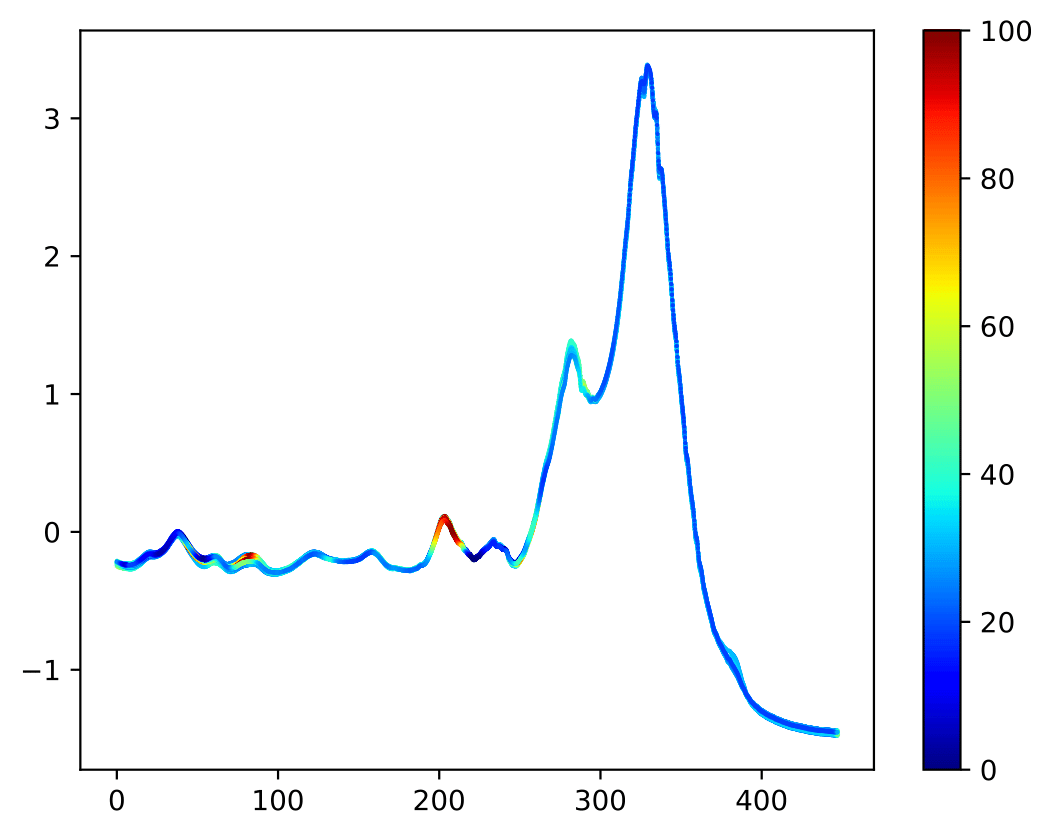}
		\label{sub-resnet-cam-Meat-class-3}}
	\caption{Highlighting with the Class Activation Map the contribution of each time series region for the three classes in Meat when using the FCN and ResNet classifiers.
		Red corresponds to high contribution and blue to almost no contribution to the correct class identification (smoothed for visual clarity and best viewed in color).}
	\label{fig-cam-meat}
	\begin{tikzpicture}[overlay]
	\node at (1.5,21.3) (s11) {\scriptsize \textsf{discriminative}};
	\node [below=-0.15 of s11] (s21) {\scriptsize \textsf{red region}};
	\node [below=0.1 of s11] (s31) {\scriptsize \textsf{detected}};
	\draw [-latex,thick,below right=0 of s31]edge[color=black]+(3.3,19.6)coordinate(x1);
	\draw [-latex,thick,below=0 of s31]edge[color=black]+(2.6,18.7)coordinate(x1);
	\draw [-latex,thick,below left=0 of s31]edge[color=black]+(-2.8,19.8)coordinate(x1);
	\node at (1.5,14) (s1) {\scriptsize \textsf{red region}};
	\node [below=-0.15 of s1] (s2) {\scriptsize \textsf{filtered-out}};
	\node [below=0.1 of s1] (s3) {\scriptsize \textsf{by ResNet}};
	\draw [-latex,thick,below right=0 of s3]edge[color=black]+(4.3,11.7)coordinate(x1);
	\draw [-latex,thick,below left=0 of s3]edge[color=black]+(-1.9,11.7)coordinate(x1);
	\node at (1.8,7) (s12) {\scriptsize \textsf{non-discriminative}};
	\node [below=-0.15 of s12] (s22) {\scriptsize \textsf{blue peak}};
	\node [below=0.1 of s12] (s32) {\scriptsize \textsf{detected}};
	\draw [-latex,thick,above right=0 of s12]edge[color=black]+(3.8,8.55)coordinate(x1);
	\draw [-latex,thick,above left=0 of s12]edge[color=black]+(-2.3,8.55)coordinate(x1);
	\end{tikzpicture}
\end{figure}
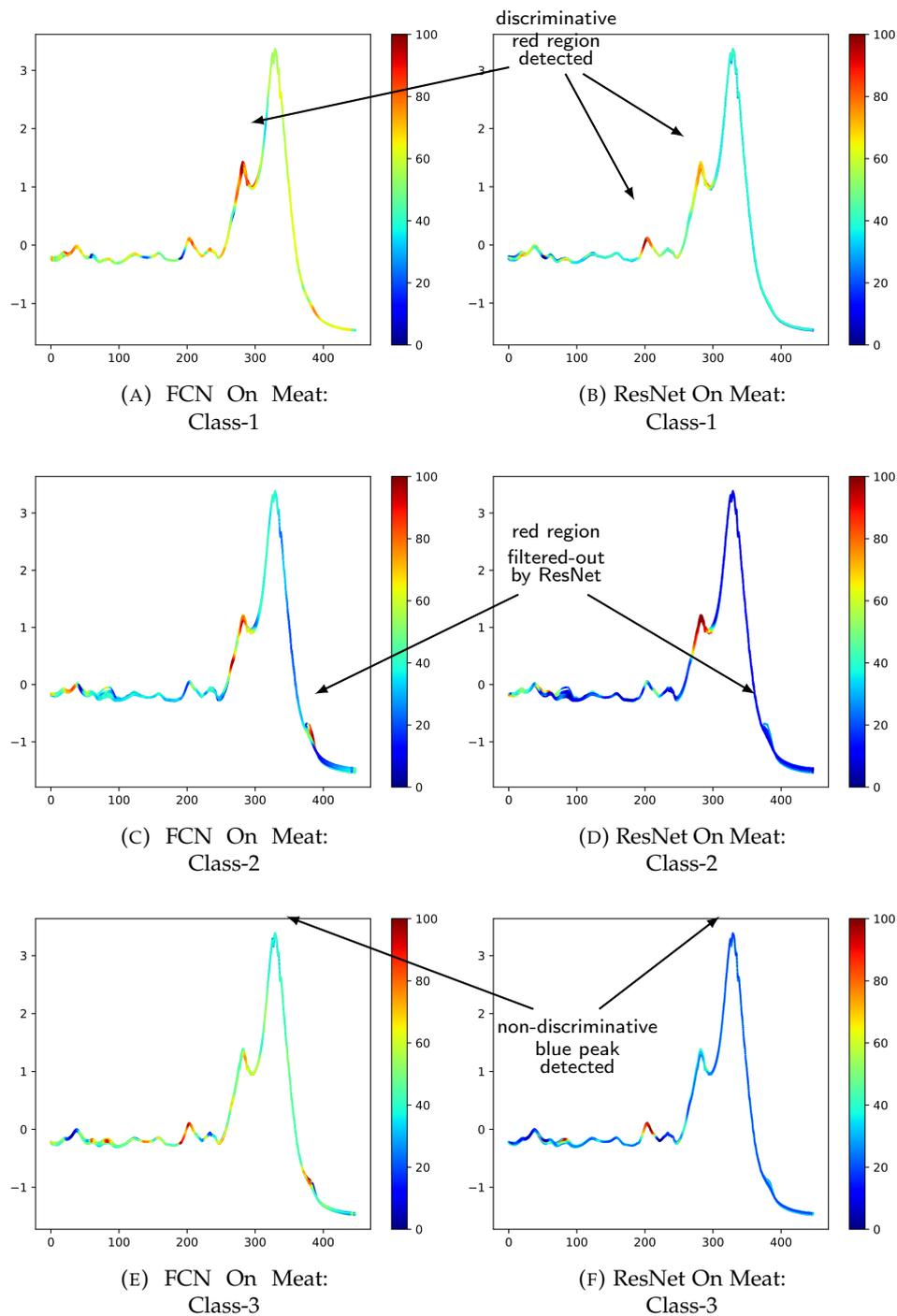 

Although the previous case study on GunPoint yielded interesting results in terms of showing that both models are localizing meaningful features, it failed to show the difference between the two most accurate deep learning classifiers: ResNet and FCN.
Therefore we decided to further analyze the CAM's result for the two models on the Meat dataset.

Meat is a food spectrograph dataset which are usually used in chemometrics to classify food types, a task that has obvious applications in food safety and quality assurance.
There are three classes in this dataset: Chicken, Pork and Turkey corresponding respectively to classes 1, 2 and 3 in Figure~\ref{fig-cam-meat}.  
\cite{alJowder1997mid} described how the data is acquired from 60 independent samples using Fourier transform infrared spectroscopy with attenuated total reflectance sampling. 

Similarly to GunPoint, this dataset is easy to visualize and does not contain very noisy time series. 
In addition, with only three classes, the visualization is possible to understand and analyze.
Finally, unlike for the GunPoint dataset, the two approaches ResNet and FCN reached significantly different results on Meat with respectively $97\%$ and $83\%$ accuracy.

Figure~\ref{fig-cam-meat} enables the comparison between FCN's CAM (left) and ResNet's  CAM (right).  
We first observe that ResNet is much more firm when it comes to highlighting the regions. 
In other words, FCN's CAM contains much more smoother regions with cyan, green and yellow regions, whereas ResNet's CAM contains more dark red and blue subsequences showing that ResNet can filter out non-discriminative and discriminative regions with a higher confidence than FCN, which probably explains why FCN is less accurate than ResNet on this dataset. 
Another interesting observation is related to the red subsequence highlighted by FCN's CAM for class 2 and 3 at the bottom right of Figure~\ref{sub-fcn-cam-Meat-class-2} and~\ref{sub-fcn-cam-Meat-class-3}.
By visually investigating this part of the time series, we clearly see that it is a non-discriminative part since the time series of both classes exhibit this bump. 
This subsequence is therefore filtered-out by the ResNet model which can be seen by the blue color in the bottom right of Figure~\ref{sub-resnet-cam-Meat-class-2} and~\ref{sub-resnet-cam-Meat-class-3}. 
These results suggest that ResNet's superiority over FCN is mainly due to the former's  ability to filter-out non-distinctive regions of the time series. 
We attribute this ability to the main characteristic of ResNet which is composed of the residual connections between the convolutional blocks that enable the model to \emph{learn to skip} unnecessary convolutions by dint of its shortcut links ~\citep{he2016deep}. 

\subsection{Multi-Dimensional Scaling}
We propose the use of MDS~\citep{kruskal1978multidimensional} with the objective to gain some insights on the spatial distribution of the input time series belonging to different classes in the dataset.
MDS uses a pairwise distance matrix as input and aims at placing each object in a N-dimensional space such as the between-object distances are preserved as well as possible. 
Using the ED on a set of input time series belonging to the test set, it is then possible to create a similarity matrix and apply MDS to display the set into a two dimensional space. 
This straightforward approach supposes that the ED is able to strongly separate the raw time series, which is usually not the case evident by the low accuracy of the nearest neighbor when coupled with ED~\citep{bagnall2017the}.

On the other hand, we propose to apply this MDS method to visualize the set of time series with its latent representation learned by the network. 
Usually in a deep neural network, we have several hidden layers and one can find several latent representation of the dataset. 
But since we are aiming at visualizing the class specific latent space, we chose to use the last latent representation of a DNN (the one directly before the softmax classifier), which is known to be a class specific layer~\citep{yosinski2014how}.  
We decided to apply this method only on ResNet and FCN for two reasons: (1) when evaluated on the UCR/UEA archive they reached the highest ranks; (2) they both employ a GAP layer before the softmax layer making the number of latent features invariant to the time series length. 

To better explain this process, for each input time series, the last convolution (for ResNet and FCN) outputs a multivariate time series whose dimensions are equal to the number of filters (128) in the last convolution, then the GAP layer averages the latter 128-dimensional  multivariate time series over the time dimension resulting in a vector of 128 real values over which the ED is computed. 
As we worked with the ED, we used metric MDS~\citep{kruskal1978multidimensional} that minimizes a cost function called \emph{Stress} which is a residual sum of squares:
\begin{equation}
Stress_D(X_1,\ldots,X_N)=\Biggl(\frac{\sum_{i,j}\bigl(d_{ij}-\|x_i-x_j\|\bigr)^2}{\sum_{i,j}d_{ij}^2}\Biggr)^{1/2}
\end{equation}
where $d_{ij}$ is the ED between the GAP vectors of time series $X_i$ and $X_j$.
Obviously, one has to be careful about the interpretation of MDS output, as the data space is highly simplified (each time series $X_i$ is represented as a single data point $x_i$).

We should note that there exists many visualization algorithms that would provide some insights about the data. 
One of the most popular techniques is called t-SNE~\citep{maaten2008visualizing}. 
This method projects high-dimensional data by giving each data point (or in our case a time series or its representation) a location in a two or three dimensional map. 
It is an optimized derivative of Stochastic Neighbor Embedding~\citep{hinton2003stochastic}. 
Although t-SNE has been used by many researchers, we decided to go with the MDS algorithm when analyzing time series representation. 
The reason behind this choice is that MDS is based on the pairwise distance matrix between the time series which is a very common way to cluster and analyze time series data~\citep{petitjean2016faster}, whereas t-SNE uses an iterative stochastic optimization process in order to preserve local structure at the expense of global structure in which we are interested. 

\begin{figure}
	\centering
	\subfloat[GunPoint-MDS-Raw]{
		\includegraphics[width=0.44\linewidth]{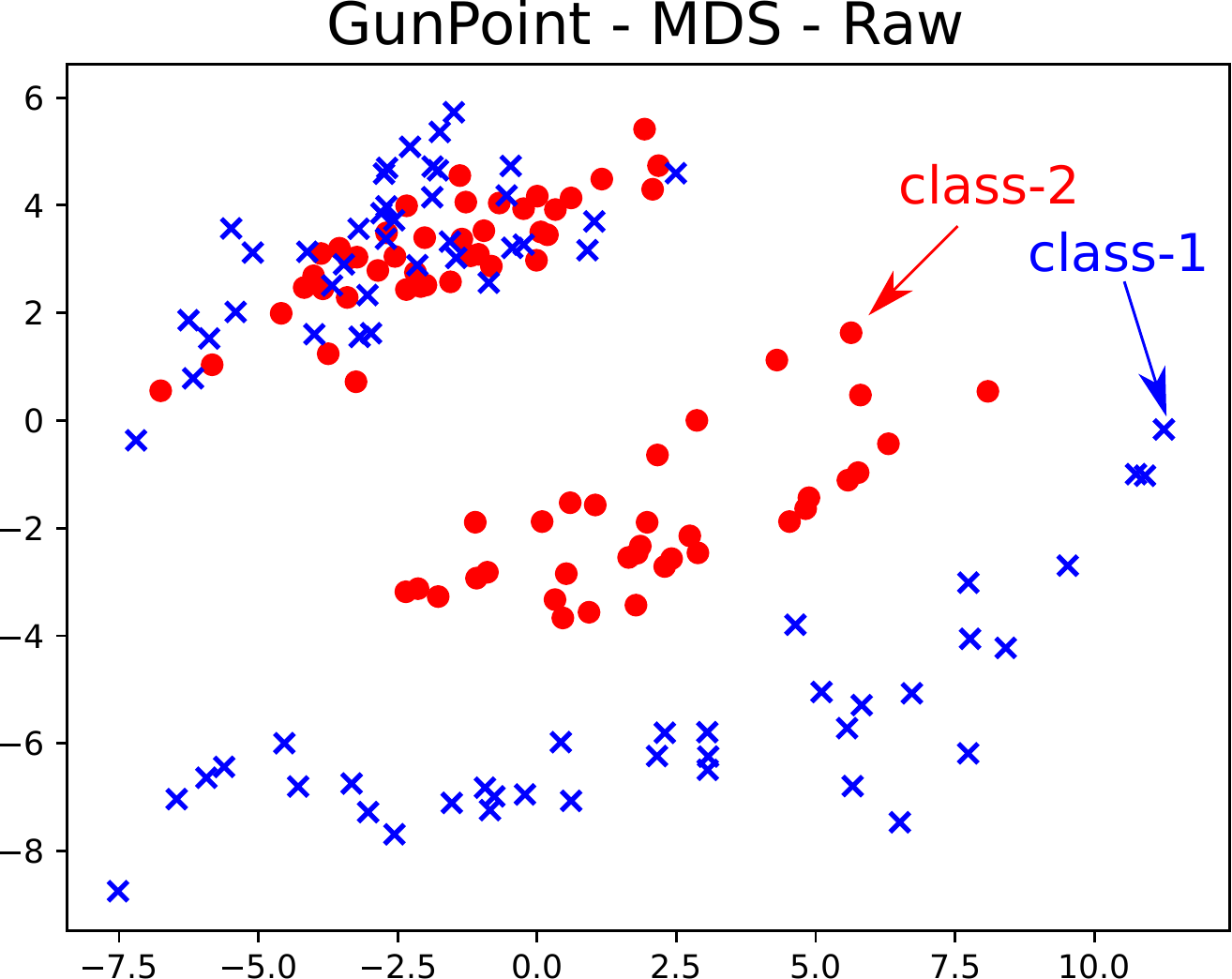}
		\label{sub-mds-gunpoint-raw}}\\
	\subfloat[GunPoint-MDS-GAP-FCN]{
		\includegraphics[width=0.44\linewidth]{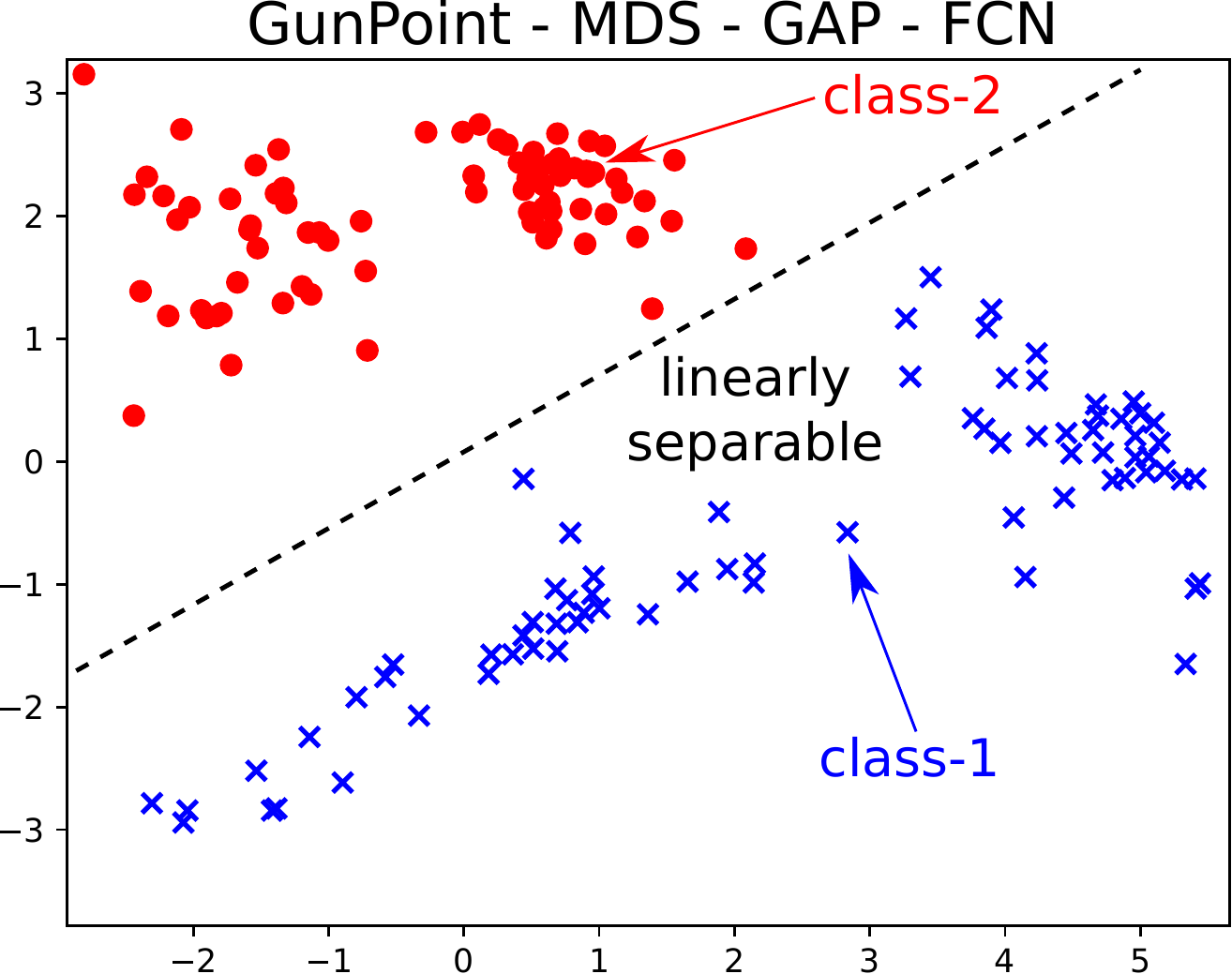}
		\label{sub-mds-gunpoint-gap-fcn}}
	\subfloat[GunPoint-MDS-GAP-ResNet]{
		\includegraphics[width=0.44\linewidth]{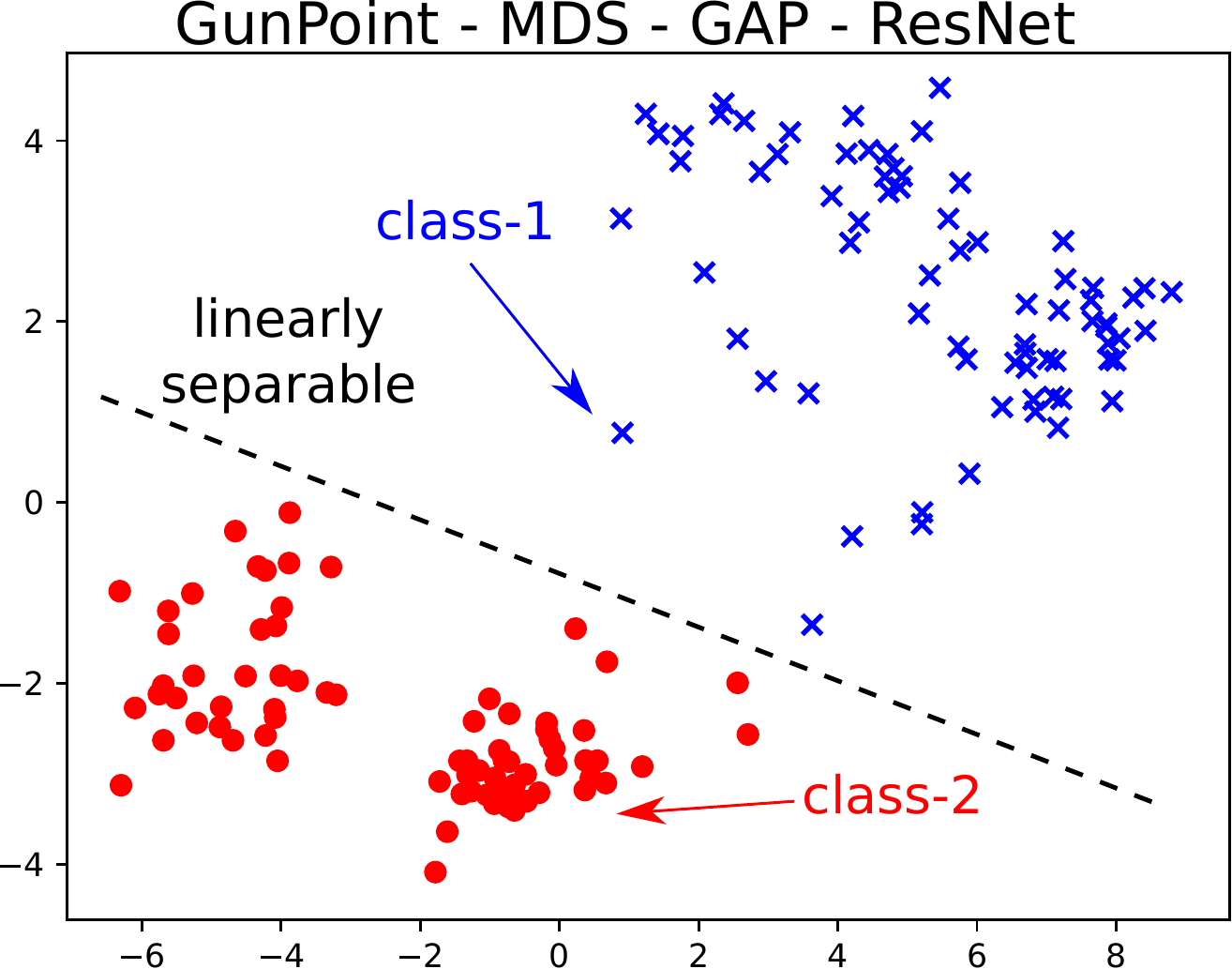}
		\label{sub-mds-gunpoint-gap-resnet}}
	\caption{Multi-Dimensional Scaling (MDS) applied on GunPoint for: (top) the raw input time series; (bottom) the learned features from the Global Average Pooling (GAP) layer for FCN (left) and ResNet (right) - (best viewed in color).
		This figure shows how the ResNet and FCN are projecting the time series from a non-linearly separable 2D space (when using the raw input), into a linearly separable 2D space (when using the latent representation).}
	\label{fig-mds-gunpoint}
\end{figure} 

Figure~\ref{fig-mds-gunpoint} shows three MDS plots for the GunPoint dataset using: (1) the raw input time series (Figure~\ref{sub-mds-gunpoint-raw}); (2) the learned latent features from the GAP layer for FCN (Figure~\ref{sub-mds-gunpoint-gap-fcn}); and (3) the learned latent features from the GAP layer for ResNet (Figure~\ref{sub-mds-gunpoint-gap-resnet}). 
We can easily observe in Figure~\ref{sub-mds-gunpoint-raw} that when using the raw input data and projecting it into a 2D space, the two classes are not linearly separable.  
On the other hand, in both Figures~\ref{sub-mds-gunpoint-gap-fcn} and~\ref{sub-mds-gunpoint-gap-resnet}, by applying MDS on the latent representation learned by the network, one can easily separate the set of time series belonging to the two classes. 
We note that both deep learning models (FCN and ResNet) managed to project the data from GunPoint into a linearly separable space which explains why both models performed equally very well on this dataset with almost 100\% accuracy. 

\begin{figure}
	\centering
	\subfloat[Wine-MDS-Raw]{
		\includegraphics[width=0.44\linewidth]{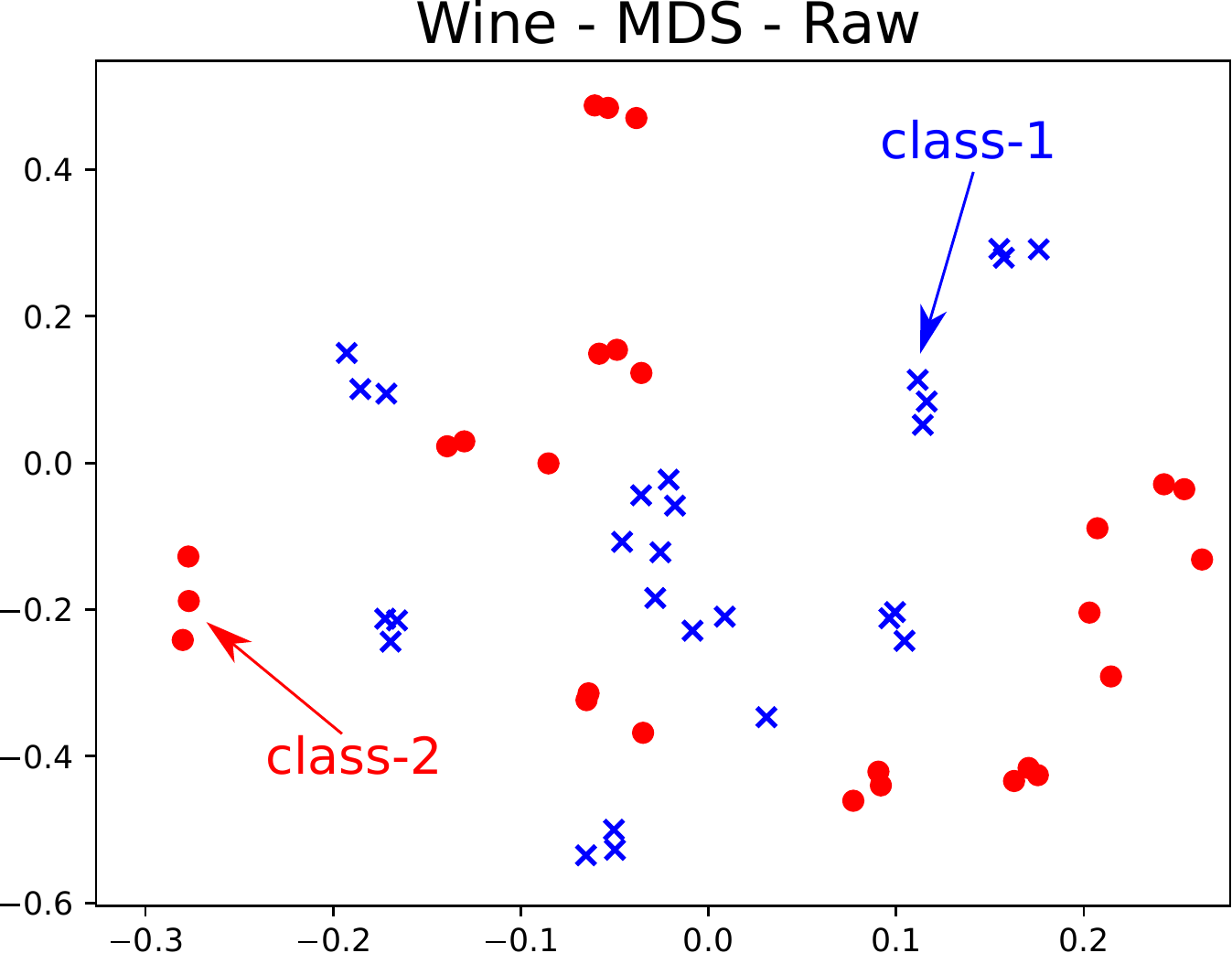}
		\label{sub-mds-wine-raw}}\\
	\subfloat[Wine-MDS-GAP-FCN]{
		\includegraphics[width=0.44\linewidth]{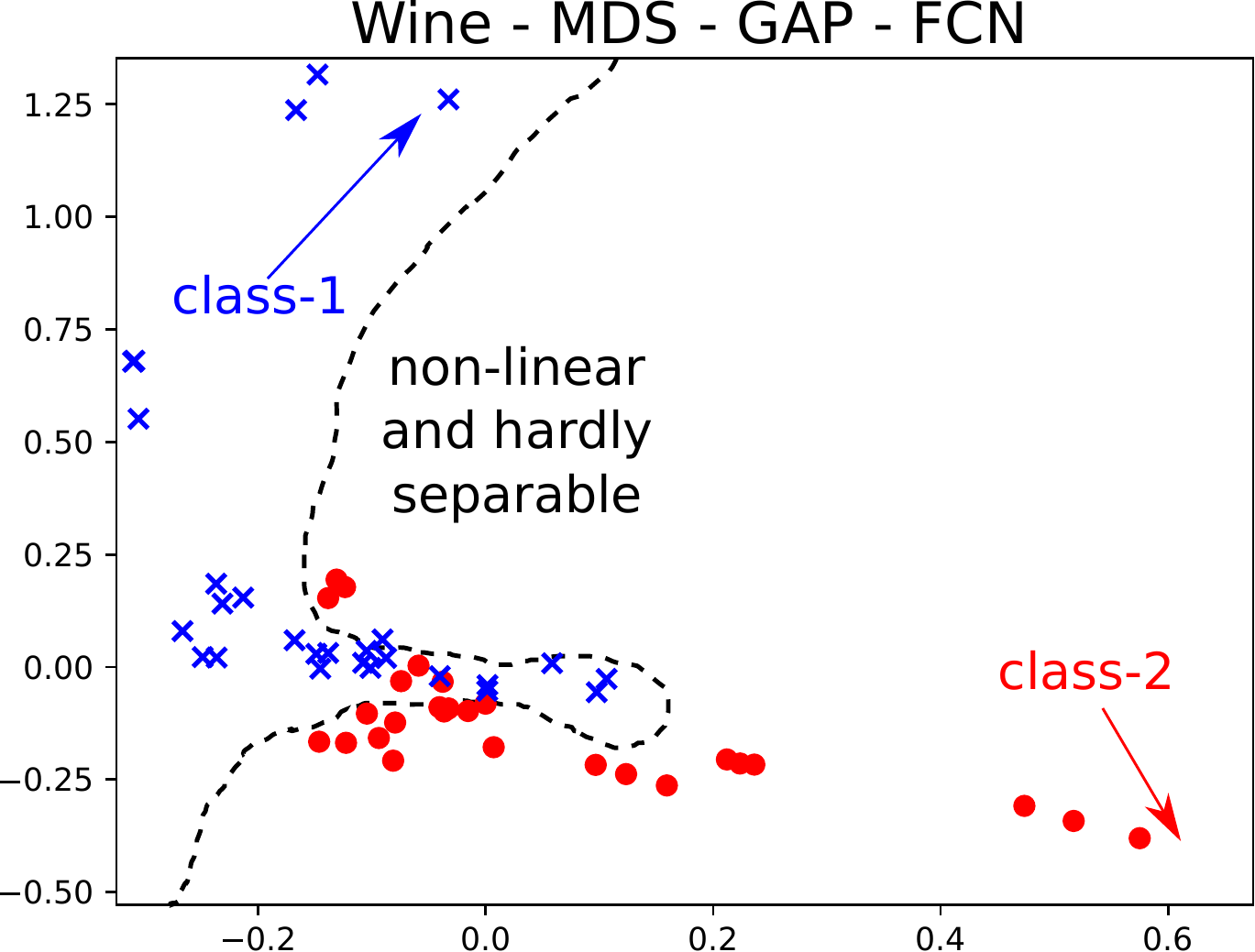}
		\label{sub-mds-wine-gap-fcn}}
	\subfloat[Wine-MDS-GAP-ResNet]{
		\includegraphics[width=0.417\linewidth]{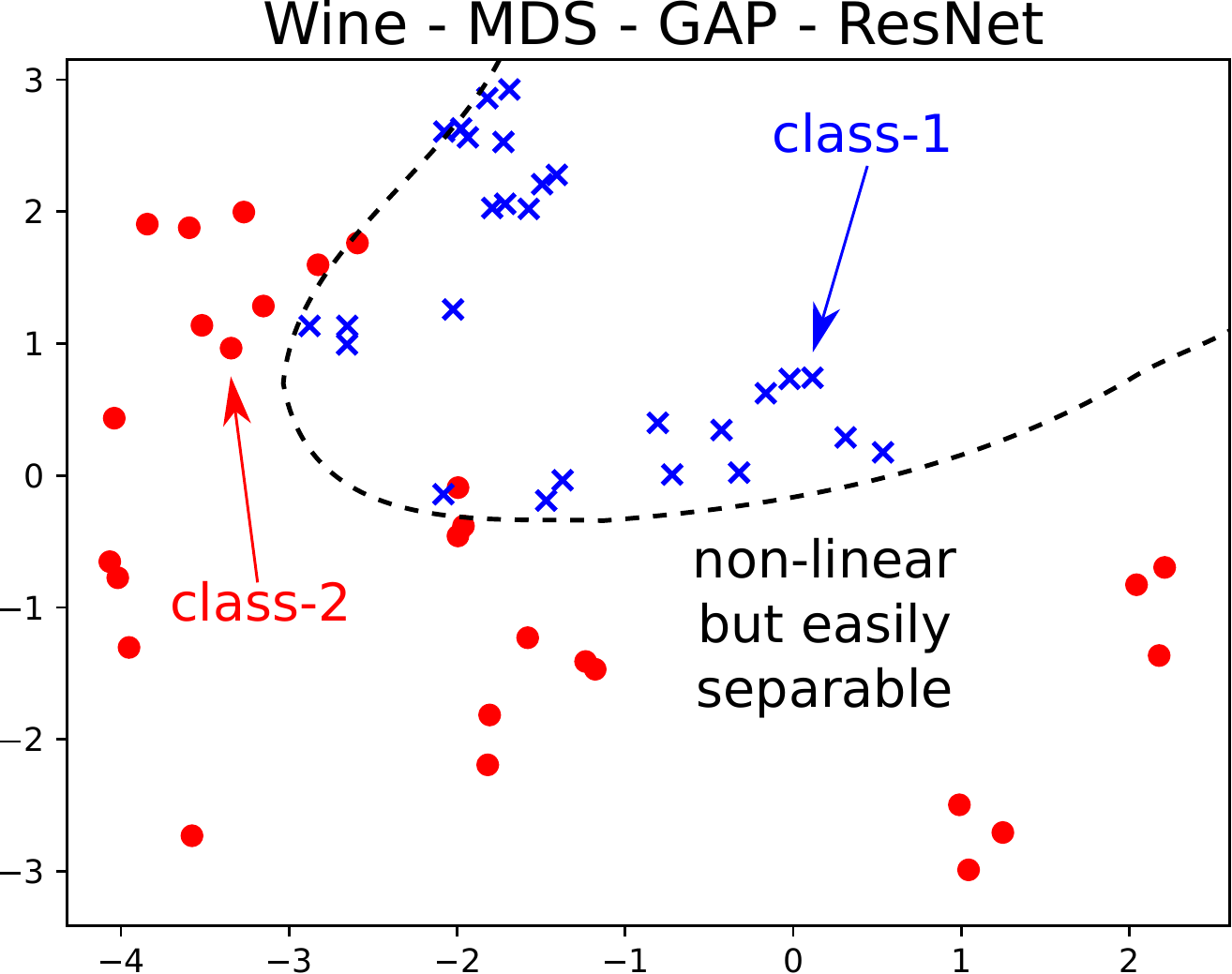}
		\label{sub-mds-wine-gap-resnet}}
	\caption{Multi-Dimensional Scaling (MDS) applied on Wine for: (top) the raw input time series; (bottom) the learned features from the Global Average Pooling (GAP) layer for FCN (left) and ResNet (right) - (best viewed in color).
		This figure shows how ResNet, unlike FCN, is able to project the data into an easily separable space when using the learned features from the GAP layer (Color figure online).}
	\label{fig-mds-wine}
\end{figure} 

Although the visualization of MDS on GunPoint yielded some interesting results, it failed to pinpoint the difference between the two deep learning models FCN and ResNet.
Therefore we decided to analyze another dataset where the accuracy of both models differed by almost 15\%. 
Figure~\ref{fig-mds-wine} shows three MDS plots for the Wine dataset using: (1) the raw input time series (Figure~\ref{sub-mds-wine-raw}); (2) the learned latent features from the GAP layer for FCN (Figure~\ref{sub-mds-wine-gap-fcn}); and (3) the learned latent features from the GAP layer for ResNet (Figure~\ref{sub-mds-wine-gap-resnet}). 
At first glimpse of Figure~\ref{fig-mds-wine}, the reader can conclude that all projections, even when using the learned representation, are not linearly separable which is evident by the relatively low accuracy of both models FCN and ResNet which is equal respectively to 58.7\% and 74.4\%. 
A thorough observation shows us that the learned hidden representation of ResNet (Figure~\ref{sub-mds-wine-gap-resnet}) separates the data from both classes in a much clearer way than the FCN (Figure~\ref{sub-mds-wine-gap-fcn}). 
In other words, FCN's learned representation has too many data points close to the decision boundary whereas ResNet's hidden features enables projecting data points further away from the decision boundary.   
This observation could explain why ResNet achieves a better performance than FCN on the Wine dataset.

\section{Conclusion} \label{ch-1-sec-conclusion}

In this chapter, we presented the largest empirical study of DNNs for TSC.
We described the most recent successful deep learning approaches for TSC in many different domains such as human activity recognition and sleep stage identification. 
Under a unified taxonomy, we explained how DNNs are separated into two main categories of generative and discriminative models.
We re-implemented nine recently published end-to-end deep learning classifiers in a unique framework which we make publicly available to the community. 
Our results show that end-to-end deep learning can achieve the current state-of-the-art performance for TSC with architectures such as Fully Convolutional Neural Networks and deep Residual Networks. 
Finally, we showed how the black-box effect of deep models which renders them uninterpretable, can be mitigated with a Class Activation Map visualization that highlights which parts of the input time series, contributed the most to a certain class identification. 

Although we have conducted an extensive experimental evaluation, deep learning for time series classification, unlike for computer vision and NLP tasks, still lacks a thorough study of data augmentation~\citep{IsmailFawaz2018data,forestier2017generating} and transfer learning~\citep{IsmailFawaz2018transfer,serra2018towards}.
In addition, the time series community would benefit from an extension of this empirical study that compares in addition to accuracy, the training and testing time of these deep learning models. 
Furthermore, we think that the effect of z-normalization (and other normalization methods) on the learning capabilities of DNNs should also be thoroughly explored. 
In our future work, we aim to investigate and answer the aforementioned limitations by conducting more extensive experiments especially on multivariate time series datasets.
In order to achieve all of these goals, one important challenge for the TSC community is to provide one large generic \emph{labeled} dataset similar to the large images database in computer vision such as ImageNet~\citep{russakovsky2015imagenet} that contains 1000 classes.

In conclusion, with data mining repositories becoming more frequent, leveraging deeper architectures that can learn automatically from annotated data in an end-to-end fashion, makes deep learning a very enticing approach.
In this chapter, we demonstrated the potential of deep neural networks for TSC, nevertheless these complex machine learning models can still benefit from many regularization techniques, which is the main focus of the following chapter. 


\chapter{Regularizing deep neural networks} \label{Chapter2}

\section{Introduction}

Deep learning models usually have significantly more trainable parameters than the number of training instances. 
Nevertheless, in the previous chapter, we showed how these artificial neural networks are able to achieve good generalization capabilities compared to traditional TSC algorithms. 
Yet most of these models require some sort of regularization in order to ensure a small generalization error (i.e. a small difference between the training and testing error).
In this chapter, we present four main techniques for regularizing DNNs for TSC: (1) transfer learning~\citep{IsmailFawaz2018transfer}; (2) ensembling~\citep{IsmailFawaz2019deep}; (3) data augmentation~\citep{IsmailFawaz2018data}; and adversarial training~\citep{IsmailFawaz2019adversarial};
Our work reveals that these techniques improve the accuracy of the previously described deep learning models such as FCN and ResNet. 

The chapter is divided into four main sections, each one describing in details the regularization technique. 
First we start by describing how transfer learning can further improve the performance of DNNs for the TSC task. 
We then showcase our findings regarding how ensembling deep learning models can significantly improve the performance of these time series classifiers. 
In the following section, we present how adopting a data augmentation method can help in improving the performance of neural networks for TSC. 
Finally, we define and investigate how DNNs are vulnerable to adversarial examples, and explore the possibility of leveraging those examples to regularize deep learning time series classifiers.

\section{Transfer learning}
\begin{figure}
	\centering
	\includegraphics[width=0.7\linewidth]{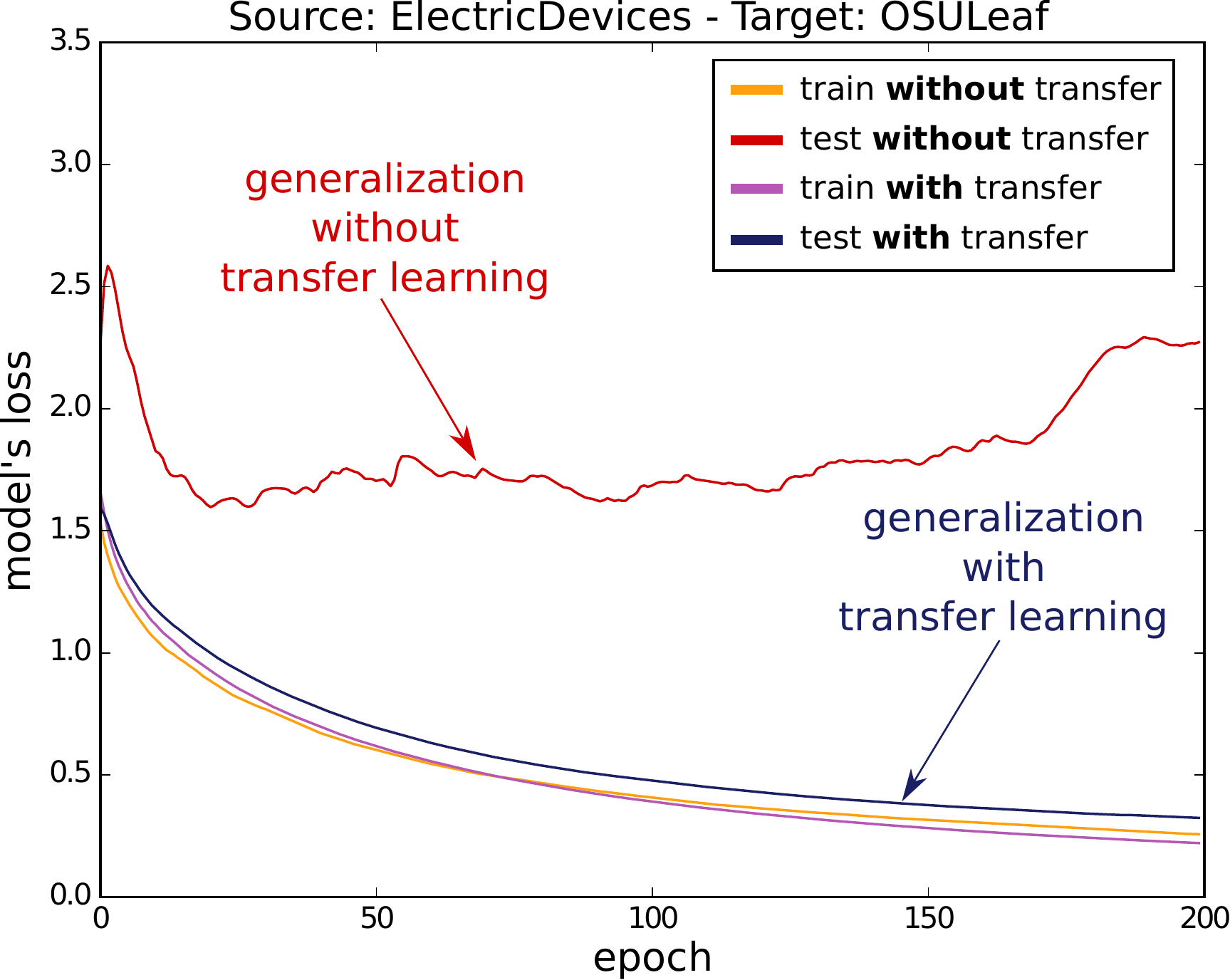}
	\caption{Evolution of model's loss (train and test) with and without the transfer learning method using ElectricDevices as source and OSULeaf as target datasets. 
		(Best viewed in color).}
	\label{fig-elect-shapelets}
\end{figure}

CNNs have recently been shown to significantly outperform the nearest neighbor approach coupled with the DTW algorithm (NN-DTW) on the UCR/UEA archive benchmark~\citep{ucrarchive} for the TSC problem~\citep{wang2017time}. 
CNNs were not only able to beat the NN-DTW baseline, but in the previous chapter, we showed how they were also able to reach results that are not significantly different than COTE~\citep{bagnal2015time} - which is an ensemble of 35 classifiers.
However, despite the high performance of these CNNs, deep learning models are still prone to overfitting. 
One example where these neural networks fail to generalize is when the training set of the time series dataset is very small.
We attribute this huge difference in accuracy to the overfitting phenomena, which is still an open area of research in the deep learning community~\citep{zhang2017understanding}.  
This problem is known to be mitigated using several regularization techniques such as transfer learning~\citep{yosinski2014transferable}, where a model trained on a source task is then fine-tuned on a target dataset. 
For example in \figurename~\ref{fig-elect-shapelets}, we trained a model on the ElectricDevices dataset~\citep{ucrarchive} and then fine-tuned this same model on the OSULeaf dataset~\citep{ucrarchive}, which significantly improved the network's generalization capability.

Transfer learning is currently used in almost every deep learning model when the target dataset does not contain enough labeled data~\citep{yosinski2014transferable}.
Despite its recent success in computer vision~\citep{csurka2017domain}, transfer learning has been rarely applied to deep learning models for time series data.
One of the reasons for this absence is probably the lack of one big general purpose dataset similar to ImageNet~\citep{russakovsky2015imagenet} or OpenImages \citep{openimages} but for time series.
Furthermore, it is only recently that deep learning was proven to work well for TSC~\citep{cui2016multi} and there is still much to be explored in building deep neural networks for mining time series data~\citep{gamboa2017deep}.

Since transferring deep learning models, between the UCR/UEA archive datasets~\citep{ucrarchive}, have not been thoroughly studied, we decided to investigate this area of research with the ultimate goal to determine in advance which dataset transfers could benefit the CNNs and improve their TSC accuracy. 

The intuition behind the transfer learning approach for time series data is also partially inspired by the observation of~\cite{cui2016multi}, where the authors showed that shapelets~\citep{ye2009time} (or subsequences) learned by the learning shapelets approach~\citep{grabocka2014learning} are related to the filters (or kernels) learned by the CNNs.
We hypothesize that these learned subsequences might not be specific to one dataset and could occur in other unseen datasets with un/related classification tasks.
Another observation for why transfer learning should work for time series data is its recent success in computer vision tasks~\citep{csurka2017domain}. 
Indeed, since time series data contain one temporal dimension (time) compared to two dimensions for images (width and height), it is only natural to think that if filters can successfully be transferred on images~\citep{yosinski2014transferable}, they should also be transferable across time series datasets. 

\begin{figure}
	\centering
	\includegraphics[width=\linewidth]{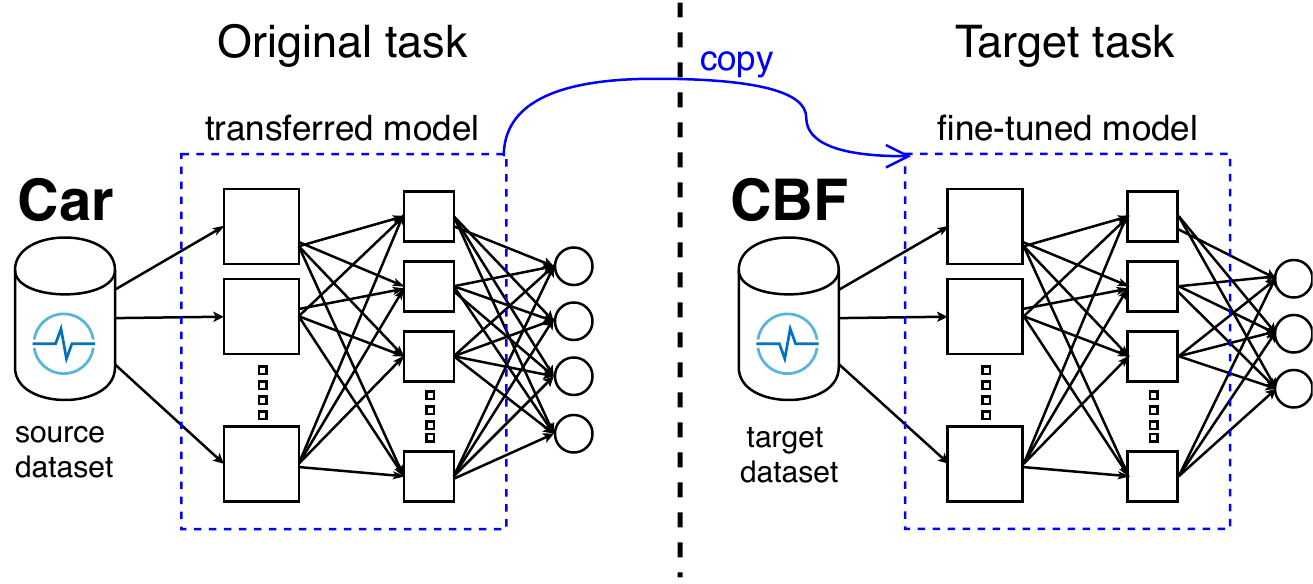}
	\caption{General deep learning training process with transfer learning for time series classification.
		In this example, a model is first pre-trained on Car (source dataset) and then the corresponding weights are fine-tuned on CBF (target dataset).}
	\label{fig-transfer_archi}
\end{figure}

To evaluate the potential of transfer learning for TSC, we performed experiments where each pair of datasets in the UCR/UEA archive was tested twice: we pre-trained a model for each dataset, then transferred and fine-tuned it on all the other datasets (a total of more than $7140$ trained models).
\figurename~\ref{fig-transfer_archi} illustrates the architecture of our proposed framework of transfer learning for TSC on two datasets. 
The obtained results show that time series do exhibit some low level features that could be used in a transfer learning approach. 
They also show that using transfer learning reduces the training time by reducing the number of epochs needed for the network to converge on the train set. 

Motivated by the consensus that transferring models between similar datasets improves the classifier's accuracy~\citep{weiss2016a}, we used the DTW algorithm as an inter-datasets similarity measure in order to quantify the relationship between the source and target datasets in our transfer learning framework.
Our experiments show that DTW can be used to predict the best source dataset for a given target dataset. 
Our method can thus identify which datasets should be considered for transfer learning given a new TSC problem.


\subsection{Background and related work}

Before getting into the details of the recent applications for transfer learning, we give a formal definition of the latter~\citep{weiss2016a}. 
\begin{definition}
	Transfer learning for deep neural networks, is the process of first training a base network on a source dataset and task, and then transfer the learned features (the network's weights) to a second network to be trained on a target dataset and task. 
\end{definition}
Throughout this section, we will refer to \emph{source} dataset as the dataset we are transferring the pre-trained model \emph{from}, and to \emph{target} dataset as the dataset we are transferring the pre-trained model \emph{to}.

Now that we have established the necessary definition, we will dive into the recent applications of transfer learning for time series data mining tasks. 
In fact, transfer learning is sometimes confused with the domain adaptation approach~\citep{pan2010a,long2015learning}. 
The main difference with the latter method is that the model is jointly trained on the source and target datasets~\citep{weiss2016a}.
The goal of using the target instances during training, is to minimize the discrepancy between the source's and target's instances.
In~\cite{ariefang2017da}, a domain adaptation approach was proposed to predict human indoor occupancy based on the carbon dioxide concentration in the room.
In~\cite{kasteren08recognizing}, hidden Markov models' generative capabilities were used in a domain adaptation approach to recognize human activities based on a sensor network.   

For time series anomaly detection, a transfer learning approach was used to determine which time series should be transferred from the source to the target dataset to be used with an NN-DTW classifier~\citep{vercruyssen2017transfer}.
Similarly,~\cite{spiegel2016} developed a method to transfer specific training examples from the source dataset to the target dataset and hence compute the dissimilarity matrix using the new training set. 
As for time series forecasting, a transfer learning approach for an auto-encoder was employed to predict the wind-speed in a farm~\citep{hu2016transfer}. 
The authors proposed first to train a model on the historical wind-speed data of an old farm and fine-tune it using the data of a new farm. 
In~\cite{banerjee2017a} restricted Boltzmann machines were first pre-trained for acoustic phoneme recognition and then fine-tuned for post-traumatic stress disorder diagnosis.    

Perhaps the recent work in~\cite{serra2018towards} is the closest to ours in terms of using transfer learning to improve the accuracy of deep neural networks for TSC. 
In this work, the authors designed a CNN with an attention mechanism to encode the time series in a supervised manner. 
Before fine-tuning a model on a target dataset, the model is first jointly pre-trained on several source datasets with themes~\citep{bagnall2017the} that are different from the target dataset's theme which limits the choice of the source dataset to only one. 
Additionally, unlike~\cite{serra2018towards}, we take a pre-designed deep learning model without modifying it nor adding regularizers.   
This enabled us to solely attribute the improvement in accuracy to the transfer learning feature, which we describe in details in the following section.  

\subsection{Method} \label{sec-method-transfer-learning}

In this subsection, we present our proposed method of transfer learning for TSC. 
The adopted neural network architecture was FCN, for the simple reason of being three times faster than ResNet while being ranked the second most accurate model in the previous chapter.  
We then thoroughly explain how we adapted the network for the transfer learning process. 
Finally, we present our DTW based method that enabled us to compute the inter-datasets similarities, which we later use to guide the transfer learning process.

\subsubsection{Network adaptation}

After training FCN on the 85 datasets in the archive, we obtain 85 different neural networks.
The only difference between these 85 neural network architectures lies in the output layer.
The rest of the layers have the same number of parameters but with different values. 
In fact the last layer, which is a softmax classifier, depends on the number of classes in the dataset.

Thus, given a source dataset $D_s$ and a target dataset $D_t$, we first train the network on $D_s$. 
We then remove the last layer and replace it with another softmax layer whose number of neurons is equal to the number of classes in the target dataset $D_t$.  
The added softmax layer's parameters are initialized randomly using Glorot's uniform initialization~\citep{glorot2010understanding}.
This new network is then re-trained (fine-tuned) on $D_t$.

We chose to fine-tune the whole network instead of training only the last newly added output layer. 
We tried to limit back-propagating the gradient to the last layer, but found that the network failed to converge. 
This is in compliance with the transfer learning literature~\citep{yosinski2014transferable}, where re-training the whole network almost always leads to better results. 

Finally, we should add that one of the advantages of using a global average pooling layer is that we do not need to re-scale the input time series when transferring models between time series of different length. 
%

\subsubsection{Inter-datasets similarity}

One of the main challenges with transfer learning is choosing the source dataset. 
In~\cite{pan2011transfer}, it was demonstrated that a learning algorithm trained with a certain source domain will not yield an optimal performance if the marginal distributions of the datasets' input are different.  
In our case, the total number of datasets in the UCR/UEA archive is 85. 
Therefore for each target dataset in the archive, we have 84 potential source datasets. 
This makes the trial and error based approach for transfer learning very costly in terms of computational resources.
Hence, we propose to use the DTW distance to compute the similarities between the datasets, thus guiding the choice of a source dataset for a given target dataset.

Note that it is practically impossible to directly estimate the performance of a model learned on a source dataset by applying it on a target dataset's train set since the last layer of the network is \emph{specific}~\citep{yosinski2014transferable} to the classes of the source dataset.

Before describing in details our method for computing inter-datasets similarity, we start by explaining the DTW distance. 

DTW was first proposed for speech recognition when aligning two audio signals~\citep{sakoe1978dynamic}. 
Suppose we want to compute the dissimilarity between two time series, $X_1=[x_{1,1},x_{1,2},\dots,x_{1,m}]$ and $X_2=[x_{2,1},x_{2,2},\dots,x_{2,n}]$.
The length of $X_1$ and $X_2$ are denoted respectively by $m$ and $n$.  

Let $M(X_1,X_2)$ be the $m\times n$ point-wise dissimilarity matrix between $X_1$ and $X_2$, where $M_{i,j}=||x_{1,i}-x_{2,j}||^2$. 
A warping path $P=((c_1,d_1),(c_2,d_2),\dots,(c_s,d_s))$ is a series of points that define a crossing of $M$. 
The warping path must satisfy three conditions: (1) $(c_1,d_1)=(1,1)$; (2) $(c_s,d_s)=(m,n)$; (3) $0\le c_{i+1}-c_i \le 1$ and $0\le d_{j+1}-d_j \le 1$ for all $i<m$ and $j<n$. 
The DTW measure between two series corresponds to the path through $M$ that minimizes the total distance. 
In fact, the distance for any path $P$ is equal to $D_P(A,B)=\sum_{i=1}^{s}P_i$.
Hence if $\textbf{P}$ is the space of all possible paths, the optimal one - whose cost is equal to $DTW(A,B)$ - is denoted by $P^*$ and can be computed using: $\min_{P\in \textbf{P}}D_P(A,B)$. 
The optimal warping path can be obtained efficiently by applying a dynamic programming technique to fill the cost matrix $M$.

Now that we have explained in details the DTW algorithm, which is usually used for computing a distance between two time series, we will describe the DBA algorithm which was first proposed by~\cite{petitjean2011a} in order to average in the DTW induced space (as opposed to the arithmetic mean in the Euclidean space).

DBA is an averaging method which consists in iteratively refining an initial series in order to minimize its squared DTW distance to the set of series we want to average. 
Technically for each refinement (or iteration), DBA consists of two main steps: 
\begin{enumerate}
	\item Computing the DTW between each individual series and the current temporary average that we want to refine. 
	This is in order to find optimal associations between the elements of the average series and elements of the set of series to be averaged. 
	\item Now that we have these associations, we update each element of the average series as the barycenter of the elements associated to it during the previous step.  
\end{enumerate}

In order to compute the similarities between the datasets, we first reduce the number of time series for each dataset to one time series (or prototype) per class.
The per class prototype is computed by averaging the set of time series in the corresponding class, using the previously described DBA algorithm as a data reduction step. 
The latter summarizing function was proposed and validated as an averaging method in the DTW induced space. 
In addition, DBA has been recently used as a data reduction technique where it was evaluated in a nearest centroid classification schema~\citep{petitjean2014dynamic}.
Therefore, to generate the similarity matrix between the UCR datasets, we computed a distance between each pair of datasets.
Finally, for simplicity and since the main goal of this project is not the inter-datasets similarity,  we chose the distance between two datasets to be equal to the minimum distance between the prototypes of their corresponding classes. 
Note that other averaging methods such as soft-DTW~\citep{cutur2017soft} and TEKA~\citep{marteau2019times} could be used instead of DBA in our framework, but we leave such exploration for our future work. 

Algorithm~\ref{algo-sim} shows the different steps followed to compute the distance matrix between the UCR datasets.
The first part of the algorithm (lines 1 through 7) presents the data reduction technique similar to~\cite{petitjean2014dynamic}.
For the latter step, we first go through the classes of each dataset (lines 1, 2 and 3) and then average the set of time series for each class. 
Following the recommendations in~\cite{petitjean2014dynamic}, the averaging method (DBA) was initialized to be equal to the medoid of the time series selected set (line 4). 
We fixed the number of iterations for the DBA algorithm to be equal to 10, for which the averaging method has been shown to converge~\citep{petitjean2011a}.

After having reduced the different sets for each time series dataset, we proceed to the actual distance computation step (lines 8 through 22).
From line 8 to 10, we loop through every possible combination of datasets pairs. 
Lines 13 and 14 show the loop through each class for each dataset (at this stage each class is represented by one average time series thanks to the data reduction steps).
Finally, lines 15 through 19 set the distance between two datasets to be equal to the minimum DTW distance between their corresponding classes. 

One final note is that when computing the similarity between the datasets, the only time series data we used came from the training set, thus eliminating any bias due to having seen the test set's distribution. 

\begin{algorithm}
	\begin{algorithmic}[1]
		\renewcommand{\algorithmicrequire}{\textbf{Input:}}
		\renewcommand{\algorithmicensure}{\textbf{Output:}}
		\Require $N$ time series datasets in an array $D$
		\Ensure  $N\times N$ datasets similarity matrix
		\\ \textit{Initialization} : matrix $M$ of size $N\times N$ 
		\\ \textit{data reduction step}
		\For {$i = 1 $ to $N$}		\State $C=D[i].classes$
		\For {$c=1$ to $ length(C)$}
		\State $avg\_init = medoid(C[c])$
		\State $C[c] = DBA(C[c],avg\_init)$ 
		\EndFor
		\EndFor
		\\ \textit{distance calculation step}
		\For {$i = 1 $ to $N$}
		\State $C_i=D[i].classes$
		\For {$j=1$ to $N$}
		\State $C_j=D[j].classes$
		\State $dist = \infty$
		\For {$c_i=1$ to $ length(C_i)$}
		\For{$c_j=1$ to $ length(C_j)$}
		\State $cdist = DTW(C_i[c_i], C_j[c_j])$
		\State $dist = minimum(dist, cdist)$ 
		\EndFor
		\EndFor
		\State $M[i,j] = dist$
		\EndFor
		\EndFor
		\Return $M$ 
	\end{algorithmic}
	\caption{Inter-datasets similarity}
	\label{algo-sim}
\end{algorithm}

\subsection{Experimental setup}

\subsubsection{Datasets}

We evaluate our developed framework thoroughly on the largest publicly available benchmark for time series analysis: the UCR/UEA archive~\citep{ucrarchive}, which consists of 85 datasets selected from various real-world domains.
The time series in the archive are already z-normalized to have a mean equal to zero and a standard deviation equal to one. 
During the experiments, we used the default training and testing set splits provided by UCR. 
For pre-training a model, we used only the train set of the source dataset.
We also fine-tuned the pre-trained model solely on the target dataset's training data. 
Hence the test sets were only used for evaluation purposes.    

\subsection{Experiments}

For each pair of datasets ($D_1$ and $D_2$) in the UCR/UEA archive we need to perform two experiments:
\begin{itemize}
	\item $D_1$ is the source dataset and $D_2$ is the target dataset. 
	\item $D_1$ is the target dataset and $D_2$ is the source dataset. 
\end{itemize}
Which makes it in total 7140 experiments for the 85 dataset in the archive. 
Hence, given the huge number of models that need to be trained, we ran our experiments on a cluster of 60 GPUs. 
These GPUs were a mix of three types of Nvidia graphic cards: GTX 1080 Ti, Tesla K20, K40 and K80. 
The total sequential running time was approximately 168 days, that is if the computation has been done on a single GPU.  
But by leveraging the cluster of 60 GPUs, we managed to obtain the results in less than one week.
We implemented our framework using the open source deep learning library Keras~\citep{chollet2015keras} with the Tensorflow~\citep{tensorflow} back-end.
For reproducibility purposes, we provide the 7140 trained Keras models (in a HDF5 format) on the companion web page of this project\footnote{\url{http://germain-forestier.info/src/bigdata2018/}}. We have also published the raw results and the full source code of our method to enable the time series community to verify and build upon our findings\footnote{\url{https://github.com/hfawaz/bigdata18}}.

\begin{figure}
	\centering
	\includegraphics[width=\textwidth]{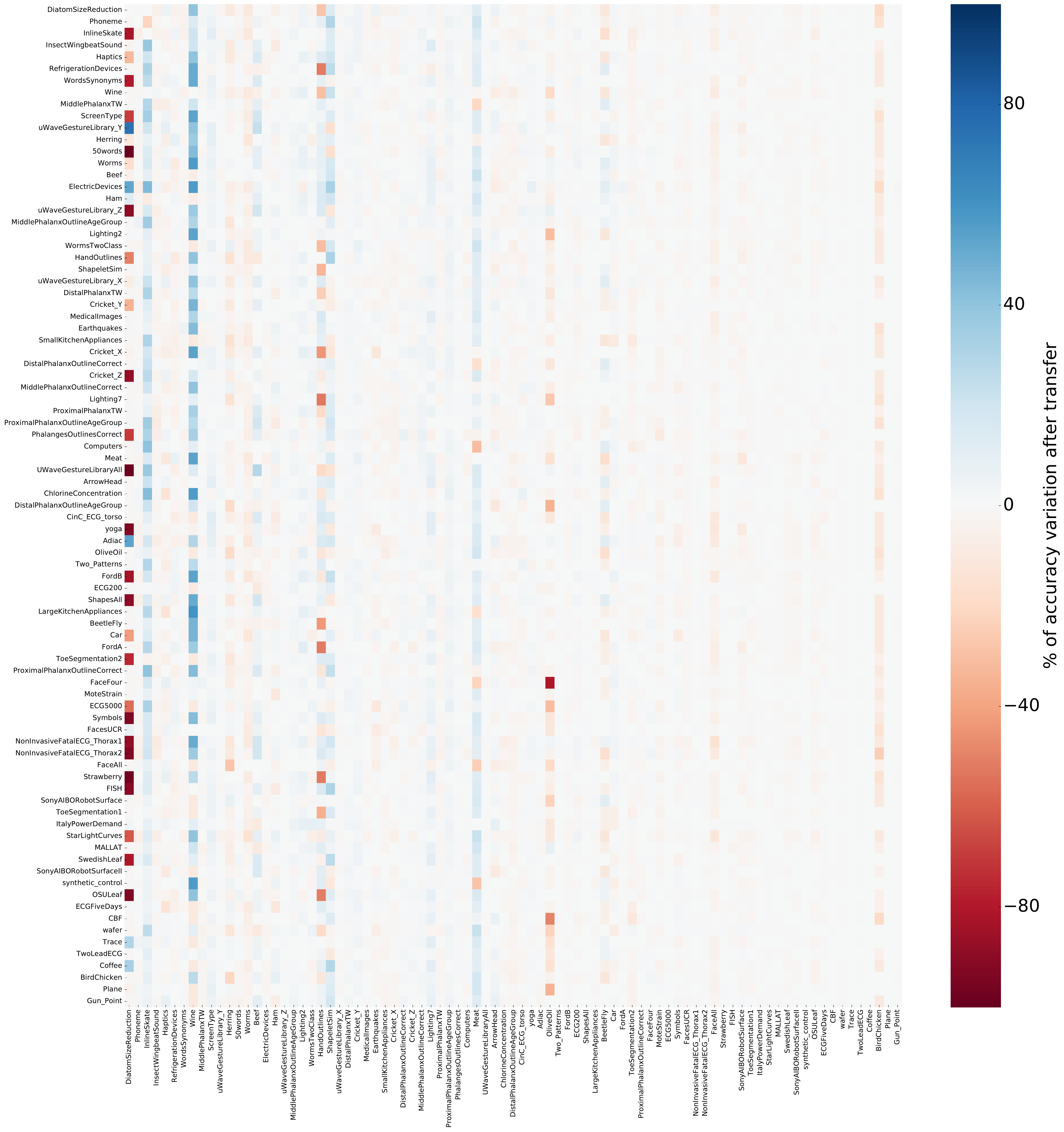}
	\caption{
		The variation in percentage over the original accuracy when fine tuning a pre-trained model.
		The rows' indexes correspond to the source datasets and the columns' indexes correspond to the target datasets.
		The \textbf{red} color shows the extreme case where the chosen pair of datasets (source and target) deteriorates the network's performance.
		Where on the other hand, the \textbf{blue} color identifies the improvement in accuracy when transferring the model from a certain source dataset and fine-tuning on another target dataset.
		The white color means that no change in accuracy has been identified when using the transfer learning method for two datasets. 
		The matrix actually has a size of $85\times85$ (instead of $85\times84$) for visual clarity with its diagonal left out of the analysis.
		(Best viewed in color).
	}
	\label{fig-heat_map}
\end{figure}

\subsection{Results}

The experiments described in the previous section yielded interesting yet hard-to-understand results. 
In this section, we first present the result of the $85\times84$ experiments in a form of a matrix (displayed as a heat map in \figurename~\ref{fig-heat_map}). 
We then empirically show how choosing the wrong source dataset for a given target dataset could decrease the network's performance. 
Therefore, we provide a DTW based solution to choose the best source dataset for a given target dataset. 
Finally, we detail a few interesting case studies where the behavior of the proposed method has a significant impact on the transfered model's accuracy. 

\subsubsection{Transfer learning accuracy variation matrix}

In order to have a fair comparison across the datasets, we illustrate the variation in the transferred model's accuracy based on the percentage of variation compared to the original accuracy (without transfer learning). 
For example, consider the original accuracy (equal to 74.6\%) when training the neural network from scratch on the target dataset HandOutlines. 
Then instead of training the model from scratch (with random initializations) we obtain a 86.5\% accuracy when initializing the network's weights to be equal to the weights of a pre-trained network on the source dataset MedicalImages.
Hence, the percentage of accuracy variation with respect to the original value is equal to $100\times(86.5-74.6)/74.6\approx+16\%$. 
Thus negative values (red in \figurename~\ref{fig-heat_map}) indicate a decrease in performance when using the transfer learning approach. 
Whereas, a positive percentage (blue in \figurename~\ref{fig-heat_map}) indicates an increase in performance when fine-tuning a pre-trained model. 

When observing the heat map in \figurename~\ref{fig-heat_map}, one can easily see that fine-tuning a pre-trained model almost never hurts the performance of the CNN.
This can be seen by the dominance of the white color in the heat map, which corresponds to almost no variation in accuracy. 

On the other hand, the results which we found interesting are the two extreme cases (red and blue) where the use of transfer learning led to high variations in accuracy.
Interestingly for a given target dataset, the choice of source dataset could deteriorate or improve the CNN's performance as we will see in the following subsection. 

\subsubsection{Naive transfer learning}

\begin{figure}
	\centering
	\includegraphics[width=0.7\linewidth]{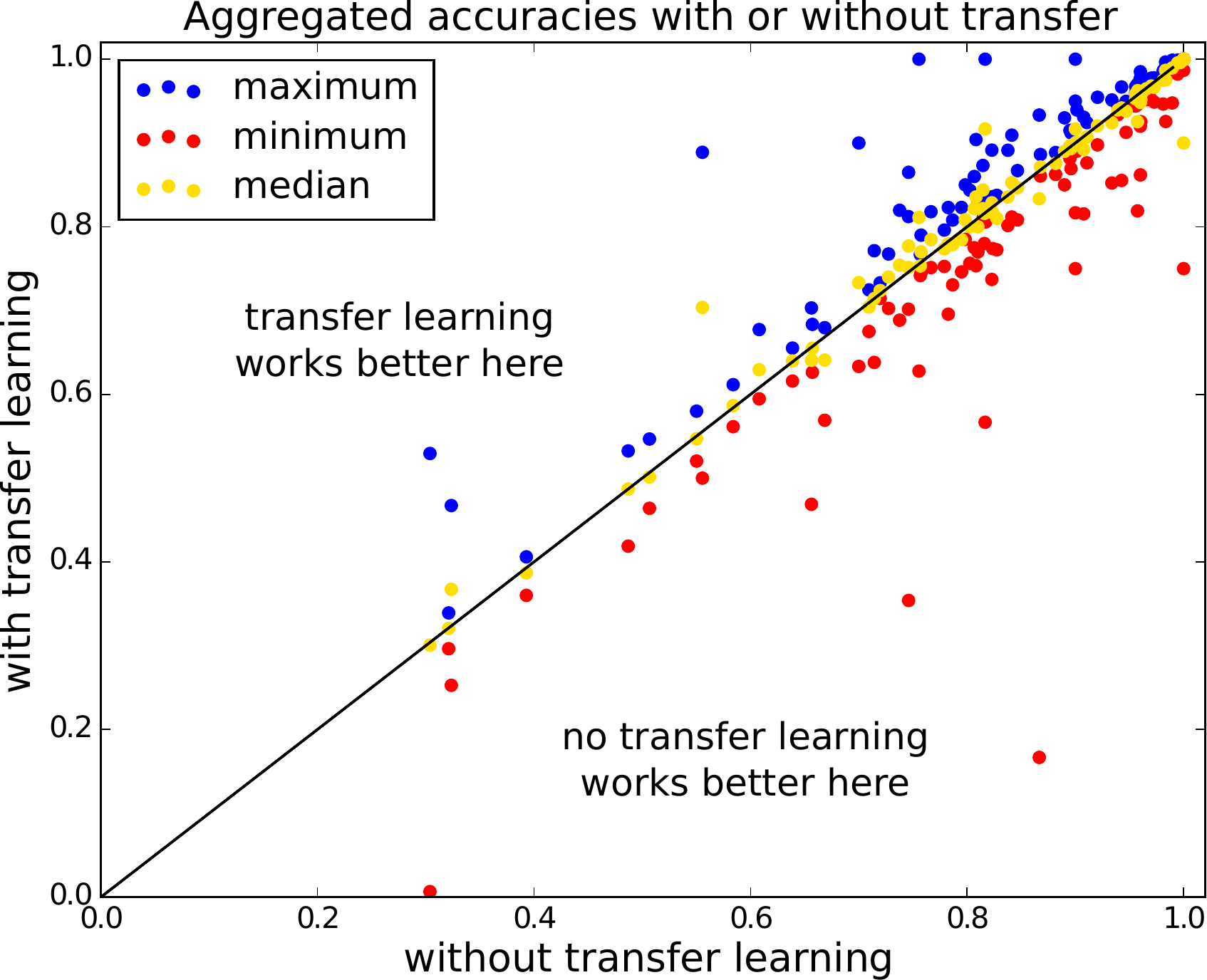}
	\caption{The three aggregated accuracies (minimum, median and maximum) of the Convolutional Neural Networks with the transfer learning approach against no transfer learning.}
	\label{fig-min_max}
\end{figure}

While observing the heat map in \figurename~\ref{fig-heat_map}, we can easily see that certain target datasets (columns) exhibit a high variance of accuracy improvements when varying the source datasets.
Therefore, to visualize the worst and best case scenarios when fine-tuning a model against training from scratch, we plotted in \figurename~\ref{fig-min_max} a pairwise comparison of three aggregated accuracies \{$minimum,median,maximum$\}. 

For each target dataset $D_t$, we took its minimum accuracy among the source datasets and plot it against the model's accuracy when trained from scratch.
This corresponds to the red dots in \figurename~\ref{fig-min_max}.  
By taking the minimum, we illustrate how one can \emph{always} find a bad source dataset for a given target dataset and decrease the model's original accuracy when fine-tuning a pre-trained network.

On the other hand, the maximum accuracy (blue dots in \figurename~\ref{fig-min_max}) shows that there is also \emph{always} a case where a source dataset increases the accuracy when using the transfer learning approach. 

As for the median (yellow dots in \figurename~\ref{fig-min_max}), it shows that on average, pre-training and then fine-tuning a model on a target dataset improves without significantly hurting the model's performance. 

One extreme case, where the choice of the source dataset had a huge impact on the model's accuracy, is the OliveOil dataset.
Precisely the accuracy decreased from 93.3\% to 16.7\% when choosing respectively MALLAT and FaceFour as source datasets. 

This analysis showed us that blindly and naively using the transfer learning approach could drastically decrease the model's performance. 
Actually, this is largely due to the fact that the initial weights of the network have a significant impact on the training~\citep{glorot2010understanding}. 
This problem has been identified as \emph{negative transfer learning} in the literature, where there still exists a need to quantify the amount of relatedness between the source and target datasets and whether an attempt to transfer knowledge from the source to the target domain should be made~\citep{weiss2016a}. 
Therefore in the following paragraph, we show how our similarity based solution can quantify this relatedness between the source and the target, thus enabling us to predict the best source dataset for a given target dataset. 

\subsubsection{Smart transfer learning}

In order to know in advance which source dataset is suited for which target dataset, we propose to leverage the similarity between two datasets. 
Our method is designed specifically for time series data without any previous domain knowledge about the datasets. 
Using the method we described in subsection~\ref{sec-method-transfer-learning}, we managed to compute a nearest neighbor for a target dataset and set this nearest neighbor to be the chosen source dataset for the current target dataset in question. 

The results showed that this proposed DTW based method will help in achieving what is called \emph{positive transfer}~\citep{weiss2016a}. 
As opposed to \emph{negative transfer}, positive transfer learning means that the learning algorithm's accuracy increases when fine-tuning a pre-trained model compared to a training from scratch approach~\citep{weiss2016a}.

\begin{figure}[htbp]
	\centering
	\includegraphics[width=0.7\linewidth]{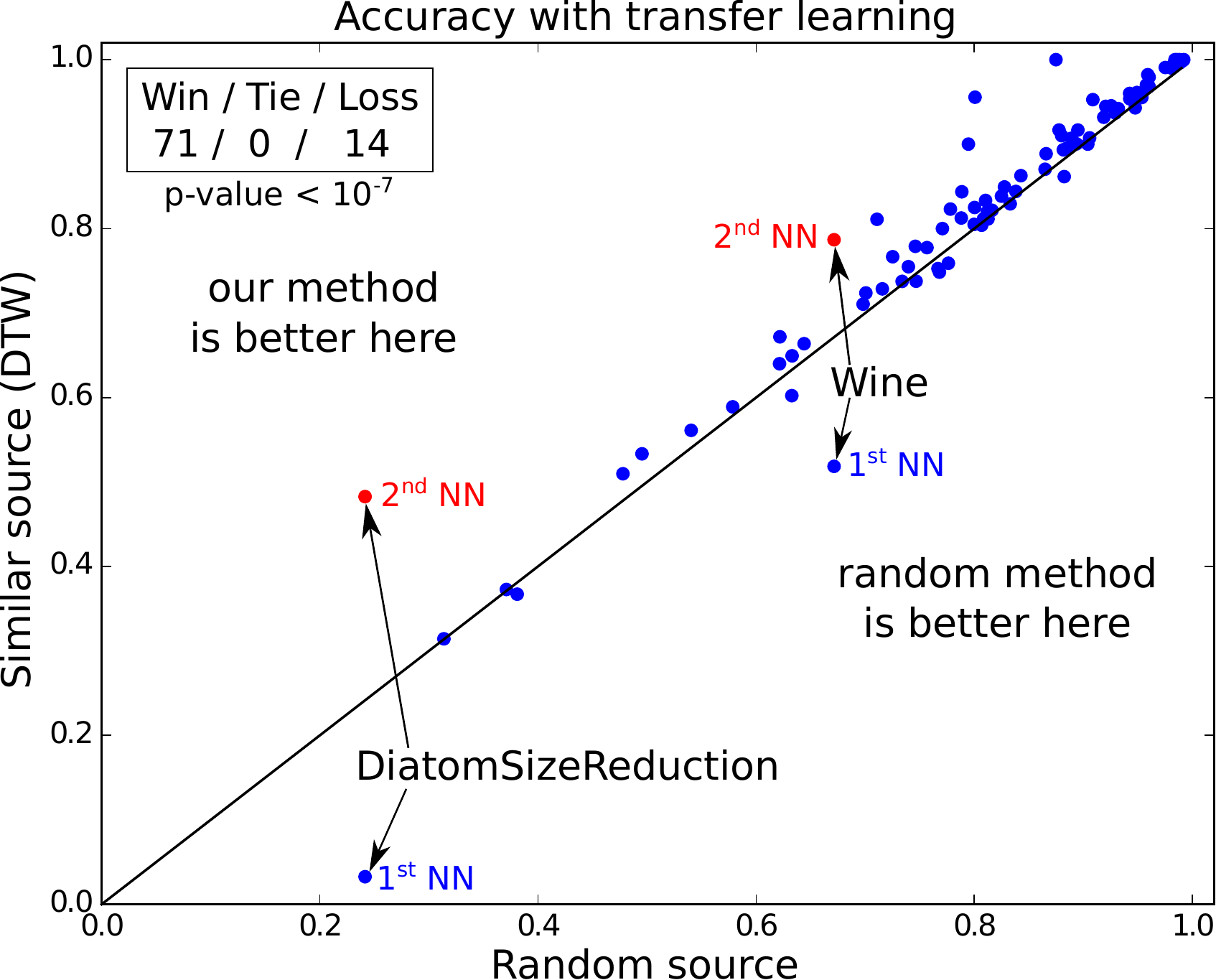}
	\caption{
		The accuracy of a fine-tuned model for two cases: 
		($x$ axis) when the source dataset is selected randomly;  
		($y$ axis) when the source dataset is selected using our Dynamic Time Warping based solution.}
	\label{fig-dtw_vs_rnd}
\end{figure}

\figurename~\ref{fig-dtw_vs_rnd} shows a pairwise accuracy plot for two approaches: a random selection process of the source dataset against a ``smart'' selection of the source dataset using a nearest neighbor algorithm with the distance calculated in algorithm~\ref{algo-sim}. 
In order to reduce the bias due to the random seed, the accuracy for the random selection approach was averaged over 1000 iterations. 
This plot shows that on average, choosing the most similar dataset using our method is significantly better than a random selection approach (with $p<10^{-7}$ for the Wilcoxon signed-rank test).
Respectively our method wins, ties and loses on 71, 0 and 14 datasets against randomly choosing the source dataset.
We should also note that for the two datasets DiatomSizeReduction and Wine, the nearest neighbor is not always the best choice.  
Actually, we found that the second nearest neighbor increases drastically the accuracy from 3.3\% to 46.7\% for DiatomSizeReduction and from 51.9\% to 77.8\% for Wine (see the $2^{nd}$ NN dots in \figurename~\ref{fig-dtw_vs_rnd}). 
This means that certain improvements could be incorporated to our inter-datasets similarity calculation such as adding a warping window~\citep{dau2017judicious} or changing the number of prototypes for each class which we aim to study in our future work.

Therefore, since in \figurename~\ref{fig-dtw_vs_rnd} the most similar dataset is the only one that is considered as a potential source for a given target, another interesting study would be to analyze the accuracy on a given target dataset as a function of how dissimilar the source dataset is.
However due to the huge number of datasets in the UCR/UEA archive compared to the space limitation, we chose to only study the most interesting cases where the results can be visually interpreted.

\subsubsection{Interesting case studies}

In this final analysis we chose to work with three interesting target datasets: \textit{ShapeletSim}, \textit{HandOutlines} and \textit{Meat}.
These datasets were chosen for different reasons such as the small size of the training set, the relatedness to shapelets and the transfer learning's accuracy variation.

\begin{figure}[htbp]
	\centerline{\includegraphics[width=0.7\linewidth]{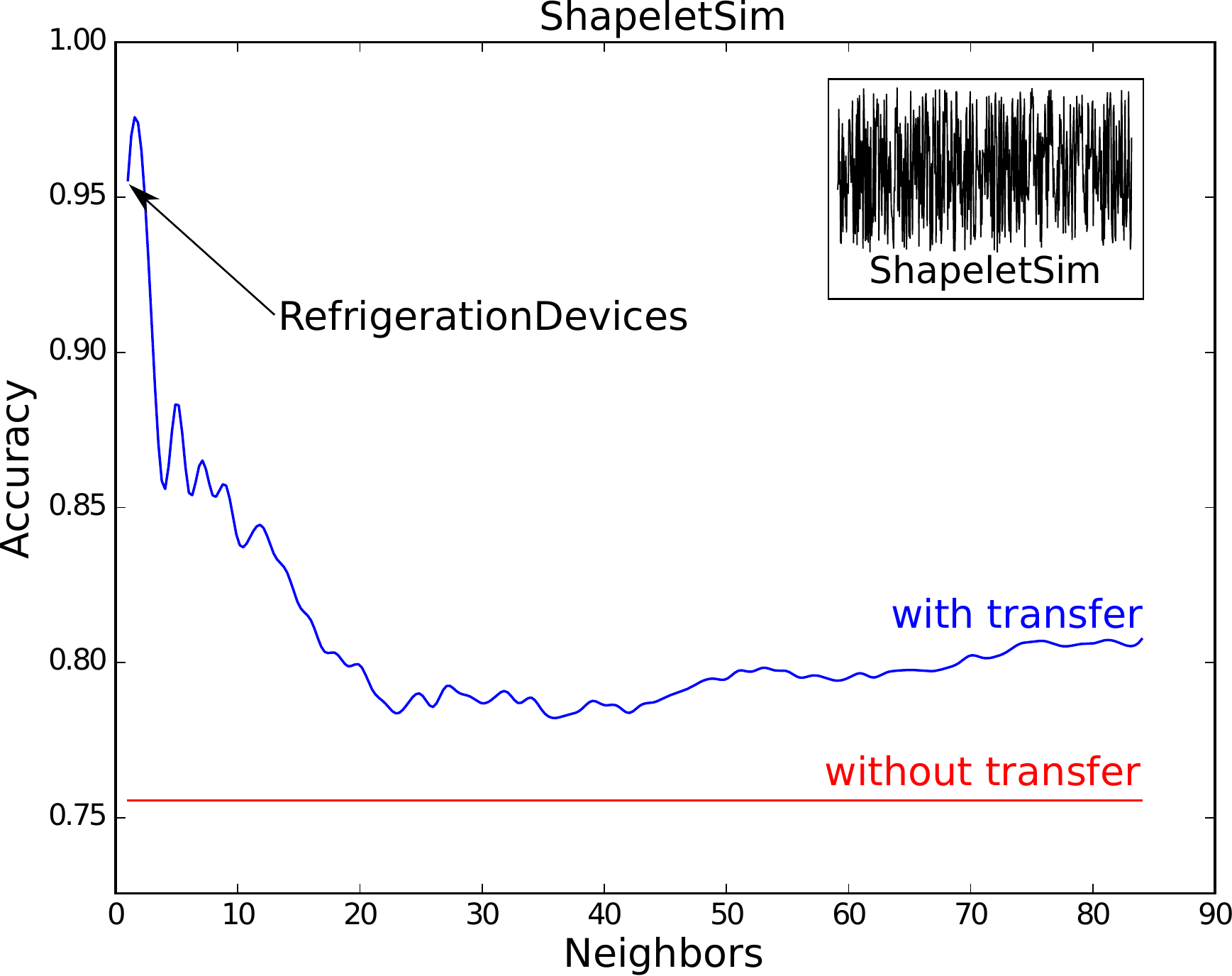}}
	\caption{
		The fine-tuned model's accuracy variation on the target dataset ShapeletSim with respect to the chosen source dataset neighbor (smoothed for visual clarity - best viewed in color).
	}
	\label{fig-ShapeletSim}
\end{figure}

\textit{ShapeletSim} contains one of the smallest training sets in the UCR/UEA archive (with 20 training instances). 
Additionally, this dataset is a simulated dataset designed specifically for shapelets which makes it interesting to see how well CNNs can fine-tune (pre-learned) shapelets~\citep{cui2016multi} when varying the source dataset.
\figurename~\ref{fig-ShapeletSim} shows how the model's accuracy decreases as we go further from the target dataset.
Precisely the average accuracy for the top 3 neighbors reaches 93\% compared to the original accuracy of 76\%. 
Actually, we found that the closest dataset to \textit{ShapeletSim} is the RefrigerationDevices dataset which contains readings from 251 households with the task to identify three classes: Fridge, Refrigerator and Upright Freezer.
This is very interesting since using other background knowledge one cannot easily predict that using RefrigerationDevices as a source for ShapeletSim will lead to better accuracy improvement. 
To understand better this source/target association, we investigated the shapes of the time series of each dataset and found that both datasets exhibit very similar spiky subsequences which is likely the cause for the transfer learning to work between these two datasets. 

\begin{figure}[htbp]
	\centerline{\includegraphics[width=0.7\linewidth]{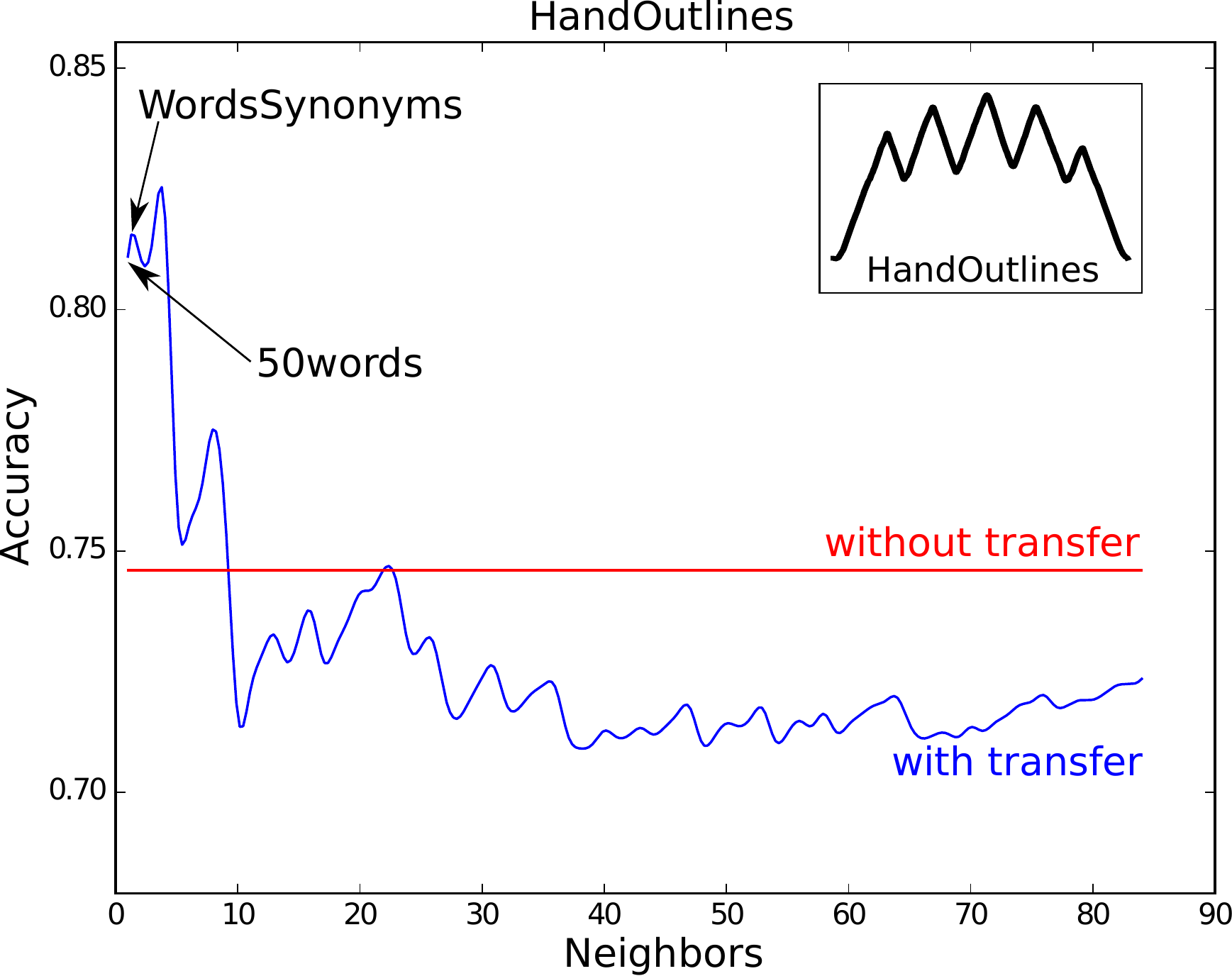}}
	\caption{
		The fine-tuned model's accuracy variation on the target dataset HandOutlines with respect to the chosen source dataset neighbor (smoothed for visual clarity - best viewed in color).
	}
	\label{fig-HandOutlines}
\end{figure}

\textit{HandOutlines} is one of the datasets where fine-tuning a pre-trained model almost never improves the accuracy.
Unlike \textit{ShapeletSim}, this dataset contains enough labeled data for the learning algorithm to learn from (with 1000 time series in the training set).
Surprisingly, we found that one could drastically increase the model's performance when choosing the best source dataset. 
\figurename~\ref{fig-HandOutlines} shows a huge difference (10\%) between the model's accuracy when fine-tuned using the most \emph{similar} source dataset and the accuracy when choosing the most \emph{dissimilar} source dataset.
\textit{HandOutlines} is a classification problem that uses the outlines extracted from hand images. 
We found that the two most similar datasets (50words and WordsSynonyms) that yielded high accuracy improvements, are also words' outlines extracted from images of George Washington's manuscripts. 

\begin{figure}[htbp]
	\centerline{\includegraphics[width=0.7\linewidth]{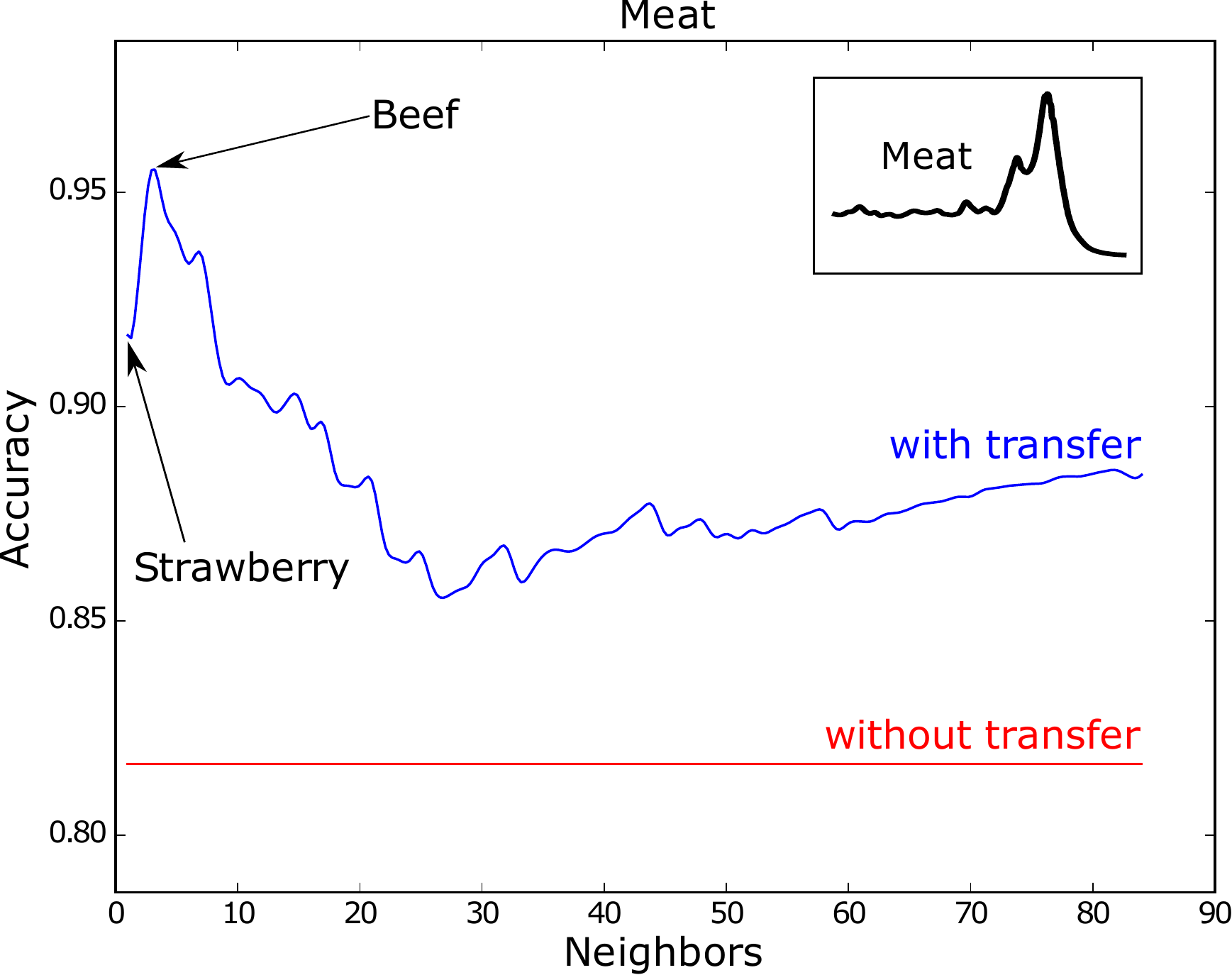}}
	\caption{
		The fine-tuned model's accuracy variation on the target dataset Meat with respect to the chosen source dataset neighbor (smoothed for visual clarity - best viewed in color).
	}
	\label{fig-Meat}
\end{figure}

\textit{Meat} is one of the smallest datasets (with 20 training instances) where the transfer learning approach was almost always beneficial.
However, we would like to examine the possibility of improving the accuracy even for the case where the transfer learning seems to be positive~\citep{weiss2016a} for any choice of source dataset.
\figurename~\ref{fig-Meat} shows that the accuracy reaches almost 95\% for the top 3 closest datasets and then decrease the less similar the source and target datasets are.
While investigating these similarities, we found the top 1 and 3 datasets to be respectively Strawberry and Beef which are all considered spectrograph datasets~\citep{bagnall2017the}. 
As for the second most similar dataset, our method determined it was 50words.
Given the huge number of classes (fifty) in 50words our method managed to find some latent similarity between the two datasets which helped in improving the accuracy of the transfer learning process.

\subsection{Conclusion}
In this section, we investigated the transfer learning approach on a state of the art deep learning model for TSC problems. 
Our extensive experiments with every possible combination of source and target datasets in the UCR/UEA archive, were evidence that the choice of the source dataset could have a significant impact on the model's generalization capabilities. 
Precisely when choosing a bad source dataset for a given target dataset, the optimization algorithm can be stuck in a local optimum. 
This phenomena has been identified in the transfer learning literature by \emph{negative transfer learning} which is still an active area of research~\citep{weiss2016a}.
Thus, when deploying a transfer learning approach, the big data practitioner should give attention to the relationship between the target and the chosen source domains.

These observations motivated us to examine the use of the well known time series similarity measure DTW, to predict the choice of the source dataset when fine-tuning a model on a time series target dataset.
After applying this transfer learning guidance, we concluded that transferring deep CNNs on a target dataset works best when fine-tuning a network that was pre-trained on a similar source dataset.
These findings are very interesting since no previous observation made the link between the space induced by the classic DTW and the features learned by the Convolutional Neural Networks. 

Our results should motivate the big data practitioners to no longer train models from scratch when classifying time series, but instead to fine-tune pre-trained models. 
Especially because CNNs, if designed properly, can be adapted across different time series datasets with varying length.  

In our future work, we aim again to reduce the deep neural network's overfitting phenomena by generating synthetic data using a Weighted DTW Barycenter Averaging method~\citep{forestier2017generating}, since the latter distance gave encouraging results in guiding a complex deep learning tool such as transfer learning.    

Finally, with big data time series repositories becoming more frequent~\citep{zoumpatianos2014indexing}, leveraging existing source datasets that are similar to, but not exactly the same as a target dataset of interest, makes a transfer learning method an enticing approach.

\section{Ensembling}
TSC tasks differ from traditional classification tasks by the natural temporal ordering of their attributes~\citep{bagnall2017the}. 
To tackle this problem, a huge amount of research was dedicated into coupling and enhancing time series similarity measures with an NN classifier~\citep{dau2017judicious,gharghabi2018ultra}. 
In~\cite{lines2015time}, ten elastic distances were compared to the traditional DTW algorithm to find out that no single measure could outperform the classic NN-DTW for TSC.
These findings motivated the authors to construct a single EE classifier that includes all eleven different similarity measures, and achieve a significant improvement compared to the individual classifiers~\citep{lines2015time}. 
Hence, recent contributions were focused on ensembling different discriminant classifiers such as decision trees~\citep{baydogan2013a} and SVMs~\citep{bostrom2015binary} on different data representation techniques such shapelet transform~\citep{bostrom2015binary} or DTW features~\citep{kate2016using}. 
These ideas gave rise to COTE~\citep{bagnall2016time} and its extended version HIVE-COTE~\citep{lines2018time} where 37 different classifiers were ensembled over multiple time series data transformation techniques in order to reach current state-of-the-art performance for TSC~\citep{bagnall2017the}. 

With the advent of deep neural networks into industrial and commercial applications such as self-driving cars~\citep{qiu2018multi} and speech recognition systems~\citep{liul2018stochastic}, time series data mining practitioners started investigating the application of deep learning to TSC problems~\citep{wang2017time}.
In the previous chapter, we showed how deep CNNs are able to achieve results that are not significantly different than current state-of-the-art algorithms for TSC problems when evaluated over the 85 time series datasets from the UCR/UEA archive~\citep{ucrarchive}.   
These results suggest that building upon deep learning based solutions for TSC could further improve the current state-of-the-art performance of deep neural networks. 

One way of improving neural network based classifiers is to build an ensemble of deep learning models. 
This idea seems very interesting for TSC tasks since the state-of-the-art is moving towards ensembled solutions~\citep{lines2018time,lines2015time,bagnall2017the,baydogan2013a}. 
In addition, deep neural network ensembles seem to achieve very promising results in many supervised machine learning domains such as skin lesions detection~\citep{goyal2018deep}, facial expression recognition~\citep{wen2017ensemble} and automatic bucket filling~\citep{dadhich2018predicting}. 

\begin{figure}
	\centering
	\includegraphics[width=0.8\linewidth]{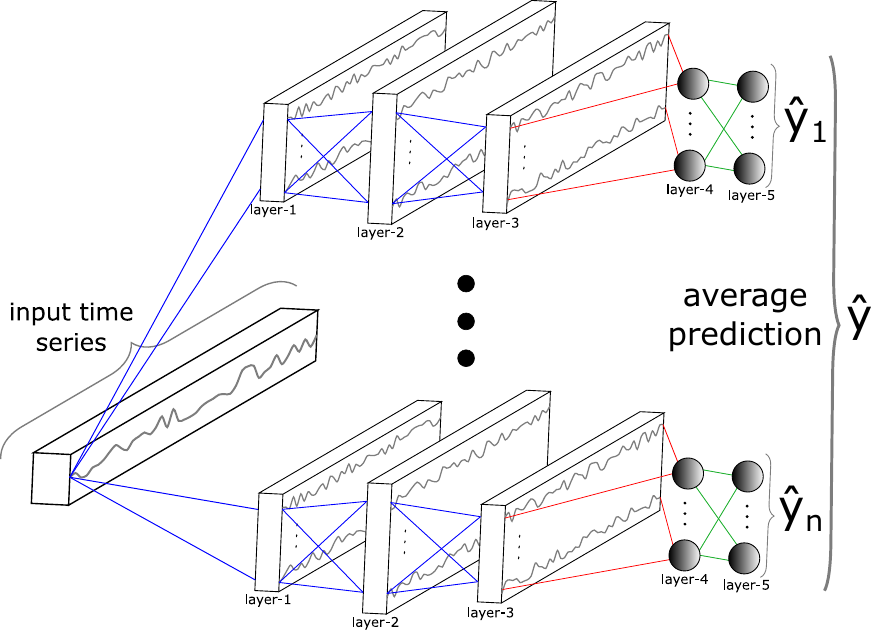}
	\caption{Ensemble of deep convolutional neural networks for time series classification.}
	\label{fig:ensemble}
\end{figure}

Therefore, we propose to ensemble the current state-of-the-art deep learning models for TSC developed in the previous chapter, by constructing one model composed of 60 different deep neural networks: 6 different architectures~\citep{wang2017time,zheng2014time,zhao2017convolutional,serra2018towards} each one with 10 different initial weight values.
By evaluating on the 85 datasets from the UCR/UEA archive, we demonstrate a significant improvement over the individual classifiers while also reaching very similar performance to HIVE-COTE: the current state-of-the-art ensemble of 37 non deep learning based time series classifiers.  
Finally, inspired by the our finding on transfer learning in the previous section, we replace ensembling randomly initialized networks with an ensemble constructed out of fine-tuned models from 84 different source datasets, which showed a significant improvement for TSC problems.


\subsection{Background}

In this subsection we present some work related to ensembling neural network classifiers, with a focus on time series data.

Constructing an ensemble of many deep learning classifiers has been shown to achieve high performance in many different fields. 
In~\cite{goyal2018deep}, an ensemble of two neural networks was adopted: (1) Inception-v4 and (2) Inception-ResNet-v2. 
Both of these classifiers are learned with a joint meta-learning approach in an end-to-end manner.
A forest CNN was proposed by~\cite{lee2017ensemble} for image classification, where similarly to random forest, the ensemble is constructed by replacing the individual nodes with a CNN and finally the classifier's decision is taken by performing a majority voting scheme over the different decisions of the individual trees in the forest. 
Another ensemble of CNNs for facial expression recognition was proposed in~\cite{wen2017ensemble} where each individual classifier was trained independently to output a probability for each class and then the network's final decision was taken using a probability-based fusion method. 
In~\cite{dadhich2018predicting}, an ensemble of neural networks was found to outperform other hybrid machine learning ensembles when solving an automatic bucket filling problem. 
Finally in~\cite{ienco2018semi}, deep auto-encoders were ensembled in order to learn an unsupervised latent representation of the input data over multiple resolutions, thus improving the quality of the produced clusters.

Although in almost all use cases ensembling deep neural networks almost always yields to better decisions, we did not find any approach using a neural network ensemble for domain agnostic TSC.
Perhaps the work in~\cite{jin2016ensemble} is the closest to ours where a neural network based ensemble was used to perform biomedical TSC, where individual architectures were constructed with some domain knowledge specific to the classification problem at hand such as choosing the filter length with local and distorted views. 
Therefore, we decided to further explore ensembling deep neural networks for TSC, by combining multiple deep learning models in different settings. 

\subsection{Methods}

In this subsection, we start by presenting the six different architectures composing our ensembles of neural networks. 
For completeness, we describe the random initialization technique adopted for all models. 
Finally, we present a transfer learning based alternative to randomly initializing the weights of the networks. 

\subsubsection{Architectures}

The average rank of the six chosen deep learning classifiers, over the 85 datasets from the UCR/UEA archive~\citep{ucrarchive,bagnall2017the} is listed in Table~\ref{tab:avg-rank}. 
All of these architectures were implemented in a common framework during our empirical study presented in the previous chapter, containing originally  9 different deep learning approaches for TSC. 
However only 6 out of these 9 approaches were probabilistic classifiers whereas the three other classifiers performed a hard prediction: meaning an input time series is assigned a specific class rather than a probability distribution over all the classes in a dataset.    
Therefore, we chose to only ensemble the 6 probabilistic models, thus allowing us to combine the networks by averaging the a posteriori probability for each class over the individual classifiers' output.  
Finally, we present a brief description of these 6 different architectures and refer the interested reader to a more thorough explanation in the corresponding papers. 
All hyperparameters can be found in~\cite{IsmailFawaz2018deep}.

\begin{table}
	\centering
	\begin{tabular}{|c|c|c|}
		\toprule
		\textbf{Approach} & \textbf{Rank} & \textbf{Wins} \\
		\midrule
		ResNet~\citep{wang2017time} & 1.88 & 41 \\
		FCN~\citep{wang2017time} & 2.49 & 18 \\
		Encoder~\citep{serra2018towards} & 3.34 & 10 \\
		MLP~\citep{wang2017time} & 4.08 & 4 \\
		Time-CNN~\citep{zhao2017convolutional} & 4.38 & 4 \\
		MCDCNN~\citep{zheng2014time} & 4.83 & 3 \\
		\bottomrule
	\end{tabular}
	\caption{Average rank of the six classifiers constituting the Neural Network Ensemble for time series classification over the 85 datasets from the UCR/UEA archive.}
	\label{tab:avg-rank}
\end{table}

\textit{Multi-Layer Perceptron} (MLP) is the simplest form of deep neural networks and was proposed in~\cite{wang2017time} as a baseline architecture for TSC. 
The architecture contains three hidden layers, with each one fully connected to the output of its previous layer.  
The main characteristic of this architecture is the use of a Dropout layer~\citep{srivastava2014a} to reduce overfitting. 
One disadvantage is that since the input time series is fully connected to the first hidden layer, the temporal information in a time series is lost~\citep{IsmailFawaz2018deep}.

\textit{Fully Convolutional Neural Network} (FCN), originally proposed in~\cite{wang2017time}, is considered a competitive architecture yielding the second best results when evaluated on the UCR/UEA archive (see Table~\ref{tab:avg-rank}).
This network is comprised of three convolutional layers, each one performing a non-linear transformation of the input time series. 
A global average pooling operation is used before the final softmax classifier, thus reducing drastically the number of parameters in a network and allowing an architecture that is invariant to the length of the input time series. 
The latter characteristic motivated us to perform a transfer learning technique in~\cite{IsmailFawaz2018transfer}, and ensembling the resulting neural networks which is later discussed in this subsection. 

\textit{Residual Network} (ResNet) was originally proposed in~\cite{wang2017time} and showed similar performance to FCN when evaluated on 44 datasets from the archive. 
However, when evaluated over the 85 datasets, ResNet significantly outperformed FCN (see Table~\ref{tab:avg-rank}). 
The main characteristic of ResNet is the addition of residual connections which enables a direct flow of the gradient~\citep{wang2017time}.

\textit{Encoder} was originally proposed in~\cite{serra2018towards} as a hybrid CNN that modifies the FCN architecture~\citep{wang2017time} by mainly adding a Dropout layer~\citep{srivastava2014a} and an attention mechanism. 
The latter operation enables Encoder to learn to localize which regions of the input time series are useful for a certain class identification. 

\textit{Multi-Channels Deep Convolutional Neural Networks} (MCDCNN) was originally proposed in~\cite{zheng2014time} for multivariate TSC and adapted to univariate data by~\cite{IsmailFawaz2018deep}.
It consists of a traditional CNN, where each convolutional layer is followed by a max pooling operation, then a traditional fully connected layer is used before the final softmax classifier.

\textit{Time Convolutional Neural Network} (Time-CNN) was originally proposed for univariate as well as multivariate TSC~\citep{zhao2017convolutional}. 
Similarly to MCDCNN, this network is a traditional CNN with one major exception: the use of the mean squared error instead of the traditional categorical cross-entropy loss function, which has been used by all the deep learning approaches we have mentioned so far.
Therefore for Time-CNN, the sum of the output class probabilities is not guaranteed to be equal to one.

\subsubsection{Ensembling models with random initial weights}

We have described in the previous subsection, the architecture of six different classifiers. 
The weights for each network are initialized randomly using Glorot's uniform initialization method~\citep{glorot2010understanding}. 
This technique ensures a uniform distribution of the initial weight values.
However due to non-convexity, networks with the same architecture but different initial weights could yield different validation accuracy. 
In~\cite{choromanska2015loss}, the authors showed that deeper networks are much more stable with respect to the randomness.
This would suggest that ensembling relatively non deep architectures would yield to a much better improvement in accuracy than ensembling deeper architectures. 
Fortunately, for low dimensional time series data, current state-of-the-art architectures are much less deeper than their counterpart networks for high dimensional images. 
Therefore, we believe that we can leverage this instability of neural networks for time series data by ensembling the decision taken by the same network but with different random initializations, using the following equation: 
\begin{equation}
\hat{y}_{i,c}=\frac{1}{n}\sum_{j=1}^{n}\sigma_c(x_i,\theta_j) ~~|~~\forall c\in [1,C]
\end{equation}
with $\hat{y}_{i,c}$ denoting the ensemble's output probability of having the input time series $x_i$ belonging to class $c$, which is equal to the logistic output $\sigma_c$ averaged over the $n$ randomly initialized models.   
We should note that training an ensemble of the same architecture with different initial weight values has been shown to improve neural network's performance on many computer vision problems~\citep{wen2017ensemble}, however, we did not encounter any previous work that combines such classifiers for TSC.  

\subsubsection{Transfer learning}

An alternative to training a deep classifier from scratch is to fine-tune a model that has been already pre-trained on a un/related task~\citep{IsmailFawaz2018transfer}, which was described in the previous section of this current chapter. 
This process is called transfer learning, where the network is first trained on a source dataset, then the final layer is removed and replaced with a new randomly initialized softmax layer whose number of neurons is equal to the number of classes in the target dataset. 
The pre-trained model is then fine-tuned or re-trained on the target dataset's training set. 
With 85 datasets in the archive, each target dataset will have 84 potential source datasets, which motivated us to ensemble the decision of these 84 FCN models.

\subsection{Results}

In this section we present the results of different ensembling schemes when evaluated on the 85 datasets from the UCR/UEA archive~\citep{ucrarchive}, which is currently the largest publicly available benchmark for time series analysis.
In order to compare multiple classifiers over several datasets, we followed the same procedure described in the previous chapter. 
All experiments were conducted on a hybrid cluster of more than 60 NVIDIA GPUs comprised of GTX 1080 Ti, Tesla K20, K40 and K80.
Note that the code and the raw results are publicly available on the project's companion repository\footnote{\url{https://github.com/hfawaz/ijcnn19ensemble}}.

\subsubsection{Ensembling randomly initialized models}

\begin{figure}
	\centering
	\includegraphics[width=\linewidth]{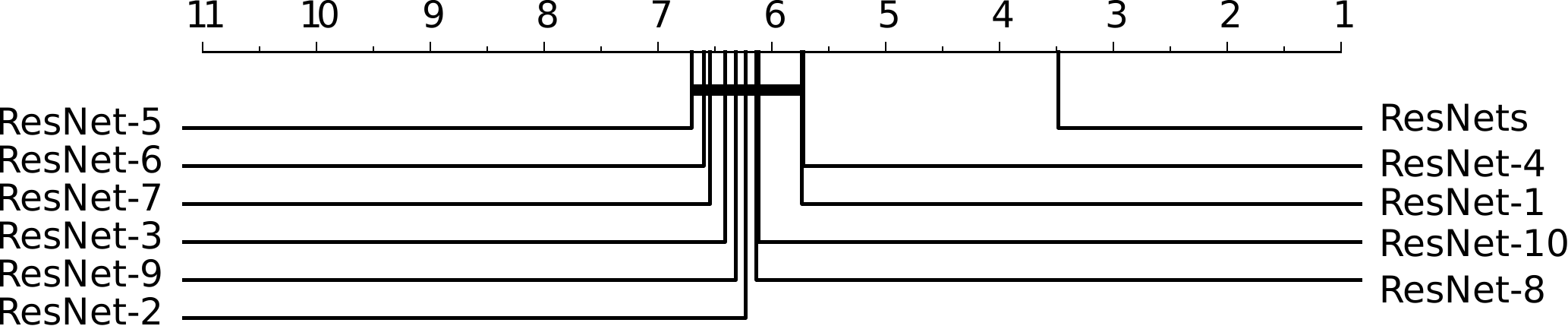}
	\caption{Critical difference diagram showing the pairwise statistical comparison of ten ResNets with random initializations as well as one ResNet ensemble composed of these ten individual neural networks.}
	\label{fig:cd-diagram-resnets}
\end{figure}

By ensembling randomly initialized networks, we are able to achieve a significant improvement in accuracy. 
Figure~\ref{fig:cd-diagram-resnets} shows a critical difference diagram where ten different random initializations of ResNet did not yield to significantly different results.
However, by ensembling these different networks, we were able to demonstrate a significant improvement in the average rank over the 85 datasets.
We should note that the latter phenomenon was also observed for the five other neural networks described in the previous subsection. 
Finally, we should emphasize that an ensembling technique will improve the stability of ResNet in terms of accuracy, in other words reducing the bias due to the initial weight values as well as the randomness induced by gradient descent based optimization. 

\subsubsection{Ensembling all neural networks}

\begin{figure}
	\centering
	\includegraphics[width=\linewidth]{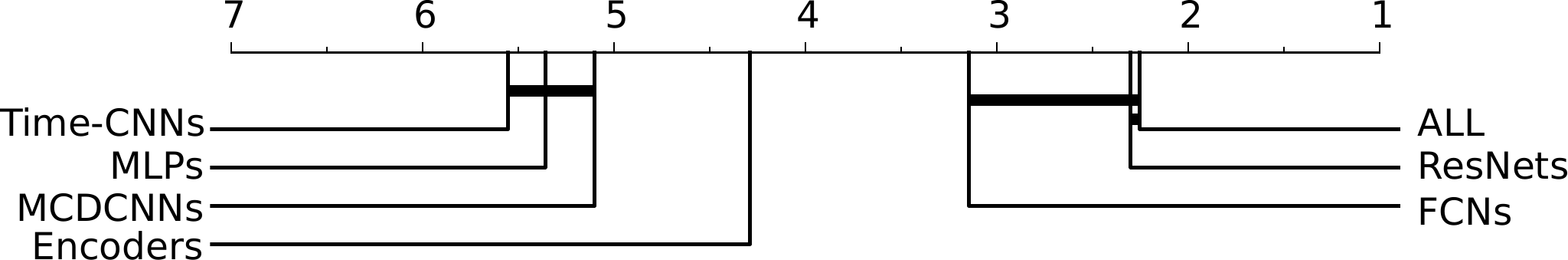}
	\caption{Critical difference diagram showing the pairwise statistical comparison of six architectures ensembled with ten different random initializations each, as well as one ensemble containing the six models.}
	\label{fig:cd-diagram-all}
\end{figure}

After demonstrating that using an ensemble of neural networks is always better than a single classifier, we sought to answer the following question: \textit{Could an ensemble of hybrid randomly initialized networks achieve even better performance?}
Figure~\ref{fig:cd-diagram-all} shows a critical difference diagram containing six ensembles of homogenized networks as well as the hybrid ensemble of \emph{all} available networks. 
The latter classifier contains sixty different networks: each architecture (six in total) is initialized with ten different random weight values.  
The results show that ensembling all networks was able to outperform all classifiers. 
However the statistical test failed to find any significant difference between the full ensemble and individual ResNet/FCN ensembles. 
This would suggest that the ensemble is highly affected by the poor performance of Time-CNN, MLP and MCDCNN. 
The latter classifiers showed the worst average rank without any significant difference, thus suggesting that removing them would yield even better performance.

\subsubsection{Neural Network Ensemble}

\begin{figure}
	\centering
	\includegraphics[width=0.7\linewidth]{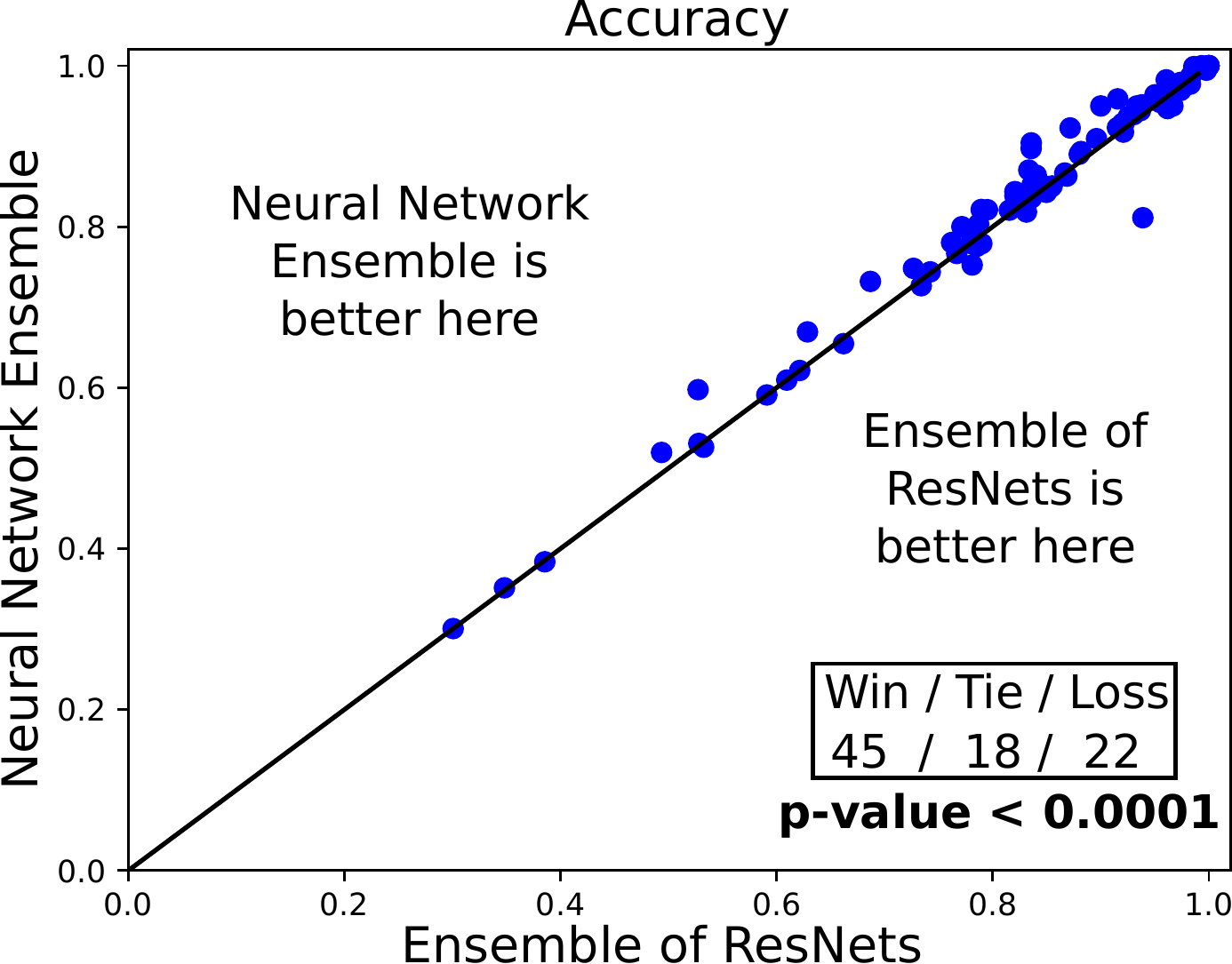}
	\caption{The Neural Network Ensemble (NNE) composed of ResNet, FCN and Encoder is significantly better than an ensemble of pure ResNets.}
	\label{fig:resnet-vs-nne}
\end{figure}

The results in the previous section, suggest that choosing carefully the classifiers in the pool would yield to a better ensemble. 
Therefore, we construct an NNE comprised solely of ResNet, FCN and Encoder. 
These three architectures were the only ones to yield significantly different results when a homogenized ensemble was adopted (Figure~\ref{fig:cd-diagram-all}).
Further investigations suggested that FCN performs better than ResNet on electrocardiography datasets~\citep{IsmailFawaz2018deep}, which would motivate researchers to combine these two classifiers in order to have a robust algorithm that improves the accuracy over the whole datasets. 
However, for small datasets such as CinCECGTorso, both FCN and ResNet overfitted the dataset very easily with almost 82\% test accuracy~\citep{IsmailFawaz2018data}, whereas Encoder managed to achieve very good performance with a 91\% accuracy, therefore implying a combination of ResNet, FCN and Encoder would yield to better accuracy on a various range of TSC datasets. 
Figure~\ref{fig:resnet-vs-nne} shows how NNE is able to outperform an ensemble of pure ResNets with 45 wins and 18 ties on 85 datasets from the archive. 
We believe that the combination of an FCN with ResNet and Encoder, enables the classifier to benefit respectively from the residual linear connections and the attention mechanism. 

\begin{figure}
	\centering
	\includegraphics[width=\linewidth]{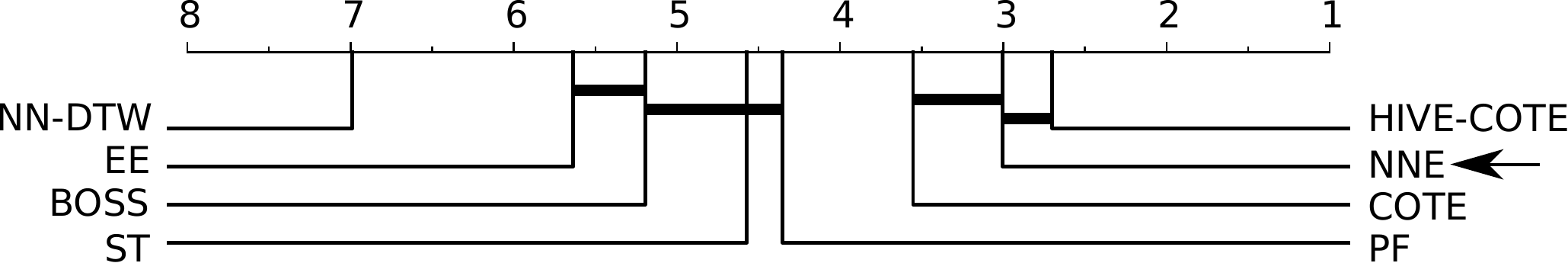}
	\caption{Critical difference diagram showing the pairwise statistical comparison of current state-of-the-art algorithms with the Neural Network Ensemble (NNE) added to the pool.}
	\label{fig:cd-diagram-bake-off}
\end{figure}

To further understand how NNE is performing with respect to current state-of-the-art TSC algorithms, we illustrate in Figure~\ref{fig:cd-diagram-bake-off} a critical difference diagram containing NNE and seven other non deep learning based classiifers: 
(1) NN-DTW corresponds to the nearest neighbor coupled with the Dynamic Time Warping distance; 
(2) EE is an ensemble of nearest neighbor classifiers with eleven elastic distances; 
(3) BOSS corresponds to the ensemble Bag-of-SFA-Symbols; 
(4) ST is another ensemble of off-the-shelf classifiers computed over the Shapelet Transform data domain; 
(5) PF or Proximity Forest is an ensemble of decision trees coupled with eleven elastic distances; 
finally (6) COTE and (7) HIVE-COTE are two ensembles of respectively 35 and 37 classifiers using multiple data transformation techniques. 
The results for these classifiers were taken from~\cite{bagnall2017the} except for PF whose results were taken from the original paper~\citep{lucas2018proximity}.  
Figure~\ref{fig:cd-diagram-bake-off} clearly shows how our NNE is able to reach state-of-the-art performance for TSC, suggesting that CNNs are able to extract one dimensional discriminant features useful for classification in an end-to-end manner, as opposed to other hand-engineered features used by HIVE-COTE such as the Discrete Fourier Transform, DTW features and the Shapelet Transform.        

\subsubsection{Ensembling fine-tuned models}

\begin{figure}
	\centering
	\includegraphics[width=0.7\linewidth]{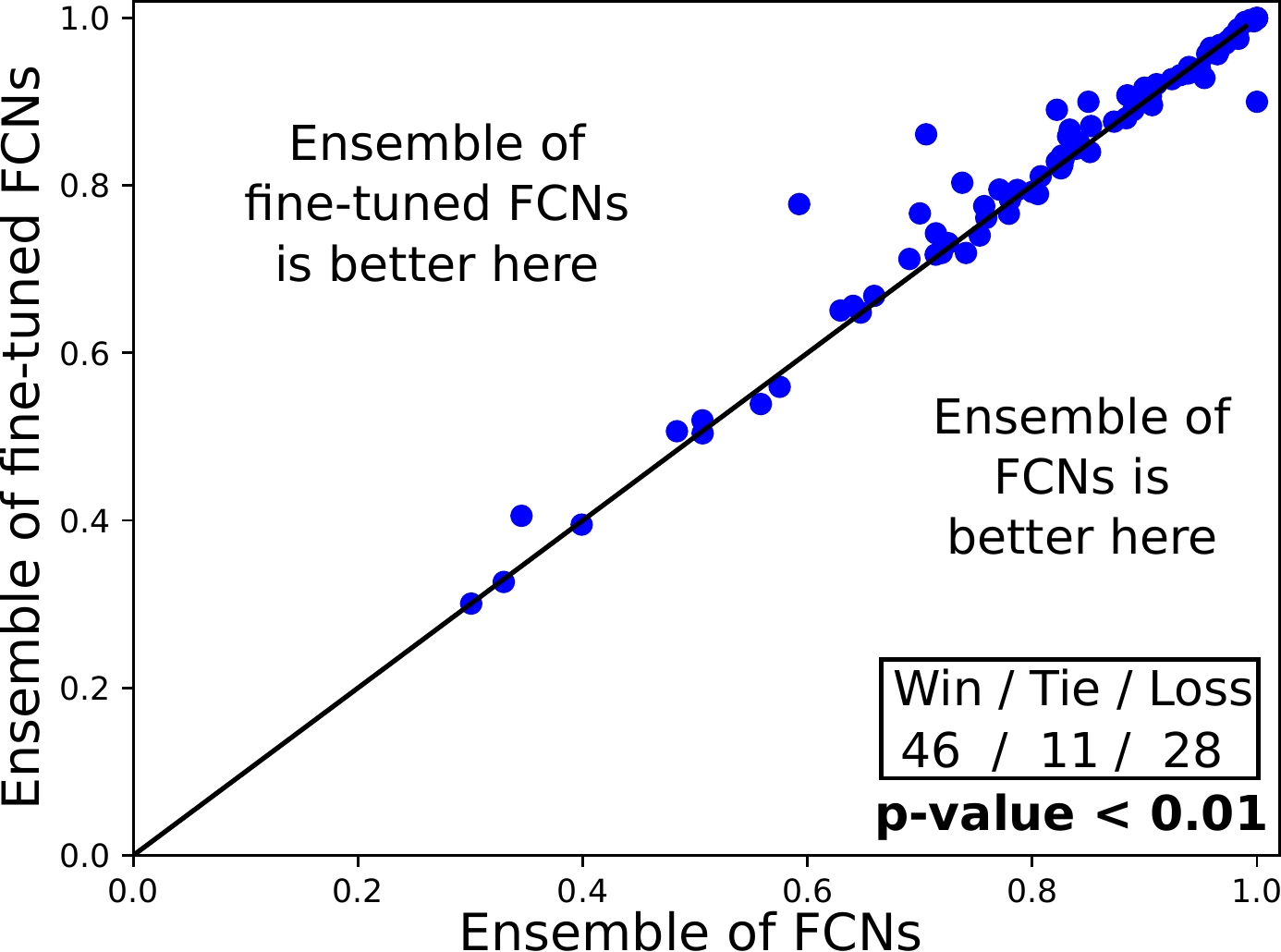}
	\caption{Ensembling fine-tuned models is significantly better than ensembling randomly initialized FCN models that are trained from scratch.}
	\label{fig:fcn-vs-transfer}
\end{figure}

Figure~\ref{fig:fcn-vs-transfer} shows that ensembling fine-tuned FCNs is significantly better than ensembling randomly initialized FCN models that are trained from scratch. 
However, this transfer learning based ensemble did not manage to outperform ResNets' ensemble nor NNE. 
These results show that the choice of architecture is very crucial and suggest that an ensemble of transferred ResNets would demonstrate even better performance than an ensemble of pure ResNets or NNE.   

\subsection{Conclusion}

In this section, we showed how ensembling deep neural networks can achieve state-of-the-art performance for time series classification. 
We showed that it would be almost always beneficial to ensemble randomly initialized models rather than choosing one trained neural network out of the ensemble. 
Finally, we investigated an ensemble of transferred deep CNNs to demonstrate even better performance than ensembling randomly initialized networks.
In the future, we would like to consider a meta-ensembling approach where the output logistics of individual deep learning models are fed to a meta-network that learns to map these inputs to the correct prediction. 

\section{Data augmentation}

Deep learning usually benefits from large training sets~\citep{zhang2017understanding}. 
However, for many applications only relatively small training data exist. 
In TSC, this phenomenon can be observed by analyzing the UCR/UEA archive's datasets~\citep{ucrarchive}, where 20 datasets have 50 or fewer training instances. 
These numbers are relatively small compared to the billions of labeled images in computer vision, where deep learning has seen its most successful applications~\citep{lecun2015deep}.

\begin{figure}
	\centering
	\subfloat[DiatomSizeReduction]{
		\includegraphics[width=0.47\linewidth]{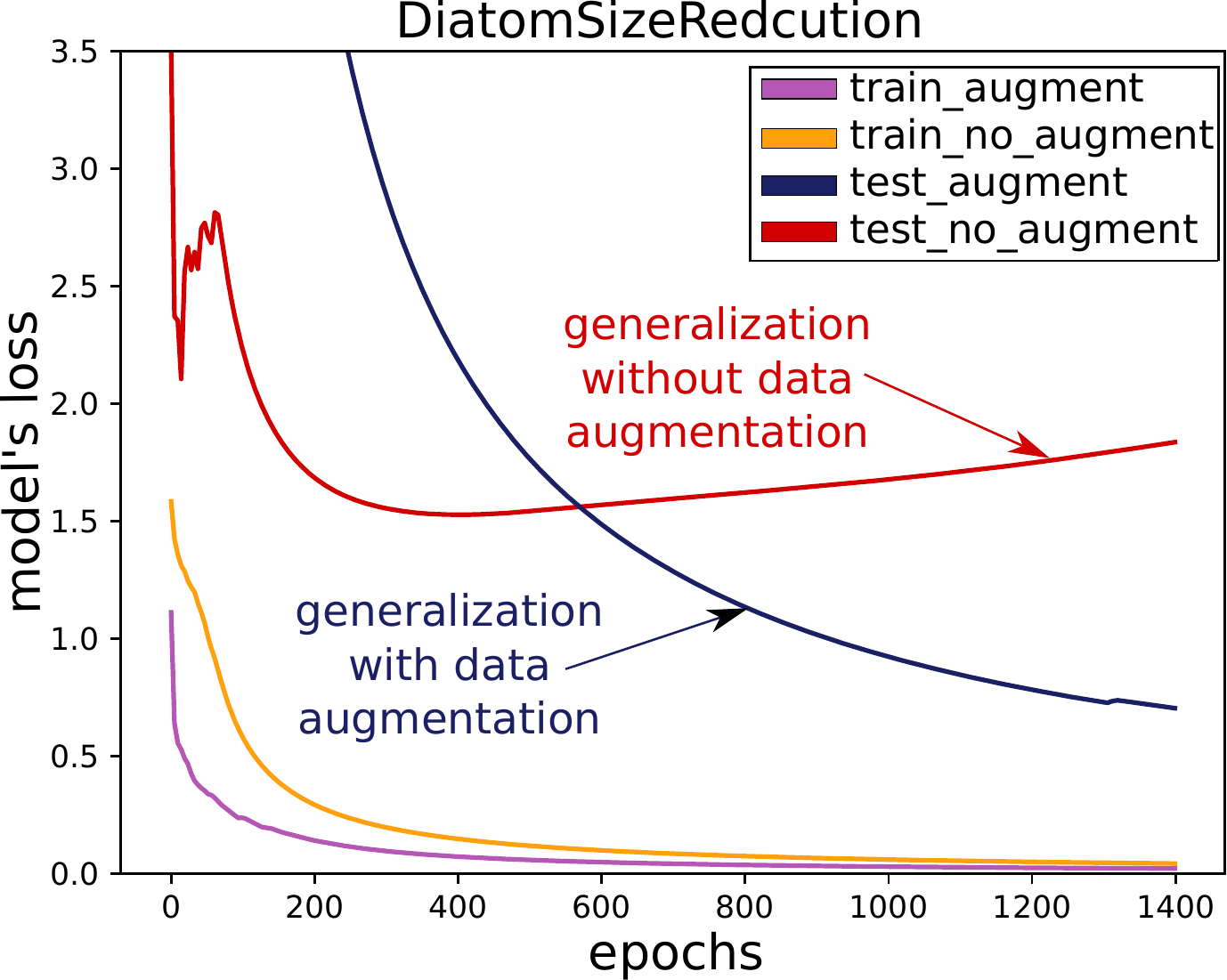}
		\label{sub-diatomsizereduction}}
	\hspace{.1cm} 
	\subfloat[Meat]{
		\includegraphics[width=0.47\linewidth]{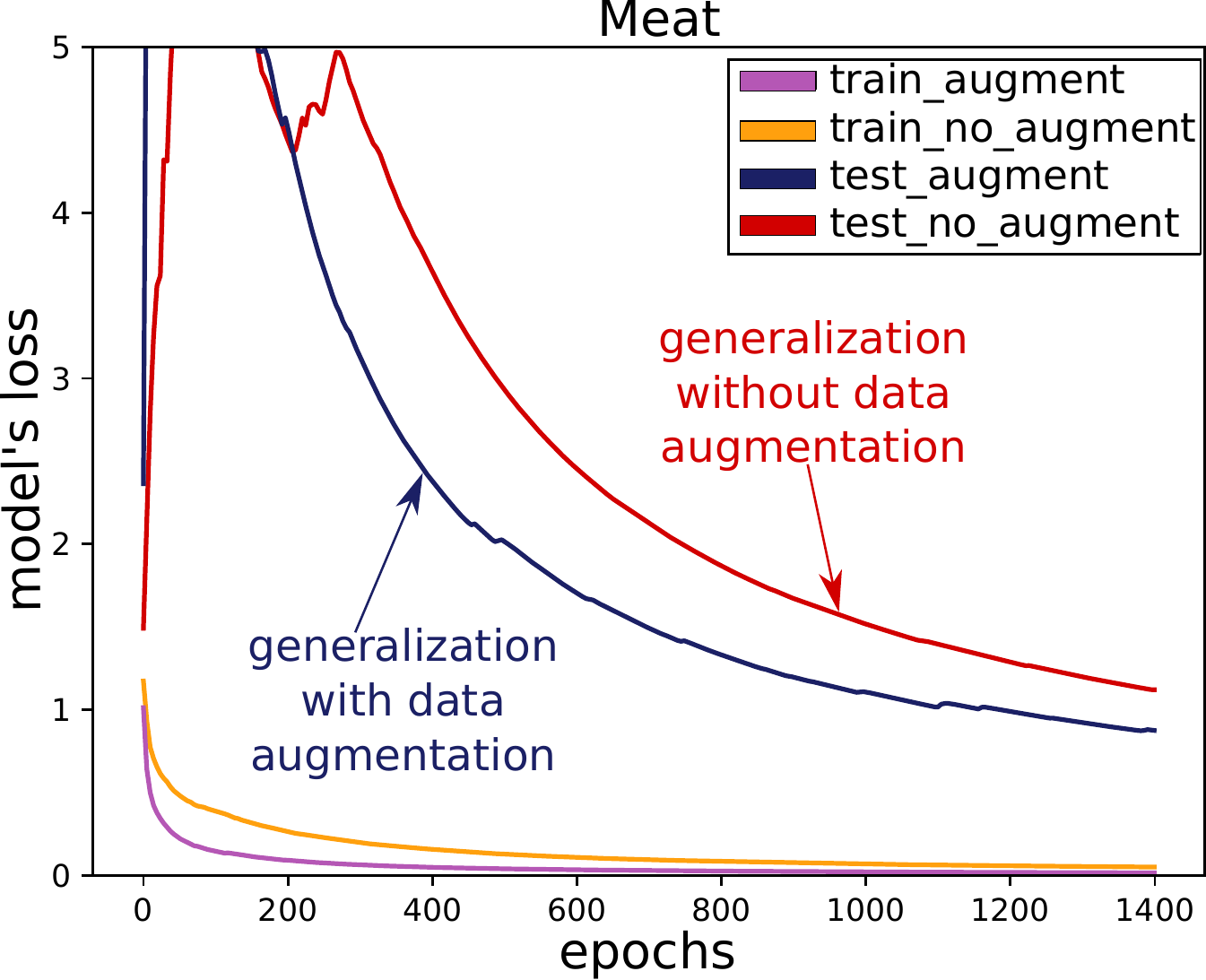}
		\label{sub-meat}}
	\caption{The model's loss with/without data augmentation on the DiatomSizeReduction and Meat datasets (smoothed and clipped for visual clarity).}
	\label{fig-plot-generalization}
\end{figure}

Although the recently proposed deep CNNs reached state of the art performance in TSC on the UCR/UEA archive~\citep{wang2017time}, they still show low generalization capabilities on some small datasets such as the CinCECGTorso dataset with 40 training instances. 
This is surprising since the NN-DTW performs exceptionally well on this dataset which shows the relative easiness of this classification task. 
Thus, inter-time series similarities in such small datasets cannot be captured by the CNNs due to the lack of labeled instances, which pushes the network's optimization algorithm to be stuck in local minimums~\citep{zhang2017understanding}. 
\figurename~\ref{fig-plot-generalization} illustrates on an example that the lack of labeled data can sometimes be compensated by the addition of synthetic data.

This phenomenon, also known as \emph{overfitting} in the machine learning community, can be solved using different techniques such as regularization or simply collecting more \emph{labeled} data~\citep{zhang2017understanding} (which in some domains are hard to obtain).  
Another well-known technique is data augmentation, where synthetic data are generated using a specific method. 
For example, images containing street numbers on houses can be slightly rotated without changing what number they actually are~\citep{krizhevsky2012imagenet}.  
For deep learning models, these methods are usually proposed for image data and do not generalize well to time series~\citep{um2017data}.
This is probably due to the fact that for images, a visual comparison can confirm if the transformation (such as rotation) did not alter the image's class, while for time series data, one cannot easily confirm the effect of such ad-hoc transformations on the nature of a time series.  
This is the main reason why data augmentation for TSC have been limited to mainly two relatively simple techniques: slicing and manual warping, which are further discussed in the next subsection.   

In this section, we propose to leverage from a DTW based data augmentation technique specifically developed for time series, in order to boost the performance of a deep ResNet for TSC.
Our preliminary experiments reveal that data augmentation can drastically increase the accuracy for CNNs on some datasets while having a small negative impact on other datasets.
We finally propose to combine the decision of the two trained models and show how it can reduce significantly the rare negative effect of data augmentation while maintaining its high gain in accuracy on other datasets.

\subsection{Related work}
The most used data augmentation method for TSC is the slicing window technique, originally introduced for deep CNNs in~\cite{cui2016multi}. 
The method was originally inspired by the image cropping technique for data augmentation in computer vision tasks~\citep{zhang2017understanding}. 
This data transformation technique can, to a certain degree, guarantee that the cropped image still holds the same information as the original image. 
On the other hand, for time series data, one cannot make sure that the discriminative information has not been lost when a certain region of the time series is cropped. 
Nevertheless, this method was used in several TSC problems, such as in~\cite{krell2018data} where it improved the SVMs accuracy for classifying electroencephalographic time series. 
In~\cite{kvamme2018predicting}, this slicing window technique was also adopted to improve the CNNs' mortgage delinquency prediction using customers' historical transactional data. 
In addition to the slicing window technique, jittering, scaling, warping and permutation were proposed in~\cite{um2017data} as generic time series data augmentation approaches.
The authors in~\cite{um2017data} proposed an additional data augmentation method specific to wearable sensor time series data that rotates the trajectory of a person's arm around an axis (e.g. the $x$ axis). 

In~\cite{leguennec2016data}, the authors proposed to extend the slicing window technique with a warping window that generates synthetic time series by warping the data through time. 
This method was used to improve the classification of their deep CNN for TSC, which was also shown to significantly decrease the accuracy of a NN-DTW classifier when compared to our adopted data augmentation algorithm~\citep{forestier2017generating}.
We should note that the use of a window slicing technique means that the model should classify each subsequence alone and then finally classify the whole time series using a majority voting approach. 
Alternatively, our method does not crop time series into shorter subsequences which enables the network to learn discriminative properties from the whole time series in an end-to-end manner.

\subsection{Method}
\subsubsection{Architecture}
We have chosen to improve the generalization capability of the deep ResNet proposed in~\cite{wang2017time} for two main reasons, whose corresponding architecture is illustrated in \figurename~\ref{ch-1-fig-resnet-archi}.  
First, by adopting an already validated architecture, we can attribute any improvement in the network's performance solely to the data augmentation technique. 
The second reason is that ResNet~\citep{wang2017time}, to the best of our knowledge, is the deepest neural network validated on large number of TSC tasks (such as the UCR/UEA archive~\citep{ucrarchive}), which according to the deep learning literature will benefit the most from the data augmentation techniques as opposed to shallow architectures~\citep{bengio2011deep}.
Deep ResNets were first proposed by~\cite{he2016deep} for computer vision tasks.
They are mainly composed of convolutions, with one important characteristic: the residual connections which acts like shortcuts that enable the flow of the gradient directly through these connections. 

The input of this network is a univariate time series with a varying length $l$. 
The output consists of a probability distribution over the $C$ classes in the dataset. 
The network's core contains three residual blocks followed by a Global Average Pooling layer and a final softmax classifier with $C$ neurons. 
Each residual block contains three 1-D convolutions of respectively 8, 5 and 3 filter lengths.
Each convolution is followed by a batch normalization~\citep{ioffe2015batch} and a ReLU as the activation function.
The residual connection consists in linking the input of a residual block to the input of its consecutive layer with the simple addition operation. 
The number of filters in the first residual blocks is set to 64 filters, while the second and third blocks contain 128 filters.  

All network's parameters were initialized using Glorot's Uniform initialization method~\citep{glorot2010understanding}. 
These parameters were learned using Adam~\citep{kingma2015adam} as the optimization algorithm. 
Following~\cite{wang2017time}, without any fine-tuning, the learning rate was set to $0.001$ and the exponential decay rates of the first and second moment estimates were set to $0.9$ and $0.999$ respectively.
Finally, the categorical cross-entropy was used as the objective cost function during the optimization process.  

\subsubsection{Data augmentation}

The data augmentation method we have chosen to test with this deep architecture, was first proposed in~\cite{forestier2017generating} to augment the training set for a 1-NN coupled with the DTW distance in a cold start simulation problem.
In addition, the 1-NN was shown to sometimes benefit from augmenting the size of the train set even when the whole dataset is available for training.
Thus, we hypothesize that this synthetic time series generation method should improve deep neural network's performance, especially that the generated examples in~\cite{forestier2017generating} were shown to closely follow the distribution from which the original dataset was sampled. 
The method is mainly based on a weighted form of DBA technique~\citep{petitjean2011a,petitjean2016faster,petitjean2014dynamic}. 
The latter algorithm averages a set of time series in a DTW induced space and by leveraging a weighted version of DBA, the method can thus create an infinite number of new time series from a given set of time series by simply varying these weights.
Three techniques were proposed to select these weights, from which we chose only one in our approach for the sake of simplicity, although we consider evaluating other techniques in our future work.
The weighting method is called Average Selected which consists of selecting a subset of close time series and fill their bounding boxes. 

We start by describing in details how the weights are assigned, which constitutes the main difference between an original version of DBA and the weighted version originally proposed in~\cite{forestier2017generating}. 
Starting with a random initial time series chosen from the training set, we assign it a weight equal to $0.5$. 
The latter randomly selected time series will act as the initialization of DBA.
Then, we search for its 5 nearest neighbors using the DTW distance. 
We then randomly select 2 out these 5 neighbors and assign them a weight value equal to 0.15 each, making thus the total sum of assigned weights till now equal to $0.5+2\times0.15=0.8$.
Therefore, in order to have a normalized sum of weights (equal to 1), the rest of the time series in the subset will share the rest of the weight $0.2$. 
We should note that the process of generating synthetic time series leveraged only the training set thus eliminating any bias due to having seen the test set's distribution. 

As for computing the average sequence, we adopted the DBA algorithm in our data augmentation framework.
Although other time series averaging methods exist in the literature, we chose the weighted version of DBA since it was already proposed as a data augmentation technique to solve the cold start problem when using a nearest neighbor classifier~\citep{forestier2017generating}.
Therefore we emphasize that other weighted averaging methods such as soft-DTW~\citep{cutur2017soft} and TEKA~\citep{marteau2019times} could be used instead of DBA in our framework, but we leave such exploration for our future work. 

We did not test the effect of imbalanced classes in the training set and how it could affect the model's generalization capabilities.  
Note that imbalanced time series classification is a recent active area of research that merits an empirical study of its own~\citep{geng2018cost}.  
At last, we should add that the number of generated time series in our framework was chosen to be equal to double the amount of time series in the most represented class (which is a hyper-parameter of our approach that we aim to further investigate in our future work).   

\subsection{Results}

We evaluated the data augmentation method for ResNet on the UCR/UEA archive~\citep{ucrarchive}, which is the largest publicly available TSC benchmark.
The archive is composed of datasets from different real world applications with varying characteristics such the number of classes and the size of the training set. 
Finally, for training the deep learning models, we leveraged the high computational power of more than 60 GPUs in one huge cluster\footnote{Our source code is available on \url{https://github.com/hfawaz/aaltd18}}
We should also note that the same parameters' initial values were used for all compared approaches, thus eliminating any bias due to the random initialization of the network's weights.

\begin{figure}
	\centering
	\subfloat[ResNet with/without augmentation]{
		
		\includegraphics[width=0.47\linewidth]{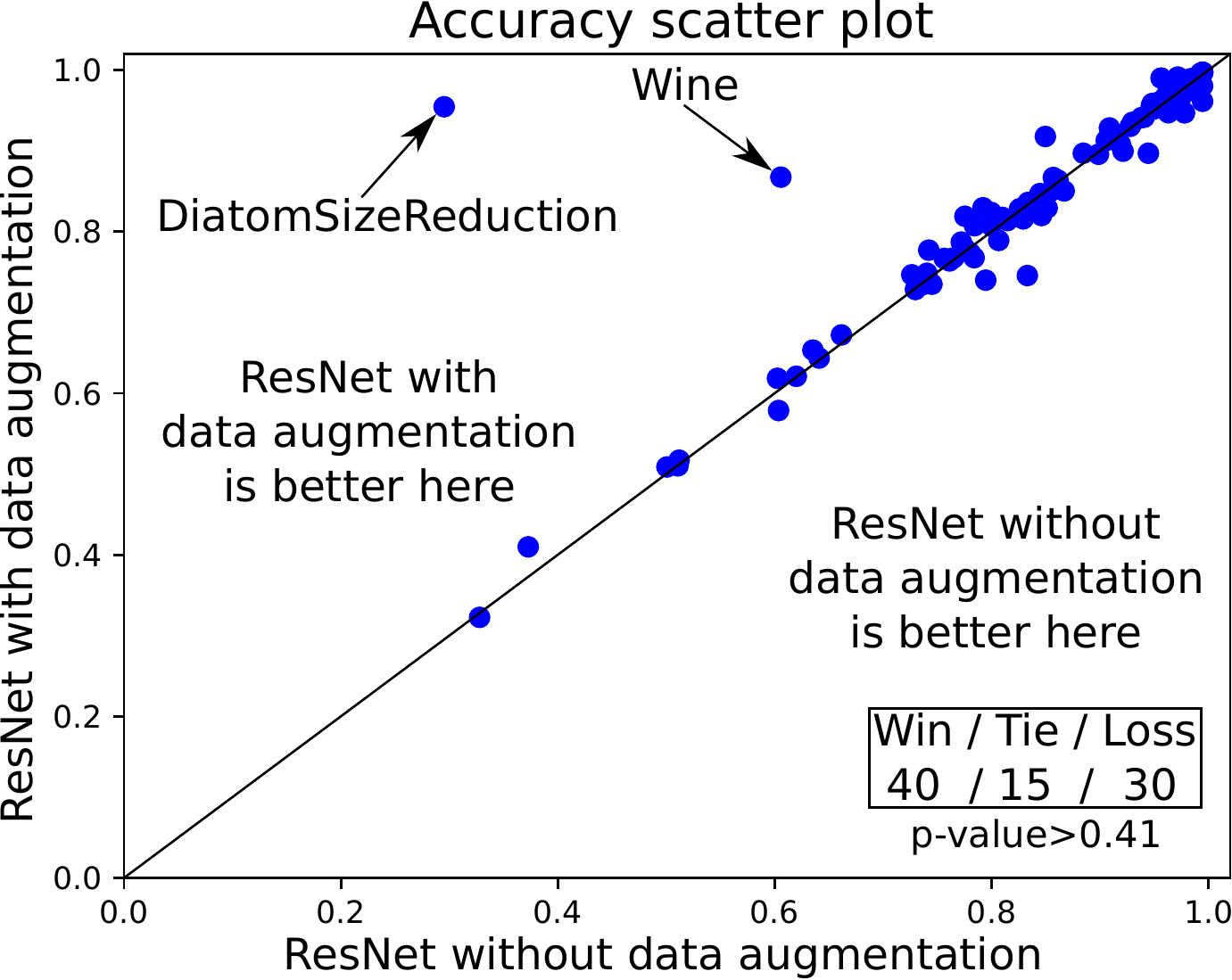}
		\label{sub-data-augment}}
	\hspace{.1cm}
	\subfloat[ResNet ensemble vs ResNet]{ 
		\includegraphics[width=0.47\linewidth]{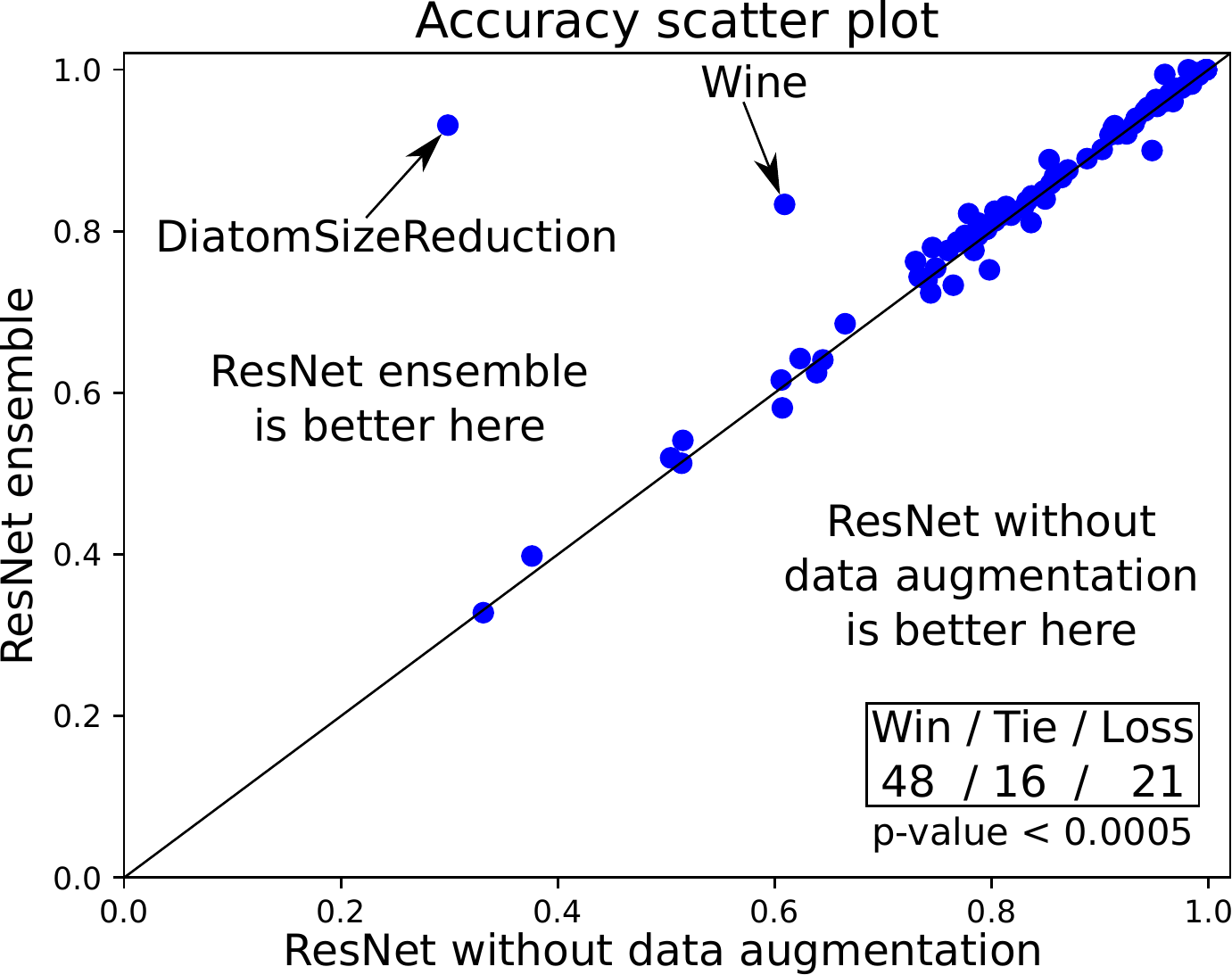}
		\label{sub-ensemble}
	}
	\caption{Accuracy of ResNet with and/or without data augmentation.}
	\label{fig-scatters}
\end{figure}

Our results show that data augmentation can drastically improve the accuracy of a deep learning model while having a small negative impact on some datasets in the worst case scenario.
\figurename~\ref{sub-data-augment} shows the difference in accuracy between ResNet with and without data augmentation, it shows that the data augmentation technique does not lead a significant decrease in accuracy. 
Additionally, we observe a huge increase of accuracy for the DiatomSizeReduction dataset (the accuracy increases from $30\%$ to $96\%$ when using data augmentation). 

This result is very interesting for two main reasons. 
First, DiatomSizeReduction has the smallest training set in the UCR/UEA archive~\citep{ucrarchive} (with 16 training instances), which shows the benefit of increasing the number of training instances by generating synthetic time series. 
Secondly, the DiatomSizeReduction dataset is the one where ResNet yield the worst accuracy without augmentation. 
On the other hand, the NN coupled with DTW (or ED) gives an accuracy of $97\%$ which shows the relative easiness of this dataset where time series exhibit similarities that can be captured by the simple Euclidean distance, but missed by the deep ResNet due to the lack of training data (which is compensated by our data augmentation technique). 
The results for the Wine dataset (57 training instances) also show an important improvement when using data augmentation.  

While we did show that deep ResNet can benefit from synthetic time series on some datasets, we did not manage to show any significant improvement over the whole UCR/UEA archive ($p$-value $> 0.41$ for the Wilcoxon signed rank test).
Therefore, we decided to leverage an ensemble technique where we take into consideration the decisions of two ResNets (trained with and without data augmentation).
In fact, we average the a posteriori probability for each class over both classifier outputs, then assign for each time series the label for which the averaged probability is maximum, thus giving a more robust approach to out-of-sample generated time series. 
The results in \figurename~\ref{sub-ensemble} show that the datasets which benefited the most from data augmentation exhibit almost no change to their accuracy improvement. 
While on the other hand the number of datasets where data augmentation harmed the model's accuracy decreased from $30$ to $21$.  
The Wilcoxon signed rank test shows a significant difference ($p$-value $< 0.0005$).
The ensemble's results are in compliance with the recent consensus in the TSC community, where ensembles tend to improve the individual classifiers' accuracy~\citep{bagnall2017the}. 

\subsection{Conclusion}
In this section, we showed how overfitting small time series datasets can be mitigated using a recent data augmentation technique that is based on DTW and a weighted version of the DBA algorithm. 
These findings are very interesting since no previous observation made a link between the space induced by the classic DTW and the features learned by the CNNs, whereas our experiments showed that by providing enough time series, CNNs are able to learn time invariant features that are useful for classification.  

In our future work, we aim to further test other variant weighting schemes for the DTW-based data augmentation technique, while providing a method that predicts when and for which datasets, data augmentation would be beneficial.

\section{Adversarial examples}



As we have previously discussed, TSC problems are encountered in various real world data mining tasks ranging from health care~\citep{abdelfattah2018augmenting,ma2018health,IsmailFawaz2018evaluating} and security~\citep{tan2017indexing,tobiyama2016malware} to food safety~\citep{briandet1996discrimination,nawrocka2013determination} and power consumption monitoring~\citep{owen2012powering,zheng2018wide}.
With deep learning models revolutionizing many machine learning fields such as computer vision~\citep{krizhevsky2012imagenet} and natural language processing~\citep{yang2018investigating,wang2018hierarchical}, we have shown in the previous chapter how researchers recently started to adopt these models for TSC tasks~\citep{IsmailFawaz2018deep}.

Following the advent of deep learning, researchers started to study the vulnerability of deep networks to adversarial attacks~\citep{yuan2017adversarial}.
In the context of image recognition, an adversarial attack consists in modifying an original image so that the changes are almost undetectable by a human~\citep{yuan2017adversarial}.
The modified image is called an adversarial image, which will be misclassified by the neural network, while the original one is correctly classified. 
One of the most famous real-life attacks consists in altering a traffic sign image so that it is misinterpreted by an autonomous vehicle~\citep{eykholt2018robust}.
Another application is the alteration of illegal content to make it undetectable by automatic moderation algorithms~\citep{yuan2017adversarial}.
The most common attacks are gradient-based methods, where the attacker modifies the image in the direction of the gradient of the loss function with respect to the input image thus increasing the misclassification rate~\citep{goodfellow2015explaining,kurakin2017adversarial,yuan2017adversarial}. 

\begin{figure}
	\centering
	\includegraphics[width=.8\linewidth]{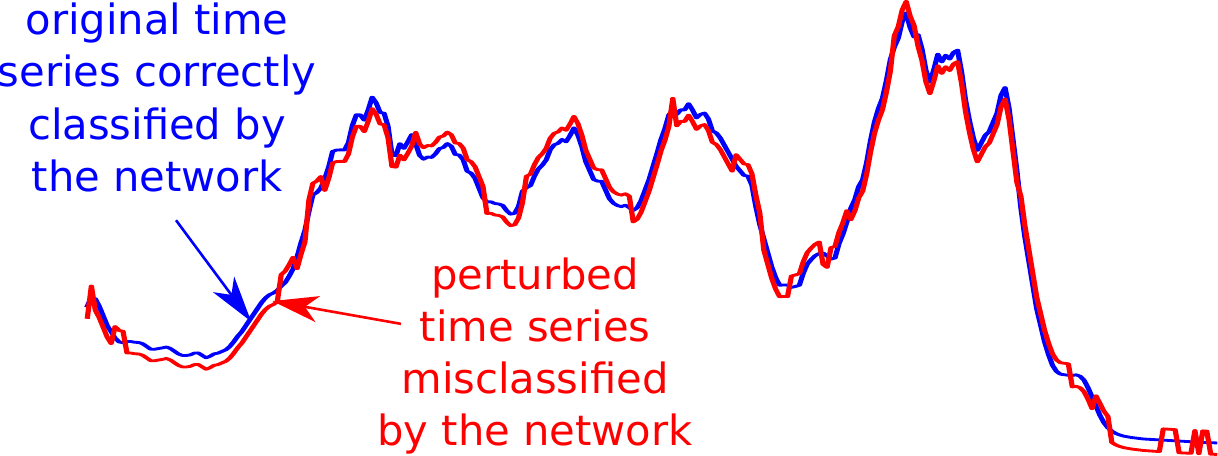}
	\caption{Example of a perturbed time series that is misclassified by a deep network after applying a small perturbation (time series from the Coffee dataset~\citep{ucrarchive} containing spectrographs of coffee beans).}
	\label{fig-adv-attack-example-coffee}
\end{figure}

While these approaches have been intensely studied in the context of image recognition (e.g. NeurIPS competition on Adversarial Vision Challenge), adversarial attacks haven not been thoroughly explored for TSC.
This is surprising as deep learning models are getting more and more popular to classify time series~\citep{IsmailFawaz2019deep,wang2017time,ma2018health,zheng2018wide,IsmailFawaz2018data,IsmailFawaz2018transfer}.
Furthermore, potential adversarial attacks are present in many applications where the use of time series data is crucial.
For example, Figure~\ref{fig-adv-attack-example-coffee} shows an original and perturbed time series of coffee beans spectrograph. 
While a deep neural network correctly classifies the original time series as Robusta beans, adding small perturbations makes it classify it as Arabica.
Therefore, since Arabica beans are more valuable than Robusta beans, this attack could be used to deceive food control tests and eventually the consumers.

In this section, we present, transfer and adapt adversarial attacks that have been shown to work well on images to time series data. 
We also present an experimental study using the 85 datasets of the UCR/UEA archive~\citep{ucrarchive} which reveals that neural networks are prone to adversarial attacks.
We highlight specific real-life use cases to stress the importance of such attacks in real-life settings, namely food quality and safety, vehicle sensors and electricity consumption.
Our findings show that deep networks for time series data are vulnerable to adversarial attacks like their computer vision counterparts.
Therefore, this work sheds the light on the need to protect against such attacks, especially when deep learning is used for sensitive TSC applications. 
We also show that adversarial time series learned using one network architecture can be transferred to different architectures.
We then discuss some mechanisms to prevent these attacks while making the models more robust to adversarial examples.
Finally, in the spirit of regularizing DNNs, we show how these perturbed time series can be leveraged in order to improve the generalization capability of a deep learning model: a technique called Adversarial Training~\citep{xie2019adversarial}. 


\subsection{Background}

In this subsection, we start with the necessary definitions for ease of understanding. 
We then follow by an overview of critical applications based on deep learning approaches for TSC where adversarial attacks could have serious and dangerous consequences.
Finally, we present a brief survey of the current state-of-the-art methods for adversarial attacks which have been mainly proposed and validated on image datasets~\citep{yuan2017adversarial}.  
%
%
%

\begin{definition}
	$f(\cdot) \in F: \mathbb{R}^T \rightarrow \hat{Y} $ represents a deep learning model for TSC.
\end{definition}

\begin{definition}
	$J_f(\cdot , \cdot)$ denotes the loss function (e.g.\ cross-entropy) of the model $f$.
\end{definition}

\begin{definition}\label{def-adv-ex}
	$X^{'}$ denotes the adversarial example, a perturbed version of $X$ (the original instance) such that $\hat{Y}\ne \hat{Y}^{'}$ and $\norm{X-X^{'}}_p\leq \epsilon$.
\end{definition}


In this section, we focus on the application of DNNs in crucial and sensitive decision making systems, thus motivating the investigation of neural network's vulnerabilities to adversarial examples.  
In~\cite{ma2018health}, CNNs were used to mine temporal electronic health data for risk prediction and disease sub-typing.  
In situations where algorithms are taking the decision for reimbursement of medical treatment, tampering medical records in an imperceptible manner could eventually lead to fraud.    
Apart from the health care industry, deep CNNs are also being used when monitoring power consumption from houses or factories. 
For example in~\cite{zheng2018wide}, time series data from smart grids were analyzed for electricity theft detection, where in such use cases perturbed data can help thieves to avoid being detected.
Other crucial decision making systems such as malware detection in smart-phones, leverage temporal data in order to classify if an Android application is malicious or not~\citep{tobiyama2016malware}. 
Using adversarial attacks, a hacker might generate synthetic data from his/her application allowing it to bypass the security systems and get it installed on the end user's smart-phone.
Finally, when deep neural networks are deployed for road anomaly detection~\citep{cabral2018an}, perturbing the data recorded by sensors placed on the road could help the entities responsible for such life threatening anomalies, to avoid being captured. 
We should note that this list of potential attacks is not exhaustive.

\begin{figure*}[th!]
	\centering
	\includegraphics[width=\linewidth]{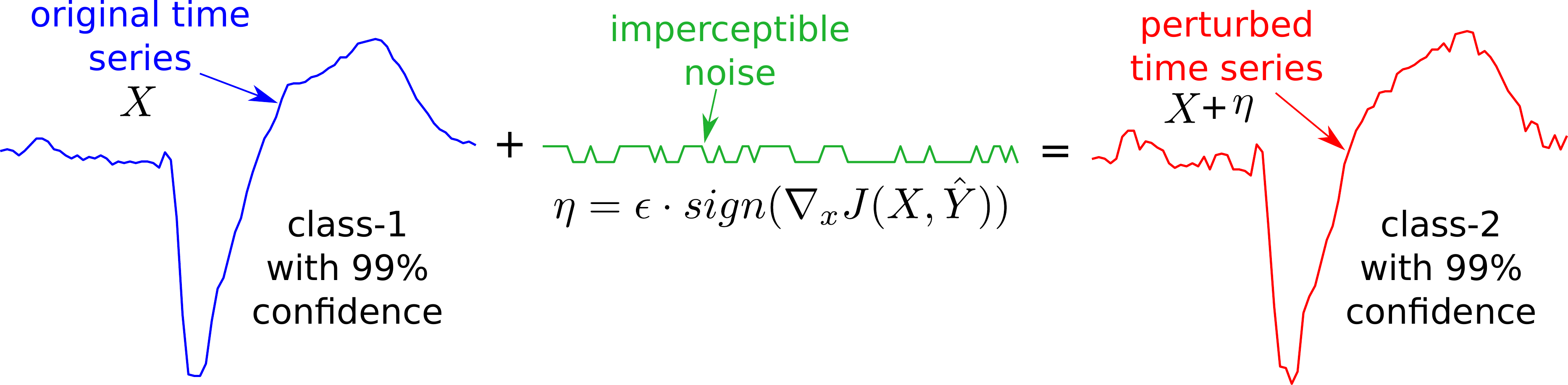}
	\caption{Example of perturbing the classification of an input time series from the TwoLeadECG~\citep{ucrarchive} dataset by adding an imperceptible noise computed using the Fast Gradient Sign Method.}
	\label{fig-pert-example}
\end{figure*}

\subsubsection{Adversarial attacks}
\cite{szegedy2014intriguing} were the first to introduce adversarial examples against deep neural networks for image recognition tasks in 2014. 
Following these intriguing findings, a huge amount of research has been dedicated to generating, understanding and preventing adversarial attacks on deep neural networks~\citep{goodfellow2015explaining,kurakin2017adversarial,eykholt2018robust}.   

Most of these methods have been proposed for image recognition tasks~\citep{yuan2017adversarial}. 
For example, in~\cite{goodfellow2015explaining}, a fast gradient-based attack was developed as an alternative to expensive optimization techniques~\citep{szegedy2014intriguing}, where the authors explained the presence of such adversarial examples with the hypothesis of linearity for deep learning models.
This kind of attack was also extended by a more costly iterative procedure~\citep{kurakin2017adversarial}, where the authors showed for the first time that even printed adversarial images (\emph{i.e.} perceived by a camera) are able to fool a pre-trained network.   
More recently, it has been shown that perturbing stop signs can trick autonomous vehicles into misclassifying it as a speed limit sign~\citep{eykholt2018robust}. 

Other fields such as Natural Language Processing have also been investigated to create adversarial attacks such as adding distracting phrases at the end of a paragraph in order to show that deep learning-based reading comprehension systems were not able to distinguish subtle differences in text~\citep{robin2017adversarial}.  
For a review on the different adversarial attacks for deep learning systems, we refer the interested readers to a recent survey in~\cite{yuan2017adversarial}.

For general TSC tasks, it is surprising how adversarial attack approaches have been ignored by the community.
The only previous work mentioning attacks for TSC is~\cite{oregi2018adversarial}. 
By adapting a soft KNN coupled with DTW, the authors showed that adversarial examples could fool the proposed nearest neighbors classifier on a single simulated dataset (synthetic\_control from the UCR/UEA archive~\cite{ucrarchive}). 
However, the fact that the KNN classifier is no longer considered as the state-of-the-art classifier for time series data~\citep{bagnall2017the}, we believe that it is important to investigate the generation of adversarial time series examples that deteriorate the accuracy of state-of-the-art classifiers such as ResNet~\citep{IsmailFawaz2018deep,wang2017time} and to validate it on the whole 85 datasets in the UCR/UEA archive.
Finally, we formally define an adversarial attack on deep neural networks for TSC.
\begin{definition}
	Given a trained deep learning model $f$ and an original input time series $X$, generating an adversarial instance $X^{'}$ can be described as a box-constrained optimization problem.
	\begin{multline}\label{eq-optim}
	\min_{X^{\theta}} \norm{X^{'}-X}~s.t.\\
	f(X^{'})=\hat{Y^{'}},~f(X)=\hat{Y}~and~\hat{Y}\ne\hat{Y^{'}}
	\end{multline}
\end{definition}
Let $\eta=X-X^{'}$ be the perturbation added to $X$, which corresponds to a very low amplitude signal. 
Figure~\ref{fig-pert-example} illustrates this process where the green time series corresponds to the added perturbation $\eta$.
The optimization problem in (\ref{eq-optim}) minimizes the perturbation while misclassifying the input time series.

\subsection{Adversarial attacks for time series}

In this subsection, we present two attack methods that we then use to generate adversarial time series examples for the ResNet model described in Chapter~\ref{Chapter1}. 
%
%
%
For testing adversarial examples, we used ResNet, which was originally proposed for TSC in~\cite{wang2017time}, where it was validated on 44 datasets from the UCR/UEA archive~\citep{ucrarchive}. 
In Chapter~\ref{Chapter1}, we identified that ResNet achieved state-of-the-art performance for TSC, with results that are not significantly different than the HIVE-COTE, the current state-of-the-art classifier, which is an ensemble of 37 classifiers~\citep{lines2018time}. 
Note that our adversarial attack methods are independent of the chosen network architecture, and that we chose ResNet for its robustness~\citep{IsmailFawaz2018deep} as well as its use in many critical domains such as malware detection~\citep{cabral2018an}.
In addition, adversarial examples are known to be transferable across different neural network architectures which enables the synthetic time series to fool other deep learning models: a technique known as black-box attack~\citep{yuan2017adversarial}.
Finally, we describe a very recent method called AdvProp~\citep{xie2019adversarial} that leverages adversarial training in order to improve image recognition systems, which we implement and apply to our TSC problem.

\subsubsection{Fast Gradient Sign Method}
FGSM was first proposed in~\cite{goodfellow2015explaining} to generate adversarial images that fooled the famous GoogLeNet model.
FGSM is considered  ``fast'' and replaces the expensive linear search method previously proposed in~\cite{szegedy2014intriguing}. 
The attack is based on a one step gradient update along the direction of the gradient's sign at each time stamp. 
The perturbation process  (illustrated in Figure~\ref{fig-pert-example}) can be expressed as:
\begin{equation}
\eta = \epsilon \cdot sign(\nabla_x J(X,\hat{Y}))  
\end{equation}
where $\epsilon$ denotes the magnitude of the perturbation (a hyperparameter). 
The adversarial time series $X^{'}$ can be easily generated with $X^{'}=X+\eta$. 
The gradient can be efficiently computed using back-propagation. 

\subsubsection{Basic Iterative Method}
BIM extends FGSM by applying it multiple times with a small step size and clip the obtained time series elements after each step to ensure that they are in an $\epsilon$-neighborhood of the original time series~\citep{kurakin2017adversarial}. 
In fact, by adding smaller changes or perturbations in an iterative manner, the method is able to generate adversarial examples that are closer to the original samples and have a better chance of fooling the network. 
Algorithm~\ref{algo-bim} shows the different steps of this iterative attack which requires setting three hyperparameters: (1) the number of iterations $I$; (2) the amount of maximum perturbation $\epsilon$ and (3) the per step small perturbation $\alpha$.  
In our experiments we have set $\epsilon=0.1$ heuristically similarly to~\cite{goodfellow2015explaining,kurakin2017adversarial} and the rest of BIM hyperparameters were left at their default value in the Cleverhans API~\citep{papernot2018cleverhans}. 

\begin{algorithm}
	\begin{algorithmic}[1]
		\caption{Iterative Adversarial Attack}
		\label{algo-bim}
		
		\renewcommand{\algorithmicensure}{\textbf{Parameter:}}
		\renewcommand{\algorithmicrequire}{\textbf{Input:}}
		\renewcommand{\algorithmicreturn}{\textbf{Output:}}
		
		\Ensure $I$, $\epsilon$, $\alpha$
		\Require original time series $X$ \& its label $\hat{Y}$
		\Return perturbed time series $X^{'}$
		
		\State $X^{'}\leftarrow X$\
		\For {$i=1$ to $I$}
		\State$\eta = \alpha \cdot sign(\nabla_{x}J(X^{'},\hat{Y}))$
		\State $X^{'}=X^{'}+\eta$
		\State $X^{'}=min\{X+\epsilon,max\{X-\epsilon,X^{'}\}\}$
		\EndFor
	\end{algorithmic}
\end{algorithm}

\subsubsection{Adversarial Training}
Adversarial training consists of generating adversarial examples during the training process of a neural network classifier. 
These adversarial instances are used as training examples in order to defend against adversarial attacks~\citep{goodfellow2015explaining}.
This technique currently constitutes the foundation of state-of-the-arts for defending against these attacks~\citep{xie2019feature}. 
Most previous approaches have witnessed a decrease in accuracy on clean original instances when the model is trained in adversarial setting~\citep{tsipras2018there}, showing an inevitable trade-off between a model's accuracy and its robustness to adversarial attacks. 
However,~\cite{xie2019adversarial} were the first to show that one can leverage adversarial training to improve the test time accuracy of a DNN. 
They managed to achieve this by adding a using two different batch normalization layers~\citep{ioffe2015batch}: one for the clean original intact training instances and a second one for the perturbed examples.
\cite{xie2019adversarial} argued that the main reason behind the deterioration in accuracy when using adversarial training was that the perturbed instances come from a distribution that is significantly different than the clean original ones. 
Thus having different normalization layers should allow the network to adapt its weights for each distribution. 
In our case, we have implemented AdvProp and tested its effect when training DNNs with adversarial training for TSC. 

\subsection{Results}

\subsubsection{Experimental setup}
To train the deep neural network, we leveraged the parallel computation of a cluster of more than 60 GPUs (a mix of GTX 1080 Ti, Tesla K20, K40 and K80). 
All of our experiments were evaluated on the 85 datasets from the publicly available UCR/UEA archive~\citep{ucrarchive}.
The model was trained/tested using the original training/testing splits provided in the archive.  
To perform the attacks, we have adapted the Cleverhans API~\citep{papernot2018cleverhans} by extending the well known attacks for time series data and perturbed only the test instances without using the test labels, similarly to the computer vision literature~\citep{yuan2017adversarial}. 

For reproducibility and to allow the time series community to verify and build on our findings, the source code for generating adversarial time series is publicly available on our GitHub repository\footnote{\url{https://github.com/hfawaz/ijcnn19attacks}}.
In addition, we provide on our companion web page\footnote{\url{https://germain-forestier.info/src/ijcnn2019/}} the raw results, our pre-trained models as well as a set of perturbed time series for each dataset in the UCR/UEA archive.
This would allow time series data mining practitioners to test the robustness of their machine learning models against adversarial attacks.

\begin{figure}[t]
	\centering
	\includegraphics[width=.7\linewidth]{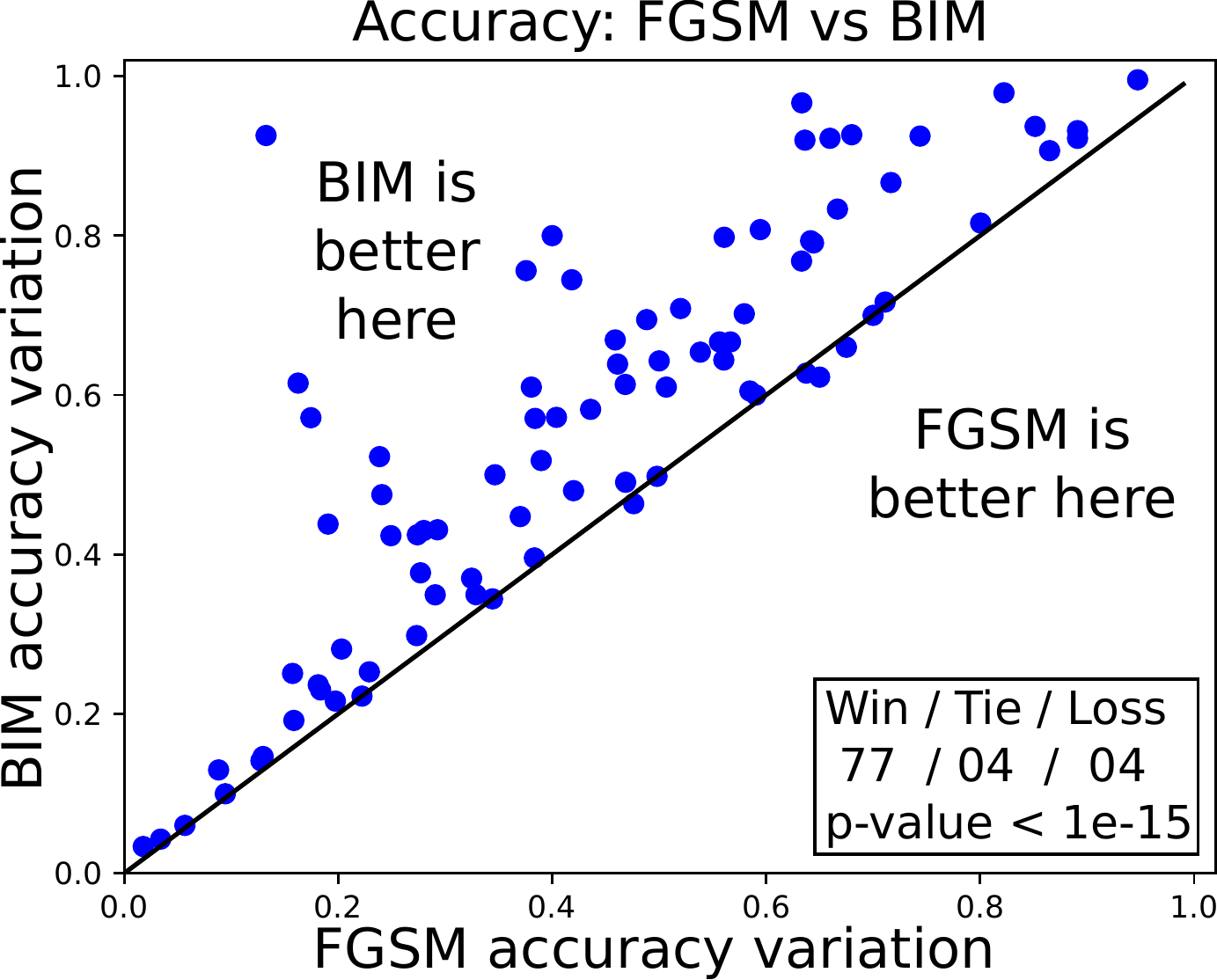}
	\caption{Accuracy variation for two attacks (FGSM and BIM) with respect to ResNet's original accuracy.}
	\label{fig-acc-plot}
\end{figure}

\subsubsection{Adversarial attacks on the whole UCR/UEA archive}
For all datasets, both attacks managed to reduce ResNet's accuracy. 
One exception is the DiatomSizeReduction dataset which is the smallest one in the archive with an already low original accuracy equal to 30\% due to overfitting~\citep{IsmailFawaz2018deep}.
Figure~\ref{fig-acc-plot} shows the accuracy variation for both attacks with respect to the network's original accuracy on the UCR/UEA archive.
On average, over the 85 datasets, FGSM and BIM managed to reduce the model's accuracy respectively by 43.2\% and 56.89\%. 
The Wilcoxon signed-rank test indicates that BIM is \emph{significantly} better than FGSM in decreasing the model's accuracy, with a $p$-value~$\le10^{-15}$.  
However, we should note that FGSM is a fast approach allowing real-time generation of adversarial time series whereas BIM is time-consuming and requires a certain number of iterations $I$. 

By analyzing the use-cases where both attacks failed to fool the classifier, we found out that the corresponding datasets have two interesting characteristics that could explain the classifier's robustness to adversarial examples. 
The first one is that 50\% of the simulated datasets (CBF, Two\_patterns and synthetic\_control) in the archive are in the top six hardest datasets to attack.
Perhaps since these are synthetic datasets generated by humans to serve some human intuition for TSC, small perturbations imperceptible by humans, are not enough to alter the classifier's decision.
The second observation is that a network trained on datasets with time series of short length is harder to fool. 
This is rather expected since the less data points we have, the less amount of perturbation the attacker is allowed to add. 
For example ItalyPowerDemand contains the shortest sequences ($T=24$) and is the second most hardest use-case for both attacks. 

\subsubsection{Multi-Dimensional Scaling}

We used Multi-Dimensional Scaling~\citep{kruskal1978multidimensional,forestier2017generating} (explained in Chapter~\ref{Chapter1}) with the objective to gain some insights on the spatial distribution of the perturbed time series compared to the original ones. 
Using the ED on a set of time series (original and perturbed), it is then possible to create a similarity matrix and apply MDS to display the set into a two dimensional space.
The latter straightforward approach supposes that the ED is able to strongly separate the raw time series, which is usually not the case evident by the low accuracy of the nearest neighbor when coupled with the ED~\citep{bagnall2017the}. 
Therefore, we decided to use the \emph{linearly} separable representation of time series from the output of the GAP layer, which is used as input to the softmax \emph{linear} classifier (multinomial logistic regression). 
More precisely, for each input time series, the last convolution outputs a multivariate time series whose dimensions are equal to the number of filters (128) in the last convolution, then the GAP layer averages the latter 128-dimensional multivariate time series over the time dimension resulting in a vector of 128 real values over which the ED is computed. 
This enables the MDS projection to be as close as possible to ResNet's latent representation of the time series. 
Obviously, one has to be careful about the interpretation of MDS output, as the data space is highly simplified (each time series $X_i$ is represented as a single data point in 2D space).

\subsubsection{Attacks on food quality and safety}

\begin{figure}[t]
	\centering
	\includegraphics[width=.7\linewidth]{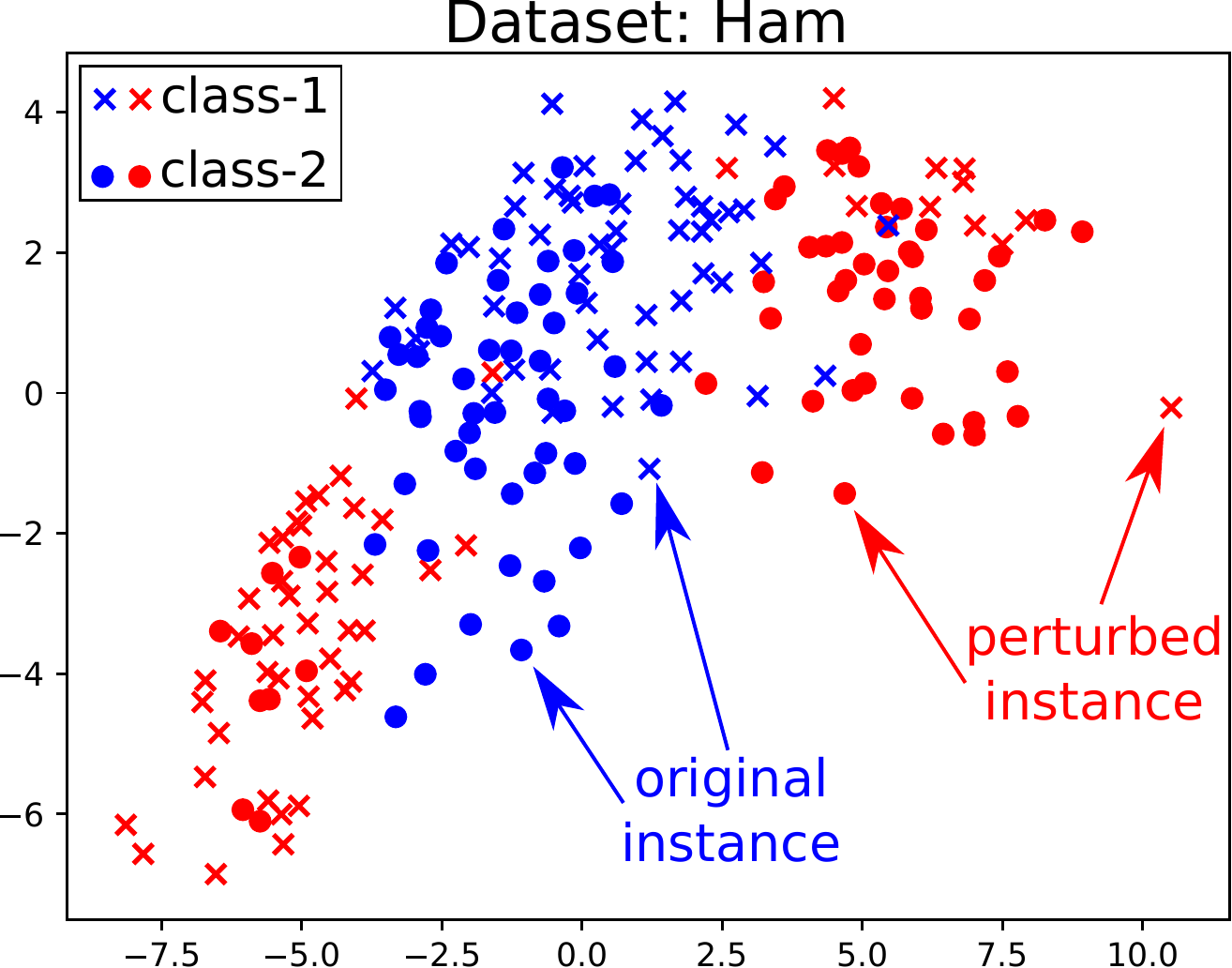}
	\caption{Multi-Dimensional Scaling showing the distribution of perturbed time series on the whole test set of the Ham dataset where the accuracy decreased from 80\% to 21\% after performing the BIM attack.}
	\label{fig-ham-mds}
\end{figure}

The determination of food quality, type and authenticity along with the detection of adulteration are major issues in the food industry~\citep{nawrocka2013determination}.
With meat related product, authenticity checking concerns for example the identification of substitution of high value raw materials with cheaper materials like less costly cuts, offal, blood, water, eggs or other types of proteins.
These substitutions not only decease the consumers but can also cause severe allergic responses as the substitute materials are hidden.
Discriminating meat that has been frozen-and-thawed from fresh meat is also an important issue, as refreezing food can result in an increased amount of bacteria.
Spectroscopic methods have been historically very successful at evaluating the quality of agricultural products, especially food~\citep{nawrocka2013determination}.
This technique is routinely used as a quality assurance tool to determine the composition of food ingredients.

In this context, an adversarial attack could be used to modify recorded spectrographs (seen as time series) in order to hide the low qualify of the food.
The Beef dataset~\citep{al2002detection} (from the UCR/UEA archive) contains four classes of beef spectrograms, from pure and adulterated beef with varying degrees of potential adulterants (heart, tripe, kidney, and liver).
An adversarial attack could thus consist in modifying an adulterated beef to make a network classify it as pure beef.
For this dataset, FGSM and BIM reduced the model's accuracy respectively by 56.7\% and 66.7\%. 


The Ham dataset~\citep{olias2006sodium} contains measurements from 19 Spanish and 18 French dry-cured hams, with the goal to distinguish the provenance of the food.
An adversarial attack could consist in perturbing the spectrograms to hide the real provenance of the food. 
Figure~\ref{fig-ham-mds} shows the MDS projection of the original and perturbed instances for Ham's test set, where one can see that the adversarial examples are \emph{pushed} toward the other class.

\begin{figure}[t]
	\centering
	\includegraphics[width=0.7\linewidth]{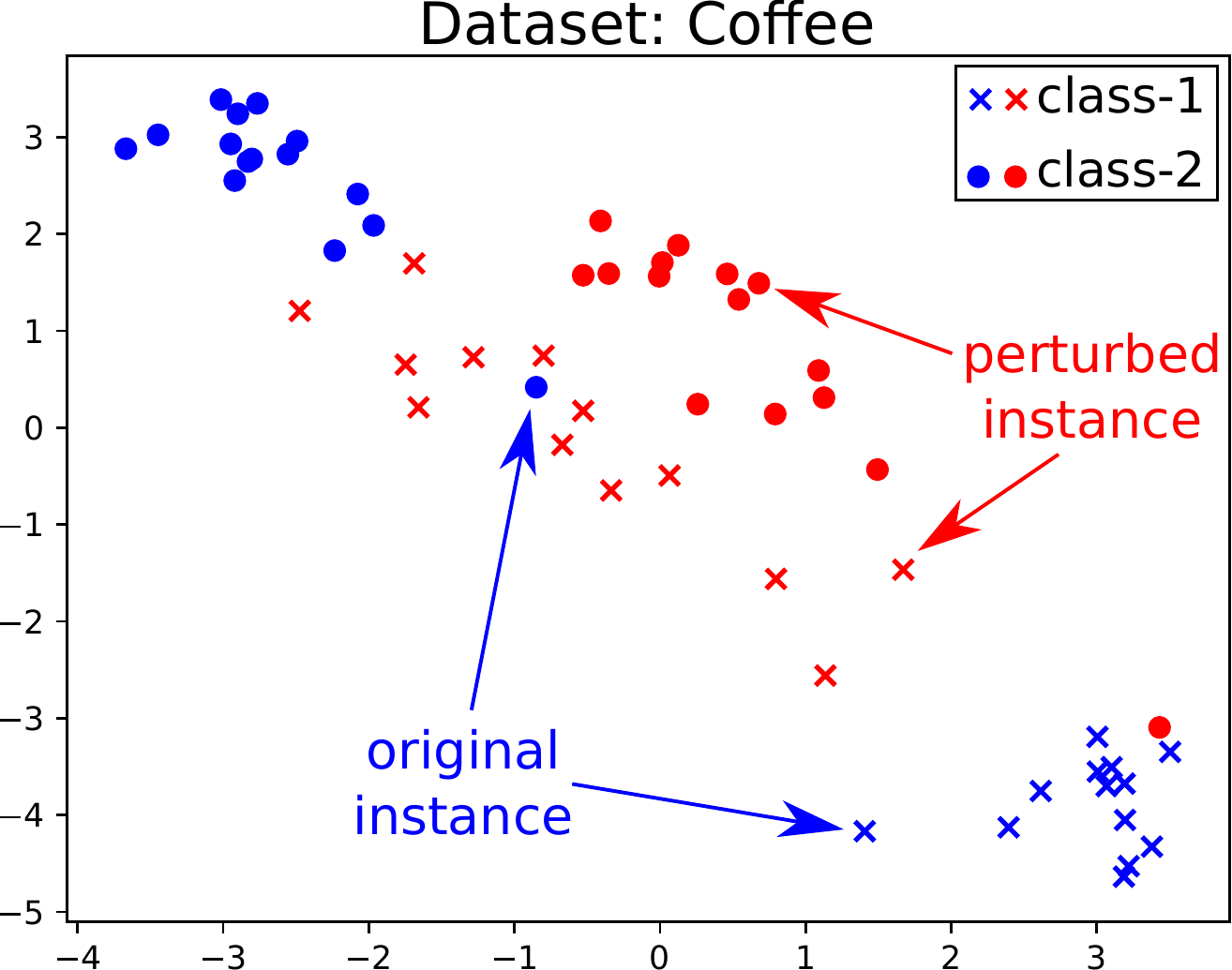}
	\caption{Multi-Dimensional Scaling showing the distribution of perturbed time series on the whole test set of the Coffee dataset where the accuracy decreased from 100\% to 50\% after performing the FGSM attack.}
	\label{fig-coffee-mds}
\end{figure}

The Coffee dataset~\citep{briandet1996discrimination} is a two class problem to distinguish between Robusta and Arabica coffee beans.
Arabica beans are valued most highly by the trade, as they are considered to have a finer flavor than Robusta.
An adversarial attack could consist in altering the spectrograms to make Robusta beans look like Arabica  beans.
Figure~\ref{fig-coffee-mds} shows the MDS representation of the original and perturbed time series from the test set. 
We can clearly see how the instances are \emph{pushed} toward the class frontiers after performing the FGSM attack. 

\subsubsection{Attacks on vehicle sensors}

The increase in the number of sensors and other electrical devices has drastically augmented the amount of data produced in the industry.
These data are now routinely used to monitor systems or to perform predictive maintenance and prevent failures~\citep{susto2015machine}.
The car industry is not an exception with the increasing number of sensors present in modern vehicles, especially for advanced driver assistance systems and autonomous driving.

Data are also used to perform diagnostic on vehicles in order to detect engine problems or compliance with environmental regulations.
In this context, an adversarial attack could consist in altering sensor readings in order to hide a specific problem or to pass a CO$_2$ emission test.
The famous ``dieselgate'' (or ``emissionsgate'')~\citep{brand2016beyond} made this kind of attack a reality as multiple automakers have been suspected of using emission control systems during laboratory emissions testing.

To illustrate this use case, we used the FordA datasets (from the UCR/UEA archive) that was was originally used in a competition in the IEEE World Congress on Computational Intelligence, 2008.
The classification problem is to diagnose whether a certain symptom exists or not in an automotive subsystem.
Each case consists of 500 measurements of engine noise and a class label.
In this context, an attack could consist in hiding an engine problem.
In practice for the FordA dataset, the model's accuracy decreased by 57.9\% and 70.2\% when applying respectively the FGSM and BIM attacks.  

\begin{figure}[t]
	\centering
	\includegraphics[width=0.7\linewidth]{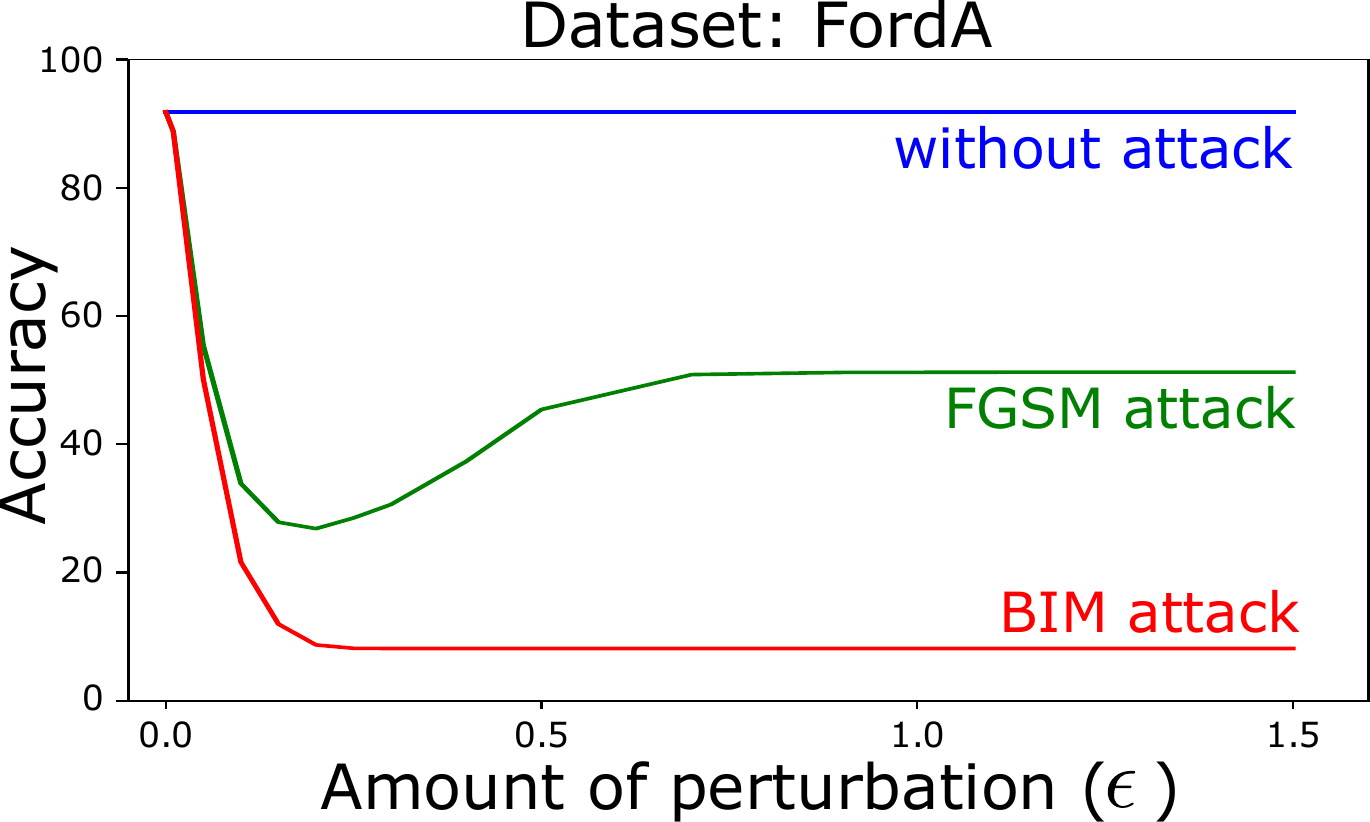}
	\caption{Accuracy variation with respect to the amount of perturbation for FGSM and BIM attacks on FordA.}
	\label{fig-plot-eps}
\end{figure}

Figure~\ref{fig-plot-eps} depicts the variations of ResNet's accuracy on FordA with respect to the amount of perturbation $\epsilon$ allowed for the FGSM and BIM attacks. 
As expected~\citep{kurakin2017adversarial}, we found that FGSM fails to generate adversarial examples that can fool the network for larger values of $\epsilon$, whereas the BIM produces perturbed time series that can reduce a model's accuracy to almost 0.0\%.
This can be explained by the fact that BIM adds a small amount of perturbation $\alpha$ on each iteration whereas FGSM adds $\epsilon$ amount of noise for each data point in the series that may not be useful for misclassifying the test sample. 

\subsubsection{Attacks on electricity consumption}

Smart meters are electronic devices that record electric power consumption while sending information to the electricity supplier for monitoring, billing and data analysis. 
These meters typically register energy hourly and report back at least once a day to the supplier by leveraging a two-way communication channel between the device and the supplier's central system. 
These smart meters have raised a set of concerns in public opinion especially because they send detailed information about how much electricity is being used for each time stamp. 
Precisely, it has been shown that it is possible to know exactly which type of electric device is or has been used from simply analyzing the electricity consumption data~\cite{owen2012powering}.
In this context, an attack could consist in modifying the electricity consumption time series of one device to make it recognized as another in order to hide which devices are actually used by a specific user.

To illustrate this use case, we used the SmallKitchenAppliances dataset from the UCR/UEA archive that was recorded as part of government sponsored study called Powering the Nation~\citep{owen2012powering}.
By collecting and analyzing behavioral data about consumers' daily use of electricity within their homes, the goal is to reduce the UK's carbon footprint. 
The dataset contains readings from 251 households recorded over a month. 
Each univariate time series has a length equal to 720 corresponding to 24 hours of readings taken every two minutes. 
The three classes are: Kettle, Microwave and Toaster.
For this dataset, FGSM and BIM managed to reduce the classifier's accuracy respectively by 38.4\% and 57\%. 

\begin{figure}[t]
	\centering
	\includegraphics[width=.8\linewidth]{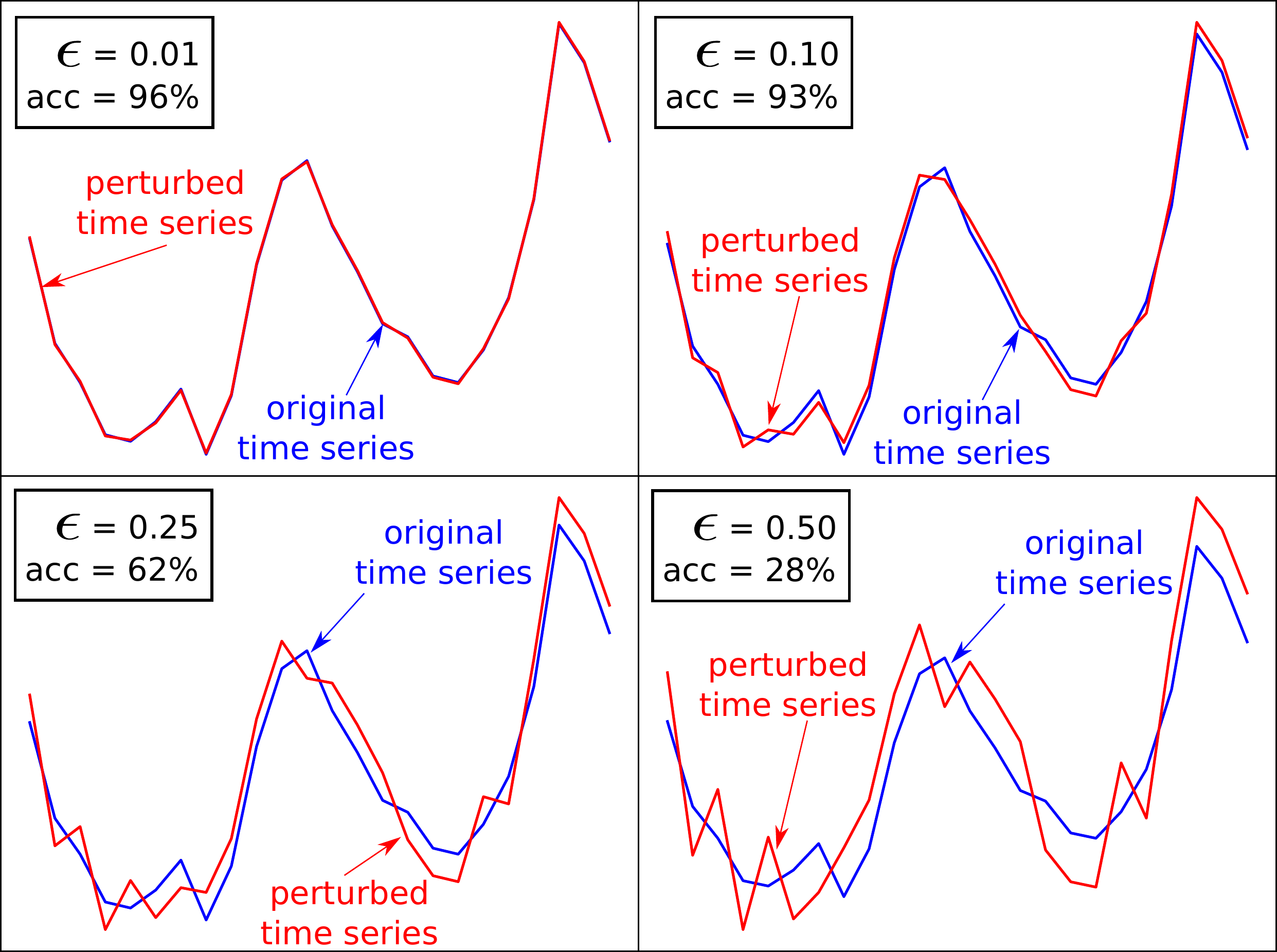}
	\caption{
		Accuracy variation for ItalyPowerDemand with respect to the perturbation $\epsilon$ where FGSM managed to fool the network with this example for $\epsilon\ge 0.3$. 
	}
	\label{fig-pert-epsilon}
\end{figure}

The ItalyPowerDemand dataset~\citep{keogh2006intelligent}, another dataset from the UCR/UEA archive, contains twelve monthly electrical power demand time series from Italy. 
The task is to differentiate between instances that correspond to winter months (October to March) and summer months (April to September). 
This dataset contains the shortest time series in the archive ($T=24$), thus requiring a higher amount of perturbation $\epsilon$ in order to be misclassified. 
Figure~\ref{fig-pert-epsilon} shows the variation of the model's accuracy as well as the shape of a time series from the ItalyPowerDemand dataset with respect to the amount of noise that is added. 
For this example, both attacks needed higher values of perturbation ($\epsilon\ge0.3$) rather than the default setting ($\epsilon=0.1$). 

\subsubsection{Are adversarial examples transferable?} 

To evaluate the transferability of perturbed time series, we used the FCN which was originally proposed in~\cite{wang2017time} and was shown in Chapter~\ref{Chapter1} to be the second most accurate deep time series classifier when evaluated on the UCR/UEA archive~\citep{ucrarchive}.
We used the test sets altered with FGSM and BIM using ResNet and try to classify it with FCN (both were originally trained on the same train set).
For both FGSM and BIM attacks, FCN's accuracy decreases respectively by 38.2\% and 42.8\% which shows that adversarial examples are capable of generalizing to a different network architecture. 
The Wilcoxon signed-rank test also shows that BIM is \emph{significantly} better than FGSM in reducing FCN's accuracy, with a $p$-value $\le10^{-7}$.
This type of attacks is known as ``black box'' where the attackers do not have access to the target model's internal parameters (FCN) yet they are able to generate perturbed time series that fool the classifier.

\subsubsection{How can we prevent such attacks?}

Countermeasures for adversarial attacks~\citep{yuan2017adversarial} follow two defense strategies: (1) \emph{reactive}: identify the perturbed instance; (2) \emph{proactive}: improve the network's robustness without generating adversarial examples. 
One of the most straightforward proactive methods is \emph{adversarial training}, which consists of (re)training the classifier with adversarial examples. 
Other reactive techniques consist of detecting the adversarial examples during testing.
However, most of these detectors are still prone to attacks that are designed specifically to fool the detectors~\citep{yuan2017adversarial}. 
Therefore, we think that the time series community would have much to offer in this area by leveraging the decades of research into non-probabilistic classifiers such as the nearest neighbor coupled with DTW~\citep{tan2018efficient}.  
Running classifiers against the adversarial examples that we provide here, is a first step toward identifying vulnerable models and making them more robust to such type of attacks.
Finally, a recent approach proposed by~\cite{abdu2020detecting} defended against adversarial attacks by framing the problem as an outlier detection task. 
By constructing a normalcy model based on information and chaos-theoretic measures,~\cite{abdu2020detecting} were able to determine whether unseen time series instances are normal or adversarial. 

\subsubsection{Can we leverage adversarial examples to improve generalization?}
\begin{figure}
	\centering
	\includegraphics[width=.7\linewidth]{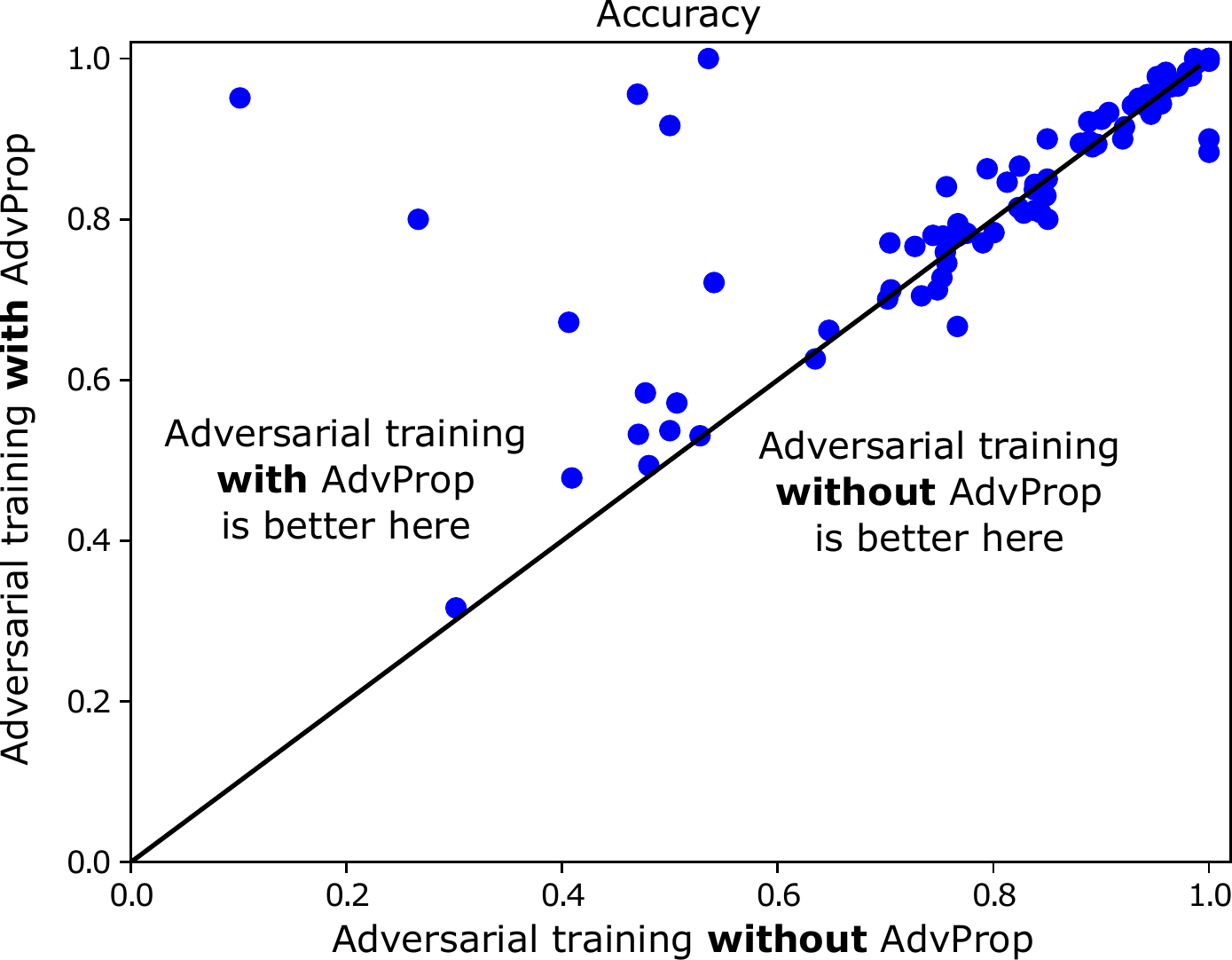}
	\caption{
		Accuracy of adversarial training with or without AdvProp.
	}
	\label{fig-advProp-pairwise-plot}
\end{figure}

\begin{figure}
	\centering
	\includegraphics[width=.7\linewidth]{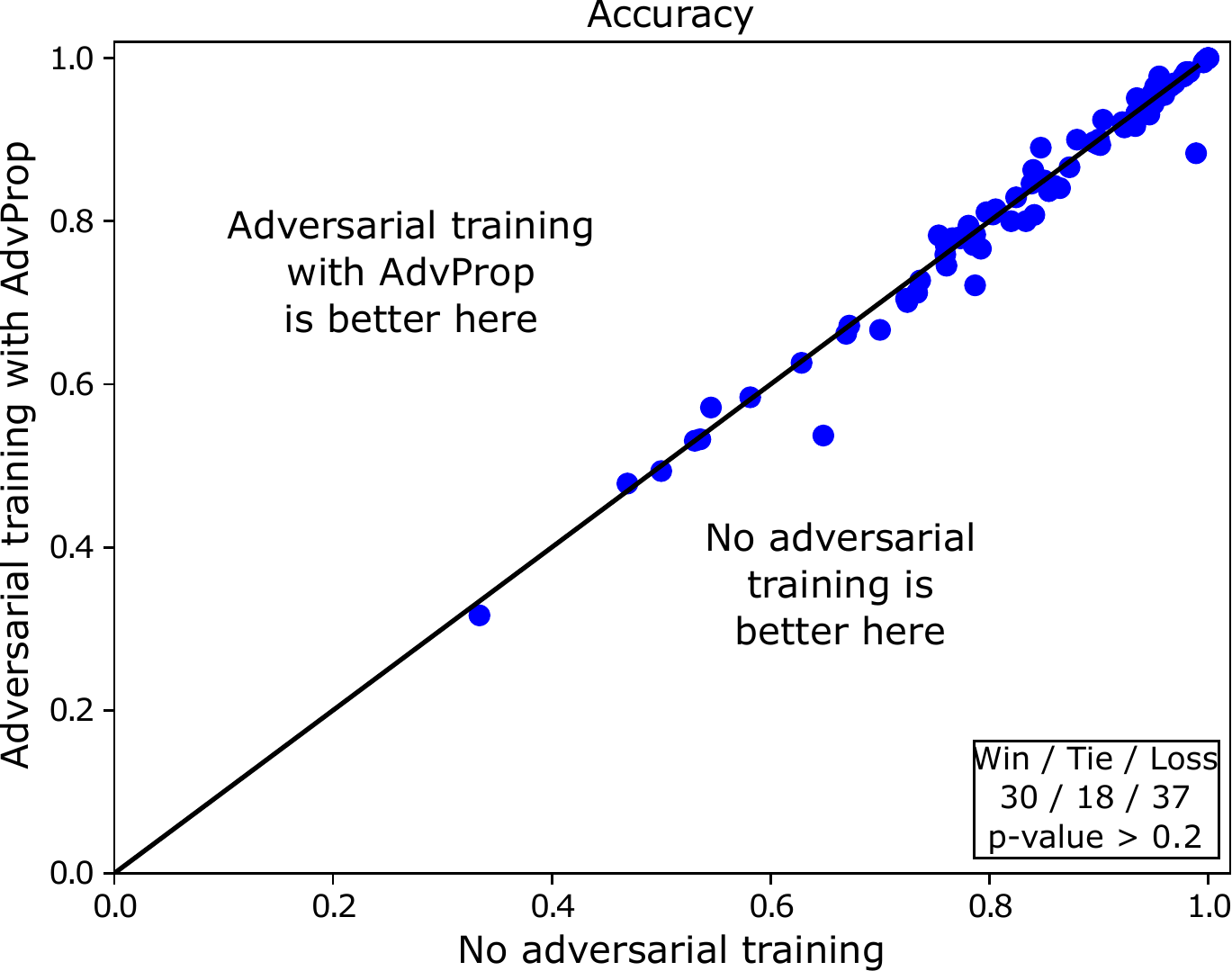}
	\caption{
		The effect of adversarial training on the model's accuracy.
	}
	\label{fig-advtraining-pairwise-plot}
\end{figure}

As we have previously mentioned,~\cite{xie2019adversarial} were among the first to show that adversarial training can be leveraged to improve the generalization capabilities of a neural network, by adopting two batch normalization layers: one for the original images and another for the perturbed ones. 
When adopting their approach for TSC, we found similar results: using AdvProp was necessary in order to avoid deteriorating the accuracy of our model. 
Figure~\ref{fig-advProp-pairwise-plot} illustrates the pairwise comparison between applying adversarial training with or without the AdvProp technique.
We can clearly see how vanilla adversarial training will deteriorate significantly the accuracy compared to using the AdvProp method. 
These results are in line with the original AdvProp paper~\citep{xie2019adversarial}. 
However, Figure~\ref{fig-advtraining-pairwise-plot} shows that using adversarial training does not improve significantly the accuracy for TSC, unlike the observations in~\cite{xie2019adversarial} on the ImageNet dataset. 
We hope that these preliminary results will give some insights to time series data mining researchers looking to leverage adversarial training as a regularization technique of DNNs for TSC. 

\subsection{Conclusion}

In this section, we introduced the concept of adversarial attacks on deep learning models for time series classification.
We defined and adapted two attacks, originally proposed for image recognition, for the TSC task.   
We showed how adversarial perturbations are able to reduce the accuracy for the state-of-the-art deep learning classifier (ResNet) when evaluated on the UCR/UEA archive benchmark. 
With deep neural networks becoming frequently adopted by time series data mining practitioners in real-life critical decision making systems, we shed the light on some crucial use cases where adversarial attacks could have serious and dangerous consequences. 
Finally we presented our initial preliminary results and showed that vanilla adversarial training will deteriorate the accuracy, which is why the AdvProp method was necessary to maintain a high accurate classifier. 

In the future, we would like to investigate countermeasures techniques to defend machine learning models against such attacks while exploring the transferability of adversarial examples to other non deep learning state-of-the-art classifiers.
Finally, we would like to further explore the dozens of adversarial attacks that are published each year in order to identify and protect vulnerable deep learning models for TSC. 

\section{Conclusion}

In this chapter, we have presented four main regularization techniques of deep learning models for TSC.
First by implementing a transfer learning approach, coupled with an inter-dataset similarity selection algorithm, we were able to significantly improve the accuracy of FCN. 
We then proposed to leverage the high variance due to the stochastic nature of the optimization process of neural networks, by ensembling the decision of more than one network. 
Unlike transfer learning, we observed that ensembling will almost always improve the classifier's accuracy. 
We then presented a data augmentation technique based on the famous DTW algorithm, allowing us to increase the number of training samples, thus ameliorating the model's accuracy. 
Finally, we focused on the vulnerabilities of neural networks to adversarial attacks and gave many use case examples where such attacks might be catastrophic.
Then in the spirit of regularizing DNNs, we showed how we can leverage these adversarial attacks to improve a network's generalization capabilities by leveraging a recent approach called AdvProp. 

We believe that this chapter should motivate researchers into further looking into many types of regularization methods that were mostly applied to images. 
We should note that this list of techniques is not exhaustive, many other types of regularization still exist and constitute currently a very hot topic in machine learning such as self-supervised pre-training~\citep{newell2020useful}. 
Nevertheless, we believe that the current state-of-the-art neural networks architectures for TSC are suboptimal and still lack behind the current advances of deep learning in computer vision. 
Therefore, we present in the following chapter a novel architecture for TSC based on the famous Inception module proposed by Google for the ImageNet competition. 
 

\chapter{InceptionTime: Finding AlexNet for Time Series Classification} \label{Chapter3}

\section{Introduction}

Recent times have seen an explosion in the magnitude and prevalence of time series data.
Industries varying from health care~\citep{forestier2018surgical,lee2018diagnosis,IsmailFawaz2019automatic} and social security~\citep{yi2018an} to human activity recognition~\citep{yuan2018muvan} and remote sensing~\citep{pelletier2019temporal}, all now produce time series datasets of previously unseen scale --- both in terms of time series length and quantity.
This growth also means an increased dependence on automatic classification of time series data, and ideally, algorithms with the ability to do this at scale.

In the previous chapters, we have shown how these TSC problems, differ significantly to traditional supervised learning for structured data, in that the algorithms should be able to handle and harness the temporal information present in the signal.
It is easy to draw parallels from this scenario to computer vision problems such as image classification and object localization, where successful algorithms learn from the spatial information contained in an image.
Put simply, the time series problem is essentially the same class of problem, just with one less dimension.
Yet despite this similarity, the current state-of-the-art algorithms from the two fields share little resemblance~\citep{IsmailFawaz2018deep}.

Deep learning has a long history (in machine learning terms) in computer vision~\citep{lecun1998efficient} but its popularity exploded with AlexNet~\citep{krizhevsky2012imagenet}, after which it has been unquestionably the most successful class of algorithms~\citep{lecun2015deep}. 
Conversely, deep learning has only recently started to gain popularity amongst time series data mining researchers (see Chapter~\ref{Chapter1}).
This is emphasized by the fact that ResNet, which is currently considered the state-of-the-art neural network architecture for TSC when evaluated on the UCR/UEA archive~\citep{ucrarchive}, was originally proposed merely as a baseline model for the underlying task~\citep{wang2017time}.
Given the similarities in the data, it is easy to suggest that there is much potential improvement for deep learning in TSC.
In Chapter~\ref{Chapter2}, we have shown how it is possible to improve the accuracy of a given deep learning architecture using various regularization techniques such as ensembling, transfer learning, data augmentation and adversarial training. 
However, we believe that there is still room for improvement in terms of network architecture, which can be considered an orthogonal task to the various DNNs regularization methods. 

In this chapter, we take an important step towards finding the equivalent of `AlexNet' for TSC by presenting \ourmethod{} --- a novel deep learning ensemble for TSC.
\ourmethod{} achieves state-of-the-art accuracy when evaluated on the UCR/UEA archive (currently the largest publicly available repository for TSC~\citep{ucrarchive}) while also possessing ability to scale to a magnitude far beyond that of its strongest competitor.

\ourmethod{} is an ensemble of five deep learning models for TSC, each one created by cascading multiple Inception modules~\citep{szegedy2015going}. 
Each individual classifier (model) will have exactly the same architecture but with different randomly initialized weight values.
The core idea of an Inception module is to apply multiple filters simultaneously to an input time series.
The module includes filters of varying lengths, which as we will show, allows the network to automatically extract relevant features from both long and short time series.
In fact, the ensemble here follows the same ensembling idea presented in Chapter~\ref{Chapter2}. 

After presenting \ourmethod{} and its results, we perform an analysis of the architectural hyperparameters of deep neural networks --- depth, filter length, number of filters --- and the characteristics of the Inception module --- the bottleneck and residual connection, in order to provide insight into why this model is so successful.
In fact, we construct networks with filters larger than have ever been explored for computer vision tasks, taking direct advantage of the fact that time series exhibit one less dimension than images.



\section{Related work}
In Chapter~\ref{Chapter1}, we have shown that deeper CNN models coupled with residual connections such as ResNet can further improve the classification performance. 
In essence, since time series data exhibit only one structuring dimension (i.e. time, as opposed to two spatial dimensions for images), it is possible to explore more complex models that are usually computationally infeasible for image recognition problems: for example removing the pooling layers that throw away valuable information in favour of reducing the model's complexity.
We therefore propose an Inception based network that applies several convolutions with various filters lengths. 
In contrast to networks designed for images, we are able to explore filters 10 times longer than recent Inception variants for image recognition tasks~\citep{szegedy2017inception}.  

Inception was first proposed by~\cite{szegedy2015going} for end-to-end image classification. 
Now the network has evolved to become Inceptionv4, where Inception was coupled with residual connections to further improve the performance~\citep{szegedy2017inception}. 
As for TSC a relatively competitive Inception-based approach was proposed in~\cite{karimi2018scalable}, where time series where transformed to images using Gramian Angular Difference Field, and finally fed to an Inception model that had been pre-trained for (standard) image recognition. 
Unlike this feature engineering approach, by adopting an end-to-end learning from raw time series data, a one-dimensional Inception model was used for Supernovae classification using the light flux of a region in space as an input MTS for the network~\citep{brunel2019a}.
However, the authors limited the conception of their Inception architecture to the one proposed by Google for ImageNet~\citep{szegedy2017inception}. 
In our work, we explore much larger filters than any previously proposed network for TSC in order to reach state-of-the-art performance on the UCR benchmark. 

\section{InceptionTime: an accurate and scalable time series classifier}

In this section, we start by describing the proposed architecture we call \ourmethod{} for classifying time series data. Specifically, we detail the main component of our network: the Inception module. 
We then present our proposed model \ourmethod{} which consists of an ensemble of 5 different Inception networks initialized randomly.
Finally, we adapt the concept of Receptive Field for time series data.

\begin{figure}
	\centering
	\includegraphics[width=\linewidth]{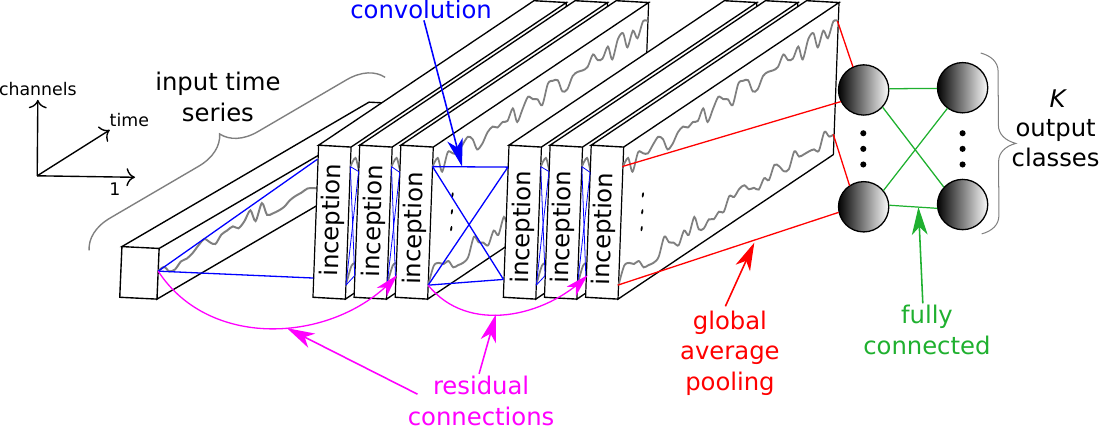}
	\caption{Our Inception network for time series classification.}
	\label{fig:inception-archi}
\end{figure}

\subsection{Inception Network: a novel architecture for TSC}
The composition of an Inception network classifier contains \emph{two} different residual blocks, as opposed to ResNet, which is comprised of \emph{three}.
For the Inception network, each block is comprised of three Inception modules rather than traditional fully convolutional layers.
Each residual block's input is transferred via a shortcut linear connection to be added to the next block's input, thus mitigating the vanishing gradient problem by allowing a direct flow of the gradient~\citep{he2016deep}. 
Following these residual blocks, we employed a GAP layer that averages the output multivariate time series over the whole time dimension.
At last, we used a final traditional fully-connected softmax layer with a number of neurons equal to the number of classes in the dataset.
\figurename~\ref{fig:inception-archi} depicts an Inception network's architecture showing 6 different Inception modules stacked one after the other.

\begin{figure*}
	\centering
	\includegraphics[width=\linewidth]{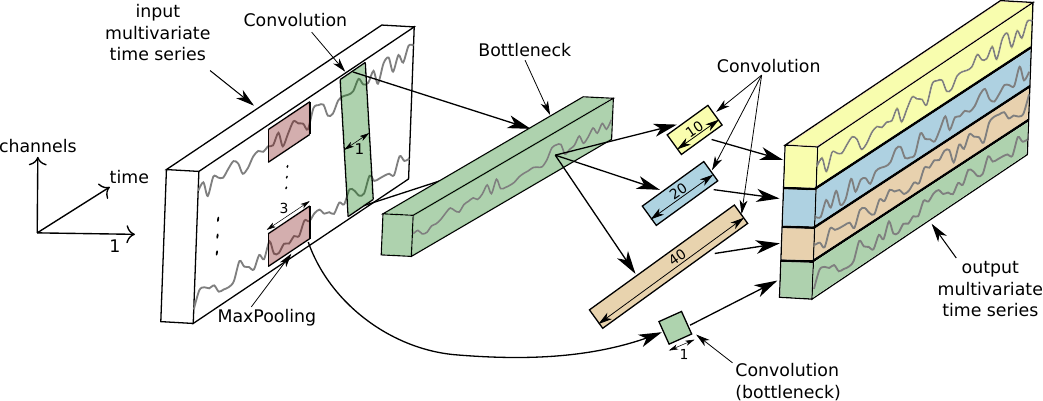}
	\caption{Inside our Inception module for time series classification. 
		For simplicity we illustrate a bottleneck layer of size $m=1$.}
	\label{fig:inception-module}
\end{figure*}

As for the Inception module, \figurename~\ref{fig:inception-module} illustrates the inside details of this operation. 
Let us consider the input to be an MTS with $M$ dimensions.
The first major component of the Inception module is called the ``bottleneck'' layer.
This layer performs an operation of sliding $m$ filters of length 1 with a stride equal to 1.
This will transform the time series from an MTS with $M$ dimensions to an MTS with $m \ll M$ dimensions, thus reducing significantly the dimensionality of the time series as well as the model's complexity and mitigating overfitting problems for small datasets.
Note that for visualization purposes, \figurename~\ref{fig:inception-module} illustrates a bottleneck layer with $m=1$. 
Finally, we should mention that this bottleneck technique allows the Inception network to have much longer filters than ResNet (almost ten times) with roughly the same number of parameters to be learned, since without the bottleneck layer, the filters will have $M$ dimensions compared to $m \ll M$ when using the bottleneck layer. 
The second major component of the Inception module is sliding multiple filters of different lengths simultaneously on the same input time series. 
For example in \figurename~\ref{fig:inception-module}, three different convolutions with length $l \in \{10,20,40\}$ are applied to the input MTS, which is technically the output of the bottleneck layer. 
Additionally, in order to make our model invariant to small perturbations, we introduce another parallel MaxPooling operation, followed by a bottleneck layer to reduce the dimensionality.
The output of sliding a MaxPooling window is computed by taking the maximum value in this given window of time series. 
Finally, the output of each independent parallel convolution/MaxPooling is concatenated to form the output MTS.
The latter operations are repeated for each individual Inception module of the proposed network.

By stacking multiple Inception modules and training the weights (filters' values) via backpropagation, the network is able to extract latent hierarchical features of multiple resolutions thanks to the use of filters with various lengths.
For completeness, we specify the exact number of filters for our proposed Inception module: 3 sets of filters each with 32 filters of length $l\in \{10,20,40\}$ with MaxPooling added to the mix, thus making the total number of filters per layer equal to $32\times4=128=M$ - the dimensionality of the output MTS. 
The default bottleneck size value was set to $m=32$.

\subsection{\ourmethod{}: a neural network ensemble for TSC}
Our proposed state-of-the-art \ourmethod{} model is an ensemble of 5 Inception networks, with each prediction given an even weight. 
In fact, during our experimentation, we have noticed that a single Inception network exhibits high standard deviation in accuracy, which is very similar to ResNet's behavior~\citep{IsmailFawaz2019deep}.
We believe that this variability comes from both the randomly initialized weights and the stochastic optimization process itself. 
This was an important finding for us, previously observed in~\cite{scardapane2017randomness}, as rather than training only one, potentially very good or very poor, instance of the Inception network, we decided to leverage this instability through ensembling, creating \ourmethod{}.
The following equation explains the ensembling of predictions made by a network with different initializations:
\begin{equation}
\hat{y}_{i,c}=\frac{1}{n}\sum_{j=1}^{n}\sigma_c(x_i,\theta_j) ~~|~~\forall c\in [1,C]
\end{equation}
with $\hat{y}_{i,c}$ denoting the ensemble's output probability of having the input time series $x_i$ belonging to class $c$, which is equal to the logistic output $\sigma_c$ averaged over the $n$ randomly initialized models.
More details on ensembling neural networks for TSC can be found in Chapter~\ref{Chapter2}. 
As for our proposed model, we chose the number of individual classifiers to be equal to $5$, which is justified in the next section.
We should note that we have opted to a neural network ensemble given the small training size of the UCR/UEA archive datasets which are not well suited to deep learning approaches, thus allowing us to control and leverage the variance of the error, which is likely to reduce when increasing the training set's size.

\subsection{Receptive field}

\begin{figure}
	\centering
	\includegraphics[width=0.7\linewidth]{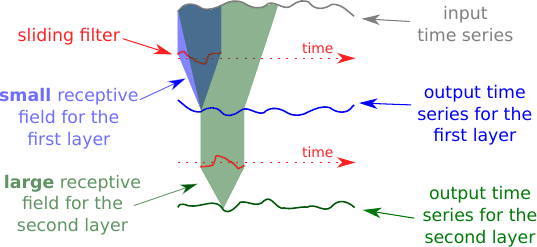}
	\caption{Receptive field illustration for a two layers CNN.}
	\label{fig:receptive-field}
\end{figure}

The concept of Receptive Field is an essential tool to the understanding of deep CNNs~\citep{luo2016understanding}.
Unlike FC networks or MLPs, a neuron in a CNN depends only on a region of the input signal. 
This region in the input space is called the receptive field of that particular neuron. 
For computer vision problems this concept was extensively studied, such as in~\cite{liu2018understanding} where the authors compared the effective and theoretical receptive fields of a CNN for image segmentation.   

For temporal data, the receptive field can be considered as a theoretical value that measures the maximum field of view of a neural network in a one-dimensional space: the larger it is, the better the network becomes (in theory) in detecting longer patterns. 
We now provide the definition of the RF for time series data, which is later used in our experiments. 
Suppose that we are sliding convolutions with a stride equal to $1$.
The formula to compute the RF for a network of depth $d$ with each layer having a filter length equal to $k_i$ with $i\in[1,d]$ is: 
\begin{equation}\label{eq-rf}
1+\sum_{i=1}^{d}(k_i-1)
\end{equation}
By analyzing equation~\ref{eq-rf} we can clearly see that adding two layers to the initial set of $d$ layers, will increase only slightly the value of $RF$. 
In fact in this case, if the old $RF$ value is equal to $RF^{'}$, the new value $RF$ will be equal to $RF^{'} + 2\times (k-1)$. 
Conversely, by increasing the filter length $k_i$, $\forall i \in [1,d]$ by 2, the new value $RF$ will be equal to $RF^{'} + 2\times d$.
This is rather expected since by increasing the filter length for all layers, we are actually increasing the $RF$ for each layer in the network.
\figurename~\ref{fig:receptive-field} illustrates the RF for a two layers CNN. 

In this chapter, we chose to focus on the RF concept since it has been known for computer vision problems, that larger RFs are required to capture more context for object recognition~\citep{luo2016understanding}.
Following the same line of thinking, we hypothesize that detecting larger patterns from very long one-dimensional time series data, requires larger receptive fields. 

\begin{figure}
	\centering
	\includegraphics[width=0.6\linewidth]{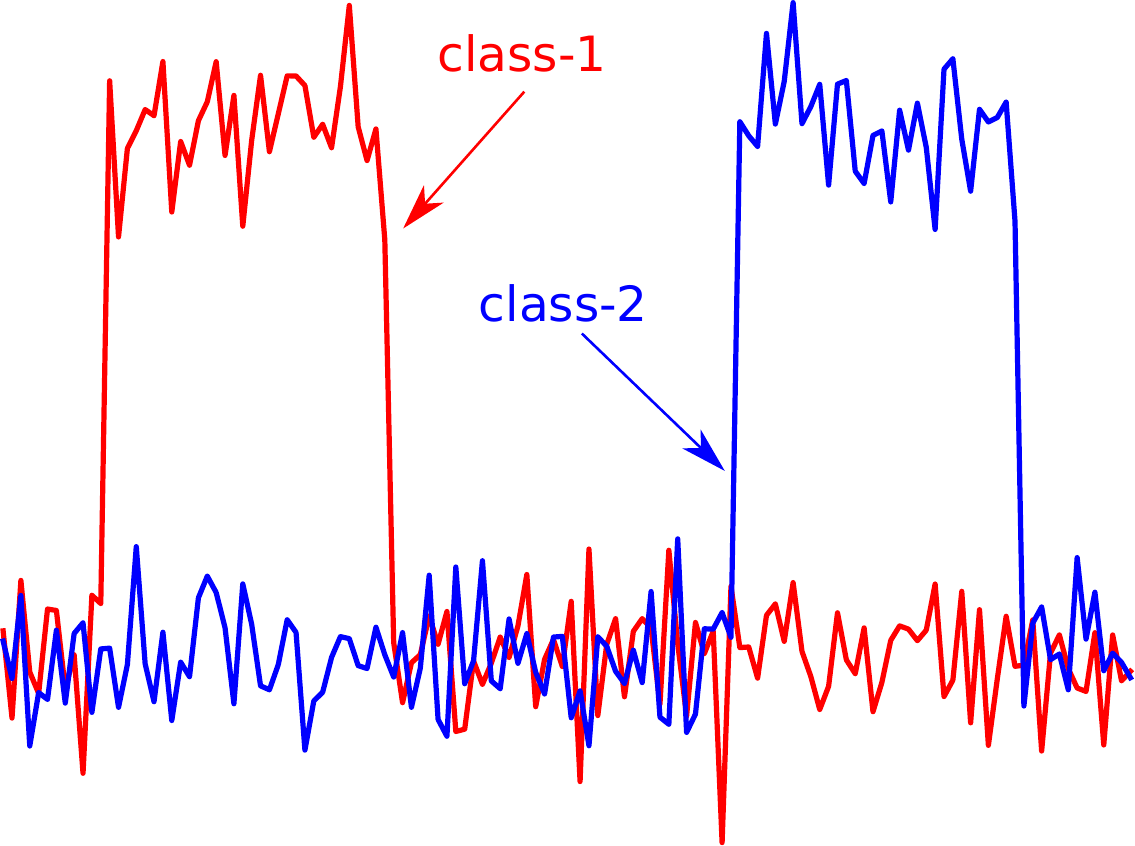}
	\caption{Example of a synthetic binary time series classification problem.}
	\label{fig:synthetic-dataset-example}
\end{figure}

\section{Experimental setup}

First, we detail the method to generate our synthetic dataset, which is later used in our architecture and hyperparameter study. 
For testing our different deep learning methods, we created our own synthetic TSC dataset. 
The goal was to be able to control the length of the time series data as well as the number of classes and their distribution in time.
To this end, we start by generating a univariate time series using uniformly distributed noise sampled between 0.0 and 0.1. 
Then in order to assign this synthetic random time series to a certain class, we inject a pattern with an amplitude equal to 1.0 in a pre-defined region of the time series. 
This region will be specific to a certain class, therefore by changing the placement of this pattern we can generate an unlimited amount of classes, whereas the random noise will allow us to generate an unlimited amount of time series instances per class.
One final note is that we have fixed the length of the pattern to be equal to 10\% the length of the synthetic time series. 
An example of a synthetic binary TSC problem is depicted in \figurename~\ref{fig:synthetic-dataset-example}.

All DNNs were trained by leveraging the parallel computation of a remote cluster of more than 60 GPUs comprised of GTX 1080 Ti, Tesla K20, K40 and K80.
Local testing and development was performed on an NVIDIA Quadro P6000. 
The latter graphics card was also used for computing the training time of a model. 
When evaluating global accuracy and computational complexity, we have used the UCR/UEA archive~\citep{ucrarchive}, which is the largest publicly available archive for TSC.
The models were trained/tested using the original training/testing splits provided in the archive.  
To study the effect of different hyperparameters and architectural designs, we used in addition to the traditional UCR benchmark for TSC, the synthetic dataset whose generation is described in details in the previous paragraph. 
All time series data were $z$-normalized (including the synthetic series) to have a mean equal to zero and a standard deviation equal to one. 
This is considered a common best-practice before classifying time series data~\citep{bagnall2017the}. 
Finally, we should note that all models are trained using the Adam optimization algorithm~\citep{kingma2015adam} and all weights are initialized randomly using Glorot's uniform technique~\citep{glorot2010understanding}.

When comparing with the state-of-the-art results published in~\cite{bagnall2017the} we used the deep learning model's median test accuracy over the different runs, similarly to what we have done in Chapter~\ref{Chapter1}. 
Following the recommendations in~\cite{demsar2006statistical} we adopted the Friedman test~\citep{friedman1940a} in order to reject the null hypothesis. 
We then performed the pairwise post-hoc analysis recommended by~\cite{benavoli2016should} where we replaced the average rank comparison by a Wilcoxon signed-rank test with Holm's alpha ($5\%$) correction~\citep{garcia2008an}. 
To visualize this type of comparison we used a critical difference diagram proposed by~\cite{demsar2006statistical}, where a thick horizontal line shows a cluster of classifiers (a clique) that are not-significantly different in terms of accuracy.

In order to allow for the time series community to build upon and verify our findings, the source code for all these experiments was made publicly available on our companion repository\footnote{\url{https://github.com/hfawaz/\ourmethod{}}}.
In addition, we are planning on providing the pre-trained deep learning models, thus allowing data mining practitioners to leverage these networks in a transfer learning setting~\citep{IsmailFawaz2018transfer}. 
 
\section{Experiments: \ourmethod{}}
\begin{figure}
	\centering
	\includegraphics[width=\linewidth]{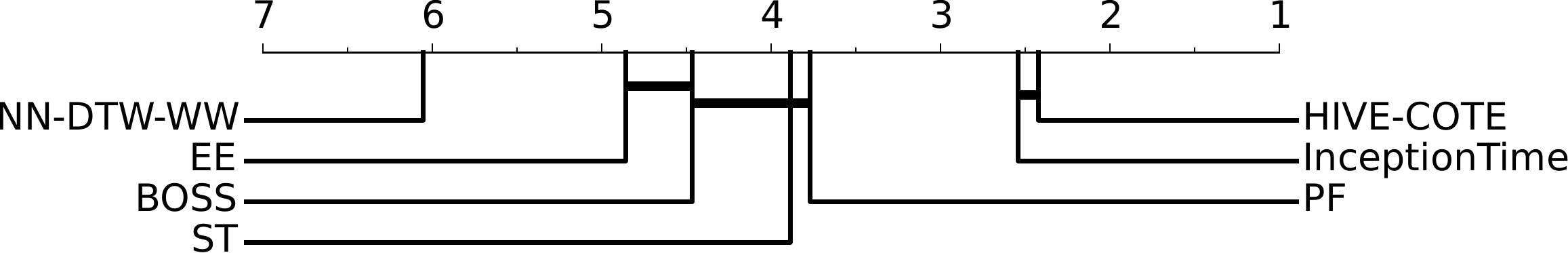}
	\caption{Critical difference diagram showing the performance of \ourmethod{} compared to the current state-of-the-art classifiers of time series data.}
	\label{fig:cd-diagram-with-inception-ensemble}
\end{figure}
In this section, we present the results of our proposed novel classifier called \ourmethod{}, evaluated on the 85 datasets of the UCR/UEA archive.
We note that throughout this chapter (unless specified otherwise) \ourmethod{} refers to an ensemble of 5 Inception networks, while the ``\ourmethod{}($n$)'' notation is used to denote an ensemble of $n$ Inception networks. 

\figurename~\ref{fig:cd-diagram-with-inception-ensemble} illustrates the critical difference diagram with \ourmethod{} added to the mix of the current state-of-the-art classifiers for time series data, whose results were taken from~\cite{bagnall2017the}. 
We can see here that our \ourmethod{} ensemble reaches competitive accuracy with the class-leading algorithm HIVE-COTE, an ensemble of 37 TSC algorithms with a hierarchical voting scheme~\citep{lines2016hive}. 
While the two algorithms share the same clique on the critical difference diagram, the trivial GPU parallelization of deep learning models makes learning our \ourmethod{} model a substantially easier task than training the 37 different classifiers of HIVE-COTE, whose implementation does not trivially leverage the GPUs' computational power. 

\begin{figure}
	\centering
	\includegraphics[width=0.7\linewidth]{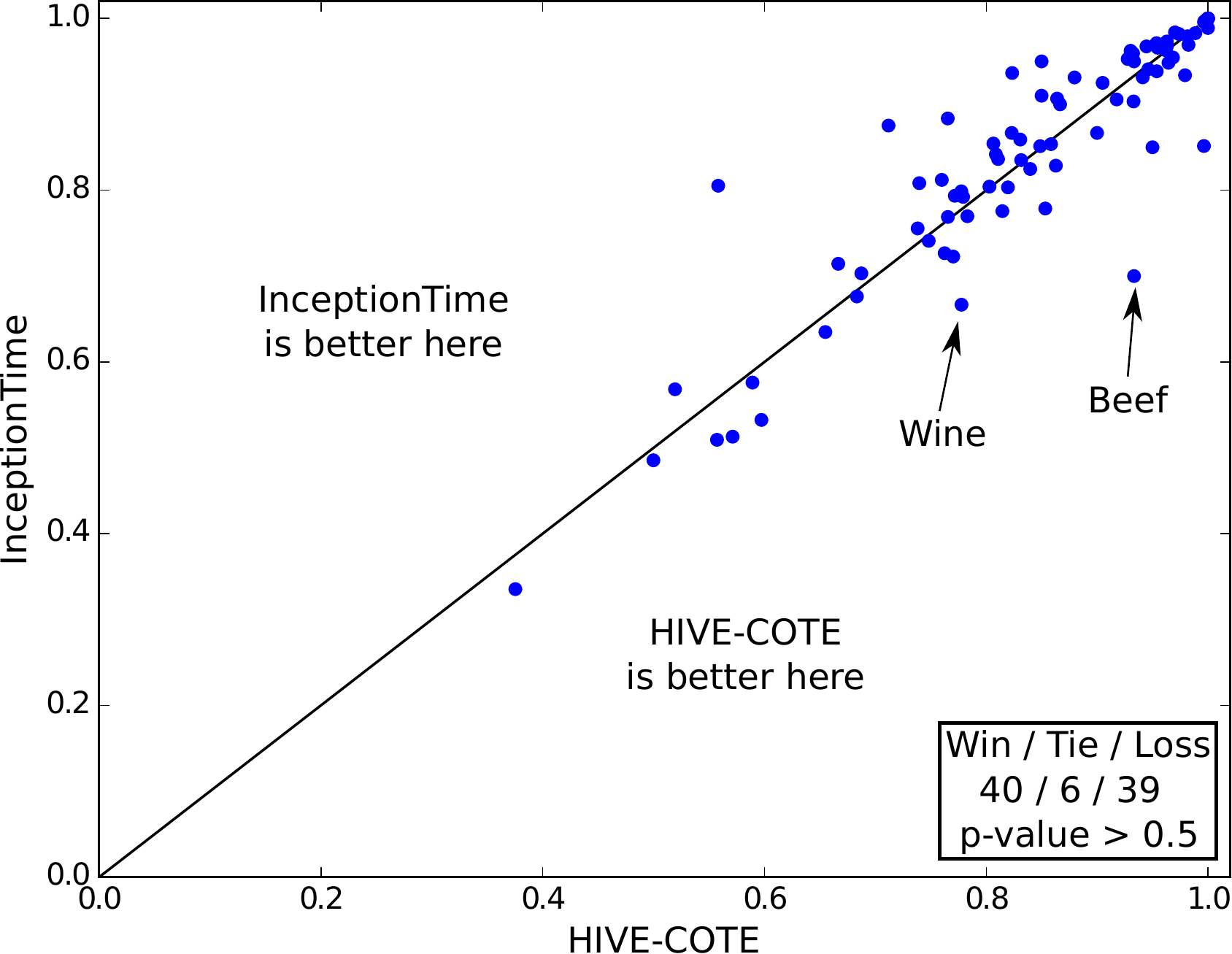}
	\caption{Accuracy plot showing how our proposed \ourmethod{} model is not significantly different than HIVE-COTE.}
	\label{fig:plot-inceptions-vs-hive-cote}
\end{figure}

To further visualize the difference between the \ourmethod{} and HIVE-COTE, \figurename~\ref{fig:plot-inceptions-vs-hive-cote} depicts the accuracy plot of \ourmethod{} against HIVE-COTE for each of the 85 UCR datasets. 
The results show a Win/Tie/Loss of 40/6/39 in favor of \ourmethod{}, however the difference is not statistically significant as previously discussed.
From \figurename~\ref{fig:plot-inceptions-vs-hive-cote}, we can also easily spot the two datasets for which  \ourmethod{} noticeably under-performs (in terms of accuracy) with respect to HIVE-COTE: Wine and Beef.
These two datasets contain spectrography data from different types of beef/wine, with the goal being to determine the correct type of meat/wine using the recorded time series data.
In Chapter~\ref{Chapter2}, we showed that transfer learning significantly increases the accuracy for these two datasets, especially when fine-tuning a dataset with similar time series data. 
Our results suggest that further potential improvements may be available for \ourmethod{} when applying a transfer learning approach, as recent discoveries in~\cite{kashiparekh2019convtimenet} show that the various filter lengths of the Inception modules have been shown to benefit more from fine-tuning than networks with a static filter length.

\begin{figure}
	\centering
	\includegraphics[width=0.7\linewidth]{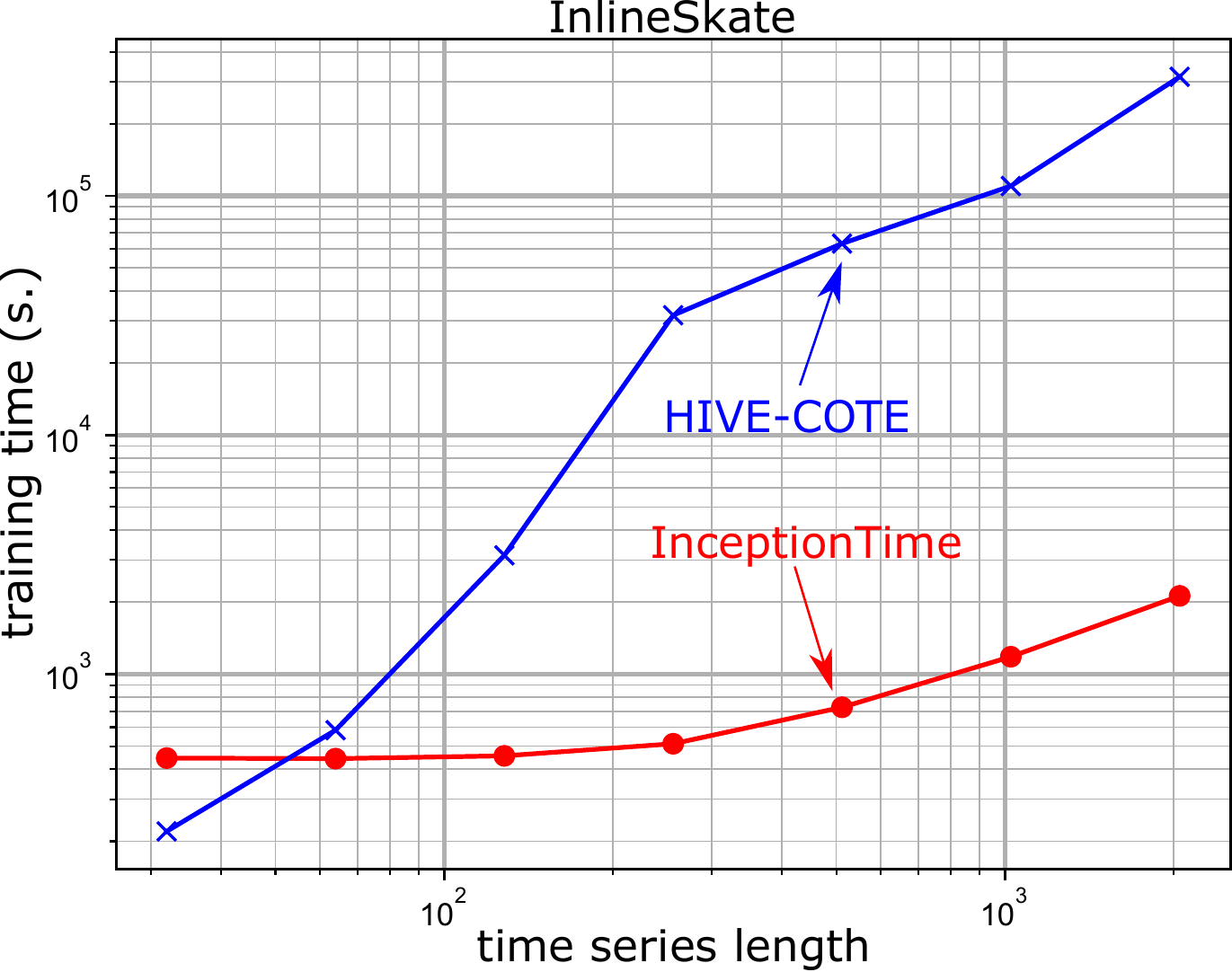}
	\caption{Training time as a function of the series length for the InlineSkate dataset.}
	\label{fig:plot-inception-vs-hive-cote-length}
\end{figure}

\begin{figure}
	\centering
	\includegraphics[width=0.7\linewidth]{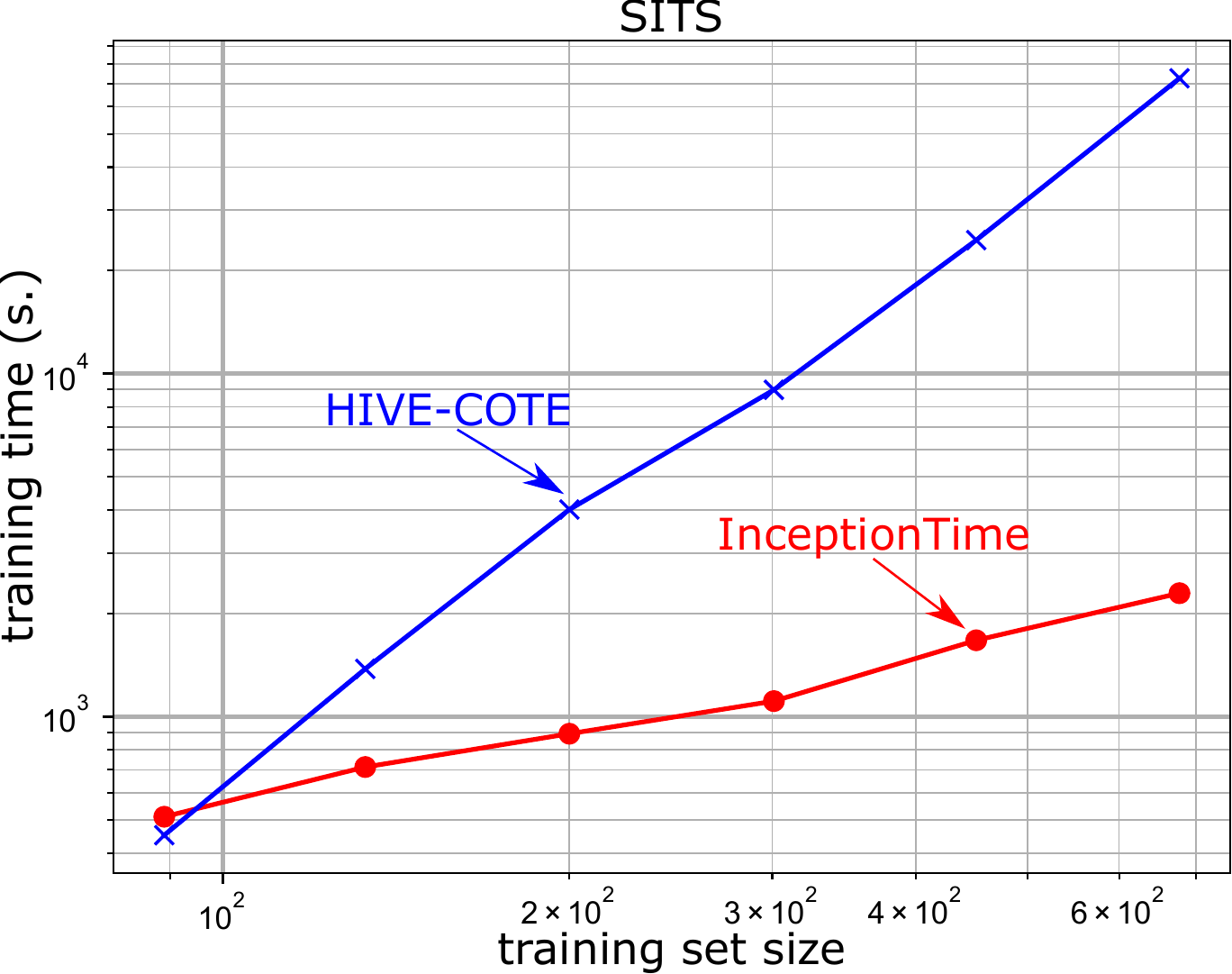}
	\caption{Training time as a function of the training set size for the SITS dataset.}
	\label{fig:plot-inception-vs-hive-cote-train-size}
\end{figure}

Now that we have demonstrated that our proposed technique is able to reach the current state-of-the-art accuracy for TSC problems, we will further investigate the time complexity of our model. 
Note that during the following experiments, we ran our ensemble on a single Nvidia Quadro P6000 in a sequential manner, meaning that for \ourmethod{}, 5 different Inception networks were trained one after the other. 
Therefore we did not make use of our remote cluster of GPUs.
First we start by investigating how our algorithm scales with respect to the length of the input time series. 
\figurename~\ref{fig:plot-inception-vs-hive-cote-length} shows the training time versus the length of the input time series. 
For this experiment, we used the InlineSkate dataset with an exponential re-sampling. 
We can clearly see that \ourmethod{}'s complexity increases almost linearly with an increase in the time series' length, unlike HIVE-COTE, whose execution is almost two order of magnitudes slower. 
Having showed that \ourmethod{} is significantly faster when dealing with long time series, we now proceed to evaluating the training time with respect to a number of time series in a dataset.
To this end, we used a Satellite Image Time Series dataset~\citep{tan2017indexing}. 
The data contain approximately one million time series, each of length 46 and labeled as one of 24 possible land-use classes (e.g.\ `wheat', `corn', `plantation', `urban').
From \figurename~\ref{fig:plot-inception-vs-hive-cote-train-size} we can easily see how our \ourmethod{} is an order of magnitude faster than HIVE-COTE, and the trend suggests that this difference will only continue to grow, rendering \ourmethod{} a clear favorite classifier in the Big Data era.  
Note that HIVE-COTE uses heuristics in its implementation, which explains why the complexity appears lower in the experiments than the expected $O(T^4)$.
To summarize, we believe that \ourmethod{} should be considered as one of the top state-of-the-art methods for TSC, given that it demonstrates equal accuracy to that of HIVE-COTE (see \figurename~\ref{fig:plot-inceptions-vs-hive-cote}) while being much faster (see \figurename~\ref{fig:plot-inception-vs-hive-cote-length} and~\ref{fig:plot-inception-vs-hive-cote-train-size}).

\begin{figure}
	\centering
	\includegraphics[width=0.7\linewidth]{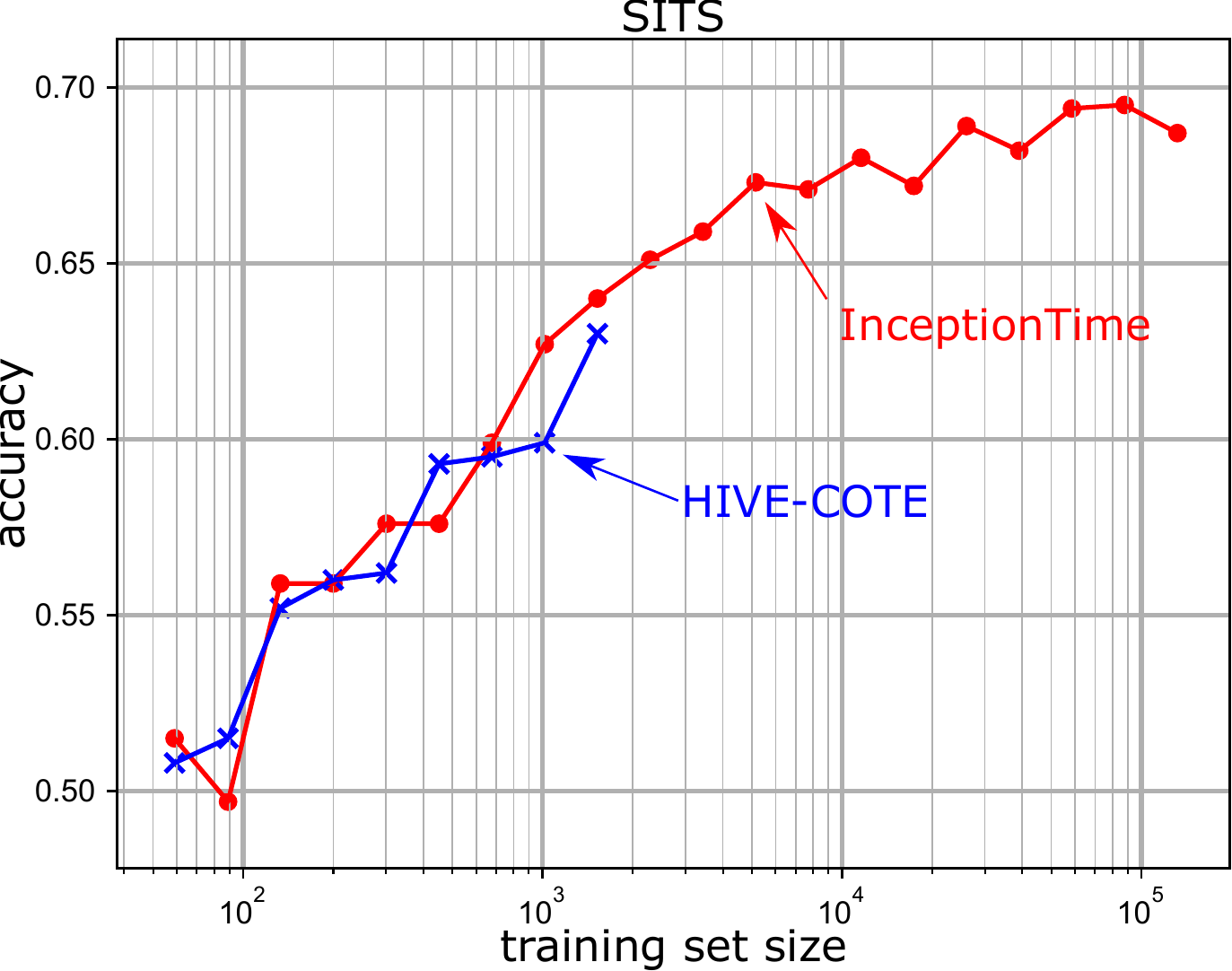}
	\caption{Accuracy as a function of the training set size for the SITS dataset.}
	\label{fig:plot-inception-vs-hive-cote-accuracy}
\end{figure}

In order to further demonstrate the capability of \ourmethod{} to handle efficiently a large amount of training samples unlike its counterpart HIVE-COTE, we show in \figurename~\ref{fig:plot-inception-vs-hive-cote-accuracy} how the accuracy continues to increase with \ourmethod{} for larger training set sizes, where HIVE-COTE would take 100 times longer to run. 

\begin{figure}
	\centering
	\includegraphics[width=0.7\linewidth]{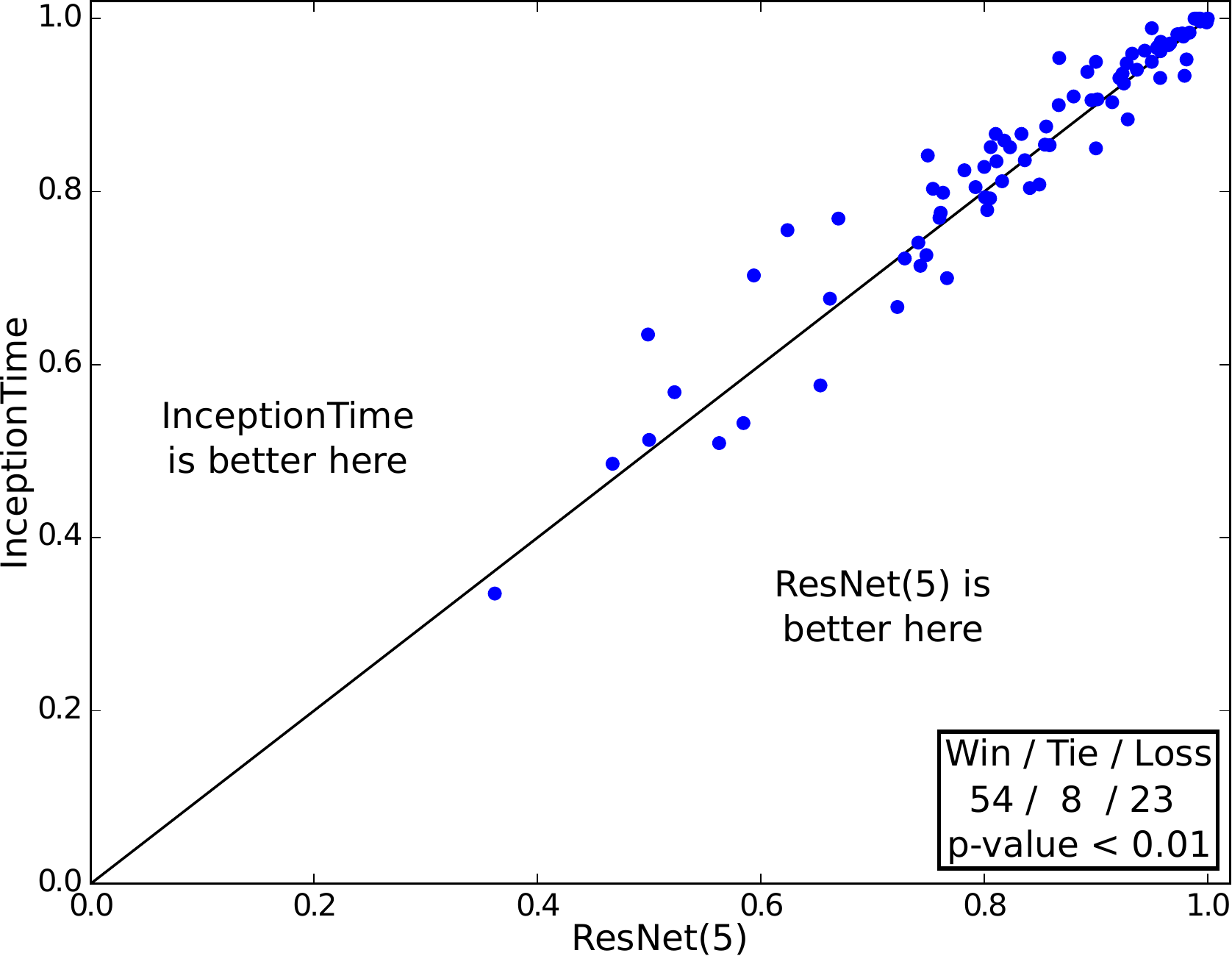}
	\caption{Plot showing how \ourmethod{} significantly outperforms ResNet(5).}
	\label{fig:plot-inception-vs-resnet}
\end{figure}

The pairwise accuracy plot in \figurename~\ref{fig:plot-inception-vs-resnet} compares \ourmethod{} to a model we call ResNet(5), which is an ensemble of 5 different ResNet networks~\citep{IsmailFawaz2019deep}. 
We found that \ourmethod{} showed a significant improvement over its neural network competitor, the previous best deep learning ensemble for TSC.
Specifically, our results show a Win/Tie/Loss of 54/8/23 in favor of \ourmethod{} against ResNet(5) with a $p$-value $<0.01$, suggesting the significant gain in performance is mainly due to improvements in our proposed Inception network architecture.
Additionally, in order to have a fair comparison between ResNet(5) and \ourmethod{}, we fixed the batch size of ResNet to 64 -- equal to the default value used for \ourmethod{}. 
This would further highlight that the improvement is mainly due to the architectural design of our proposed network, and not due to some other optimization hyperparameter such as the batch size.
Finally, we would like to note that when using the original batch size value proposed by~\cite{wang2017time} for ResNet, we observed similar results: \ourmethod{} was significantly better than the original ResNet(5) with a Win/Tie/Loss of 53/7/25.

\begin{figure}
	\centering
	\includegraphics[width=\linewidth]{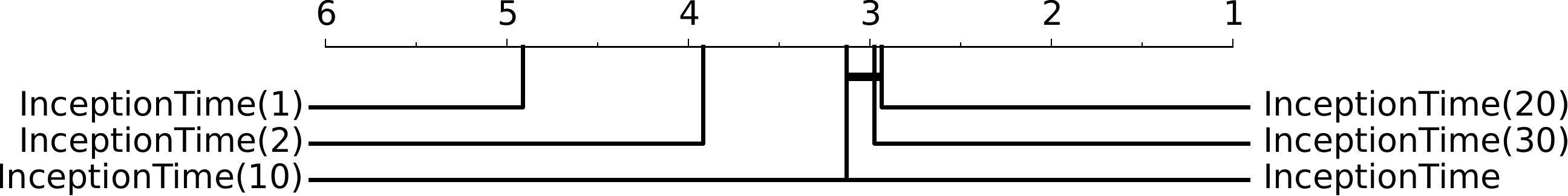}
	\caption{Critical difference diagram showing the effect of the number of individual classifiers in the \ourmethod{} ensemble.}
	\label{fig:cd-diagram-ensembles}
\end{figure}

In order to better understand the effect of the randomness on the accuracy of our neural networks, we present in \figurename~\ref{fig:cd-diagram-ensembles} the critical difference diagram of different \ourmethod{}($x$) ensembles with $x\in \{1,2,5,10,20,30\}$ denoting the number of individual networks in the ensemble.
Note that \ourmethod{}(1) is equivalent to a single Inception network and \ourmethod{} is equivalent to \ourmethod{}(5).
By observing \figurename~\ref{fig:cd-diagram-ensembles} we notice how there is no significant improvement when $x\ge5$, which is why we chose to use an ensemble of size 5, to minimize the classifiers' training time.

\section{Architectural Hyperparameter study}

In this section, we will further investigate the hyperparameters of our deep learning architecture and the characteristics of the Inception module in order to provide insight for practitioners looking at optimizing neural networks for TSC.
First, we start by investigating the batch size hyperparameter, since this will greatly influence training time of all of our models.
Then we investigate the effectiveness of residual and bottleneck connections, both of which are present in \ourmethod{}.
After this we will experiment on model depth, filter length, and number of filters.
In all experiments the default values for \ourmethod{} are: batch size 64; bottleneck size 32; depth 6; filter length \{10,20,40\}; and, number of filters 32.
Finally, since the train/test split (provided in the archive) does not help in estimating the generalization ability of our approach, we have conducted a sensitivity analysis that evaluates the second best value for each of the network's hyperparameters.


\subsection{Batch size}
\begin{figure}
	\centering
	\includegraphics[width=\linewidth]{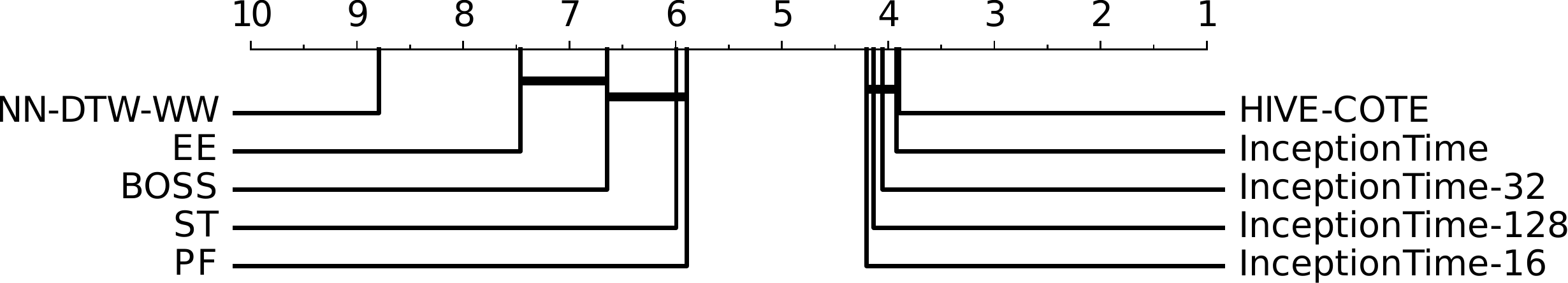}
	\caption{Critical difference diagram showing the effect of the batch size hyperparameter value over \ourmethod{}'s average rank.}
	\label{fig:cd-diagram-inception-batch-size}
\end{figure}

We started by investigating the batch size hyperparameter on the UCR/UEA archive, since this will greatly influence training time of our models.
The critical difference diagram in \figurename~\ref{fig:cd-diagram-inception-batch-size} shows how the batch size affects the performance of \ourmethod{}. 
The horizontal thick line between the different models shows a non significant difference between them when evaluated on the 85 datasets, with a small superiority to \ourmethod{} (batch size equal to 64).
Finally, we should note that as we did not observe any significant impact on accuracy we did not study the effect of this hyperparameter on the simulated dataset and we chose to fix the batch size to 64 (similarly to \ourmethod{}) when experimenting on the simulated dataset below.

\subsection{Bottleneck and residual connections}

\begin{figure}
	\centering
	\includegraphics[width=0.7\linewidth]{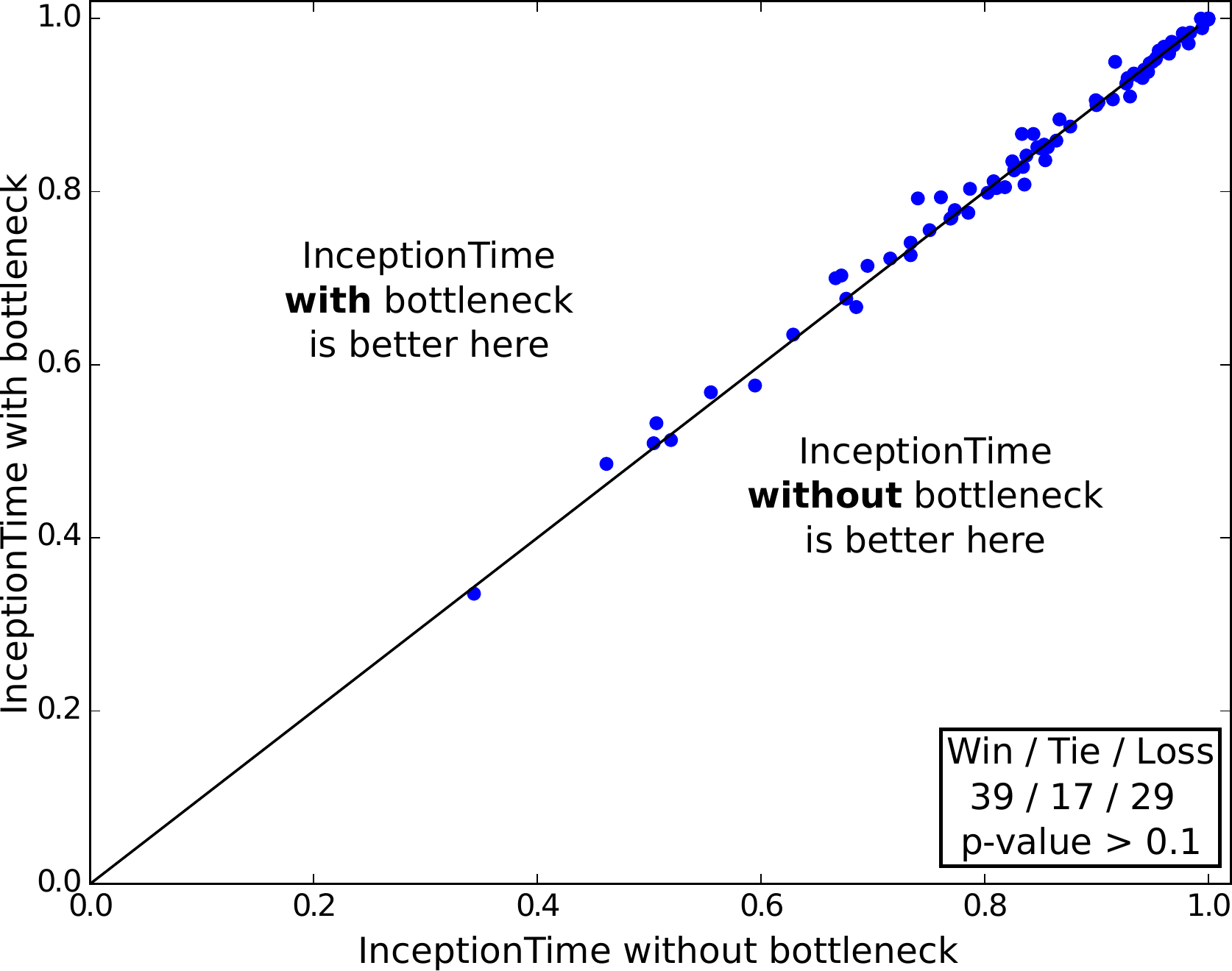}
	\caption{Accuracy plot for \ourmethod{} with/without the bottleneck layer.}
	\label{fig:plot-inception-with-without-bottleneck}
\end{figure}

In Chapter~\ref{Chapter1}, compared to other deep learning classifiers, ResNet achieved the best classification accuracy when evaluated on the 85 datasets and as a result we chose to look at the specific characteristic of this architecture --- its residual connections. 
Additionally, we tested one of the defining characteristics of Inception --- the bottleneck feature. 
For the simulated dataset, we did not observe any significant impact of these two connections, we therefore proceed with experimenting on the 85 datasets from the UCR/UEA archive.

\figurename~\ref{fig:plot-inception-with-without-bottleneck} shows the pairwise accuracy plot comparing \ourmethod{} with/without the bottleneck.
Similar to the experiments on the simulated dataset, we did not find any significant variation in accuracy when adding or removing the bottleneck layer.

\begin{figure}
	\centering
	\includegraphics[width=\linewidth]{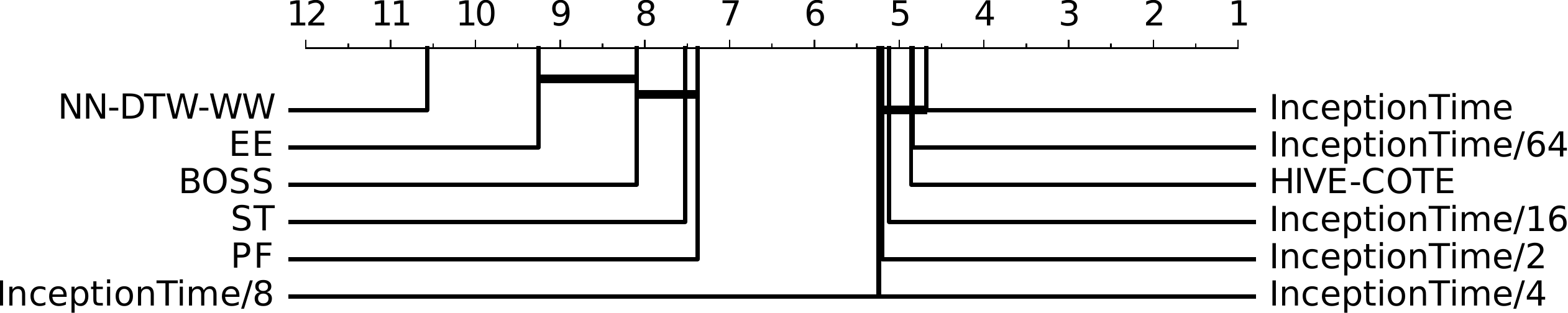}
	\caption{Critical difference diagram showing how the network's bottleneck size affects \ourmethod{}' average rank.}
	\label{fig:cd-diagram-inceptiontime-bottleneck-size-with-bake-off}
\end{figure}

In fact, using a Wilcoxon Signed-Rank test we found that \ourmethod{} with the bottleneck layer is only slightly better than removing the bottleneck layer ($p$-value $>0.1$).
In terms of accuracy, these results all suggest not to use a bottleneck layer, however we should note that the major benefit of this layer is to significantly decrease the number of parameters in the network. 
In this case, \ourmethod{} with the bottleneck contains almost half the number of parameters to be learned, and given that it does not significantly decrease accuracy, we chose to retain its usage.
In a more general sense, these experiments suggest that choosing whether or not to use a bottleneck layer is actually a matter of finding a balance between a model's accuracy and its complexity.
The latter observation is evident in \figurename~\ref{fig:cd-diagram-inceptiontime-bottleneck-size-with-bake-off} where choosing smaller bottleneck size in order to reduce \ourmethod{}'s runtime will result in small yet insignificant decrease in accuracy.

\begin{figure}
	\centering
	\includegraphics[width=0.7\linewidth]{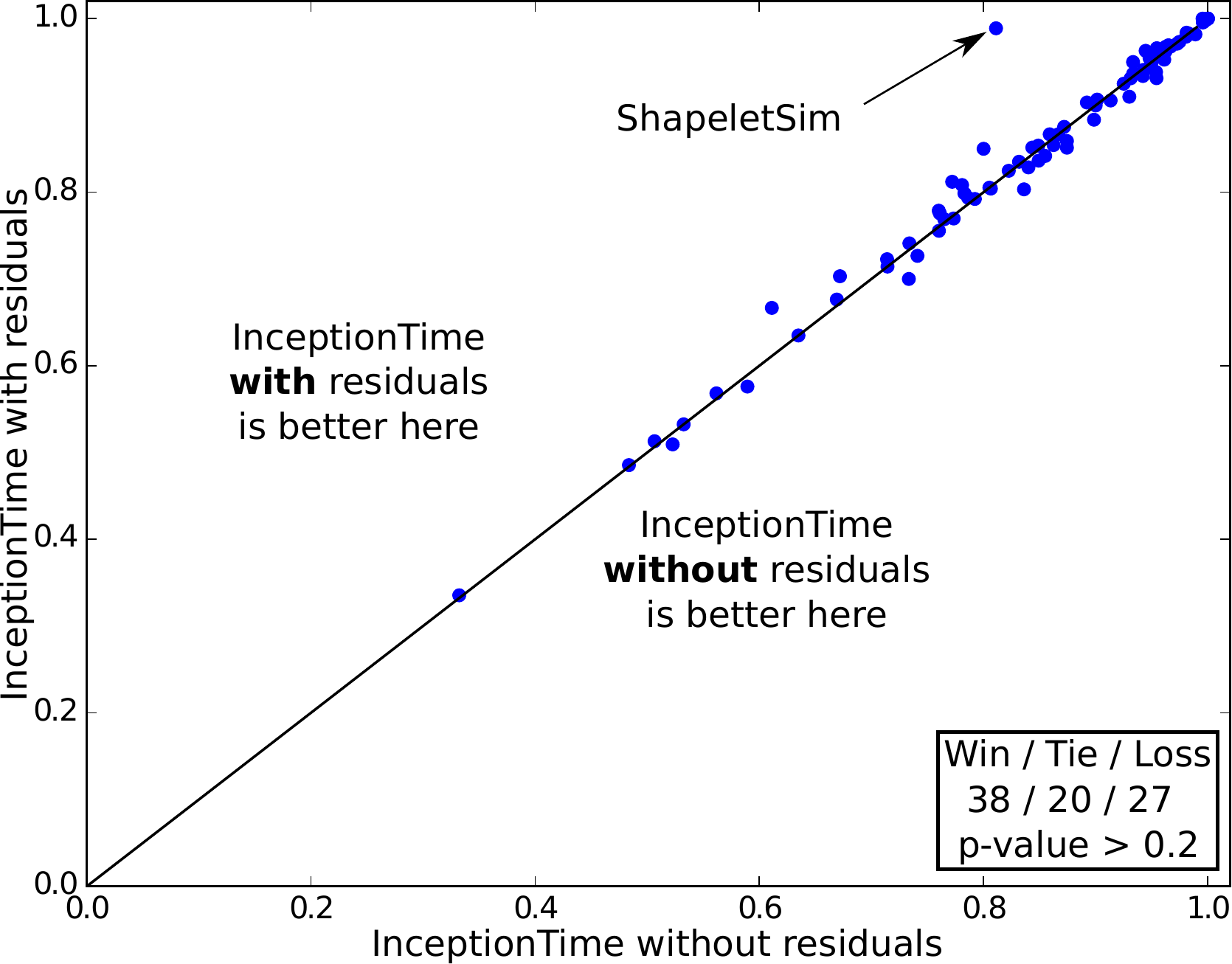}
	\caption{Accuracy plot for \ourmethod{} with/without the residual connections.}
	\label{fig:plot-inception-with-without-residual}
\end{figure}

To test the residual connections, we simply removed the residual connection from \ourmethod{}. 
Thus, without any shortcut connection, \ourmethod{} will simply become a deep convolutional neural network with stacked Inception modules.
\figurename~\ref{fig:plot-inception-with-without-residual} shows how the residual connections have a minimal effect on accuracy when evaluated over the whole 85 datasets in the UCR/UEA archive with a $p$-value $>0.2$.  

This result was unsurprising given that for computer vision tasks residual connections are known to improve the convergence rate of the network but not alter its test accuracy~\citep{szegedy2017inception}. 
However, for some datasets in the archive, the residual connections did not show any improvement nor deterioration of the network's convergence either. 
This could be linked to other factors that are specific to these data, such as the complexity of the dataset.

One example of interest that we noticed was a significant decrease in \ourmethod{}'s accuracy when removing the residual component for the ShapeletSim dataset. 
This is a synthetic dataset, designed specifically for shapelets discovery algorithms, with shapelets (discriminative subsequences) of different lengths~\citep{hills2014classification}.
Further investigations on this dataset indicated that \ourmethod{} without the residual connections suffered from a severe overfitting. 

While not the case here, some research has observed benefits of skip, dense or residual connections~\citep{huang2017densely}.
Given this, and the small amount of labeled data available in TSC compared to computer vision problems, we believe that each case should be independently study whether to include residual connections.
The latter observation suggests that a large scale general purpose labeled dataset similar to ImageNet~\citep{russakovsky2015imagenet} is needed for TSC.
Finally, we should note that the residual connection has a minimal impact on the network's complexity~\citep{szegedy2017inception}. 

\subsection{Depth}

Most of deep learning's success in image recognition tasks has been attributed to how `deep' the architectures are~\citep{lecun2015deep}. 
Consequently, we decided to further investigate how the number of layers affects a network's accuracy.
Unlike the previous hyperparameters, we present here the results on the simulated dataset. 
Apart from the depth parameter, we used the default values of \ourmethod{}.
For this dataset we fixed the number of training instances to 256 and the number of classes to 2 (see \figurename~\ref{fig:synthetic-dataset-example} for an example).
The only dataset parameter we varied was the length of the input time series.

\begin{figure}
	\centering
	\includegraphics[width=.7\linewidth]{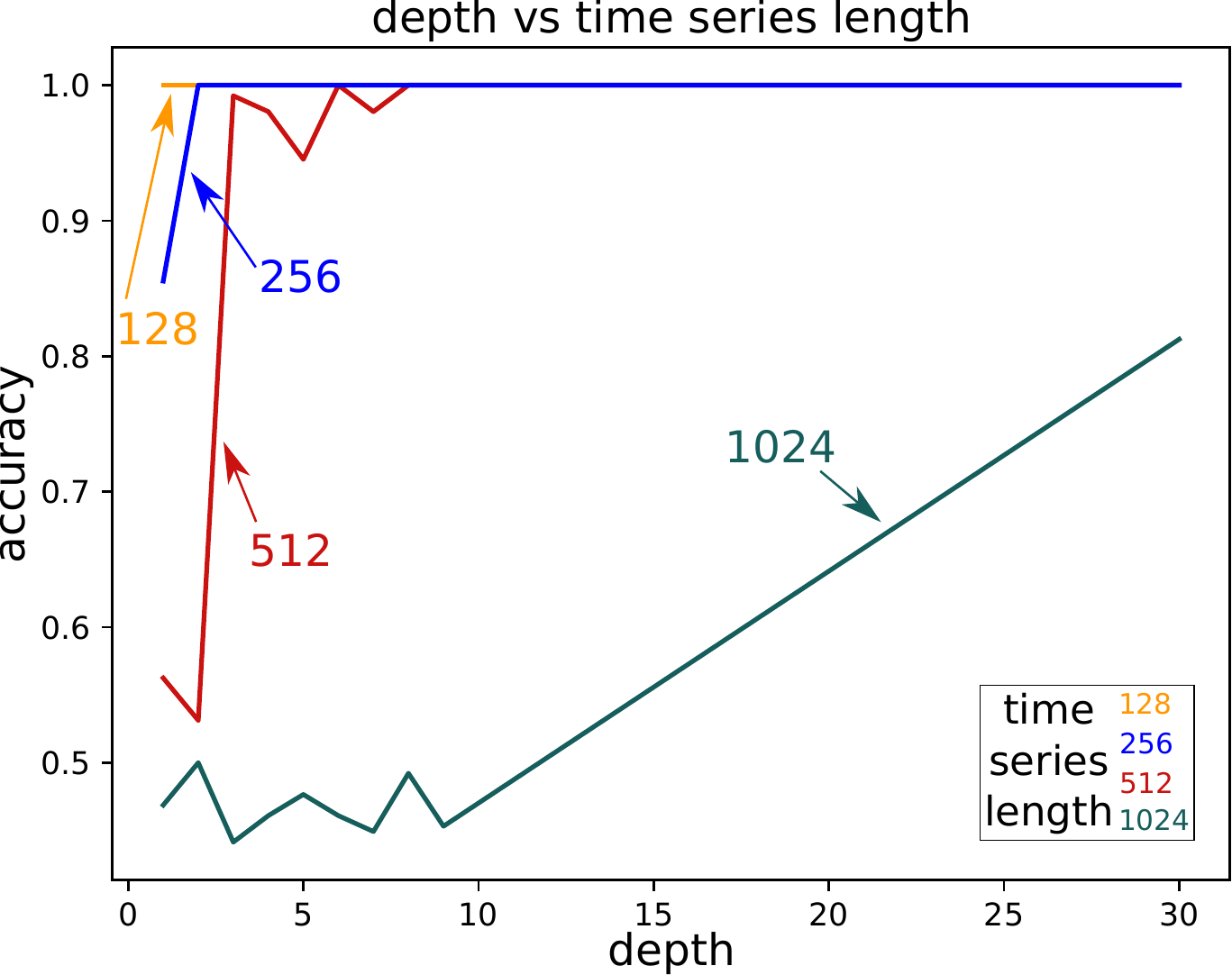}
	\caption{Inception network's accuracy over the simulated dataset, with respect to the network's depth as well as the length of the input time series.}
	\label{fig:depth-vs-length}
\end{figure}

\figurename~\ref{fig:depth-vs-length} illustrates how the model's accuracy varies with respect to the network's depth when classifying datasets of time series with different lengths. 
Our initial hypothesis was that as longer time series can potentially contain longer patterns and thus should require longer receptive fields in order for the network to separate the classes in the dataset. 
In terms of depth, this means that longer input time series will garner better results with deeper networks. 
And indeed, when observing \figurename~\ref{fig:depth-vs-length}, one can easily spot this trend: deeper networks deliver better results for longer time series.

\begin{figure}
	\centering
	\includegraphics[width=\linewidth]{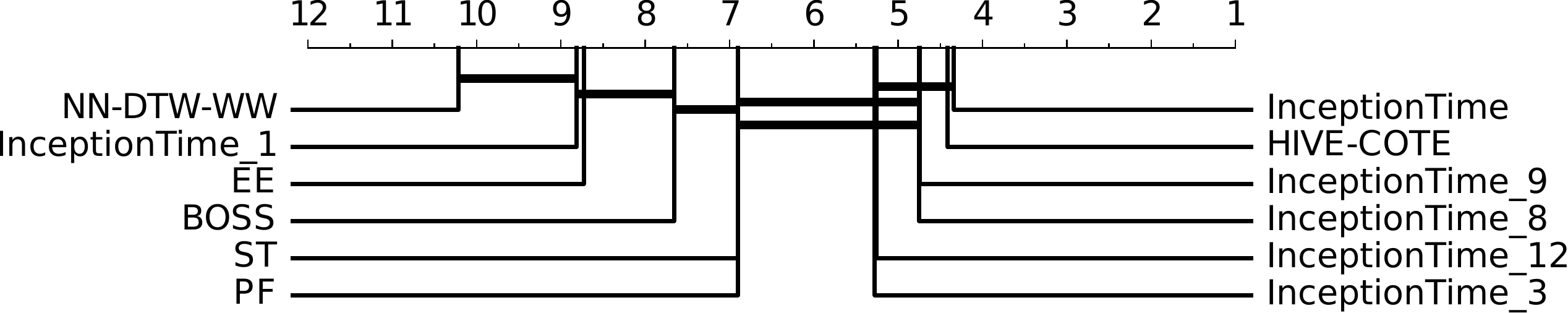}
	\caption{Critical difference diagram showing how the network's depth affects \ourmethod{}' average rank.}
	\label{fig:cd-diagram-inceptiontime-depth-with-bake-off}
\end{figure}

In order to further see how much effect the depth of a model has on real TSC datasets, we decided to implement deeper and shallower \ourmethod{} models, by varying the depth between 1 layer and 12 layers.
In fact, compared with the original architecture proposed by~\cite{wang2017time}, the deeper (shallower) version of \ourmethod{} will contain one additional (fewer) residual blocks each one comprised of three inception modules. 
By adding these layers, the deeper (shallower) \ourmethod{} model will contain roughly double (half) the number of parameters to be learned.
\figurename~\ref{fig:cd-diagram-inceptiontime-depth-with-bake-off} depicts the critical difference diagram comparing the deeper and shallower \ourmethod{} models to the original \ourmethod{}.

Unlike the experiments on the simulated dataset, we did not manage to improve the network's performance by simply increasing its depth. 
This may be due to many reasons, however it is likely due to the fact that deeper networks need more data to achieve high generalization capabilities~\citep{lecun2015deep}, and since the UCR/UEA archive does not contain datasets with a huge number of training instances, the deeper version of \ourmethod{} was overfitting the majority of the datasets and exhibited a small insignificant decrease in performance.
On the other hand, the shallower version of \ourmethod{} suffered from a significant decrease in accuracy (see \ourmethod{}\_3 and \ourmethod{}\_1 in \figurename~\ref{fig:cd-diagram-inceptiontime-depth-with-bake-off}). 
This suggests that a shallower architecture will contain a significantly smaller RF, thus achieving lower accuracy on the overall UCR/UEA archive.

From these experiments we can conclude that increasing the RF by adding more layers will not necessarily result in an improvement of the network's performance, particularly for datasets with a small training set.
However, one benefit that we have observed from increasing the network's depth, is to choose an RF that is long enough to achieve good results without suffering from overfitting.

We therefore proceed by experimenting with varying the RF by changing the filter length.

\subsection{Filter length}

\begin{figure}
	\centering
	\includegraphics[width=0.7\linewidth]{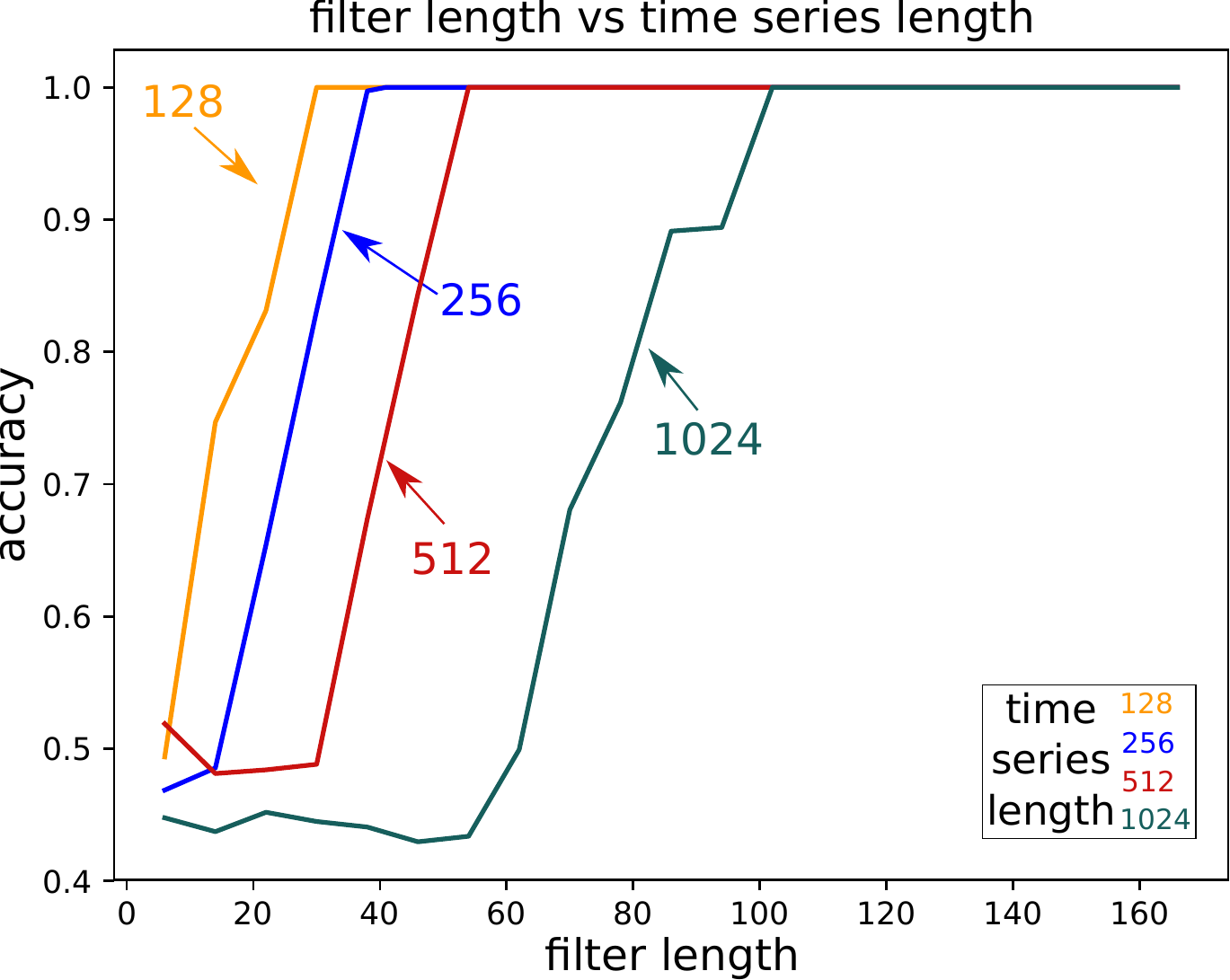}
	\caption{Inception network's accuracy over the simulated dataset, with respect to the filter length as well as the input time series length.}
	\label{fig:filter-vs-length}
\end{figure}

In order to test the effect of the filter length, we start by analyzing how the length of a time series influences the accuracy of the model when tuning this hyperparameter. 
In these experiments we fixed the number of training time series to 256 and the number of classes to 2. \figurename~\ref{fig:filter-vs-length} illustrates the results of this experiment.

We can easily see that as the length of the time series increases, a longer filter is required to produce accurate results. 
This is explained by the fact that longer kernels are able to capture longer patterns, with higher probability, than shorter ones can.
Thus, we can safely say that longer kernels almost always improve accuracy.

\begin{figure}
	\centering
	\includegraphics[width=0.7\linewidth]{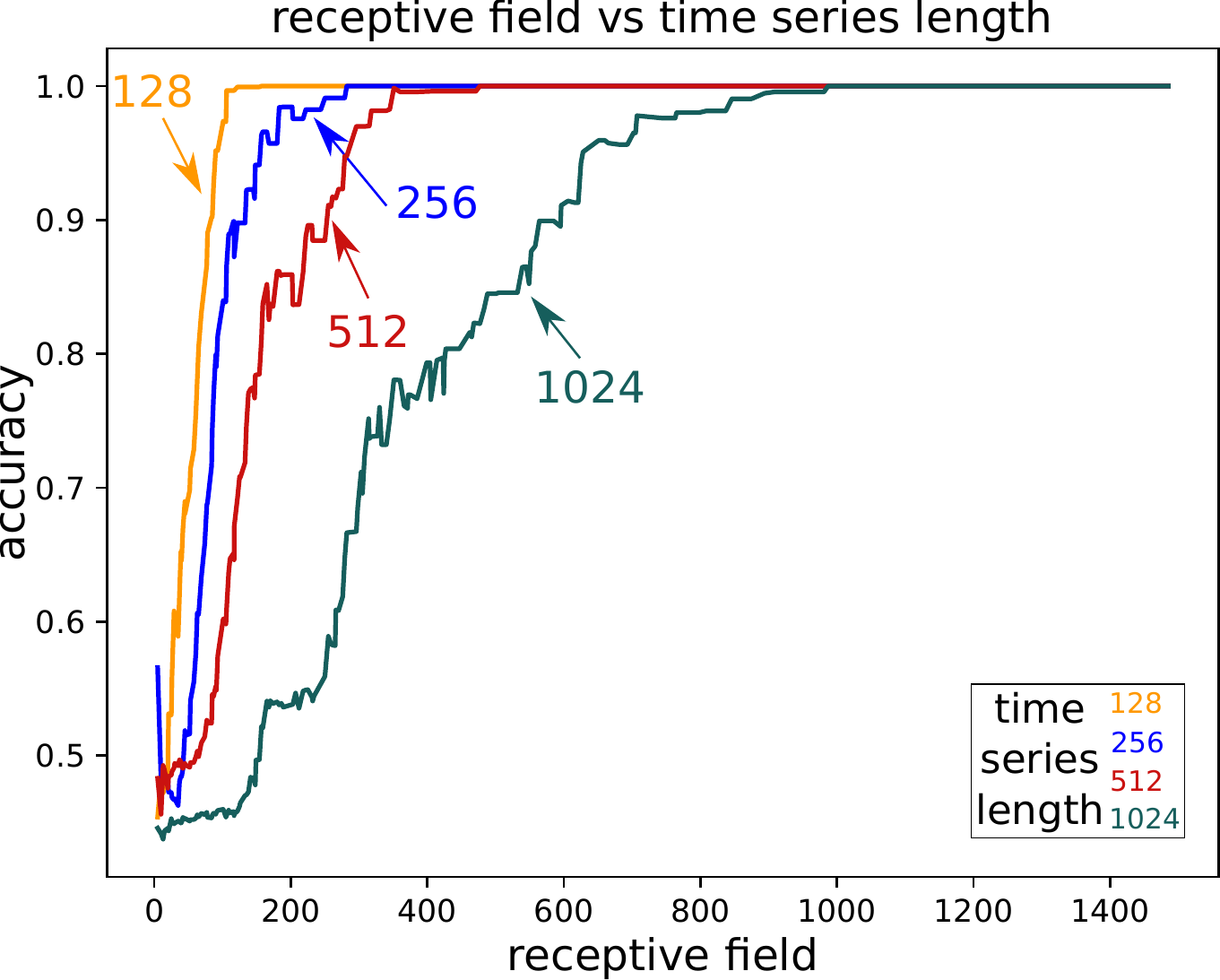}
	\caption{Inception network's accuracy over the simulated dataset, with respect to the receptive field as well as the input time series length.}
	\label{fig:plot-receptive-field}
\end{figure}

In addition to having visualized the accuracy as a function of both depth (\figurename~\ref{fig:depth-vs-length}) and filter length (\figurename~\ref{fig:filter-vs-length}), we proceed by plotting the accuracy as function of the RF for the simulated time series dataset with various lengths.
By observing \figurename~\ref{fig:plot-receptive-field} we can confirm the previous observations that longer patterns require longer RFs, with length clearly having a higher impact on accuracy compared to the network's depth.  
Moreover, by using a large enough RF to cover the whole input time series, the usage of a GAP layer won't affect \ourmethod{}'s ability to discriminate between the two patterns, because performing a GAP does not affect the model's RF.

\begin{figure}
	\centering
	\includegraphics[width=\linewidth]{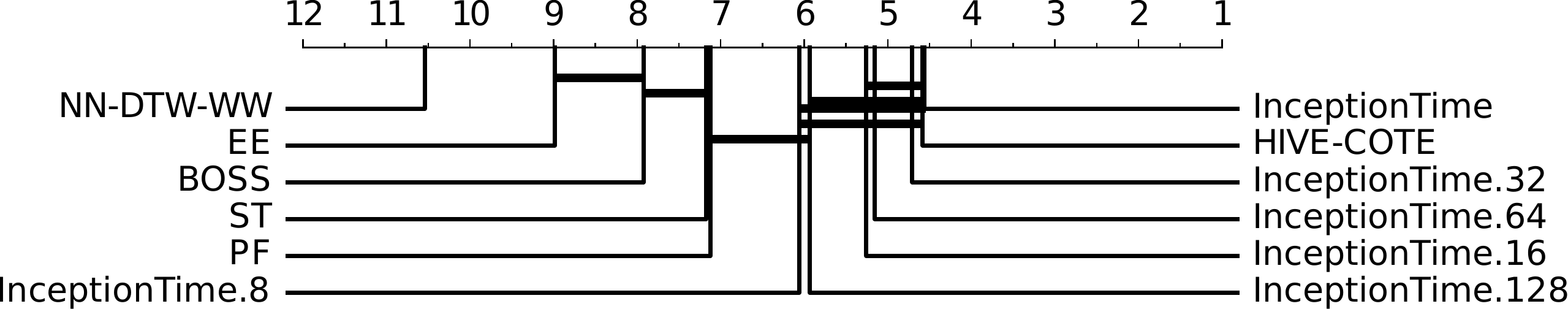}
	\caption{Critical difference diagram showing the effect of the filter length hyperparameter value over \ourmethod{}' average rank.}
	\label{fig:cd-diagram-inceptiontime-filtersize-with-bake-off}
\end{figure}

\begin{table}
	\centering
	\setlength\tabcolsep{4pt}
	{ \scriptsize
		\begin{tabularx}{0.7\textwidth}{Xccccccccc}
			\hline \\
			\small Length     & \small \ourmethod{}.8  & \small \ourmethod{}.64  & \small \ourmethod{}  \\
			\midrule \\ 
			\small $<$81   & \textbf{1.71} & 2.21 & 1.79   \\
			\small 81-250   & 1.89 & 2.11 & \textbf{1.42}   \\
			\small 251-450  & 2.45 & \textbf{1.32} & 1.86   \\
			\small 451-700  & 2.08 & 1.85 & \textbf{1.62} \\
			\small 701-1000 & \textbf{1.50}  & 2.60  & 1.80  \\
			\small $>$1000 & 2.14 & 2.00 & \textbf{1.71} \\
			\hline
		\end{tabularx}
	}
	\caption{Filter length variants of \ourmethod{} with their corresponding average ranks grouped by the datasets' length.
		Bold indicates the best model.}\label{tab-perf-lengths-inception}
\end{table}

There is a downside to longer filters however, in the potential for overfitting small datasets, as longer filters significantly increase the number of parameters in the network.
To answer this question, we again extend our experiments to the real data from the UCR/UEA archive, allowing us to verify whether long kernels tend to overfit the datasets when a limited amount of training data is available.
Therefore, we decided to train and evaluate \ourmethod{} versions containing both long and short filters on the UCR/UEA archive.
Where the original \ourmethod{} contained filters of length \{10,20,\textbf{40}\}, the five models we are testing here contain filters of length \{2,4,\textbf{8}\}; \{4,8,\textbf{16}\}; \{8,16,\textbf{32}\}; \{16,32,\textbf{64}\}; \{32,64,\textbf{128}\}.
\figurename~\ref{fig:cd-diagram-inceptiontime-filtersize-with-bake-off} illustrates a critical difference diagram showing how \ourmethod{} with longer filters will slightly decrease the network's performance in terms of accurately classifying the time series datasets.
We also investigate the relationship between the length of the time series and the length of the network's filter. 
\tablename~\ref{tab-perf-lengths-inception} depicts the average rank of each variant of \ourmethod{} over the UCR/UEA archive grouped by the datasets' lengths (with about 15 datasets in each group).
Similarly to \figurename~\ref{fig:cd-diagram-inceptiontime-filtersize-with-bake-off}, we observe that almost for all time series length, \ourmethod{} with its default filter length (32) achieves the best or the second best overall accuracy.
We can therefore summarize that the results from the simulated dataset do generalize (to some extent) to real datasets: longer filters will improve the model's performance as long as there is enough training data to mitigate the overfitting phenomena.

In summary, we can confidently state that increasing the receptive field of a model by adopting longer filters will help the network in learning longer patterns present in longer time series. However there is an accompanying disclaimer that it may negatively impact the accuracy for some datasets due to overfitting.

\subsection{Number of filters}

\begin{figure}
	\centering
	\includegraphics[width=0.7\linewidth]{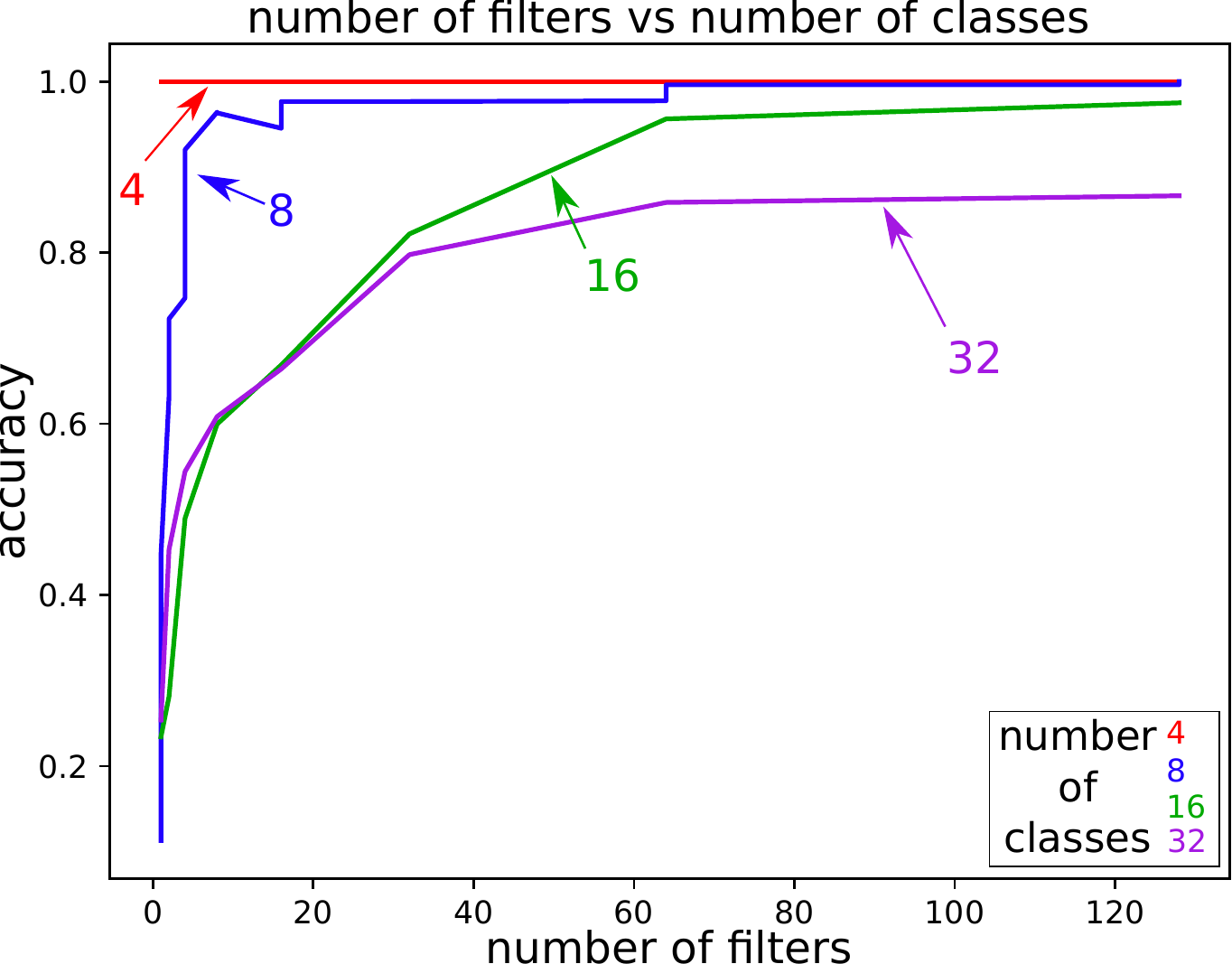}
	\caption{Inception network's accuracy over the simulated dataset, with respect to the number of filters as well as the number of classes.
	}
	\label{fig:nb-filters-vs-nb-classes}
\end{figure}

To provide some directions on how the number of filters affects the performance of the network, we experimented with varying this hyperparameter with respect to the number of classes in the dataset. 
To generate new classes in the simulated data, we varied the position and length of the patterns; for example, to create data with three classes, we inject patterns of the same length at three different positions.
For this series of experiments, we fixed the length of the time series to 256 and the number of training examples to 256.

\figurename~\ref{fig:nb-filters-vs-nb-classes} depicts the network's accuracy with respect to the number of filters for datasets with a differing number of classes.
Our prior intuition was that the more classes, or variability, present in the training set, the more features are required to be extracted in order to discriminate the different classes, and this will necessitate a greater number of filters. 
This is confirmed by the trend displayed in \figurename~\ref{fig:nb-filters-vs-nb-classes}, where the datasets with more classes require more filters to be learned in order to be able to accurately classify the input time series.

\begin{figure}
	\centering
	\includegraphics[width=\linewidth]{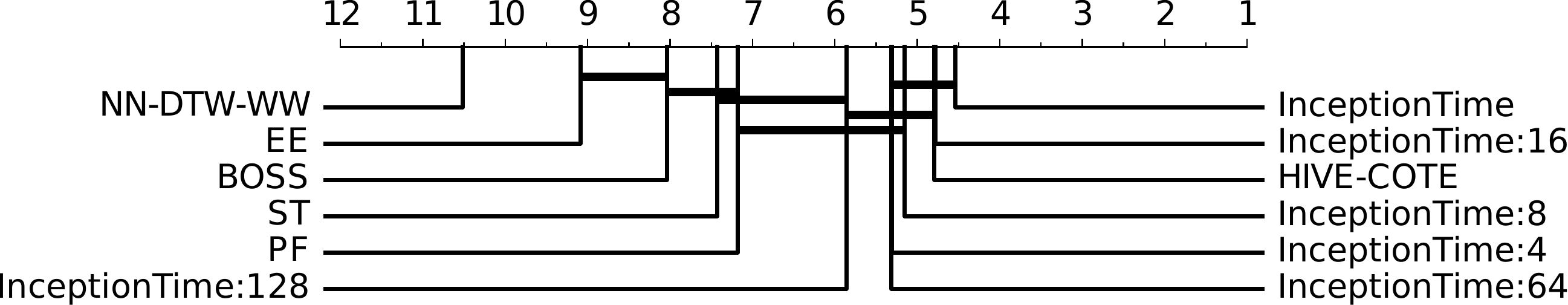}
	\caption{Critical difference diagram showing how the network's width affects \ourmethod{}' average rank.}
	\label{fig:cd-diagram-inceptiontime-nb-filters-with-bake-off}
\end{figure}

After observing on the synthetic dataset that the number of filters significantly affects the performance of the network, we asked ourselves if the current implementation of \ourmethod{} could benefit/lose from a naive increase/decrease in the number of filters per layer.
Our proposed \ourmethod{} model contains 32 filters per Inception module's component, while for these experiments we tested six ensembles, by varying the hyperparameter with a power of two.
\figurename~\ref{fig:cd-diagram-inceptiontime-nb-filters-with-bake-off} illustrates a critical difference diagram showing how increasing the number of filters per layer significantly deteriorated the accuracy of the network, whereas decreasing the number of filters did not significantly affect the accuracy. 
It appears that our \ourmethod{} model contains enough filters to separate the classes of the 85 UCR datasets, of which some have up to 60 classes (ShapesAll dataset).

Increasing the number of filters also has another side effect: it causes an explosion in the number of parameters in the network. 
The wider \ourmethod{} contains four times the number of parameters than the original implementation.
We therefore conclude that naively increasing the number of filters is actually detrimental, as it will drastically increase the network's complexity and eventually cause overfitting. 

\subsection{Sensitivity analysis}
Working with open benchmarks such as the UCR/UEA archive has pushed the community towards publishing high quality TSC algorithms. 
The UCR/UEA archive provides a train/test split for the data, which has allowed researchers to directly benchmark their works with the ones of others, as well as providing splits that were potentially more challenging and realistic than assuming that both train and test data were sampled from the same population. 
Having the train/test split available has however also led to the potential issue that the techniques designed on this benchmark archive might overfit it. 
This is especially true of deep learning classifiers that contain dozens of optimization and architectural hyperparameters. 

\begin{figure}
	\centering
	\includegraphics[width=\linewidth]{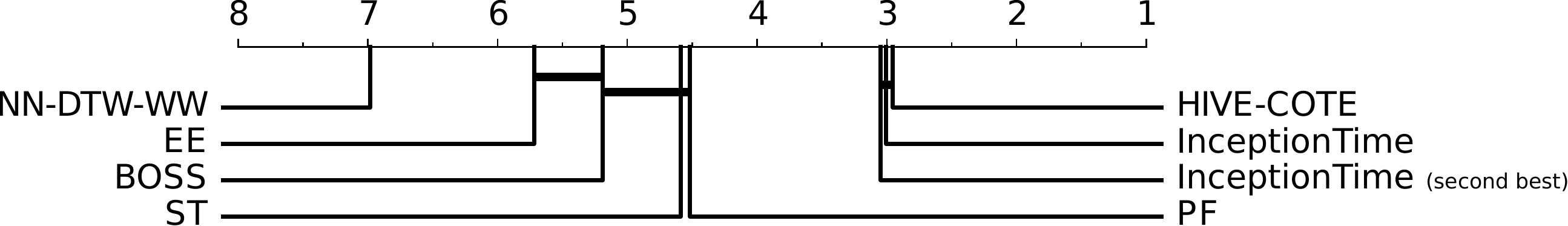}
	\caption{Critical difference diagram showing how choosing the second best hyperparameters affects \ourmethod{}'s average rank.}
	\label{fig:cd-diagram-inceptiontime-second-best}
\end{figure}

In an effort to give an idea of the sensitivity of \ourmethod{} to changes in its parameters, we have evaluated the performance of having chosen the second-best value of each of its parameters, that is the second-best value for the depth of the network (i.e. a value of 9 instead of the best value of 6), for its width (i.e. 16 instead of 32), for the length of the convolutions (final value of 32 instead of 40), for the batch size (i.e. 32 instead of 64), and for the bottleneck size (i.e. 64 instead of the default one 32). 
This gave us a new architecture ---~$\text{\ourmethod{}}_{\scriptsize{\text{(second best)}}}$~--- which we then compared with \ourmethod{} and also other algorithms. 
\figurename~\ref{fig:cd-diagram-inceptiontime-second-best} depicts the average rank of current state-of-the-art TSC algorithms with both \ourmethod{}'s default and second best hyperparameters added to the mix. 
We can clearly see that the effect is minimal: the ranking is a tiny bit lower but they are all well within the critical difference with HIVE-COTE (a post-hoc statistical test fails to reject the null hypothesis (p-value $\approx 0.71$) making the difference between the default and second best hyperparameters non significant). 

\section{Conclusion}
Deep learning for time series classification still lags behind neural networks for image recognition in terms of experimental studies and architectural designs. 
In this chapter, we fill this gap by introducing \ourmethod{}, inspired by the recent success of Inception-based networks for various computer vision tasks. 
We ensemble these networks to produce new state-of-the-art results for TSC on the 85 datasets of the UCR/UEA archive. 
Our approach is highly scalable, two orders of magnitude faster than current state-of-the-art models such as HIVE-COTE. 
The magnitude of this speed up is consistent across both Big Data TSC repositories as well as longer time series with high sampling rate. 
We further investigate the effects on overall accuracy of various hyperparameters of the CNN architecture. 
For these, we go far beyond the standard practices for image data, and design networks with long filters. 
We look at these by using a simulated dataset and frame our investigation in terms of the definition of the receptive field for a CNN for TSC.
In the future, we would like to explore how to design DNNs for multivariate TSC while investigating more recent architectural advancements that are being published each year for computer vision tasks.


\chapter{Time series analysis for surgical training} \label{Chapter4}

\section{Introduction}
Over the past one hundred years, the classic teaching methodology of ``see one, do one, teach one'' has governed the surgical education systems worldwide.
With the advent of Operation Room 2.0, recording video, kinematic and many other types of data during the surgery became an easy task, thus allowing artificial intelligence systems to be deployed and used in surgical and medical practice.
Recently, motion sensor data (e.g. kinematics) as well as surgical videos has been shown to provide a structure for peer coaching enabling novice trainees to learn from experienced surgeons by replaying those videos and/or kinematic trajectories.
In this chapter, we tackle two problems present in the current surgical training curriculum:
\begin{enumerate}
	\item Manual feedback from senior surgeons observing less experienced trainees is a laborious task that is very expensive, time-consuming and prone to subjectivity. 
	\item The high inter-operator variability in surgical gesture duration and execution renders learning from comparing novice to expert surgical videos a very difficult task.
\end{enumerate}
For the first problem, we designed a CNN (inspired by the FCN architecture explained in Chapter~\ref{Chapter1}) to predict surgical skills by extracting latent patterns in the trainees' motions performed during robotic surgery. 
As for the second problem, we propose to align multiple videos based on the alignment of their corresponding kinematic multivariate time series data, by leveraging the multiple alignment procedure based on DBA (explained in Chapter~\ref{Chapter2}). 
This chapter is divided into two main sections, each one tackling the aforementioned problem, before concluding with our future work. 

\section{Deep learning for surgical skills evaluation}



Over the last century, the standard training exercise of Dr. William Halsted has dominated surgical education in various regions of the world~\citep{polavarapu2013100}.
His training methodology of ``see one, do one, teach one'' is still one of the most adopted approaches to date~\citep{ahmidi2017a}.
The main idea is that the student could become an experienced surgeon by observing and participating in mentored surgeries~\citep{polavarapu2013100}.
These training techniques, although widely used, lack of an objective surgical skill evaluation method~\citep{kassahun2016surgical}. 
Standard assessment of surgical skills is presently based on checklists that are filled by an expert watching the surgical task~\citep{ahmidi2017a}.
In an attempt to predict a trainee's skill level without using on an expert surgeon's judgement, OSATS was proposed and is currently adopted for clinical practice~\citep{niitsu2013using}.
Alas, this type of observational rating still suffers from several external and subjective factors such as the inter-rater reliability, the development process and the bias of respectively the checklist and the evaluator~\citep{hatala2015constructing}. 

Further studies demonstrated that a vivid relationship occurs between a surgeon's technical skill and the postoperative outcomes~\citep{bridgewater2003surgeon}. 
The latter approach suffers from the fact that the aftermath of a surgery hinges on the physiological attributes of the patient~\citep{kassahun2016surgical}. 
Furthermore, obtaining this type of data is very strenuous, which renders these skill evaluation techniques difficult to carry out for surgical education.
Recent progress in surgical robotics such as the \emph{da Vinci} surgical system~\citep{davinci} enabled the recording of video and kinematic data from various surgical tasks.
Ergo, a substitute for checklists and outcome-based approaches is to generate, from these kinematics, GMFs such as the surgical task's speed, time completion, motion smoothness, curvature and other holistic characteristics~\citep{zia2017automated,kassahun2016surgical}. 
While most of these techniques are efficacious, it is not perspicuous how they could be leveraged to support the trainee with a detailed and constructive feedback, in order to go beyond a naive classification into a skill level (i.e., expert, intermediate, etc.). 
This is problematic as feedback on medical practice enables surgeons to reach higher skill levels while improving their performance~\cite{islam2016affordable}. 

Lately, a field entitled \emph{Surgical Data Science}~\citep{maier-hein2017surgical} has emerged by dint of the increasing access to a huge amount of complex data which pertain to the staff, the patient and sensors for capturing the procedure and patient related data such as kinematic variables and images~\citep{gao2014jhu}.
Instead of extracting GMFs, recent inquiries have a tendency to break down surgical tasks into finer segments called ``gestures'', manually before training the model, and finally estimate the trainees' performance based on their assessment during these individual gestures~\citep{lingling2012sparse}. 
Even though these methods achieved promising and accurate results in terms of evaluating surgical skills, they necessitate labeling a huge amount of gestures before training the estimator~\citep{lingling2012sparse}.
We pointed out two major limits in the actual existing techniques that estimate surgeons' skill level from their corresponding kinematic variables: firstly, the absence of an interpretable result of the skill prediction that can be used by the trainees to reach higher surgical skill levels; secondly, the requirement of gesture boundaries that are pre-defined by annotators which is prone to inter-annotator reliability and time-consuming~\citep{vedula2016analysis}. 

In this section, we design a novel architecture based on FCNs, dedicated to evaluating surgical skills.
By employing one-dimensional kernels over the kinematic time series, we avoid the need to extract unreliable and sensitive gesture boundaries.
The original hierarchical structure of our model allows us to capture global information specific to the surgical skill level, as well as to represent the gestures in latent low-level features.
Furthermore, to provide an interpretable feedback, instead of using a dense layer like most traditional deep learning architectures~\citep{zhou2016learning}, we place a GAP layer which allows us to take advantage from the class activation map, proposed originally by~\cite{zhou2016learning}, to localize which fraction of the trial impacted the model's decision when evaluating the skill level of a surgeon.
Using a standard experimental setup on the largest public dataset for robotic surgical data analysis: the JHU-ISI Gesture and Skill Assessment Working Set~\citep{gao2014jhu}, we show the precision of our FCN model. 
Our main contribution is to demonstrate that deep learning can be leveraged to understand the complex and latent structures when classifying surgical skills and predicting the OSATS score of a surgery, especially since there is still much to be learned on what does exactly constitute a surgical skill~\citep{kassahun2016surgical}.

\subsection{Background}
Here we turn our attention to the recent advances leveraging the kinematic data for surgical skills evaluation.
The problem we are interested in requires an input that consists of a set of time series recorded by the da Vinci's motion sensors representing the input surgery and the targeted task is to attribute a skill level to the surgeon performing a trial.
One of the earliest work focused on extracting GMFs from kinematic variables and training off-the-shelf classifiers to output the corresponding surgical skill level~\citep{kassahun2016surgical}.
Although these methods yielded impressive results, their accuracy depends highly on the quality of the extracted features. 
As an alternative to GMF-based techniques, recent studies tend to break down surgical tasks into smaller segments called surgical gestures, manually before the training phase, and assess the skill level of the surgeons based on their fine-grained performance during the surgical gestures, for example, using sparse hidden Markov model~\citep{lingling2012sparse}. 
Although the latter technique yields high accuracy, it requires manual segmentation of the surgical trial into fine-grained gestures, which is considered expensive and time-consuming. 
Hence, recent surgical skills evaluation techniques have focused on algorithms that do not require this type of annotation and are mainly data driven~\citep{IsmailFawaz2018evaluating,zia2017automated,wang2018deepsurgery,forestier2017discovering}.
For surgical skill evaluation, we distinguish two tasks. 
The first one is to output the discrete skill level of a surgeon such as novice (N), intermediate (I) or expert (E). 
For example, \cite{zia2017automated} adopted the approximate entropy algorithm to extract features from each trial which are later fed to a nearest neighbor classifier.
More recently,~\cite{wang2018deepsurgery} proposed a CNN-based approach to classify sliding windows of time series; therefore, instead of outputting the class for the whole surgery, the network is trained to output the class in an online setting for each window. 
In~\cite{forestier2018surgical}, the authors emphasized the lack of explainability for these latter approaches, by highlighting the fact that interpretable feedback to the trainees is important for a novice to become an expert surgeon~\citep{islam2016affordable}. 
Therefore, the authors proposed an approach that uses a sliding window technique with a discretization method that transforms the time series into a bag of words and trains a nearest neighbor classifier coupled with the cosine similarity.
Then, using the weight of each word, the algorithm is able to provide a degree of contribution for each sliding window and therefore give some sort of useful feedback to the trainees that explains the decision taken by the classifier. 
Although the latter technique showed interesting results, the authors did sacrifice the accuracy in favor of interpretability. 
On the other hand, using our fully convolutional neural network, we provide the trainee with an interpretable yet very accurate model by leveraging the class activation map algorithm, originally proposed for computer vision tasks by~\cite{zhou2016learning}.
The second type of problem in surgical skill evaluation is to train a model that predicts the OSATS score for a certain surgical trial.
For example, \cite{zia2017automated} extended their ApEn model to predict the OSATS score, also known as global rating score. 
Interestingly, the latter extension to a regression model instead of a classification one enabled the authors to propose a technique that provides interpretability of the model's decision, whereas our neural network provides an explanation for both classification and regression tasks. 

We present briefly the dataset used in this project as we rely on the features' definitions to describe our method.
The JIGSAWS dataset, first published by~\cite{gao2014jhu}, has been collected from eight right-handed subjects with three different surgical skill levels: novice (N), intermediate (I) and expert (E), with each group having reported, respectively, less than 10 h, between 10 and 100 h and more than 100 h of training on the Da Vinci. 
Each subject performed five trials of each one of the three surgical tasks: suturing, needle passing and knot tying.  
For each trial, the video and kinematic variables were registered.
In this project, we focused solely on the kinematics which are numeric variables of four manipulators: right and left masters (controlled by the subject's hands) and right and left slaves (controlled indirectly by the subject via the master manipulators). 
These 76 kinematic variables are recorded at a frequency of 30 Hz for each surgical trial.  
Finally, we should mention that in addition to the three self-proclaimed skill levels (N,I,E), JIGSAWS also contains the modified OSATS score~\citep{gao2014jhu}, which corresponds to an expert surgeon observing the surgical trial and annotating the performance of the trainee. 
The main goal of this work is to evaluate surgical skills by considering either the self-proclaimed discrete skill level (classification) or the OSATS score (regression) as our target variable.
We conceive each trial as a multivariate time series and designed a one-dimensional CNN dedicated to learn automatically useful features for surgical skill evaluation in an end-to-end manner~\citep{IsmailFawaz2018deep}.

\begin{figure}
	\centering
	\includegraphics[width=1.0\linewidth]{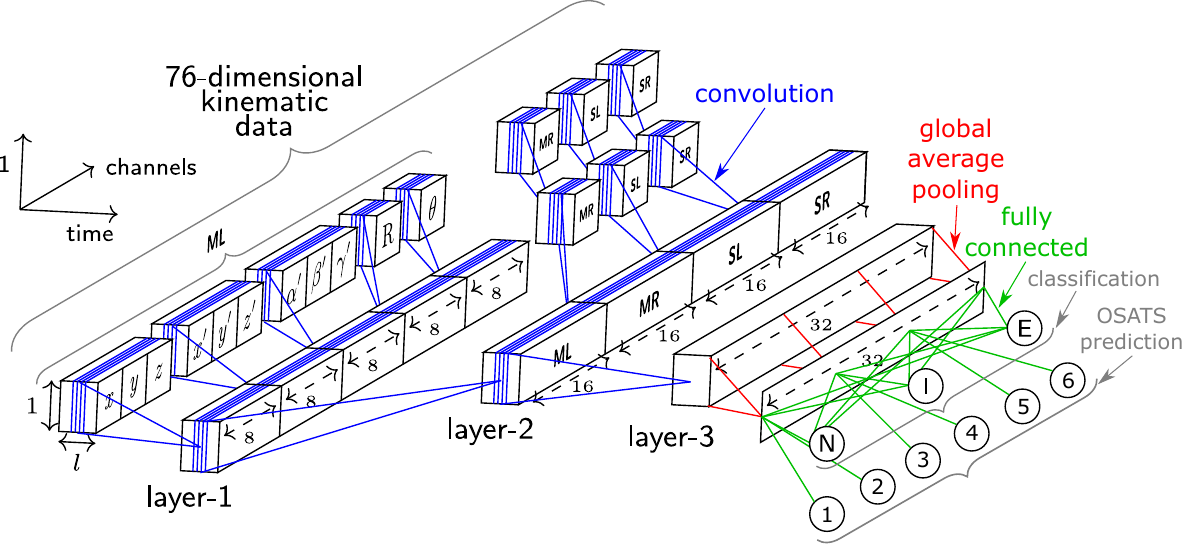}
	\caption{Fully convolutional network for surgical skill evaluation.}
	\label{fig:archi}
\end{figure}

\subsection{Method}

Our approach takes inspiration of the recent success of CNNs for time series classification shown in Chapter~\ref{Chapter1}. 
Figure~\ref{fig:archi} illustrates the fully convolutional neural network architecture, which we have designed specifically for surgical skill evaluation using temporal kinematic data.
The network's input is an MTS with a variable length $l$ and 76 channels. 
For the classification task, the output layer contains a number of neurons equal to three (N,I,E) with the softmax activation function, whereas for the regression task (predicting the OSATS score), the number of neurons in the last layer is equal to six: 
(1) ``Respect for tissue'';
(2) ``Suture/needle handling'';
(3) ``Time and motion'';
(4) ``Flow of operation'';
(5) ``Overall performance'';
(6) ``Quality of final product''~\citep{gao2014jhu}, with a linear activation function.

Compared with convolutions for image recognition, where usually the model's input exhibits two spatial dimensions (height and width) and three channels (red, green and blue), the input to our network is a time series with one spatial dimension (surgical task's length $l$) and 76 channels (denoting the 76 kinematics: $x,y,z,\dots$).
One of the main challenges we have encountered when designing our architecture was the large number of channels (76) compared to the traditional red, green and blue channels (3) for the image recognition problem. 
Hence, instead of applying the filters over the whole 76 channels at once, we propose to carry out different convolutions for each group and subgroup of channels.
We used domain knowledge when grouping the different channels, in order to decide which channels should be clustered together.

Firstly, we separate the 76 channels into four distinct groups, such as each group should contain the channels from one of the manipulators: the first, second, third and fourth groups correspond to the four manipulators (ML: master left, MR: master right, SL: slave left and SR: slave right) of the \emph{da Vinci} surgical system. 
Thus, each group assembles $19$ of the total kinematic variables.  
Next, each group of 19 channels is divided into five different subgroups each containing variables that we believe should be semantically clustered together.
For each cluster, the variables are grouped into the following five sub-clusters: 
\begin{itemize}
	\item First sub-cluster with three variables for the Cartesian coordinates ($x,y,z$);
	\item Second sub-cluster with three variables for the linear velocity ($x^{\prime},y^{\prime},z^{\prime}$);
	\item Third sub-cluster with three variables for the rotational velocity ($\alpha^{\prime},\beta^{\prime},\gamma^{\prime}$);
	\item Fourth sub-cluster with nine variables for the rotation matrix R;
	\item Fifth sub-cluster with one variable for the gripper angular velocity ($\theta$).  
\end{itemize}
Figure~\ref{fig:archi} illustrates how the convolutions in the first layer are different for each subgroup of kinematic variables. 
Following the same line of thinking, the convolutions in the second layer are different for each group of variables (SL, SR, ML and MR). 
However, in the third layer, the same filters are applied for all dimensions (or channels), which corresponds to the traditional CNN. 

To take advantage from the CAM method while reducing the number of parameters (weights) in our network, we employed a global average pooling operation after the last convolutional layer. 
In other words, the convolution's output (the MTS) will shrink from a length $l$ to 1, while maintaining the same number of dimensions in the third layer.
Without any sort of validation, we choose the following default hyperparameters.
We used $8$ kernels for the first convolution, and then we doubled the number of kernels, thus allowing us to balance the number of parameters for each layer as a function of its depth.
We used ReLU as the nonlinear hidden activation function for all convolutional layers with a stride of $1$ and a kernel length equal to $3$. 

We fixed our objective loss function to be the categorical cross-entropy to learn the network's parameters in an end-to-end manner for the classification task, and the mean squared error when learning a regressor to predict the OSATS score, which can be written as: 
\begin{equation}
\mathrm{MSE}=\frac{1}{n}\sum_{i=1}^{n}(Y_i-\hat{Y}_i)^2.
\end{equation}
The network's weights were optimized using the Adam optimization algorithm~\citep{kingma2015adam}. 
The default value of the learning rate was fixed to $0.001$ as well as the first and second moment estimates were set to $0.9$ and $0.999$ respectively~\citep{chollet2015keras}. 
We initialized the weights using Glorot's uniform initialization~\citep{glorot2010understanding}.
We randomly shuffled the training set before each epoch, whose maximum number was set to $1000$ epochs.
We then saved the model at each training iteration by choosing the network's state that minimizes the loss function on a random (non-seen) split from the training data.
This process is also referred to as ``model checkpoint'' by the deep learning community~\citep{chollet2015keras}, allowing us to choose the best number of epochs based on the validation loss.
Finally, to avoid overfitting, we added an $l2$ regularization parameter whose default value was fixed to $10^{-5}$.
For each surgical task, we have trained a different network, resulting in three different models.\footnote{\scriptsize Our source code is available at \url{https://github.com/hfawaz/ijcars19}}
We adopted for both classification and regression tasks a leave-one-super-trial-out scheme~\citep{ahmidi2017a}.

The use of a GAP layer allows us to employ the CAM algorithm, which was originally designed for image classification tasks by~\cite{zhou2016learning} and later introduced for time series data in~\cite{wang2017time}. 
Using the CAM, we are able to highlight which fractions of the surgical trial contributed highly to the classification.
This method was discussed in details in Chapter~\ref{Chapter1}.
Finally, for the regression task, the CAM can be extended in a trivial manner: Instead of computing the contribution to a classification, we are computing the contribution to a certain score prediction (1 out of 6 in total). 

\begin{table}
	\centering
	\setlength\tabcolsep{2.0pt}
	\begin{tabularx}{1.0\textwidth}{l|ccc|ccc|ccc}
		\toprule
		\multirow{2}{*}{\scriptsize Method} &
		\multicolumn{3}{c}{\scriptsize Suturing} &
		\multicolumn{3}{c}{\scriptsize Needle passing} &
		\multicolumn{3}{c}{\scriptsize Knot tying} \\
		& {\scriptsize Micro} & {\scriptsize Macro} & {\scriptsize $\rho$} & {\scriptsize Micro} & {\scriptsize Macro} & {\scriptsize $\rho$} & {\scriptsize Micro} & {\scriptsize Macro} & {\scriptsize $\rho$} \\
		\midrule
		\small S-HMM~\citep{lingling2012sparse} & 97.4 & n/a & n/a  & 96.2 & n/a& n/a & 94.4 & n/a& n/a \\
		\small ApEn~\citep{zia2017automated} & \textbf{100} & n/a & 0.59 & \textbf{100} & n/a & 0.45 & \textbf{99.9} & n/a & \textbf{0.66}\\
		\small Sax-Vsm~\citep{forestier2017discovering} & 89.7 & 86.7& n/a & 96.3 & 95.8& n/a & 61.1 & 53.3& n/a \\
		\small CNN~\citep{wang2018deepsurgery} & 93.4 & n/a& n/a & 89.9 & n/a& n/a & 84.9 & n/a& n/a \\
		\small FCN (proposed) & \textbf{100} & \textbf{100} & \textbf{0.60} & \textbf{100} & \textbf{100} &\textbf{0.57} & 92.1 & \textbf{93.2} &0.65 \\
		\bottomrule
	\end{tabularx}
	\caption{Micro, macro and Spearman's coefficient $\rho$ for surgical skill evaluation.}
	\label{tab:res-class}
\end{table}

\subsection{Results}

The first task consists in assigning a skill level for an input surgical trial out of the three possible levels: novice (N), intermediate (I) and expert (E).
In order to compare with current state-of-the-art techniques, we adopted the \emph{micro} and \emph{macro} measures defined in~\cite{ahmidi2017a}.
The \emph{micro} measure refers simply to the traditional \emph{accuracy} metric. 
However, the \emph{macro} takes into consideration the support of each class in the dataset, which boils down to computing the \emph{precision} metric. 
Table~\ref{tab:res-class} reports the \emph{macro} and \emph{micro} metrics of five different models for the surgical skill classification of the three tasks: suturing, knot tying and needle passing. 
For the proposed FCN model, we average the accuracy over 40 runs to reduce the bias induced by the randomness of the optimization algorithm. 
From these results, it appears that FCN is much more accurate than the other approaches with 100\% accuracy for the needle passing and suturing tasks.
As for the knot tying task, we report 92.1\% and 93.2\%, respectively, for the \emph{micro} and \emph{macro} configurations.
When comparing the other four techniques, for the knot tying surgical task, FCN exhibits relatively lower accuracy, which can be explained by the minor difference between the experts and intermediates for this task: Mean OSATS score is 17.7 and 17.1 for expert and intermediate, respectively.  

An S-HMM was designed to classify surgical skills~\citep{lingling2012sparse}.
Although this approach does leverage the gesture boundaries for training purposes, our method is much more accurate without the need to manually segment each surgical trial into finer gestures. 
\cite{zia2017automated} introduced ApEn to generate characteristics from each surgical task, which are later given to a classical nearest neighbor classifier with a cosine similarity metric. 
Although ApEn and FCN achieved state-of-the-art results with 100\% accuracy for the first two surgical tasks, it is still not obvious how ApEn could be used to give feedback for the trainee after finishing his/her training session.
\cite{forestier2017discovering} introduced a sliding window technique with a discretization method to transform the MTS into bag of words.
To justify their low accuracy, \cite{forestier2017discovering} insisted on the need to provide \emph{explainable} surgical skill evaluation for the trainees. 
On the other hand, FCN is equally \emph{interpretable} yet much more accurate; in other words, we do not sacrifice accuracy for interpretability. 
Finally,~\cite{wang2018deepsurgery} designed a CNN whose architecture is dependent on the length of the input time series.
This technique was clearly outperformed by our model which reached better accuracy by removing the need to pre-process time series into equal length thanks to the use of GAP.

We extended the application of our FCN model~\citep{IsmailFawaz2018evaluating} to the regression task: predicting the OSATS score for a given input time series. 
Although the community made a huge effort toward standardizing the comparison between different surgical skills evaluation techniques~\citep{ahmidi2017a}, we did not find any consensus over which evaluation metric should be adopted when comparing different regression models. 
However,~\cite{zia2017automated} proposed the use of Spearman's correlation coefficient (denoted by $\rho$) to compare their 11 combination of regression models. 
The latter is a nonparametric measure of rank correlation that evaluates how well the relationship between two distributions can be described by a monotonic function.
In fact, the regression task requires predicting six target variables; therefore, we compute $\rho$ for each target and finally report the corresponding mean over the six predictions.
By adopting the same validation methodology proposed by~\cite{zia2017automated}, we are able to compare our proposed FCN model to their best performing method. 
Table~\ref{tab:res-class} reports also the $\rho$ values for the three tasks, showing how FCN reaches higher $\rho$ values for two out of three tasks. 
In other words, the prediction and the ground truth OSATS score are more correlated when using FCN than the ApEn-based solution proposed by~\cite{zia2017automated} for the second task and equally correlated for the other two tasks.

\begin{figure}
	\centering
	\subfloat[Suturing task for an expert]{
		
		\includegraphics[width=.35\linewidth]{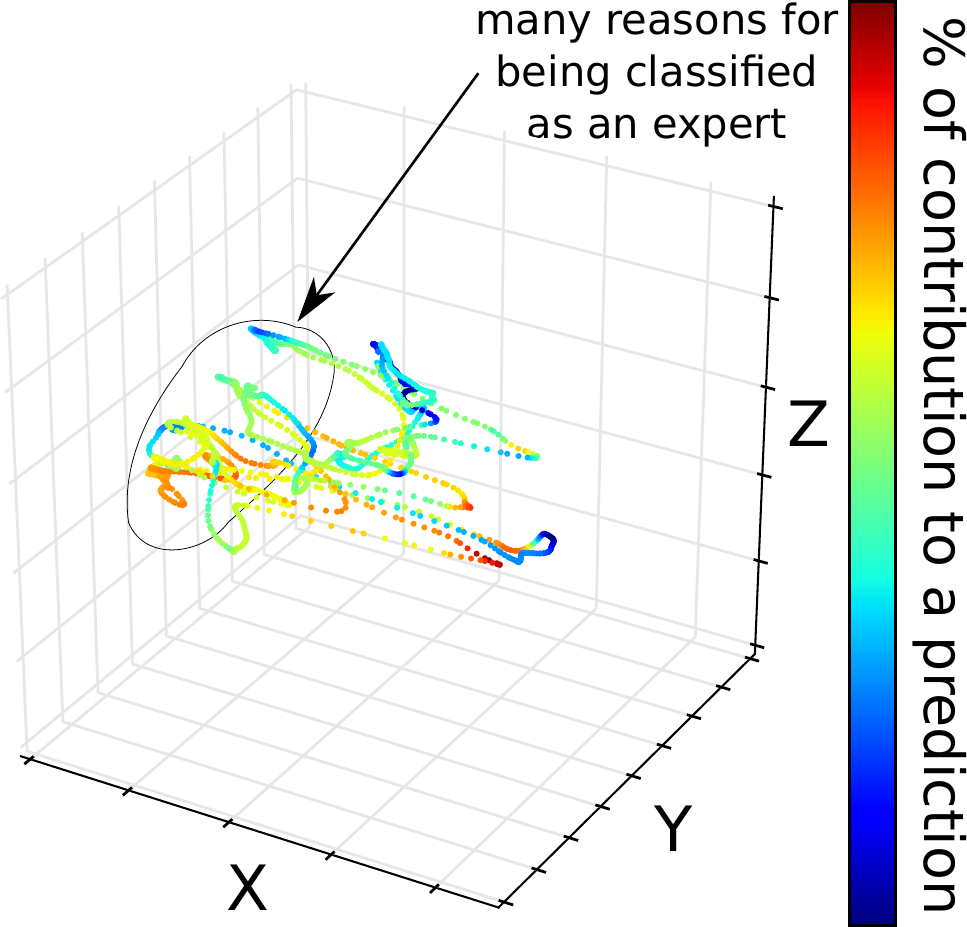}
		\label{sub:screenshot}}
	\hspace{.1cm}
	\subfloat[Suturing task for a novice]{
		\includegraphics[width=.35\linewidth]{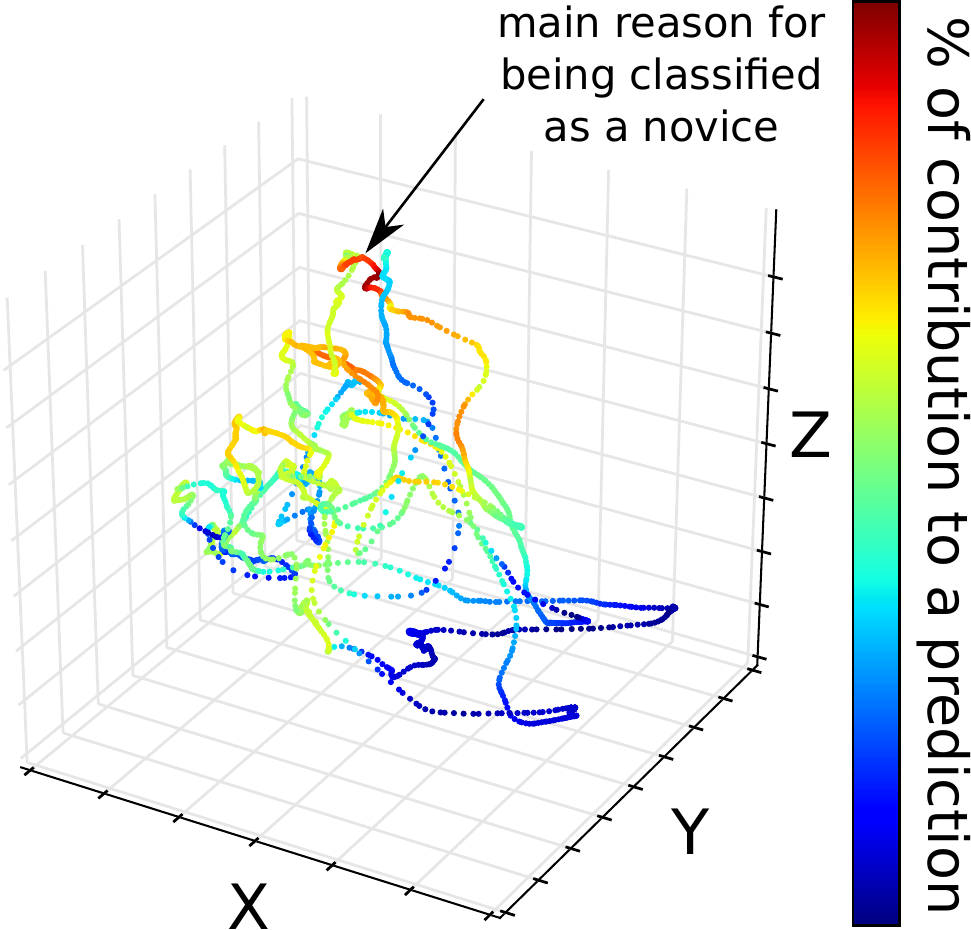}
		\label{sub:feedback}
	}
	\caption{Using class activation map to provide explainable classification.}
	\label{fig:trials}
\end{figure}

\begin{figure}
	\centering
	\subfloat[Suture/needle handling]{
		\includegraphics[width=0.35\linewidth]{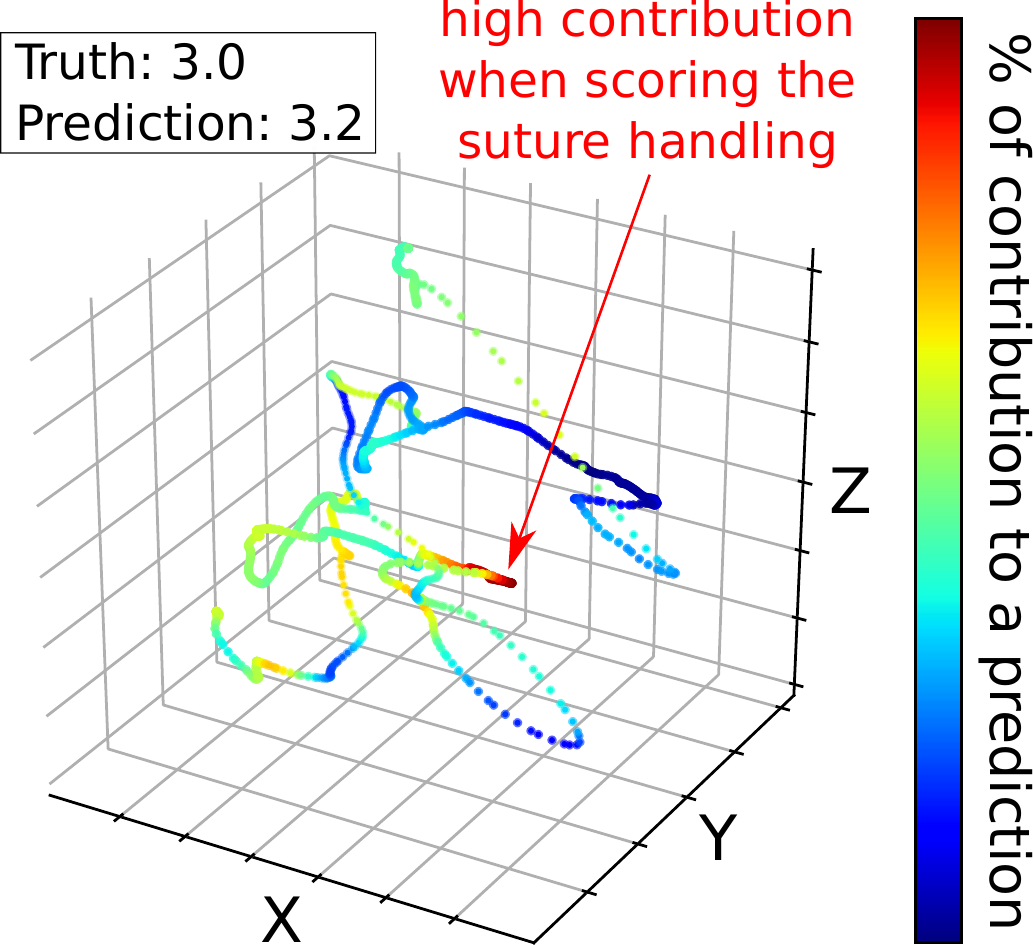}
		\label{sub-reg-2-kt-e002}}
	\subfloat[Quality of the final product]{
		\includegraphics[width=0.35\linewidth]{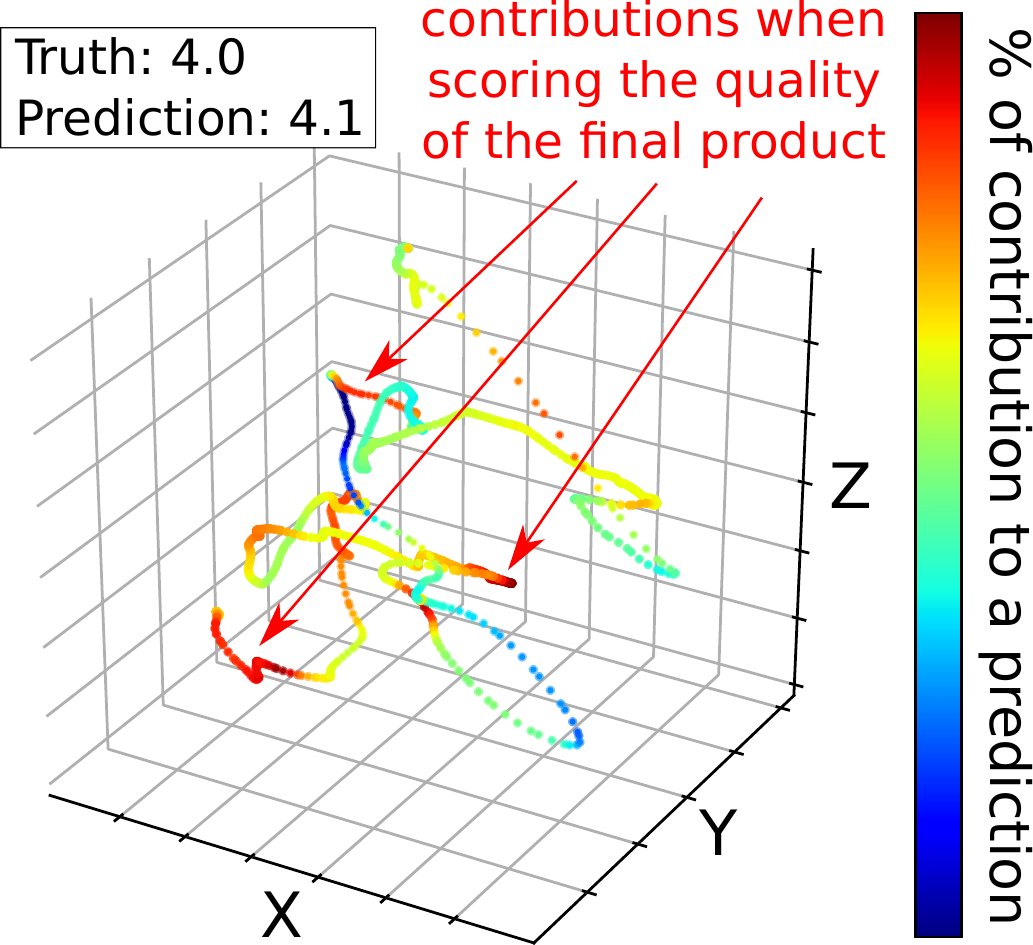}
		\label{sub-reg-6-kt-e002}}
	\caption{Feedback using the CAM on subject E's second knot-tying trial.}
	\label{fig-cam-reg-kt}
\end{figure}

The CAM technique allows us to visualize which parts of the trial contributes the most to a skill classification. 
By localizing, for example, discriminative behaviors specific to a skill level, observers can start to understand motion patterns specific to certain class of surgeons. 
To further improve themselves (the novice surgeons), the model, using the CAM's result, can pinpoint to the trainees their good/bad motor behaviors. 
This would potentially enable novices to achieve greater performance and eventually become experts. 

By generating a heatmap from the CAM, we can see in \figurename~\ref{fig:trials} how it is indeed possible to visualize the feedback for the trainee. 
In fact, we examine a trial of an expert and novice surgeon: The expert's trajectory is illustrated in \figurename~\ref{sub:screenshot} while the novice's trajectory is depicted in \figurename~\ref{sub:feedback}.
In this example, we can see how the model was able to identify which motion (red subsequence) is the main reason for identifying a subject as a novice. 
Concretely, we can easily spot a pattern that is being recognized by the model when outputting the classification of subject H's skill level: The orange and red 3D subsequences correspond to same surgical gesture ``pulling suture'' and are exhibiting a high influence over the model's decision. 
This feedback could be used to explain to a young surgeon which movements are classifying him/her as a novice and which ones are classifying another subject as an expert. 
Thus ultimately, this sort of feedback could guide the novices into becoming experts.

After having shown how our \emph{classifier} can be interpreted to provide feedback to the trainees, we now present the result of applying the same visualization (based on the CAM algorithm) in order to explain the OSATS score prediction. 
Figure~\ref{fig-cam-reg-kt} depicts the trajectory with its associated heatmaps for subject E performing the second trial of the knot-tying task. 
Figure~\ref{sub-reg-2-kt-e002} and \ref{sub-reg-6-kt-e002} illustrates the trajectory's heatmap, respectively, for ``suture/needle handling'' and ``quality of the final product'' OSATS score predictions. 
At first glimpse, one can see how a prediction that requires focusing on the whole surgical trial leverages more than one region of the input surgery---this is depicted by the multiple red subsequences in \figurename~\ref{sub-reg-6-kt-e002}.
However, when outputting a rating for a specific task such as ``suture/needle handling''---the model is focusing on less parts of the input trajectory which is shown in \figurename~\ref{sub-reg-2-kt-e002}. 



\subsection{Conclusion}

In this project, we proposed a deep learning-based method for surgical skills evaluation from kinematic data. 
We achieved state-of-the-art accuracy by designing a specific FCN, while providing explainability that justifies a certain skill evaluation, thus allowing us to mitigate the CNN's black-box effect.
Furthermore, by extending our architecture we were able to provide new state-of-the-art performance for predicting the OSATS score from the input kinematic time-series data. 
In the following section, we present a technique for automatic alignment of surgical videos from kinematic time series~\citep{IsmailFawaz2019automatic} for surgical skills evaluation. 

\section{Automatic alignment of surgical videos using kinematic time series data}



Educators have always searched for innovative ways of improving apprentices' learning rate.
While classical lectures are still most commonly used, multimedia resources are becoming more and more adopted~\citep{smith2000digital} especially in Massive Open Online Courses~\citep{means2009evaluation}.
In this context, videos have been considered as especially interesting as they can combine images, text, graphics, audio and animation.
The medical field is no exception, and the use of video-based resources is intensively adopted in medical curriculum~\citep{masic2008learning} especially in the context of surgical training~\citep{kneebone2002innovative}.
The advent of robotic surgery also simulates this trend as surgical robots, like the Da Vinci~\citep{davinci}, generally record video feeds during the intervention.
Consequently, a large amount of video data has been recorded in the last ten years~\citep{rapp2016youtube}.
This new source of data represent an unprecedented opportunity for young surgeons to improve their knowledge and skills~\citep{gao2014jhu}.
Furthermore, video can also be a tool for senior surgeons during teaching periods to assess the skills of the trainees.
In fact, a recent study by~\cite{mota2018video} showed that residents spend more time viewing videos than specialists, highlighting the need for young surgeons to fully benefit from the procedure.
In~\cite{herrera2016the}, the authors showed that knot-tying scores and times for task completion improved significantly for the subjects that watched the videos of their own performance. 

However, when the trainees are willing to asses their progress over several trials of the same surgical task by re-watching their recorded surgical videos simultaneously, the problem of videos being out-of-synch makes the comparison between different trials very difficult if not impossible.  
This problem is encountered in many real life case studies, since experts on average complete the surgical tasks in less time than novice surgeons~\citep{mcNatt2001}. 
Thus, when trainees do enhance their skills, providing them with a feedback that pinpoints the reason behind the surgical skill improvement becomes problematic since the recorded videos exhibit different duration and are not perfectly aligned.  

Although synchronizing videos has been the center of interest for several computer vision research venues, contributions are generally focused on a special case where multiple simultaneously recorded videos (with different characteristics such as viewing angles and zoom factors) are being processed~\citep{wolf2002sequence,wedge2005trajectory,padua2010linear}.
Another type of multiple video synchronization uses hand-engineered features (such as points of interest trajectories) from the videos~\citep{wang2014videosnapping,evangelidis2011efficient}, making the approach highly sensitive to the quality of the extracted features.  
This type of techniques was highly effective since the raw videos were the only source of information available, whereas in our case, the use of robotic surgical systems enables capturing an additional type of data: the kinematic variables such as the $x,y,z$ Cartesian coordinates of the Da Vinci's end effectors~\citep{gao2014jhu}. 

\begin{figure}[t]
	\centering
	\subfloat[Video without alignment]{
		\includegraphics[width=.47\linewidth]{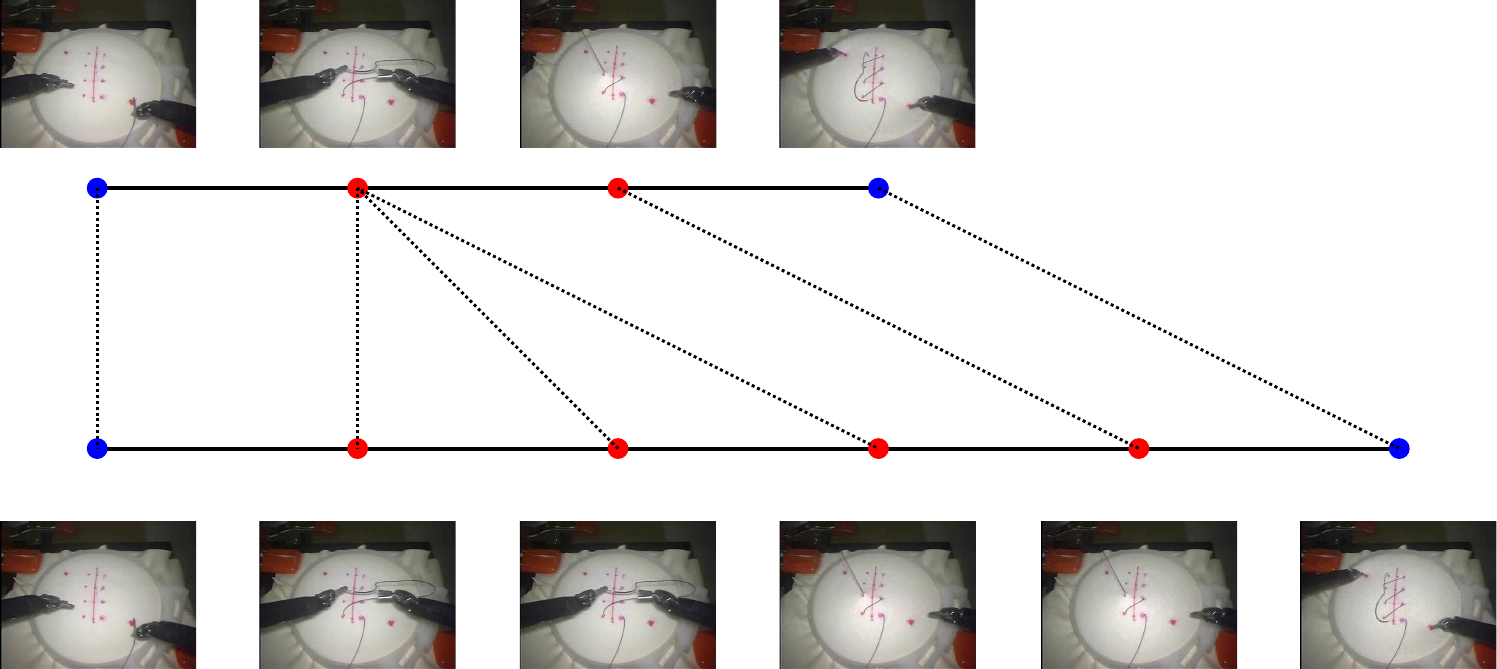}
		\label{sub:vid}
	}
	\subfloat[Video with alignment]{
		\includegraphics[width=.47\linewidth]{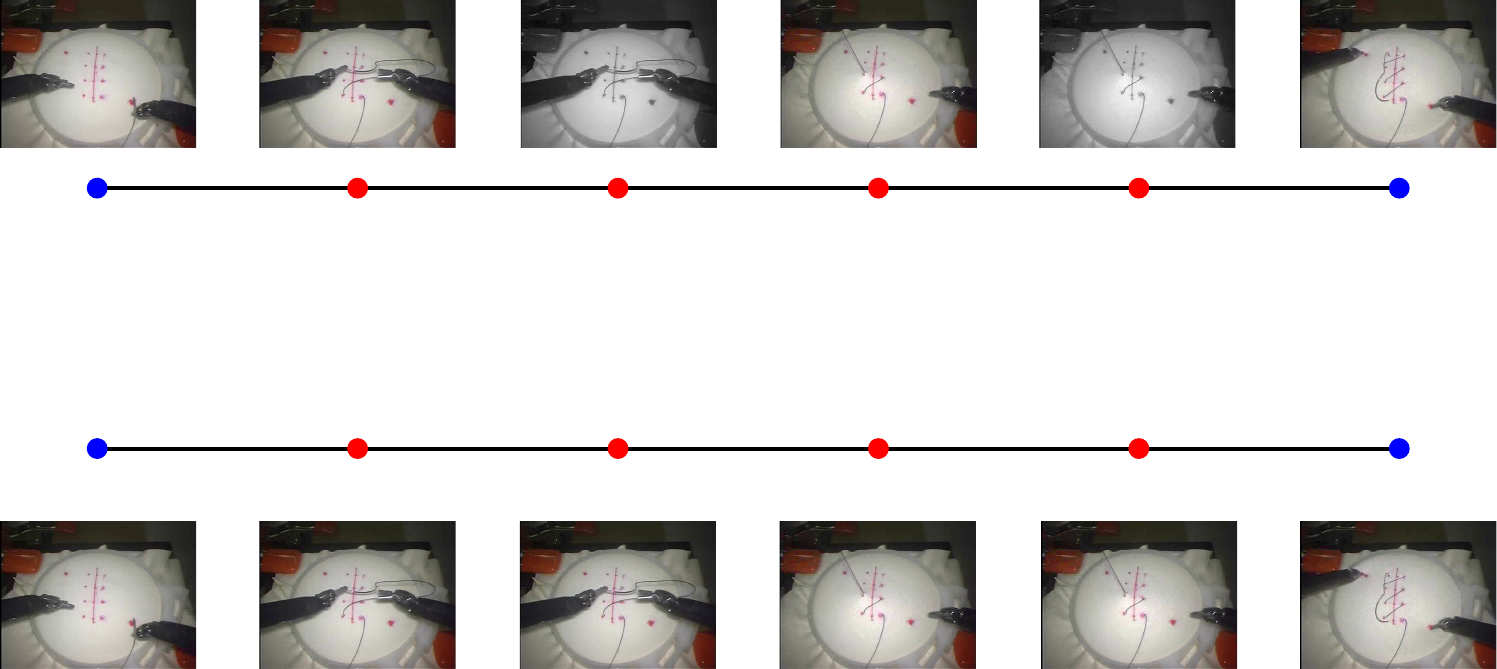}
		\label{sub:vid-synch}
	}
	\caption{Example on how a time series alignment is used to synchronize the videos by duplicating the gray-scale frames. Best viewed in color.}
	\label{fig:vid-synch}
\end{figure}

\begin{figure}[t]
	\centering
	\subfloat[Original time series without alignment]{
		\includegraphics[width=.65\linewidth]{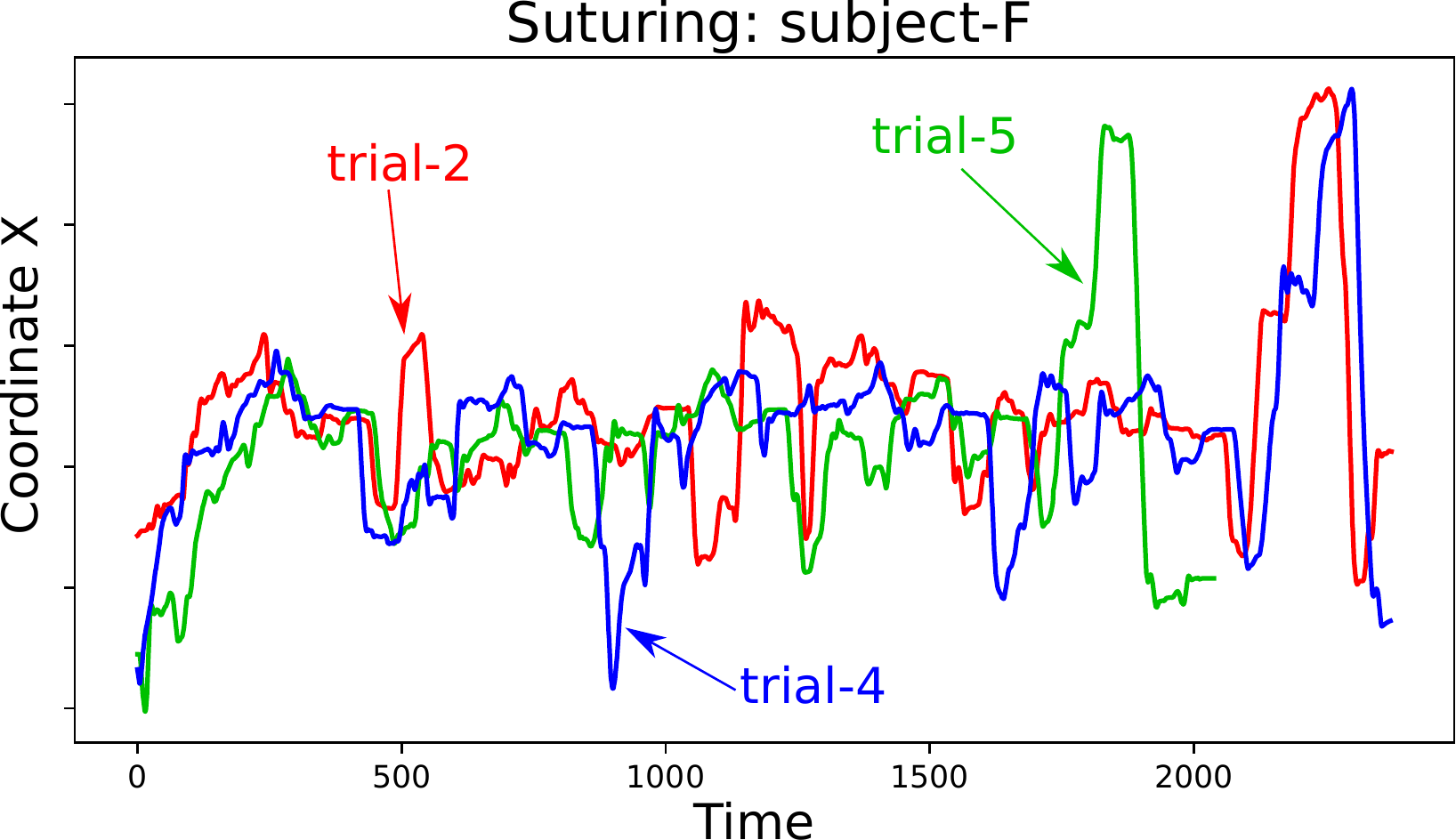}
	}\\
	\subfloat[Warped time series with alignment]{
		\includegraphics[width=.65\linewidth]{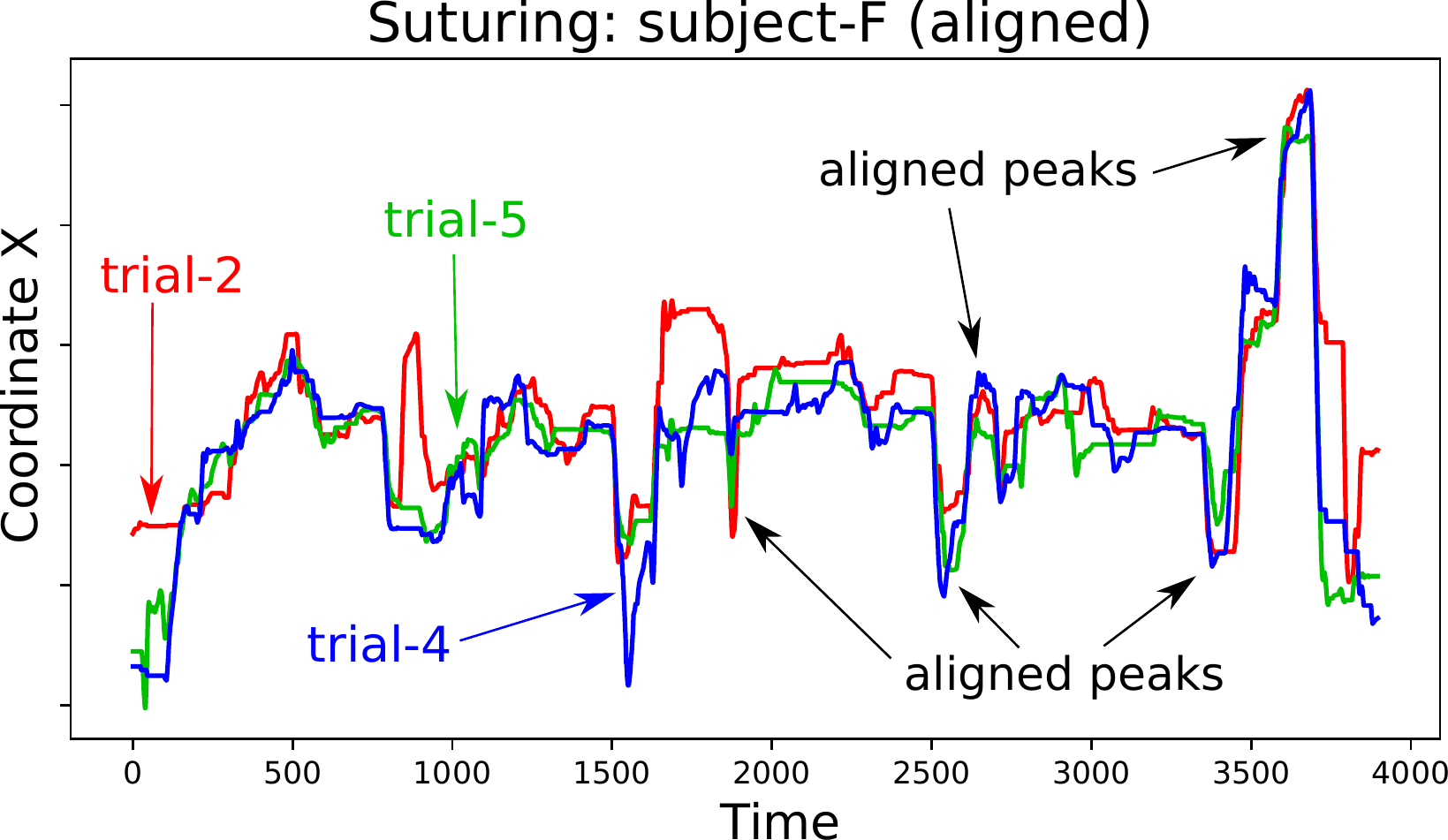}
	}
	\caption{Example of aligning coordinate X's time series for subject F, when performing three trials of the suturing surgical task.}
	\label{fig:trials-align}
\end{figure}

In this section, we propose to leverage the sequential aspect of the recorded kinematic data from the Da Vinci surgical system, in order to synchronize their corresponding video frames by aligning the time series data (see Figure~\ref{fig:vid-synch} for an example). 
When aligning two time series, the off-the-shelf algorithm is DTW~\citep{sakoe1978dynamic} which we indeed used to align two videos. 
However, when aligning multiple sequences, the latter technique does not generalize in a straightforward and computationally feasible manner~\citep{petitjean2014dynamic}. 
Hence, for multiple video synchronization, we propose to align their corresponding time series to the average time series, computed using the DBA algorithm (explained in details in Chapter~\ref{Chapter2}).
This process is called Non-Linear Temporal Scaling and has been originally proposed to find the multiple alignment of a set of discretized surgical gestures~\citep{forestier2014non}, which we extend in this work to continuous numerical kinematic data.
Figure~\ref{fig:trials-align} depicts an example of stretching three different time series using the NLTS algorithm.   
Examples of the synchronized videos and the associated code can be found on our GitHub repository\footnote{\url{https://github.com/hfawaz/aime19}}, where we used the JHU-ISI Gesture and Skill Assessment Working Set~\citep{gao2014jhu} to validate our work. 


\subsection{Methods}

Here, we detail each step of our video synchronization approach. 
We start by describing the DTW algorithm which allows us to align two videos. 
Then, we describe how NLTS enables us to perform multiple video synchronization with respect to the reference average time series computed using the DBA algorithm. 

\subsubsection{Dynamic Time Warping}

Dynamic Time Warping was first proposed for speech recognition when aligning two audio signals~\citep{sakoe1978dynamic}. 
Suppose we want to compute the dissimilarity between two time series, for example two different trials of the same surgical task, $A=(a_1,a_2,\dots,a_m)$ and $B=(b_1,b_2,\dots,b_n)$.
The length of $A$ and $B$ are denoted respectively by $m$ and $n$, which in our case correspond to the surgical trial's duration.  
Here, $a_i$ is a vector that contains six real values, therefore $A$ and $B$ can be seen as two distinct MTS. 

To compute the DTW dissimilarity between two MTS, several approaches were proposed by the time series data mining community~\citep{shokoohi2017generalizing}, however in order to apply the subsequent algorithm NLTS, we adopted the ``dependent'' variant of DTW where the Euclidean distance is used to compute the difference between two instants $i$ and $j$. 
Let $M(A,B)$ be the $m\times n$ point-wise dissimilarity matrix between $A$ and $B$, where $M_{i,j}=||a_i-b_j||^2$. 
A warping path $P=((c_1,d_1),(c_2,d_2),\dots,(c_s,d_s))$ is a series of points that define a crossing of $M$. 
The warping path must satisfy three conditions: (1) $(c_1,d_1)=(1,1)$; (2) $(c_s,d_s)=(m,n)$; (3) $0\le c_{i+1}-c_i \le 1$ and $0\le d_{j+1}-d_j \le 1$ for all $i<m$ and $j<n$. 
The DTW measure between two series corresponds to the path through $M$ that minimizes the total distance. 
In fact, the distance for any path $P$ is equal to $D_P(A,B)=\sum_{i=1}^{s}P_i$.
Hence if $\textbf{P}$ is the space of all possible paths, the optimal one - whose cost is equal to $DTW(A,B)$ - is denoted by $P^*$ and can be computed using: $\min_{P\in \textbf{P}}D_P(A,B)$. 

The optimal warping path can be obtained efficiently by applying a dynamic programming technique to fill the cost matrix $M$. 
Once we find this optimal warping path between $A$ and $B$, we can deduce how each time series element in $A$ is linked to the elements in $B$. 
We propose to exploit this link in order to identify which time stamp should be duplicated in order to align both time series, and by duplicating a time stamp, we are also duplicating its corresponding video frame. 
Concretely, if elements $a_i$, $a_{i+1}$ and $a_{i+2}$ are aligned with the element $b_j$ when computing $P^*$, then by duplicating twice the video frame in $B$ for the time stamp $j$, we are dilating the video of $B$ to have a length that is equal to $A$'s. 
Thus, re-aligning the video frames based on the aligned Cartesian coordinates: if subject $S_1$ completed ``\textit{inserting the needle}'' gesture in 5 seconds, whereas subject $S_2$ performed the same gesture within 10 seconds, our algorithm finds the optimal warping path and duplicates the frames for subject $S_1$ in order to synchronize with subject $S_2$ the corresponding gesture.     
Figure~\ref{fig:vid-synch} illustrates how the alignment computed by DTW for two time series can be used in order to duplicate the corresponding frames and eventually synchronize the two videos.  

\subsubsection{Non-Linear Temporal Scaling}

The previous DTW based algorithm works perfectly when synchronizing only two surgical videos. 
The problem arises when aligning three or more surgical trials simultaneously, which requires a multiple series alignment. 
The latter problem has been shown to be NP-Complete~\citep{wang1994} with the exact solution requiring $O(L^N)$ operations for $N$ sequences of length $L$. 
This is clearly not feasible in our case where $L$ varies between $10^3$ and $10^4$ and $N\ge3$, which is why we ought to leverage an approximation of the multiple sequence alignment solution provided by the DBA algorithm which we detailed in the Chapter~\ref{Chapter2}. 

%

NLTS was originally proposed for aligning discrete sequences of surgical gestures~\citep{forestier2014non}.
In this project, we extend the technique for numerical continuous sequences (time series). 
The goal of this final step is to compute the approximated multiple alignment of a set of sequences $D$ which will eventually contain the precise information on how much a certain frame from a certain series should be duplicated. 
We first start by computing the average sequence $T$ (using DBA) for a set of time series $D$ that we want to align simultaneously. 
Then, by recomputing the Compact Multiple Alignment between the refined average $T$ and the set of time series $D$, we can extract an alignment between $T$ and each sequence in $D$. 
Thus, for each time series in $D$ we will have the necessary information (extracted from the multiple alignment) in order to dilate the time series appropriately to have a length that is equal to $T$'s, which also corresponds to the length of the longest time series in $D$.   
Figure~\ref{fig:trials}, depicts an example of aligning three different time series using the NLTS algorithm.

\subsection{Experiments}

We start by describing the JIGSAWS dataset we have used for evaluation, before presenting our experimental study. 

\subsubsection{Dataset}

\begin{figure*}[t]
	\centering
	\includegraphics[width=.325\linewidth]{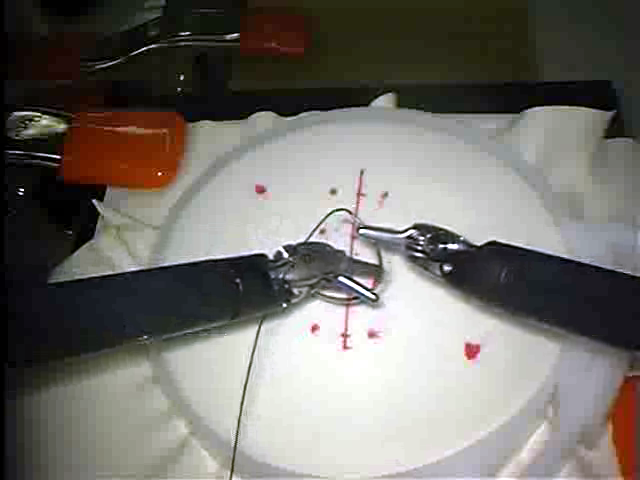}
	\includegraphics[width=.325\linewidth]{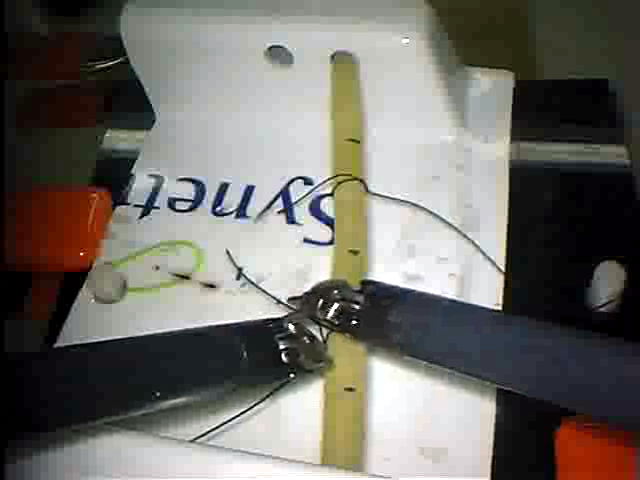}
	\includegraphics[width=.325\linewidth]{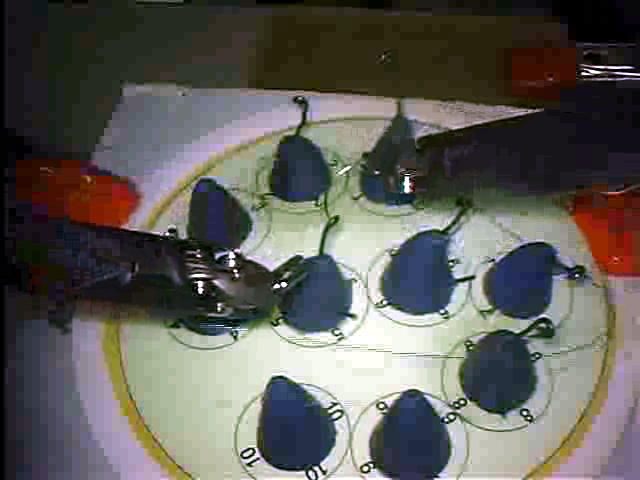}
	\caption{Snapshots of the three surgical tasks in the JIGSAWS dataset (from left to right): suturing, knot-tying, needle-passing \citep{gao2014jhu}.}
	\label{fig:jig}
\end{figure*}

The JIGSAWS dataset~\citep{gao2014jhu} includes data for three basic surgical tasks performed by study subjects (surgeons). 
The three tasks (or their variants) are usually part of the surgical skills training program. 
Figure~\ref{fig:jig} shows a snapshot example for each one of the three surgical tasks (Suturing, Knot Tying and Needle Passing). 
The JIGSAWS dataset contains kinematic and video data from eight different subjects with varying surgical experience: two experts (E), two intermediates (I) and four novices (N) with each group having reported respectively more than 100 hours, between 10 and 100 hours and less than 10 hours of training on the Da Vinci. 
All subjects were reportedly right-handed. 

The subjects repeated each surgical task five times and for each trial the kinematic and video data were recorded. 
When performing the alignment, we used the kinematic data which are numeric variables of four manipulators: left and right masters (controlled directly by the subject)  and  left  and  right slaves (controlled indirectly by the subject via the master manipulators).
These kinematic variables (76 in total) are captured at a frequency equal to 30 frames per second for each trial.
Out of these 76 variables, we only consider the Cartesian coordinates ($x,y,z$) of the left and right slave manipulators, thus each trial will consist of an MTS with 6 temporal variables. 
We chose to work only with this subset of kinematic variables to make the alignment coherent with what is visible in the recorded scene: the robots' end-effectors which can be seen in Figure~\ref{fig:jig}. 
However other choices of kinematic variables are applicable, which we leave the exploration for our future work.
Finally we should mention that in addition to the three self-proclaimed skill levels (N,I,E) JIGSAWS contains the modified OSATS score~\citep{gao2014jhu}, which corresponds to an expert surgeon observing the surgical trial and annotating the performance of the trainee. 

\subsubsection{Results}

\begin{figure}[t]
	\centering
	\subfloat[Videos synchronization process]{
		\includegraphics[width=.58\linewidth]{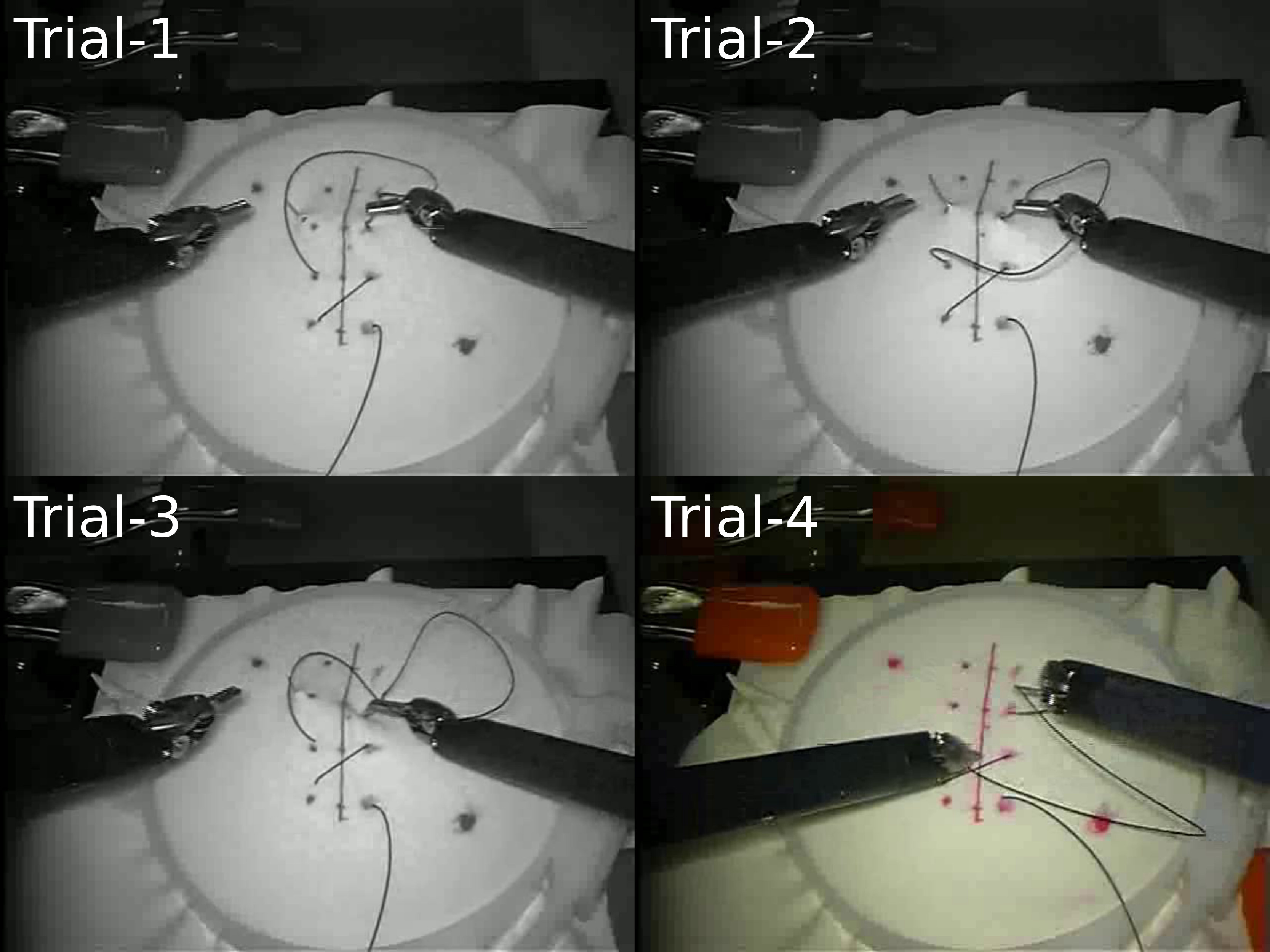}
		\label{sub:video-4x4-unsynched}
	}\\
	\subfloat[Perfectly aligned videos]{
		\includegraphics[width=.58\linewidth]{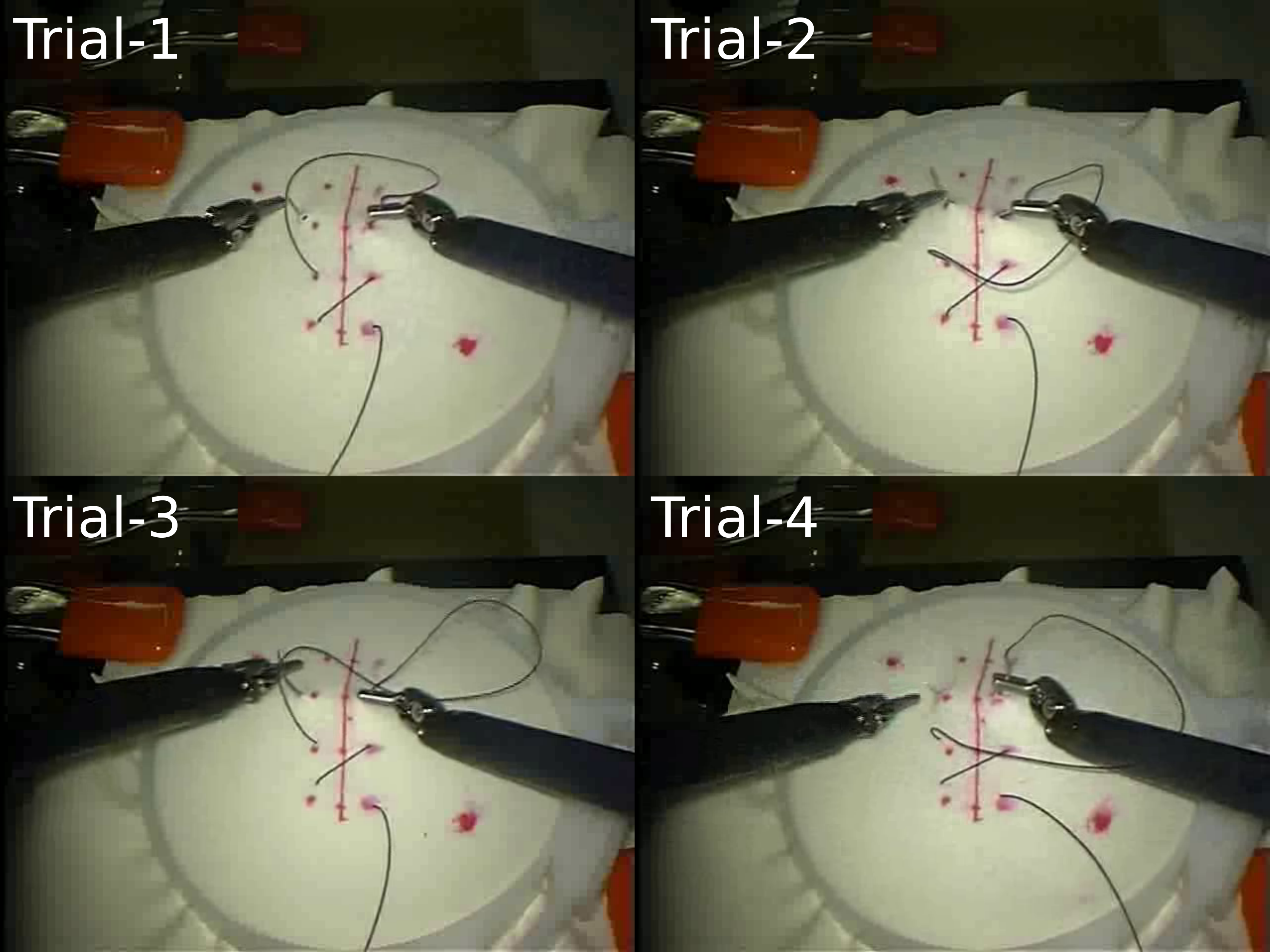}
		\label{sub:video-4x4-synched}
	}
	\caption{Video alignment procedure with duplicated (gray-scale) frames.}
	\label{fig:video-4x4}
\end{figure}

We have created a companion web page\footnote{\url{https://germain-forestier.info/src/aime2019/}} to our project where several examples of synchronized videos can be found. 
Figure~\ref{fig:video-4x4} illustrates the multiple videos alignment procedure using our NLTS algorithm, where gray-scale images indicate duplicated frames (paused video) and colored images indicate a surgical motion (unpaused video).  
In Figure~\ref{sub:video-4x4-unsynched} we can clearly see how the gray-scale surgical trials are perfectly aligned. 
Indeed, the frozen videos show the surgeon ready to perform ``\textit{pulling the needle}'' gesture~\citep{gao2014jhu}.
On the other hand, the colored trial (bottom right of Figure~\ref{sub:video-4x4-unsynched}) shows a video that is being played, where the surgeon is performing ``\textit{inserting the needle}'' gesture in order to catch up with the other paused trials in gray-scale.
Finally, the result of aligning simultaneously these four surgical trials is depicted in Figure~\ref{sub:video-4x4-synched}. 
By observing the four trials, one can clearly see that the surgeon is now performing the same surgical gesture ``\textit{pulling the needle}'' simultaneously for the four trials.
We believe that this type of observation will enable a novice surgeon to locate which surgical gestures still need some improvement in order to eventually become an expert surgeon. 

\begin{figure}
	\centering
	\includegraphics[width=0.7\linewidth]{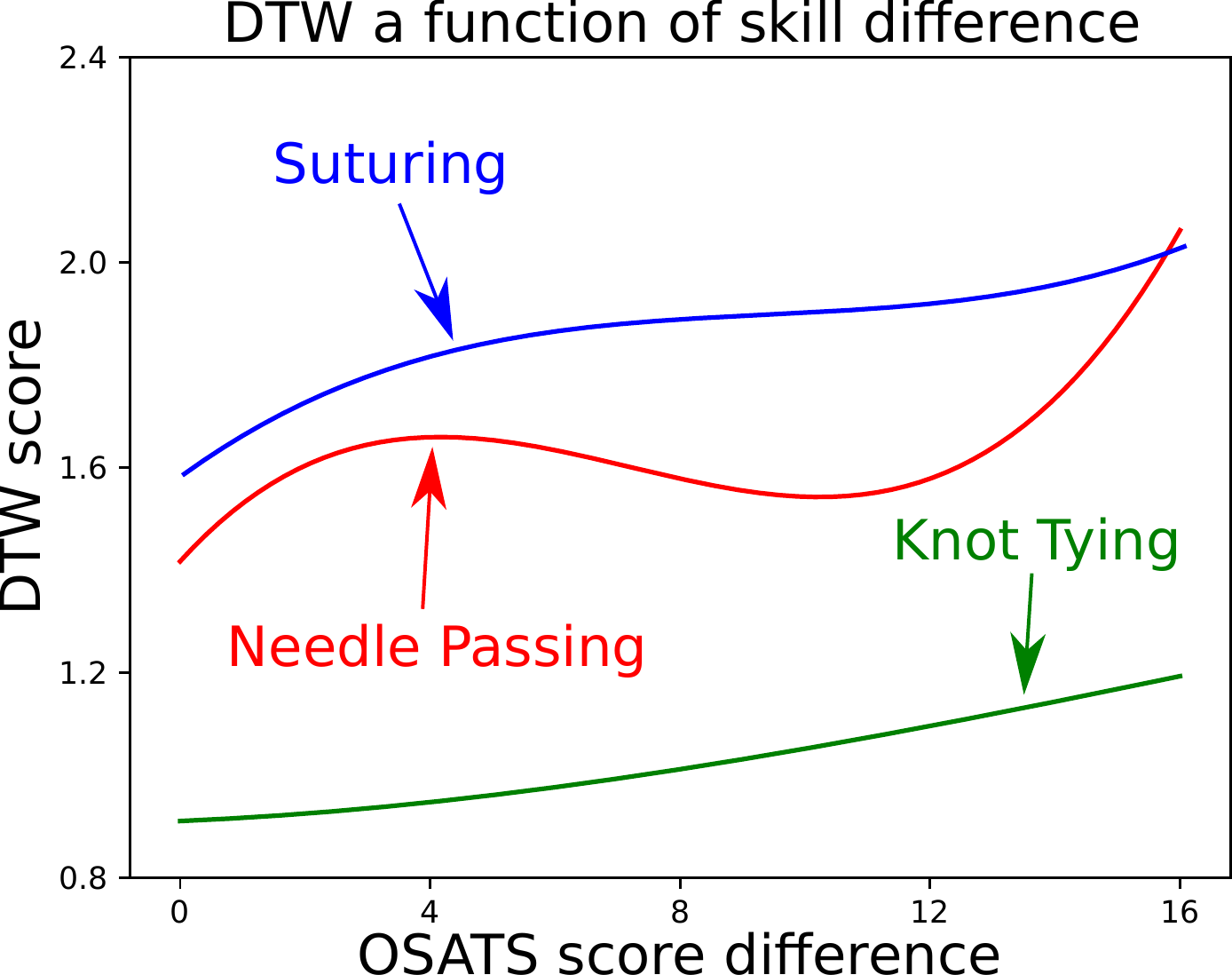}
	\caption{A polynomial fit (degree 3) of DTW dissimilarity score (y-axis) as a function of the OSATS score difference between two surgeons (x-axis).}
	\label{fig:dtw-osats}
\end{figure}

Furthermore, in order to validate our intuition that DTW is able to capture characteristics that are in relationship with the motor skill of a surgeon, we plotted the DTW distance as a function of the OSATS~\citep{gao2014jhu} score difference. 
For example, if two surgeons have both an OSATS score of 10 and 16 respectively, the corresponding difference is equal to $|10-16|=6$. 
In Figure~\ref{fig:dtw-osats}, we can clearly see how the DTW score increases whenever the OSATS score difference increases. 
This observation suggests that the DTW score is low when both surgeons exhibit similar dexterity, and high whenever the trainees show different skill levels. 
Therefore, we conclude that the DTW score can serve as a heuristic for estimating the quality of the alignment (whenever annotated skill level is not available) - especially since we observed low quality alignments for surgeons with very distinct surgical skill levels.

Finally, we should note that this work is suitable for many research fields involving motion kinematic data with their corresponding video frames. 
Examples of such medical applications are assessing mental health from videos~\citep{yamada2017detecting} where wearable sensor data can be seen as time series kinematic variables and leveraged in order to synchronize a patient's videos and compare how well the patient is responding to a certain treatment. 
Following the same line of thinking, this idea can be further applied to kinematic data from wearable sensors coupled with the corresponding video frames when evaluating the Parkinson's disease evolution~\citep{criss2011} as well as infant grasp skills~\citep{li2019manipulation}.

\subsection{Conclusion}\label{sec-conc}
In this section, we showed how kinematic time series data recorded from the Da Vinci's end effectors can be leveraged in order to synchronize the trainee's videos performing a surgical task. 
With personalized feedback during surgical training becoming a necessity~\citep{IsmailFawaz2018e,forestier2018surgical}, we believe that replaying \emph{synchronized} and well \emph{aligned} videos would benefit the trainees in understanding which surgical gestures did or did not improve after hours of training, thus enabling them to further reach higher skills and eventually become experts. 
We acknowledge that this work needs an experimental study to quantify how beneficial is replaying synchronized videos for the trainees versus observing non-synchronized trials.
Therefore, we leave such exploration and clinical try outs to our future work.  

\section{Conclusion}

In this chapter, we tackled two different surgical skills related problems. 
First by designing a one dimensional FCN, we were able to reach state-of-the-art results for the surgical skills evaluation (classification and regression). 
Doing so, we were able to mitigate the DNNs' black-box effect, using the CAM technique in order to highlight the reason behind a certain surgical skill identification.
The second surgical skill related problem was due to surgical training videos being our-of-synch, thus making it hard for trainees to understand and compare videos between different surgeons with various skill levels. 
We tackled the latter problem by proposing the use of NLTS in order to align and synchronize multiple videos simultaneously. 
These two projects were orthogonal in the sense that they could also complement each other: perhaps video and time series synchronization could be a pre-processing step that would enhance the performance surgical skills evaluation models. 
We leave such exploration to our future work when extending these two projects.

\addchap{Conclusion and future works}

\label{conclusion}

\section*{Overview of contributions}
\addcontentsline{toc}{section}{Overview of contributions}

The work presented in this manuscript first examined the current benchmark approaches in the area of deep learning for time series classification. 
Using the knowledge gained from this study, a number of new methods have been proposed to contribute to the deep learning for TSC literature, leading up to the proposal for the final classifier \ourmethod{}, which to the best of our knowledge constitutes the first time series classifier to achieve similar results to the famous HIVE-COTE ensemble, while outperforming it significantly in terms of running-time. 
We should emphasize that in order to foster reproducibility, the corresponding code for each paper is published on the following GitHub profile: \url{https://github.com/hfawaz}.

The main goal of this research was to answer the question outlined in Chapter~\ref{Chapter1}: Is there a current DNN approach that reaches state-of-the-art performance for TSC and is less complex than HIVE-COTE?
Most previous work on TSC haven't considered deep learning models as potential strong classifiers, which led to researchers proposing neural networks as baselines with little to no competitive performance in terms of accuracy when compared to other non deep learning based techniques such as HIVE-COTE. 
Arriving to our benchmark in 2018, we showed how neural networks can compete with state-of-the-art TSC algorithms, while providing interpretability using the CAM method. 

Furthermore, in a series of research projects around regularizing DNNs for TSC, we were able to showcase how much more accuracy we can squeeze from a vanilla neural network architecture designed originally as a baseline. 
We started by proposing a novel DBA-based method for predicting the best source dataset for a given target dataset when fine-tuning a neural network classifier in a transfer learning setting. 
Following the ensembles trend of time series classifiers, we showed how we can leverage the variance in a DNN due to the stochastic nature of its optimization process, in order to gain a significant boost in accuracy by ensembling various neural networks with the same or different architectures. 
We then proposed to further build again upon the DBA algorithm to generate synthetic time series, allowing us to augment the training set and eventually improve the generalization capability of a deep learning classifier. 
Finally, we turned our attention to a very hot and trending topic in machine learning: adversarial attacks. 
We showed how vulnerable DNNs are to adversarial examples and highlighted several use case studies where such attacks could be detrimental, while showing how the latter technique can be employed as part of a regularization method called adversarial training. 

Building upon this knowledge gained from studying the field of TSC, we identified a main bottleneck that hinders the practical usage of many published ensembles in real life time series data mining problems. 
This downside stems from focusing on developing very accurate classifiers, while ignoring the running time of the classifier. 
We were therefore keen on developing the first neural network ensemble (called \ourmethod{}) that is able to reach similar results to the current state-of-the-art ensemble HIVE-COTE, while providing scalability in terms of long and large time series dataset.
The magnitude of this speed up is consistent across both Big Data TSC repositories as well as longer time series with high sampling rate. 

Motivated originally by the problem of evaluating surgical skills from kinematic time series sensor data, we designed a special CNN network by leveraging our understanding of DNNs gained during the initial study conducted on domain agnostic TSC problems. 
For this specific surgical data science project, we encountered many challenges that were not present in other TSC problems: such as having very high sampling rates and a huge number of variables to work with. 
For this task, we provided an FCN based classifier while focusing on its interpretability in order to help surgeons using the system to improve their skill set. 

Having summarized the aforementioned projects, we will present in the next section the different limitations of our approaches as well as the potential research area that could be of interest to many TSC researchers looking deeper into artificial neural networks. 

\section*{Discussion of future works}
\addcontentsline{toc}{section}{Discussion of future works}

The deep learning benchmark for TSC published during this thesis has taken a big leap towards introducing time series data mining practitioners to the potential of neural networks when classifying sequential temporal data. 
While still being one of the largest studies of deep learning for TSC to date, our review focused on a small subset of discriminative end-to-end DNNs. 
The latter means that we have excluded many types of approaches based on self-supervised learning such as training a neural network on a pre-text task with large unlabeled dataset then using the learned latent representation as input features to an off-the-shelf classifier~\citep{franceschi2019unsupervised}. 
Since self-supervised learning has allowed computer vision researchers to leverage efficiently the large amount of unlabeled images and videos on the web~\citep{jing2020self}, we believe that the TSC community would benefit from benchmarking this type of approaches, allowing the exploitation of a large quantity of unlabeled raw time series data.

Following the review in Chapter~\ref{Chapter1}, we have presented various regularization methods that help in improving the generalization capabilities of a given neural network for TSC. 
We believe that there exist much more research potential in developing specific time series data augmentation techniques that make use of the temporal aspect of the data. 
For example we could leverage a recent proposed approach that allows the network to learn the optimal warping by applying continuous piecewise affine transformations~\citep{weber2019diffeomorphic}, and thus generating an infinite number of warped time series training examples. 
Furthermore, restricting transfer learning to a single architecture limits the potential of this famous implicit regularization technique, therefore it would be interesting to study how various architectures could benefit / harm the neural network's accuracy when used in a fine-tuning setting. 
In addition, having showed that ensembling DNNs by averaging the output class provabilities leads to a certain increase in the network's performance, we believe that there is still room for improvement, since averaging the a posteriori probability for each class is a very basic approach and perhaps a more robust meta-ensembling technique would demonstrate further improvements~\citep{boubrahimi2018neuro}. 
Finally, with adversarial attacks exposing the vulnerabilities of neural networks, we believe that more and more researchers should focus on developing defenses for TSC models. 
In fact, the work done in this thesis on crafting adversarial examples has already inspired researchers world wide to propose temporal defense mechanisms~\citep{abdu2020detecting}. 

In Chapter~\ref{Chapter3} we focused on the scalability of current state-of-the-art methods and proposed an Inception based network for TSC. 
We were inspired by the recent neural architecture advances in the field of computer vision. 
Following the same line of thinking, researchers would further adapt the current state-of-the-art machine learning approaches such as designing algorithms that themselves would design a neural network. 
\cite{rakhshani2020neural} spotted the potential of meta-heuristics - specifically differential evolution - when building neural networks that are specific to each TSC problem.
We therefore believe that \ourmethod{} is not the ultimate solution to all TSC datasets, but rather a strong and robust starting point to any data mining practitioner that would like to design a DNN for solving their TSC problem. 
Furthermore, perhaps DNNs are not the sole answer when choosing the best TSC algorithm. 
In fact, when referring to the "no-free-lunch theorem" developed by~\cite{wolpert1995no}, and based on our results we believe that perhaps the best approach, is a meta-algorithm that would predict the best classifier based on characteristics extracted from the data. 

Going back to our initial motivation for solving TSC problems in Chapter~\ref{Chapter4}, we developed an FCN based classifier for MTS classification, allowing us to leverage the CAM technique to provide a level of model interpretability. 
We would like to first highlight the fact that our feedback technique would benefit from an extended real use-case validation process, for example verifying with expert surgeons if indeed the model is able to detect the main reason for classifying a surgical skill.
However, this is usually expensive and was not possible during this thesis, but perhaps a research project in collaboration with expert and novice surgeons would be extremely interesting to the community.
In addition, the fact that we are performing only a leave-one-super-trial-out setup means that a surgeon should be present in the training set in order to make a prediction. 
However, since only two experts exist in the dataset, this suggests that performing a leave-one-user-out setup would mean having only one expert in the training set. 
This constitutes a huge problem originating from the limited dataset size. 
Therefore, we believe that our approach should be validated on a larger dataset.
Nevertheless, another potential solution is to perform adversarial learning in order to force the network to detect patterns that are discriminative in terms of surgical skills but not invariant in terms of subject ID. 
This would be inspired from the recent success of adversarial learning for speaker invariant speech recognition systems~\citep{adi2019to}.

Finally, with deep learning showing successful results in various machine learning fields such text mining, image segmentation and speech recognition, we believe that there is still much more to be explored for the TSC research area, especially since the deep learning discipline is moving very fast. 
We hope that this thesis, with its accompanied code and published models, could be a cornerstone for future deep learning research on time series classification.



\appendix 




\printbibliography[heading=bibintoc]


\end{document}